\documentclass[nohyperref]{article}

\usepackage{xcolor}
\definecolor{tolblue}{HTML}{77AADD}
\definecolor{tolorange}{HTML}{EE8866}
\definecolor{tolpink}{HTML}{FFAABB}
\definecolor{tolcyan}{HTML}{99DDFF}
\definecolor{tolmint}{HTML}{44BB99}
\definecolor{tolpear}{HTML}{BBCC33}
\definecolor{tololive}{HTML}{AAAA00}
\definecolor{tolgrey}{HTML}{DDDDDD}
\definecolor{recycle}{RGB}{75,113,43}

\usepackage{xurl}
\usepackage{graphicx}
\definecolor{mydarkblue}{rgb}{0,0.08,0.45}
\usepackage[pagebackref=true,colorlinks=true,linkcolor=tolblue,citecolor=mydarkblue,urlcolor =mydarkblue]{hyperref}
\usepackage{amsmath}

\usepackage{pifont}%
\usepackage{cleveref}
\usepackage{scalerel}
\usepackage{caption}
\usepackage{wrapfig}
\usepackage{subcaption}
\usepackage{bbding}
\usepackage{placeins}
\usepackage{microtype}
\usepackage{csquotes}
\usepackage{enumitem}
\usepackage{xspace}
\usepackage{booktabs}
\usepackage{mathtools}
\usepackage{amsthm}

\usepackage{tikz}
\usepackage{multirow}
\usepackage{rotating}
\usepackage{amssymb}
\usepackage[export]{adjustbox}
\usepackage{subcaption}

\usetikzlibrary{shapes.geometric, positioning, arrows}

\newcommand{\xmark}{\ding{55}}%
\newcommand{\urlcode}{\url{https://github.com/facebookresearch/ModelRatatouille}}

\definecolor{vividviolet}{rgb}{0.62, 0.0, 1.0}

\newcommand*\U{\mathcal{U}}

\newcommand*{\eg}{e.g.,\@\xspace}

\newcommand*{\ie}{i.e.,\@\xspace}
\newcommand*{\iid}{ID\@\xspace}
\newcommand*{\sota}{SoTA\@\xspace}
\newcommand*{\ood}{OOD\@\xspace}
\newcommand*{\wrt}{w.r.t.\@\xspace}
\newcommand*{\aka}{a.k.a.\@\xspace}

\let\originalleft\left
\let\originalright\right
\renewcommand{\left}{\mathopen{}\mathclose\bgroup\originalleft}
\renewcommand{\right}{\aftergroup\egroup\originalright}

\def\eqref#1{eq.~\ref{#1}}

\def\1{\bm{1}}

\def\U{\mathcal{U}}

\DeclareMathAlphabet{\mathsfit}{\encodingdefault}{\sfdefault}{m}{sl}
\SetMathAlphabet{\mathsfit}{bold}{\encodingdefault}{\sfdefault}{bx}{n}

\usepackage[accepted]{icml2023}

\renewcommand*{\backrefalt}[4]{%
    \ifcase #1 \footnotesize{(Not cited.)}%
    \or        \footnotesize{(p.~#2)}%
    \else      \footnotesize{(pp.~#2)}%
    \fi}

\newtheorem{observation}{Observation}
\newtheorem{hypothesis}{Hypothesis}
\Crefname{observation}{Observation}{Observations}
\Crefname{hypothesis}{Hypothesis}{Hypotheses}
\newtheorem{remark}{Remark}
\Crefname{remark}{Remark}{Remarks}

\icmltitlerunning{Model Ratatouille: Recycling Diverse Models for Out-of-Distribution Generalization}

\begin{document}
\twocolumn[
    \icmltitle{Model Ratatouille:\\ Recycling Diverse Models for Out-of-Distribution Generalization}

    \begin{icmlauthorlist}
        \icmlauthor{Alexandre Ramé}{1,2}
        \icmlauthor{Kartik Ahuja}{1}
        \icmlauthor{Jianyu Zhang}{1,3}
        \icmlauthor{Matthieu Cord}{2,4}
        \icmlauthor{Léon Bottou}{1,3}
        \icmlauthor{David Lopez-Paz}{1}
    \end{icmlauthorlist}

    \icmlaffiliation{1}{Meta AI, Paris, France}
    \icmlaffiliation{2}{Sorbonne Université, CNRS, ISIR, Paris, France}
    \icmlaffiliation{3}{NYU, New-York, USA}
    \icmlaffiliation{4}{Valeo.ai, Paris, France}
    \icmlcorrespondingauthor{Alexandre Ramé}{alexandre.rame@isir.upmc.fr}

    \icmlkeywords{Deep Learning, ICML}

    \vskip 0.3in
]

\printAffiliationsAndNotice  %

\begin{abstract}
    Foundation models are redefining how AI systems are built.
    Practitioners now follow a standard procedure to build their machine learning solutions: from a pre-trained foundation model, they fine-tune the weights on the target task of interest.
    So, the Internet is swarmed by a handful of foundation models fine-tuned on many diverse tasks:
    these individual fine-tunings exist in isolation without benefiting from each other.
    In our opinion, this is a missed opportunity, as these specialized models contain \emph{rich and diverse} features.
    In this paper, we thus propose \emph{model ratatouille}, a new strategy to recycle the multiple fine-tunings of the same foundation model on diverse auxiliary tasks.
    Specifically, we repurpose these auxiliary weights as initializations for multiple parallel fine-tunings on the target task; then, we average all fine-tuned weights to obtain the final model.
    This recycling strategy aims at maximizing the diversity in weights by leveraging the diversity in auxiliary tasks.
    Empirically, it improves the state of the art on the reference DomainBed benchmark for out-of-distribution generalization.
    Looking forward, this work contributes to the emerging paradigm of \emph{updatable machine learning} where, akin to open-source software development, the community collaborates to reliably update machine learning models.
    Our code is released \href{https://github.com/facebookresearch/ModelRatatouille}{here}.
\end{abstract}
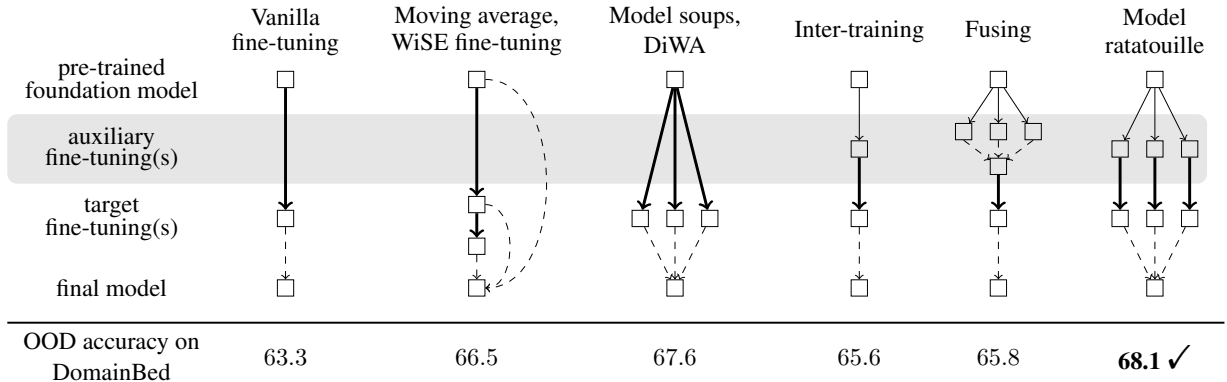
\begin{figure*}[t!]
  \begin{center}
    \resizebox{0.95\textwidth}{!}{\begin{tikzpicture}
  \newcommand{\YA}{-2.5}
  \newcommand{\YB}{0}
  \newcommand{\YC}{2.75}

  \newcommand{\YD}{8.25}

  \newcommand{\YEa}{5.1}
  \newcommand{\YE}{5.6}
  \newcommand{\YEc}{6.1}

  \newcommand{\YF}{10.25}
  \newcommand{\YFa}{9.75}
  \newcommand{\YFc}{10.75}

  \newcommand{\YG}{12.5}
  \newcommand{\YGa}{12}
  \newcommand{\YGc}{13}

  \fill[rounded corners, fill=black!10!white] (-4, -1.5) rectangle (13.25, -2.5);

 \node at (\YA, -1.15) (a) {foundation model};
  \node at (\YA, -0.85) (a) {pre-trained};

  \node at (\YA, -2.15) (a) {fine-tuning(s)};
  \node at (\YA, -1.85) (a) {auxiliary};

  \node at (\YA, -3.15) (a) {fine-tuning(s)};
  \node at (\YA, -2.85) (a) {target};

  \node at (\YA, -4) (a) {final model};

  \node at (\YA, -5.2) (a) {DomainBed};
  \node at (\YA, -4.8) (a) {OOD accuracy on};

  \draw[-, thick] (-4, -4.5) -- (13.5, -4.5);
  \node at (\YB, -5) (a) {$63.3$};
  \node at (\YC, -5) (a) {$66.5$};
  \node at (\YD, -5) (a) {$65.6$};
  \node at (\YE, -5) (a) {$67.6$};
  \node at (\YF, -5) (a) {$65.8$};
  \node at (\YG, -5) (a) {$\textbf{68.1}$ \Checkmark};

  \node at (\YB, -0.10) (a) {Vanilla};
\node at (\YB, -0.50) (a) {fine-tuning};
\node at (\YC, -0.10) (a) {Moving average,};
 \node at (\YC, -0.50) (a) {WiSE fine-tuning};
 \node at (\YD, -0.30) (a) {Inter-training};
 \node at (\YE, -0.10) (a) {Model soups,};
 \node at (\YE, -0.50) (a) {DiWA};
 \node at (\YF, -0.30) (a) {Fusing};
  \node at (\YG, -0.10) (a) {Model};
  \node at (\YG, -0.50) (a) {ratatouille};

  \node[draw] at (\YB, -1) (a1) {};
  \node[draw] at (\YC, -1) (a2) {};
  \node[draw] at (\YD, -1) (a3) {};
  \node[draw] at (\YE, -1) (a4) {};
  \node[draw] at (\YF, -1) (a5) {};
  \node[draw] at (\YG, -1) (a6) {};

  \node[draw] at (\YD, -2) (b3) {};

  \node[draw] at (\YFa, -1.75) (b5a) {};
  \node[draw] at (\YF, -1.75) (b5b) {};
  \node[draw] at (\YFc, -1.75) (b5c) {};
  \node[draw] at (\YF,  -2.25) (e5) {};

  \node[draw] at (\YGa, -2) (b6a) {};
  \node[draw] at (\YG,   -2) (b6b) {};
  \node[draw] at (\YGc, -2) (b6c) {};

  \node[draw] at (\YB, -3) (c1) {};
  \node[draw] at (\YC, -3.4) (c2) {};
 \node[draw] at (\YC, -2.8) (c21) {};
  \node[draw] at (\YD, -3) (c3) {};
  \node[draw] at (\YF, -3) (c5) {};

  \node[draw] at (\YEa, -3) (c4a) {};
  \node[draw] at (\YE, -3) (c4b) {};
  \node[draw] at (\YEc, -3) (c4c) {};

  \node[draw] at (\YGa, -3)  (c6a) {};
  \node[draw] at (\YG,  -3)   (c6b) {};
  \node[draw] at (\YGc, -3) (c6c) {};

  \node[draw] at (\YB, -4) (d1) {};
  \node[draw] at (\YC, -4) (d2) {};
  \node[draw] at (\YD, -4) (d3) {};
  \node[draw] at (\YE, -4) (d4) {};
  \node[draw] at (\YF, -4) (d5) {};
  \node[draw] at (\YG, -4) (d6) {};

  \draw[->, very thick] (a1) -- (c1);
  \draw[->, dashed] (c1) -- (d1);

  \draw[->, very thick] (a2) -- (c21);
  \draw[->, very thick] (c21) -- (c2);
  \draw[->, dashed] (c2) -- (d2);

  \draw [->,dashed] (c21) to [out=0,in=0] (d2);
  \draw [->,dashed] (a2) to [out=0,in=0] (d2);

  \draw[->] (a3) -- (b3);
  \draw[->, very thick] (b3) -- (c3);
  \draw[->, dashed] (c3) -- (d3);

  \draw[->, very thick] (a4) -- (c4a);
  \draw[->, very thick] (a4) -- (c4b);
  \draw[->, very thick] (a4) -- (c4c);

  \draw[->, dashed] (c4a) -- (d4);
  \draw[->, dashed] (c4b) -- (d4);
  \draw[->, dashed] (c4c) -- (d4);

  \draw[->] (a5) -- (b5a);
  \draw[->] (a5) -- (b5b);
  \draw[->] (a5) -- (b5c);

  \draw[->, dashed] (b5a.south) -- (e5);
  \draw[->, dashed] (b5b) -- (e5);
  \draw[->, dashed] (b5c.south) -- (e5);

  \draw[->, very thick] (e5) -- (c5);
  \draw[->, dashed] (c5) -- (d5);

  \draw[->] (a6) -- (b6a);
  \draw[->] (a6) -- (b6b);
  \draw[->] (a6) -- (b6c);

  \draw[->, very thick] (b6a) -- (c6a);
  \draw[->, very thick] (b6b) -- (c6b);
  \draw[->, very thick] (b6c) -- (c6c);

  \draw[->, dashed] (c6a) -- (d6);
  \draw[->, dashed] (c6b) -- (d6);
  \draw[->, dashed] (c6c) -- (d6);
\end{tikzpicture}}
  \end{center}
  \vskip -0.4cm
  \caption{The different fine-tuning strategies discussed in this paper: vanilla fine-tuning~\cite{oquab2014learning}, moving average~\cite{izmailov2018} and variants~\cite{Wortsman2022robust}, model soups~\cite{Wortsman2022ModelSA} and DiWA~\cite{rame2022diwa}, inter-training \cite{phang2018sentence}, fusing \cite{choshen2022fusing} and our proposed \emph{model ratatouille}. They all start with a pre-trained foundation model.
    Some strategies fine-tune the pre-trained model on auxiliary tasks (thin solid arrows \protect\tikz[baseline]\protect\draw[->](0ex,0.8ex) -- (3ex,0.8ex);): these auxiliary fine-tunings can be performed by different contributors of the community on their own data.
    Then, all strategies perform fine-tuning on the target task of interest (thick solid arrows \protect\tikz[baseline]\protect\draw[->, very thick](0ex,0.8ex) -- (3ex,0.8ex);). Finally, the weights fine-tuned on the target task are used as is, or are averaged (dashed arrows \protect\tikz[baseline]\protect\draw[->, dashed](0ex,0.8ex) -- (3ex,0.8ex);) into a final model.
    Ratatouille (i) enables compute parallelism throughout training, (ii) maximizes the amount of diversity in models' predictions, (iii) achieves state-of-the-art performance in DomainBed \cite{gulrajani2021in}, the standard computer vision benchmark for \ood generalization and (iv) does not incur any inference or training overhead compared to a traditional hyperparameter search.}
  \label{fig:intro}%
\end{figure*}

\section{Introduction}
\label{sec:introduction}

The framework of \emph{foundation models}~\cite{bommasani2021opportunities} is fueling a spectacular adoption of machine learning solutions for real-world applications:
also known as pre-trained models, these machine learning systems are trained on large-and-diverse data \cite{pmlr-v162-fang22a,nguyen2022quality,exploringlimits51104} and easy to adapt to downstream tasks.
Having ditched the \enquote{training from scratch} mentality, practitioners now follow a standardized two-step transfer learning strategy \cite{oquab2014learning}.
From some foundation model, they fine-tune on their target task with usually a limited amount of in-house data.
Unfortunately, each of these fine-tunings risks latching onto specific patterns from the practitioners' training data~\cite{arjovsky2019invariant,pmlr-v119-miller20a,NEURIPS2020_6cfe0e61}.
Thus, these shortsighted models struggle to generalize on out-of-distribution (\ood) samples \cite{hendrycks2018benchmarking,NEURIPS2020_d8330f85,gulrajani2021in,hendrycks2021many}, negatively impacting human lives~\cite{taylor2016alignment,Zech2018}.
Increased \ood generalization would enable the responsible use of machine learning in real-world applications where robustness and safety are critical, such as medical imaging \cite{degrave2021ai} and autonomous driving \cite{kuutti2020survey}.

How to best fine-tune foundation models for \ood generalization is thus becoming a central topic of research.
In particular, the recently discovered ability to average neural networks' weights~\cite{izmailov2018, Neyshabur2020} has inspired a plethora of modern fine-tuning approaches.
We illustrate some of them in \Cref{fig:intro}, such as moving averages~\cite{izmailov2018}, WiSE fine-tuning~\cite{Wortsman2022robust}, model soups~\cite{Wortsman2022ModelSA} and DiWA~\cite{rame2022diwa}.
However, these strategies cannot accommodate the swarms of specialized fine-tunings of the same foundation model increasingly available in the Internet.
Recent inter-training~\cite{phang2018sentence,pruksachatkun2020intermediate} and fusing \cite{choshen2022fusing,choshen2022cold} strategies recycle intermediate fine-tunings on auxiliary tasks to enrich the features before fine-tuning on the target task.
However, the success of these recycling strategies usually depend on the similarity between the auxiliary and target tasks.
We also argue in \Cref{sec:finetuning} that these strategies fail to fully leverage the diversity in auxiliary tasks, even though feature diversity improves \ood generalization~\cite{laakom2021within,nayman2022diverse,pmlr-v162-jain22b,zhang2022rich}.

Thus, the central question of this paper is:
\begin{center}
  \emph{How can we best recycle diverse fine-tunings of a given foundation model towards strong out-of-distribution performance on our target task?}
\end{center}
Our answer is a simple fine-tuning strategy we named \emph{model ratatouille},\footnote{We named our method after this traditional French dish for two main reasons. Firstly, the ratatouille is often used as a way to recycle leftover vegetables. Secondly, the ratatouille is better prepared by cooking each ingredient separately before mixing them: this technique ensures that each ingredient “will taste truly of itself”, as noted by chef Joël Robuchon \cite{ratatouille2020}.} illustrated in \Cref{fig:intro} and described in \Cref{sec:recycling}.
In a similar fashion to converting waste into reusable material for new uses, we take fine-tunings of the same foundation model on diverse auxiliary tasks and repurpose them as initializations to start multiple fine-tunings on the target downstream task.
Specifically, we (i) fine-tune a copy of the foundation model on each auxiliary task, (ii) fine-tune each auxiliary model on the target task, and (iii) return as the final model the average of all target fine-tuned weights.
In brief, while model soups~\cite{Wortsman2022ModelSA} averages multiple weights fine-tuned from a shared initialization, model ratatouille averages multiple weights fine-tuned from different initializations each inter-trained \cite{phang2018sentence} on different auxiliary tasks.
As we will see, ratatouille works because the fine-tunings remain linearly connected \cite{Frankle2020,mirzadeh2021linear} in the loss landscape (despite having different initializations) and thus can be averaged for improved performance.

We show the efficacy of model ratatouille in \Cref{sec:experiments}, where we set a new state of the art on DomainBed~\cite{gulrajani2021in}, the reference benchmark evaluating \ood generalization.
We will show how we leverage the diversity across the auxiliary tasks to construct a final model with decreased over-fitting to task-specific patterns.
As we discuss in our closing \Cref{sec:discussion},
this work contributes to the emerging paradigm of \emph{updatable machine learning}~\cite{updatablemachinelearning}, where practitioners work in collaboration towards incrementally and reliably updating the capabilities of a machine learning model.
As also highlighted in recent works~\cite{mergefisher21,li2022branch}, we envision a future where deep neural networks are trained by following similar pipelines to the ones in open-source development with version control systems.

\section{Fine-Tuning for \ood Generalization}
\label{sec:finetuning}

We start by describing our setup.
We train a deep model $f_\theta = f_w \circ f_\phi$, where the featurizer $f_\phi$ is parametrized by the weights $\phi$, the classifier $f_w$ is parametrized by the weights $w$, and the joint model $f_\theta$ is parametrized by the concatenation weights $\theta = (w, \phi)$.
We are dealing with out-of-distribution (OOD) generalization, and our aim is to find $\theta$ maximizing the test accuracy $\mathrm{acc}_{\mathrm{te}}(\theta)$. Specifically, while both train and test data correspond to the same target task---classifying images into a fixed set of classes---we allow a diversity \cite{ye2021odbench} (\aka covariate) distribution shift between the two, \ie that the input distributions may change at test time.
We highlight that this \ood generalization is critical in real-life applications, where the model needs to predict on samples from a new domain.

\textbf{Vanilla fine-tuning.}
For \ood generalization, transfer learning~\cite{oquab2014learning,plex2022,assayingoodwenzel2022} with empirical risk minimization \citep[ERM]{NIPS1991_ff4d5fbb} is frustratingly difficult to beat~\cite{gulrajani2021in}, as measured on real-world datasets \cite{fang2023does} such as PACS \cite{li2017deeper}, VLCS \cite{fang2013unbiased}, OfficeHome \cite{venkateswara2017deep}, TerraIncognita \cite{beery2018recognition} or DomainNet \cite{peng2019moment}.
The recipe is (i) download a pre-trained featurizer with parameters $\phi^\mathrm{pt}$, (ii) plug a classifier $w^\mathrm{lp}$ compatible with the target task, and (iii) fine-tune the network with ERM on the target task.
While the classifier $w^\mathrm{lp}$ could be initialized at random, linear probing (\ie first learning only the classifier with frozen featurizer) improves results by preventing feature distortion~\cite{kumar2022finetuning}.
For most users, particularly those with modest computation resources, the standard strategy is thus to transfer the knowledge from models pre-trained on large dataset such as ImageNet~\cite{russakovsky2015imagenet}, downloaded from public repositories such as \texttt{torchvision} \cite{torchvision2010}, \texttt{huggingface} \cite{wolf-etal-2020-transformers} or \texttt{timm} \cite{rw2019timm}.
The users usually launch multiple fine-tunings with different hyperparameters, and select the best based on some validation metric \cite{gulrajani2021in}.

\textbf{Weight averaging over epochs.}
Recently, \emph{weight averaging} strategies came to the foreground~\citep{szegedy2016rethinking,izmailov2018,Draxler2018}.
While fine-tuning a pre-trained model, they saved and averaged checkpoints every few epochs to build the final model.
Due to the nonlinear nature of deep neural networks, the efficacy of weight averaging was a surprising observation, that \citet{Frankle2020} latter called the linear mode connectivity.
\begin{observation}[LMC with different epochs~\cite{izmailov2018}]
  Two weights $\theta_a$ and $\theta_b$, obtained at two different epochs of the same fine-tuning, satisfy the linear mode connectivity (LMC): for all $\lambda \in [0, 1]$,
  \begin{equation}
    \begin{split}
      \mathrm{acc}_\mathrm{te}(\left(1 - \lambda\right) & \cdot \theta_a + \lambda \cdot \theta_b) \gtrsim \\
      & \left(1 - \lambda\right) \cdot \mathrm{acc}_\mathrm{te}(\theta_a) + \lambda \cdot \mathrm{acc}_\mathrm{te}(\theta_b).
    \end{split}
  \end{equation}
  \label{obs:1}%
\end{observation}%
The LMC holds if the accuracy of the interpolated weights is above the interpolated accuracy.
This definition is more restrictive than in the literature; for example, \citet{Frankle2020} only required less than $2\%$ in error increase with regard to the worst endpoints.
Consistently with \Cref{obs:1}, recent works \cite{arpit2021ensemble,cha2021wad,Wortsman2022robust,kaddour2022stop}  weight average checkpoints along training to improve accuracies.%

\textbf{Weight averaging over runs.}
Perhaps motivated by these results, \citet{Neyshabur2020} (along with similar works \cite{nagarajan2019uniform,Frankle2020}) pushed the envelope of weight averaging techniques, and stated:
\begin{quote}
  \begin{small}
    there is no performance barrier between two instances of models trained from pre-trained weights, which suggests that the pre-trained weights guide the optimization to a flat basin of the loss landscape [\ldots] Moreover, interpolating two random solutions from the same basin could generally produce solutions closer to the center of the basin, which potentially have better generalization performance than the endpoints.%
  \end{small}%
\end{quote}%
Two \emph{independent} fine-tunings---pre-trained similarly but differing in hyperparameter choices, data orders or other stochastic factors---also satisfy the LMC! More formally,%
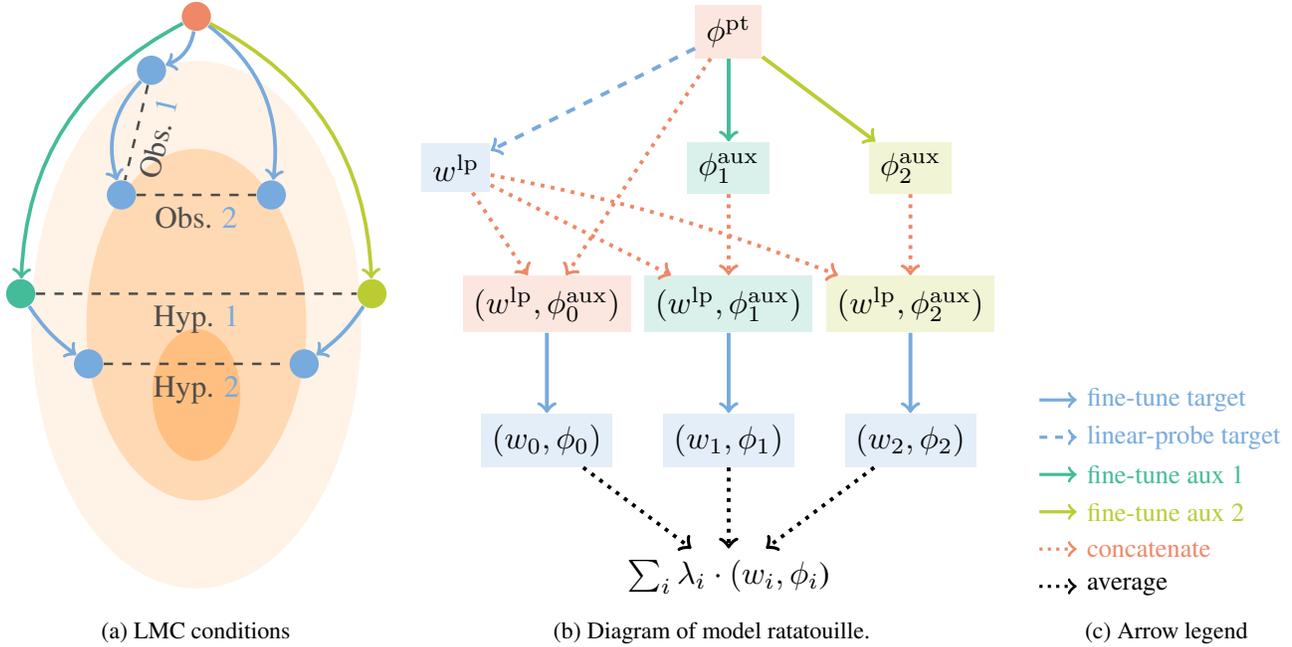
\begin{figure*}[t]
  \begin{center}
    \begin{subfigure}[b]{0.295\textwidth}
      \resizebox{\textwidth}{!}{\begin{tikzpicture}[x=1cm, y=1cm]
  \node[ellipse, minimum height = 7cm, minimum width = 4.00cm, fill=orange!0] (vbig) at (0,0) {};
  \node[ellipse, minimum height = 6cm, minimum width = 3.75cm, fill=orange!10] (ebig) at (0,0) {};
  \node[ellipse, minimum height = 4cm, minimum width = 2.5cm, fill=orange!30] (esmall) at (0,0) {};
  \node[ellipse, minimum height = 1.5cm, minimum width = 1.cm, fill=orange!50] (vsmall) at (0,-0.8) {};

  \node[circle, fill=black, radius=1pt, tolorange] (pretrain) at (vbig.90) {};
  \node[circle, fill=black, radius=1pt, tolblue] (epoch0) at (ebig.100) {};
  \node[circle, fill=black, radius=1pt, tolblue] (epoch1) at (esmall.120) {};

  \node[circle, fill=black, radius=1pt, tolblue] (soup2) at (esmall.60) {};

  \node[circle, fill=black, radius=1pt, tolmint] (aux1) at (vbig.170) {};
  \node[circle, fill=black, radius=1pt, tolblue] (fine1) at (esmall.200) {};

  \node[circle, fill=black, radius=1pt, tolpear] (aux2) at (vbig.10) {};
  \node[circle, fill=black, radius=1pt, tolblue] (fine2) at (esmall.340) {};

  \draw[->, very thick, tolblue] (pretrain) to [bend left] (soup2);
  \draw[dashed, thick, black!70] (epoch1) -- (soup2) node [pos=.5, below, sloped] (obs2) {Obs. \ref{obs:2}};

  \draw[->, very thick, tolblue] (pretrain) to [bend left] (epoch0);
  \draw[->, very thick, tolblue] (epoch0) to [bend right] (epoch1);
  \draw[dashed, thick, black!70] (epoch0) -- (epoch1) node [pos=.5, below, sloped] (obs1) {Obs. \ref{obs:1}};

  \draw[dashed, thick, black!70] (aux1) -- (aux2) node [pos=.5, below, sloped] (hyp1) {Hyp. \ref{hyp:1}};
  \draw[dashed, thick, black!70] (fine1) -- (fine2) node [pos=.5, below, sloped] (hyp2) {Hyp. \ref{hyp:2}};

  \draw[->, very thick, tolmint] (pretrain) to [bend right] (aux1);
  \draw[->, very thick, tolpear] (pretrain) to [bend left] (aux2);
  \draw[->, very thick, tolblue] (aux1)   to [bend right=10] (fine1);
  \draw[->, very thick, tolblue] (aux2)   to [bend left=10] (fine2);
\end{tikzpicture}%
}
      \vspace{-0.8cm}
      \subcaption[b]{LMC conditions}%
      \label{fig:recycling:landscape}%
    \end{subfigure}%
    \hfill%
    \begin{subfigure}[b]{0.45\textwidth}%
      \resizebox{\textwidth}{!}{\begin{tikzpicture}[x=1cm, y=1cm]
  \newcommand{\YA}{-2}
  \newcommand{\YB}{0}
  \newcommand{\YC}{2}
  \newcommand{\YL}{-6}
  \newcommand{\YG}{-5.5}
  \node[fill=tolorange!20] (pt)    at (\YB, -1) {$\phi^\mathrm{pt}$};
  \node[fill=tolblue!20] (init0) at (-3, -2.5) {$w^\mathrm{lp}$};
  \node[fill=tolmint!20] (init1) at (\YB, -2.5) {$\phi^\mathrm{aux}_1$};
  \node[fill=tolpear!20] (init2) at (\YC, -2.5) {$\phi^\mathrm{aux}_2$};

  \node[fill=tolorange!20] (lp0)   at (\YA, -4) {$(w^\mathrm{lp}, \phi^\mathrm{aux}_0)$};
  \node[fill=tolmint!20] (lp1)   at (\YB, -4) {$(w^\mathrm{lp}, \phi^\mathrm{aux}_1)$};
  \node[fill=tolpear!20] (lp2)   at (\YC, -4) {$(w^\mathrm{lp}, \phi^\mathrm{aux}_2)$};

  \node[fill=tolblue!20] (ft0)   at (\YA, -5.5) {$(w_0, \phi_0)$};
  \node[fill=tolblue!20] (ft1)   at (\YB, -5.5) {$(w_1, \phi_1)$};
  \node[fill=tolblue!20] (ft2)   at (\YC, -5.5) {$(w_2, \phi_2)$};

  \node(final) at (\YB, -7) {$\sum_i \lambda_i \cdot (w_i, \phi_i$)};

  \draw[->, very thick, tolmint] (pt) -- (init1);
  \draw[->, very thick, tolpear] (pt) -- (init2);
  \draw[->, very thick, dashed, tolblue] (pt) -- (init0);

  \draw[->, very thick, tolblue] (lp0) -- (ft0);
  \draw[->, very thick, tolblue] (lp1) -- (ft1);
  \draw[->, very thick, tolblue] (lp2) -- (ft2);

  \draw[->, very thick, dotted] (ft0) -- (final);
  \draw[->, very thick, dotted] (ft1) -- (final);
  \draw[->, very thick, dotted] (ft2) -- (final);

  \draw[->, very thick, dotted, tolorange] (init0) -- (lp0);
  \draw[->, very thick, dotted, tolorange] (init0) -- (lp1);
  \draw[->, very thick, dotted, tolorange] (init0) to [bend left=5] (lp2);

  \draw[->, very thick, dotted, tolorange] (pt) -- (lp0);
  \draw[->, very thick, dotted, tolorange] (init1) -- (lp1);
  \draw[->, very thick, dotted, tolorange] (init2) -- (lp2);
\end{tikzpicture}}%
      \subcaption[b]{Diagram of model ratatouille.}%
      \label{fig:recycling:diagram}%
    \end{subfigure}%
    \hfill
    \begin{subfigure}[b]{0.20\textwidth}%
      \resizebox{\textwidth}{!}{\begin{tikzpicture}[x=1cm, y=1cm]
  \newcommand{\YL}{0}
  \newcommand{\YG}{0.5}
  \draw[->, very thick, tolblue]           (\YL, -4) -- (\YG, -4) node [pos=1, right, sloped]     (lab1) {fine-tune target};
  \draw[->, very thick, dashed, tolblue]           (\YL, -4.5) -- (\YG, -4.5) node [pos=1, right, sloped] (lab1) {linear-probe target};

  \draw[->, very thick, tolmint]           (\YL, -5) -- (\YG, -5) node [pos=1, right, sloped] (lab1) {fine-tune aux 1};
  \draw[->, very thick, tolpear]           (\YL, -5.5) -- (\YG, -5.5) node [pos=1, right, sloped]     (lab1) {fine-tune aux 2};

  \draw[->, very thick, dotted, tolorange] (\YL, -6) -- (\YG, -6) node [pos=1, right, sloped]     (lab3) {concatenate};
  \draw[->, very thick, dotted]            (\YL, -6.5) -- (\YG, -6.5) node [pos=1, right, sloped] (lab4) {average};
\end{tikzpicture}}%
      \caption{Arrow legend}%
      \label{fig:recycling:legend}%
    \end{subfigure}%
    \hfill%
  \end{center}%
  \caption{Illustrations of (a) different linear mode connectivity (LMC) conditions, and (b) model ratatouille.
    In subplot (a), we illustrate
    \Cref{obs:1}, about LMC between two checkpoints along the same target fine-tuning;
    \Cref{obs:2}, about LMC between two target fine-tunings;
    \Cref{hyp:1}, about LMC between two auxiliary fine-tunings; and
    \Cref{hyp:2}, about LMC between two target fine-tunings initialized from auxiliary weights satisfying \Cref{hyp:1}.
    In subplot (b), we offer a diagram of our proposed recycling strategy, where we (i) fine-tune a pre-trained model on auxiliary tasks,
    (ii) plug a linear probe on the pre-trained model and the auxiliary fine-tunings, (iii) fine-tune on the target task from each auxiliary weights, and (iv) return their weight average as the final model.}%
  \label{fig:recycling}%
\end{figure*}%
\begin{observation}[LMC with different runs~\cite{Neyshabur2020}]
  The LMC holds between $\theta_a$ and $\theta_b$ fine-tuned on the target task initialized from a shared pre-trained model.
  \label{obs:2}
\end{observation}
See \Cref{fig:recycling:landscape} for an illustration of \Cref{obs:1,obs:2}.
\Cref{obs:2} was extended to reinforcement learning \cite{rame2023rewarded} and multimodal \cite{shukor2023unified} setups. This LMC inspired model soups~\cite{Wortsman2022ModelSA} and DiWA \cite{rame2022diwa}---the current state-of-the-art approaches for \ood generalization---to average all the weights obtained from a standard ERM hyperparameter search.
However, the shared initialization constraint limits models diversity~\cite{kuncheva2003measures,aksela2003comparison}, especially when compared to methods that can combine arbitrary networks, for example via prediction averaging in deep ensembles~\cite{Lakshminarayanan2017}.%

\textbf{Weight averaging over tasks.}
All the methods described so far fine-tune only on the target task:
could auxiliary datasets, increasingly available online, be incorporated into the learning process to learn richer features?
Such tasks could be an opportunity to recruit specialized features \cite{li2021Universal} that match our target task, ease optimization~\cite{zhang2022rich, anonymous2023learning}, or \enquote{offer some high-level guidance to bridge the gaps between the pre-training and fine-tuning phases}~\cite{chang2021rethinking}.
Following these ideas, \emph{inter-training}~\cite{phang2018sentence, pruksachatkun2020intermediate,choshen2022start} performs an intermediate fine-tuning of the pre-trained model on some auxiliary task, before tackling the target task.
However, the sequential nature of inter-training leads to catastrophic forgetting~\cite{rebuffi2017icarl} of useful knowledge contained in the original pre-trained model.
Moreover, the choice of the auxiliary task plays a determinant role, since \enquote{when the wrong task is chosen, inter-training hurts results}~\cite{choshen2022fusing}.
To address the shortcomings of inter-training, recent works \cite{choshen2022fusing,choshen2022cold,li2022branch,mergefisher21,2022arXiv221204089I,ilharco2022patching}
proposed to recycle weights fine-tuned on various auxiliary tasks.
In particular, concurrent \citet{choshen2022fusing} operates \emph{fusing} at initialization;
they (i) fine-tune one copy of the pre-trained model on each auxiliary task, (ii) average the auxiliary fine-tuning weights, and (iii) use such averaged model as the initialization for the target fine-tuning.
By interpolation in weights, fusing combines into one single initialization the knowledge from multiple auxiliary tasks; yet fusing empirically provides only  marginal gains in \Cref{sec:exps:ood} for OOD generalization on DomainBed.

We posit that model fusing is performing weight averaging prematurely, destroying most diversity from auxiliary tasks even before the target task can benefit from it.
To address these issues, next we propose \emph{ratatouille}, a new recycling strategy that performs one target fine-tuning per auxiliary weights, and averages weights only as the very last step.

\section{Model Ratatouille}
\label{sec:recycling}
\subsection{Recycling Diverse Initializations}
Our model ratatouille is a proposal to recycle diverse auxiliary fine-tunings of the same pre-trained model; it is compared against other fine-tuning strategies in \Cref{fig:intro} and outlined in detail in \Cref{fig:recycling:diagram}.
Ratatouille recycles these fine-tunings as diverse initializations to parallel fine-tunings on the target task. Compared to fusing, we delay the weight averaging, and in turn the destruction of diversity. Ratatouille follows this five-step recipe.
\begin{enumerate}
  \item Download a featurizer $\phi^{\mathrm{pt}}$ pre-trained on task $T_0$.
  \item Fine-tune $\phi^{\mathrm{pt}}$ on each auxiliary task $T_i$, obtaining $(w^\mathrm{aux}_i, \phi^\mathrm{aux}_i)$ for $i = 0, \ldots, M-1$.
  \item Replace each $w^\mathrm{aux}_i$ by $w^\mathrm{lp}$, obtained by linear probing the original pre-trained model $\phi^{\mathrm{pt}}$ on the target task $T$.
  \item Fine-tune each $(w^\mathrm{lp}, \phi^\mathrm{aux}_i)$ on the target task $T$, obtaining $\theta_i = (w_i, \phi_i)$ for $i = 0, \ldots, M-1$.
  \item Return as final model $\sum_{i=0}^{M-1} \lambda_i \cdot \theta_i$.
  To select the interpolating coefficients, we use two strategies. The first \enquote{uniform} averages all weights with $\lambda_i = \frac{1}{M}$. The second \enquote{greedy} sorts the $\theta_i$ by descending accuracy on the in-distribution (\iid) validation set, before greedily constructing an uniform average containing $\theta_i$ if and only if its addition lowers the \iid validation accuracy.
\end{enumerate}
If the weights from step 2 are made available online, ratatouille is without any training overhead compared to a traditional hyperparameter search.
When compared to inter-training~\cite{phang2018sentence} and fusing~\cite{choshen2022fusing}, model ratatouille avoids the difficult choice of choosing one single initialization~\cite{choshen2022start}.
The shared linear probe classifier facilitates LMC by preventing feature distortions \cite{kumar2022finetuning}.
Note that we consider the pre-training task as the auxiliary task \enquote{number zero} $T_0$; this resembles WiSE fine-tuning~\cite{Wortsman2022robust} and aims at preserving the general-purpose knowledge contained in the original pre-trained model.
The two selection strategies are those from model soups~\cite{Wortsman2022ModelSA,rame2022diwa}.

Successful weight averaging requires three conditions \cite{rame2022diwa}.
First, the weights must be individually accurate; by inter-training, ratatouille enriches the features and thus increases individual accuracies when the auxiliary tasks are well-chosen \cite{choshen2022start}.
Second, the weights should be sufficiently diverse to reduce variance. By removing the shared initialization constraint from model soups, ratatouille benefits from the additional diversity brought by specialization on various auxiliary tasks.
In essence, auxiliary tasks helps in two ways: through their similarity with the target task, and their diversity.
Third, the weights should be averageable; thus, for ratatouille to work, it requires a relaxation of the conditions under which the LMC holds, that we detail below.

\subsection{Novel Linear Mode Connectivity Hypotheses}

First, we introduce \Cref{hyp:1} that posits LMC between two models whose featurizers were fine-tuned on different auxiliary tasks.%
\begin{hypothesis}[LMC with different tasks] The LMC holds between $\left(w, \phi_a^{\mathrm{aux}}\right)$ and $\left(w, \phi_b^{\mathrm{aux}}\right)$ if $\phi_a^{\mathrm{aux}}$ and $\phi_b^{\mathrm{aux}}$ are featurizers fine-tuned on two auxiliary tasks initialized from the same pre-trained featurizer $\phi^{\mathrm{pt}}$. Here, $w$ is the linear probe of $\phi^{\mathrm{pt}}$ on the target task.\label{hyp:1}
\end{hypothesis}
Though this \Cref{hyp:1} was never formulated explicitly, it underlies fusing \cite{choshen2022fusing} and other strategies averaging auxiliary weights.
Ratatouille relies on the following \Cref{hyp:2}, which adds on top of \Cref{hyp:1} independent fine-tuning steps on the target task.
\begin{hypothesis}[LMC with different auxiliary initializations] The LMC holds between $\theta_a$ and $\theta_b$ fine-tuned on the target task starting from initializations $\left(w, \phi_a^{\mathrm{aux}}\right)$ and $\left(w, \phi_b^{\mathrm{aux}}\right)$ satisfying \Cref{hyp:1}.\label{hyp:2}%
\end{hypothesis}%
\Cref{hyp:2} is the first to posit the LMC between weights trained from different initializations.
It hints towards a more general inheritance property: if two initializations satisfy LMC, then the two final weights would too.

We expect \Cref{hyp:1,hyp:2} to hold as long as the pre-training, auxiliary and target tasks are sufficiently similar, and if hyperparameters remain in a mild range.
If they hold, we expect ratatouille to improve generalization abilities.
But this, we can only answer empirically through proper experimentation.

\section{Experiments}
\label{sec:experiments}
\begin{table*}[h!]%
    \begin{center}
        \caption{Accuracies ($\%,\uparrow$) on the DomainBed \cite{gulrajani2021in} benchmark evaluating \ood generalization. Ratatouille sets a new \sota by leveraging auxiliary tasks' diversity. The selection column indicates the weight selection strategy. The symbol \enquote{$*$} indicates inference overhead in functional ensembling. The symbol \enquote{$\dagger$} indicates the averaging of all weights across $3$ data splits.}
        \resizebox{\textwidth}{!}{%
            \begin{tabular}{lllccccccc}
                \toprule
                                                           & \textbf{Algorithm}                            & \textbf{Selection}    & \textbf{PACS}           & \textbf{VLCS}              & \textbf{OfficeHome}        & \textbf{TerraInc}  & \textbf{DomainNet}      & \textbf{Avg}       \\
                \midrule
                                                           & Vanilla fine-tuning                           & \iid val              & $85.5 \pm 0.2$          & $77.5 \pm 0.4$             & $66.5 \pm 0.3$             & $46.1 \pm 1.8$     & $40.9 \pm 0.1$          & $63.3$             \\
                                                           & CORAL \cite{coral216aaai}                     & \iid val              & $86.2 \pm 0.3$          & $78.8 \pm 0.6$             & $68.7 \pm 0.3$             & $47.6 \pm 1.0$     & $41.5 \pm 0.1$          & $64.6$             \\
                                                           & SWAD \cite{cha2021wad}                        & Loss-aware trajectory & $88.1 \pm 0.1$          & $\textbf{79.1} \pm 0.1$    & $70.6 \pm 0.2$             & $50.0 \pm 0.3$     & $46.5 \pm 0.1$          & $66.9$             \\
                                                           & MA \cite{arpit2021ensemble}                   & Uniform trajectory    & $87.5 \pm 0.2$          & $78.2 \pm 0.2$             & $70.6 \pm 0.1$             & $50.3 \pm 0.5$     & $46.0 \pm 0.1$          & $66.5$             \\
                                                           & Deep ensembles$^{*}$ \cite{arpit2021ensemble} & Uniform               & $87.6$                  & $78.5$                     & $70.8$                     & $49.2$             & $\textbf{47.7}$         & $66.8$             \\
                \midrule
                \multirow{5}{*}{\begin{turn}{90}{\small DiWA runs}\end{turn}} & Vanilla fine-tuning                           & \iid val              & $85.9 \pm 0.6$          & $78.1 \pm 0.5$             & $69.4 \pm 0.2$             & $50.4 \pm 1.8$     & $44.3 \pm 0.2$          & $65.6$             \\
                                                           & Ensemble$^{*}$                                & Uniform               & $88.1 \pm 0.3$          & $78.5 \pm 0.1$             & $71.7 \pm 0.1$             & $50.8 \pm 0.5$     & $47.0 \pm 0.2$          & $67.2$             \\
                                                           & Model soups                                   & Uniform               & $88.7 \pm 0.2$          & $78.4 \pm 0.2$             & $72.1 \pm 0.2$             & $51.4 \pm 0.6$     & $47.4 \pm 0.2$          & $67.6$             \\
                                                           & Model soups                                   & Greedy                & $88.0 \pm 0.3$          & $78.5 \pm 0.1$             & $71.5 \pm 0.2$             & $51.6 \pm 0.9$     & $\textbf{47.7} \pm 0.1$ & $67.5$             \\
                                                           & Model soups$^\dagger$                         & Uniform$^\dagger$     & $89.0$                  & $78.6$                     & $72.8$                     & $\underline{51.9}$ & $\textbf{47.7}$         & $68.0$             \\
                \midrule
                \multirow{6}{*}{\begin{turn}{90}{\small Our runs}\end{turn}} & Inter-training \cite{phang2018sentence}       & \iid val              & $89.0 \pm 0.0$          & $77.7 \pm 0.0$             & $69.9 \pm 0.6$             & $46.7 \pm 0.1$     & $44.5 \pm 0.1$          & $65.6$             \\
                                                           & Ensemble$^{*}$ of inter-training              & Uniform               & $89.2 \pm 0.1$          & $\underline{79.0} \pm 0.2$ & $72.7 \pm 0.1$             & $51.1 \pm 0.3$     & $47.2 \pm 0.1$          & $67.8$             \\
                                                           & Fusing \cite{choshen2022fusing}               & \iid val              & $88.0 \pm 1.0$          & $78.5 \pm 0.8$             & $71.5 \pm 0.5$             & $46.7 \pm 1.8$     & $44.4 \pm 0.2$          & $65.8$             \\
                                                           & Model ratatouille                             & Uniform               & $89.5 \pm 0.1$          & $78.5 \pm 0.1$             & $73.1 \pm 0.1$             & $51.8 \pm 0.4$     & $47.5 \pm 0.1$          & $\underline{68.1}$ \\
                                                           & Model ratatouille                             & Greedy                & $\textbf{90.5} \pm 0.2$ & $78.7 \pm 0.2$             & $\underline{73.4} \pm 0.3$ & $49.2 \pm 0.9$     & $\textbf{47.7} \pm 0.0$ & $67.9$             \\
                                                           & Model ratatouille$^{\dagger}$                 & Uniform$^\dagger$     & $\underline{89.8}$      & $78.3$                     & $\textbf{73.5}$            & $\textbf{52.0}$    & $\textbf{47.7}$         & $\textbf{68.3}$    \\
                \bottomrule
                \label{table:domainbed}
            \end{tabular}}%
    \end{center}%
\end{table*}%
Our numerical experiments support four main claims, sorted in decreased granularity.
First, \Cref{sec:exps:ood} showcases the state-of-the-art (\sota) results of ratatouille in DomainBed~\cite{gulrajani2021in}.
Second, \Cref{sec:exps:div} illustrates how such gains arise from increased diversity across averaged models.
Third, \Cref{sec:exps:lmc} empirically supports \Cref{hyp:1,hyp:2}, the technical conditions enabling weight averaging's success.
Finally, \Cref{sec:exps:iid} discusses the impact of ratatouille for in-domain tasks.
We invite the curious reader to consult our supplementary material. Among other experiments, we ablate in \Cref{app:trainingcost} the different components of ratatouille's procedure such as the number of auxiliary tasks, and propose in \Cref{app:robustintertraining} a robust ratatouille to further improve performance. Our code is released at \urlcode.
\subsection{\sota Performance on DomainBed}
\label{sec:exps:ood}
\textbf{Setup.}
\Cref{table:domainbed} shows our main experiment comparing the various fine-tuning strategies on DomainBed~\cite{gulrajani2021in}, the reference benchmark evaluating \ood generalization.
DomainBed contains five real-world datasets: PACS \cite{li2017deeper}, VLCS \cite{fang2013unbiased}, OfficeHome \cite{venkateswara2017deep}, TerraIncognita \cite{beery2018recognition} and DomainNet \cite{peng2019moment}.
Each contains multiple domains about the same classification task: for example, the domains in OfficeHome are \enquote{Art}, \enquote{ClipArt}, \enquote{Product} and \enquote{Photo}.
Each domain is successively considered as the test while others are for training; we report the $22$ per-domain results in \Cref{app:domainbedperdataset} but here analyze the averaged accuracy over the test domains. Standard deviations are obtained on $3$ different random data splits.
The network is a ResNet-50 \cite{he51deep} pre-trained on ImageNet \cite{russakovsky2015imagenet}.
Following DomainBed standards, each strategy leverages $20$ runs with hyperparameters sampled from \Cref{tab:hyperparam}.

\textbf{Approaches.}
Model soups \cite{Wortsman2022ModelSA,rame2022diwa} only differs from vanilla fine-tuning by the selection strategy: rather than selecting the model with highest \iid validation accuracy out of the $20$ runs, model soups either uniformly averages all weights or greedily selects some---as described in \Cref{sec:recycling}.
For strategies leveraging auxiliary trainings, given a target dataset, we consider the other DomainBed's datasets as the auxiliary tasks.
For example when tackling OfficeHome, out of the $20$ runs, $4$ are inter-trained on PACS, $4$ on VLCS, $4$ on TerraIncognita, $4$ on DomainNet and $4$ are directly transferred from ImageNet.
Then, \emph{model ratatouille is to inter-training as model soups is to vanilla fine-tuning}. In other words, while inter-training selects the best run based on \iid accuracy, ratatouille applies the uniform or the greedy selection. Thus ratatouille provides a single weight averaged network without any inference overhead.
For real-world applications, auxiliary weights may be shared by the community; in that case, ratatouille is without training overhead, except when marked by the \enquote{$\dagger$}. Indeed, \enquote{$\dagger$} symbol marks methods averaging $60=20\times3$ weights from $3$ data splits, and thus benefiting from larger training budget.
We further discuss ratatouille's training cost in \Cref{app:trainingcost}, and show in \Cref{app:numberruns} that ratatouile already performs well with only $5$ runs.
Ensembling strategies (marked by the symbol \enquote{$*$}) average predictions with large inference overhead. For example, \enquote{ensemble$^{*}$ of inter-training} averages the predictions of the $M=20$ models ratatouille averages in weights; we also report the scores from \citet{arpit2021ensemble} for the deep ensembles$^{*}$ \cite{Lakshminarayanan2017} of $M=6$ models with different classifier initializations.
For fusing, each run is initialized from $\sum_{i=0}^{4} \lambda_i \phi_{i}^{\mathrm{aux}}$ where the $\lambda_i$ hyperparameters sum to $1$ and $\phi_{i}^{\mathrm{aux}}$ are inter-trained on one the $4$ other DomainBed's datasets or directly transferred from ImageNet.
Finally, CORAL \cite{coral216aaai} is the best invariance approach; SWAD \cite{cha2021wad} and MA \cite{arpit2021ensemble} average weights along one training trajectory but differ in their selection strategy.
The experimental setup and the approaches are further described in \Cref{app:domainbed}.

\begin{figure*}[h!]
    \begin{center}
        \begin{subfigure}[b]{0.24\textwidth}
            \includegraphics[width=\textwidth]{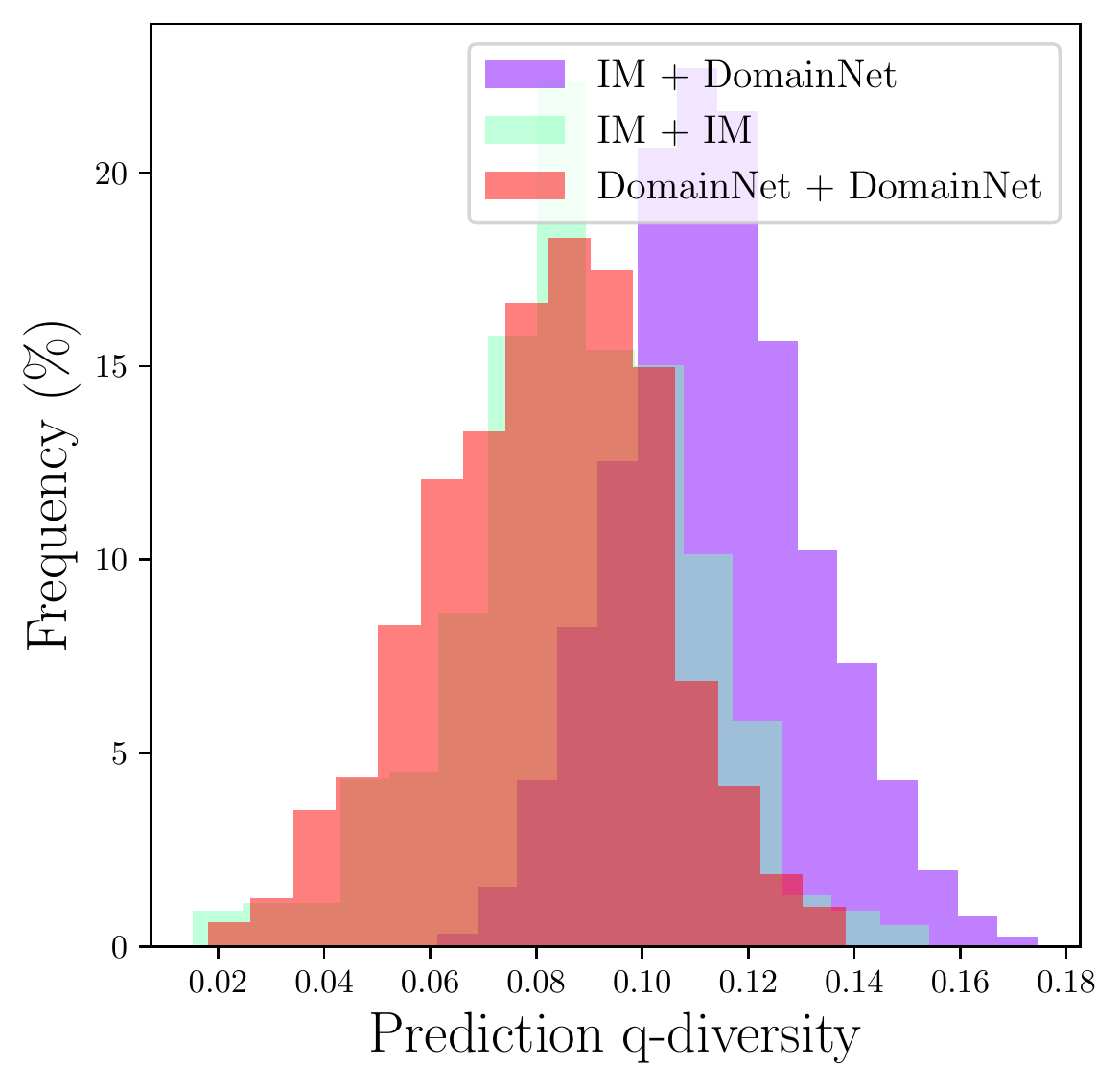}
            \caption{Diversity frequency.}
            \label{fig:diversity:a}
        \end{subfigure}
        \hfill
        \begin{subfigure}[b]{0.24\textwidth}
            \includegraphics[width=\textwidth]{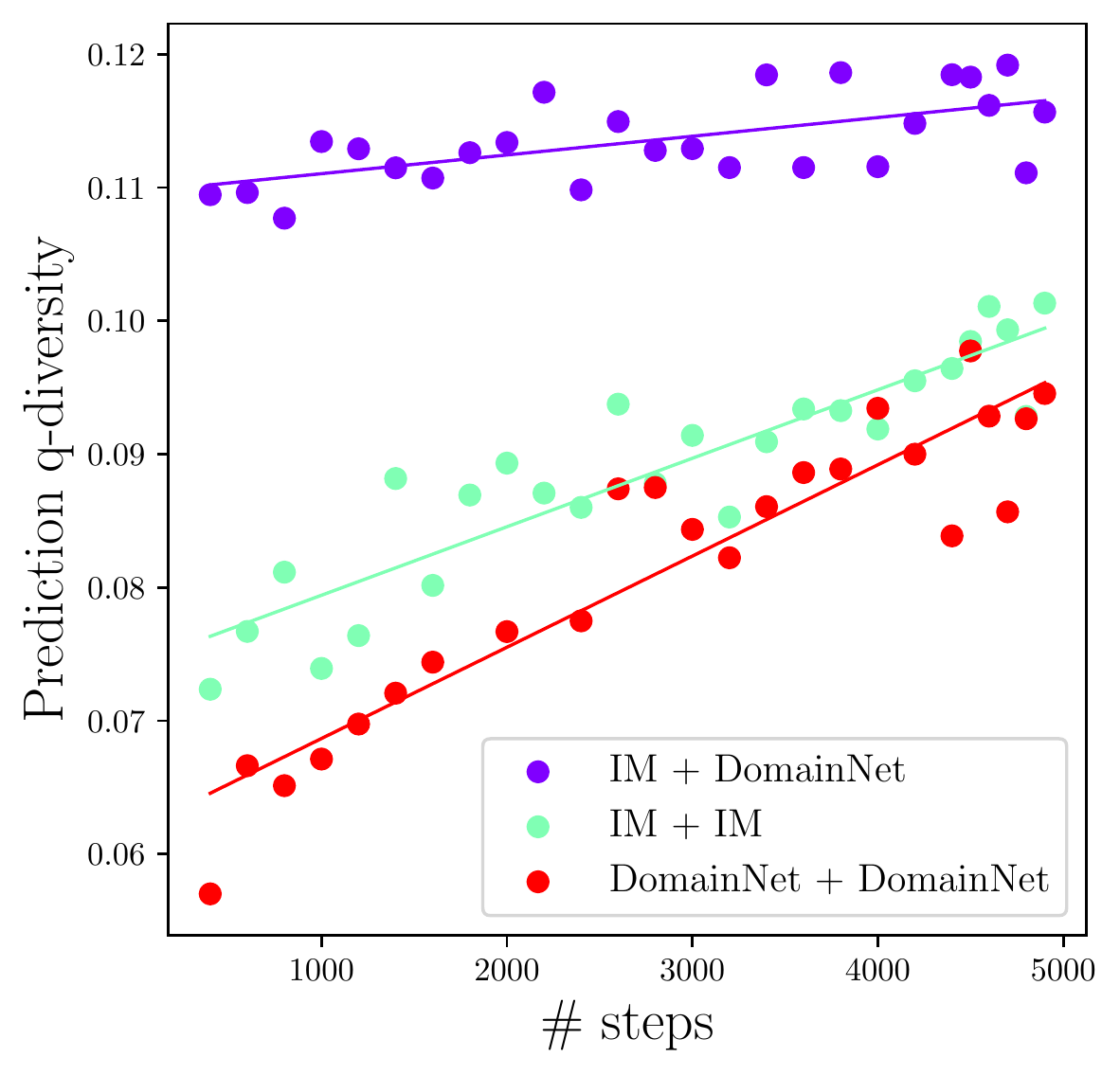}
            \caption{Diversity vs. ft steps.}
            \label{fig:diversity:b}
        \end{subfigure}
        \hfill
        \begin{subfigure}[b]{0.24\textwidth}
            \includegraphics[width=\textwidth]{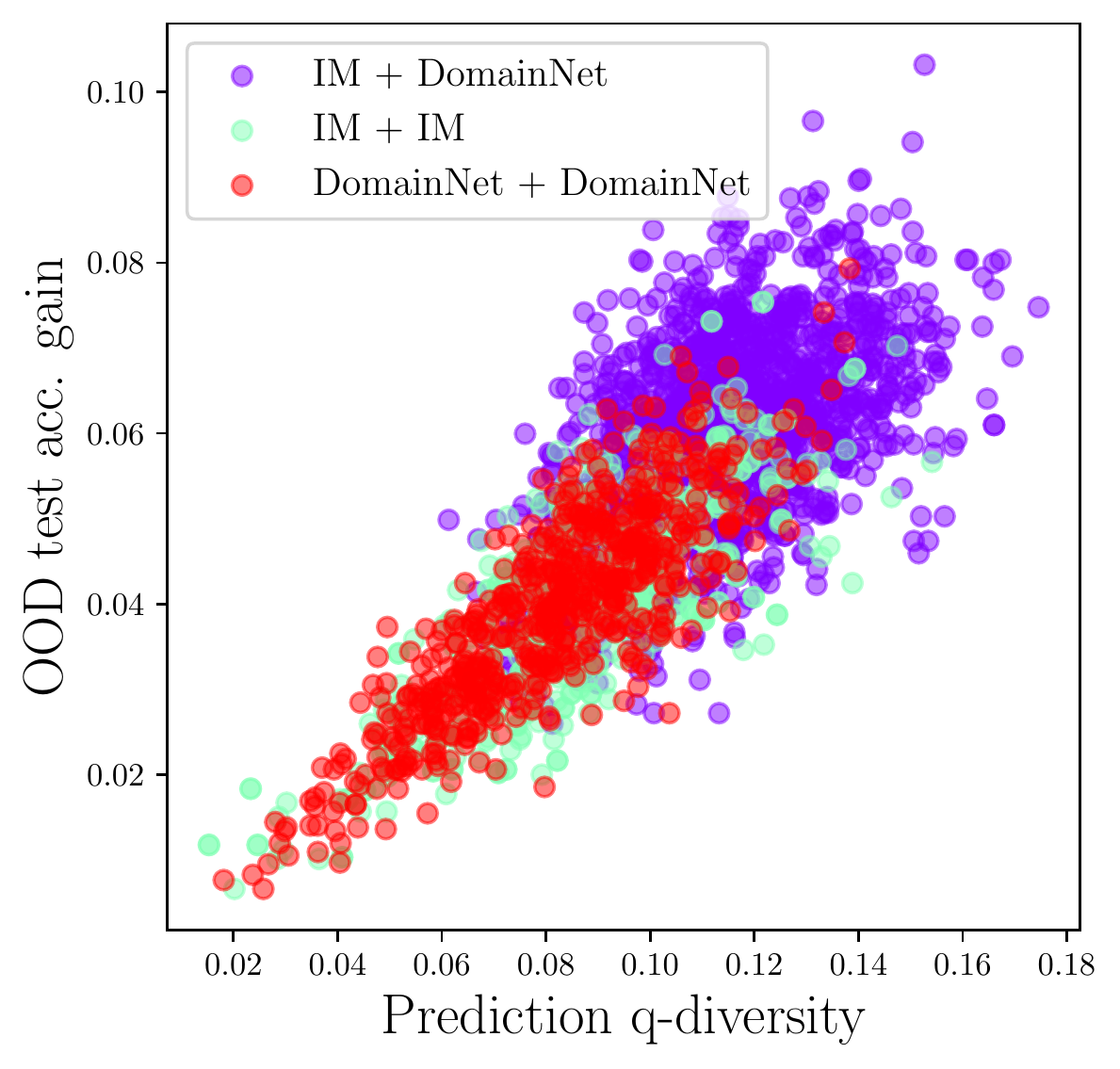}
            \caption{Acc. gain vs. diversity.}
            \label{fig:diversity:c}
        \end{subfigure}
        \hfill
        \begin{subfigure}[b]{0.24\textwidth}
            \includegraphics[width=\textwidth]{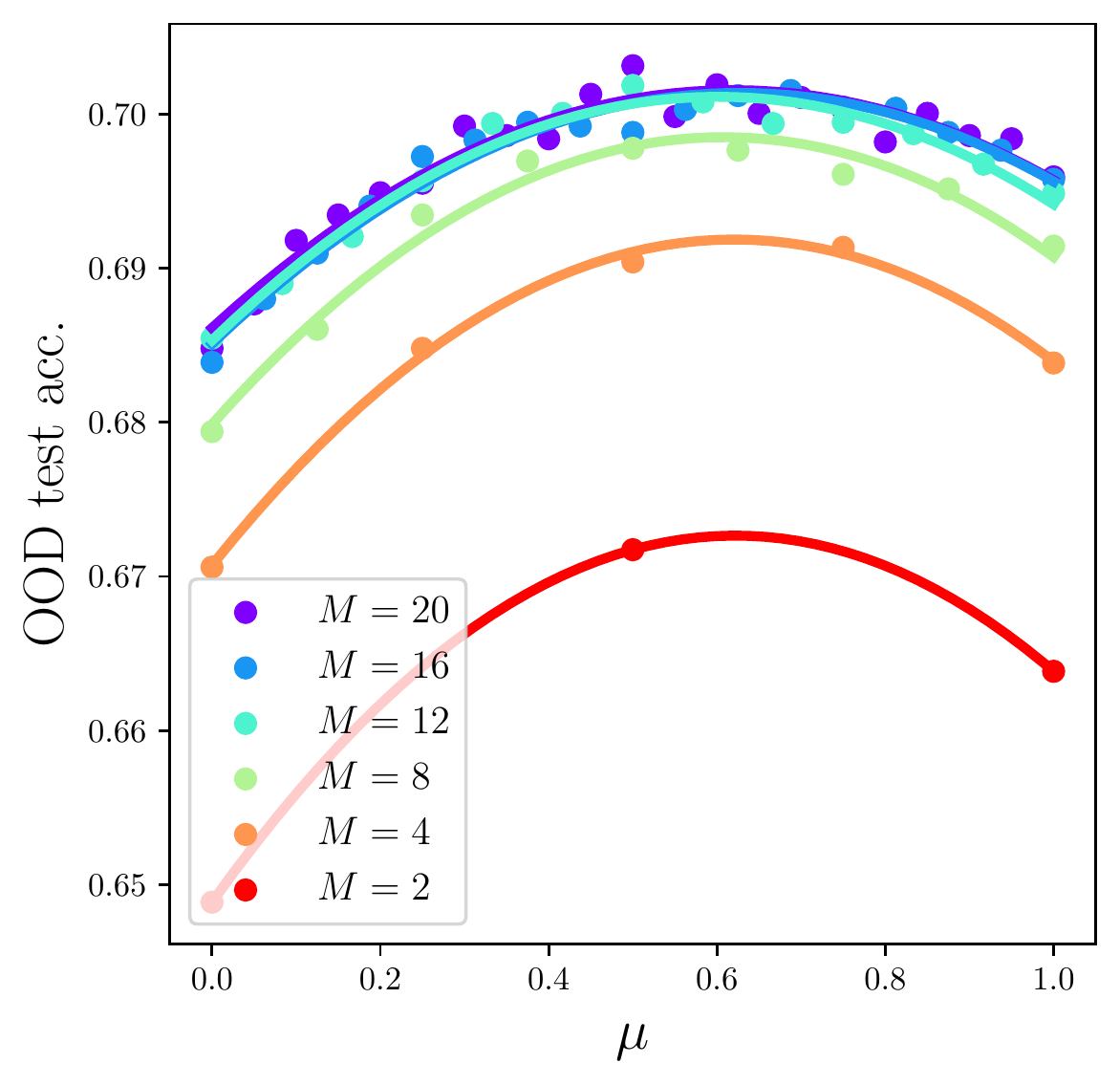}
            \caption{Acc. vs. mixing ratio.}
            \label{fig:diversity:d}
        \end{subfigure}
    \end{center}
    \caption{Explorations on q-diversity \cite{kuncheva2003measures} and its positive impact on accuracy for the \ood test domain \enquote{Art} from OfficeHome.
        In (a), we compute the diversity between pairs of models either directly fine-tuned from ImageNet, either inter-trained on DomainNet: having one model from each initialization increases diversity.
        In (b), we plot this diversity along the 5k training steps.
        In (c), we observe that the more diverse the models, the higher the accuracy gain of their weight average compared to the average of their individual accuracies.
        In (d), we average $M$ models: a proportion $(1 - \mu)$ start directly from ImageNet, the others $\mu$ are inter-trained on DomainNet. The accuracy of the weight average is maximized when $\mu\approx 0.5$.}
    \label{fig:diversity}
\end{figure*}

\textbf{Results.}
\Cref{table:domainbed} shows that ratatouille achieves a new \sota on DomainBed: with uniform selection, it achieves $68.1$ and improves model soups by $0.5$ points after averaging over all datasets.
Precisely, model ratatouille beats model soups by $0.8$ and $1.0$ points on PACS and OfficeHome with uniform selection, and by $2.5$ and $1.9$ with greedy selection.
On these two datasets, inter-training and fusing also succeed, yet they fail on TerraIncognita (both reach $46.7\%$) as all auxiliary tasks are distant from photos of animals in the wild; in contrast  on TerraIncognita, ratatouille ($51.8\%$) with uniform selection matches model soups ($51.4\%$).
This highlights the key strength of our ratatouille \wrt other recycling strategies such as fusing: namely, the robustness to the choice of auxiliary tasks.
On VLCS, ratatouille is also generally beneficial (as visible in the per-domain results from \Cref{app:domainbedperdataset}), except on one domain where the LMC breaks (as shown in \Cref{fig:vlcs1_lmc_hyp2_ood} from \Cref{app:lmc}).
For DomainNet, ratatouille is \sota though the gains are small \wrt model soups: we suspect this is because the initialization strategy becomes less critical for larger datasets \cite{chang2021rethinking} with more training epochs (see \Cref{fig:diversity:b}).
In conclusion, ratatouille consistently improves generalization on DomainBed, and works best with appropriate auxiliary tasks: we remove the need to select only the \textit{best} initialization. This is similar to model soups, that works best with appropriate hyperparameter ranges; they remove the need to select only the best set of hyperparameters.

\FloatBarrier
\begin{figure*}[t!]
    \begin{center}
        \begin{subfigure}{.19\textwidth}
            \centering
            \includegraphics[width=.95\linewidth]{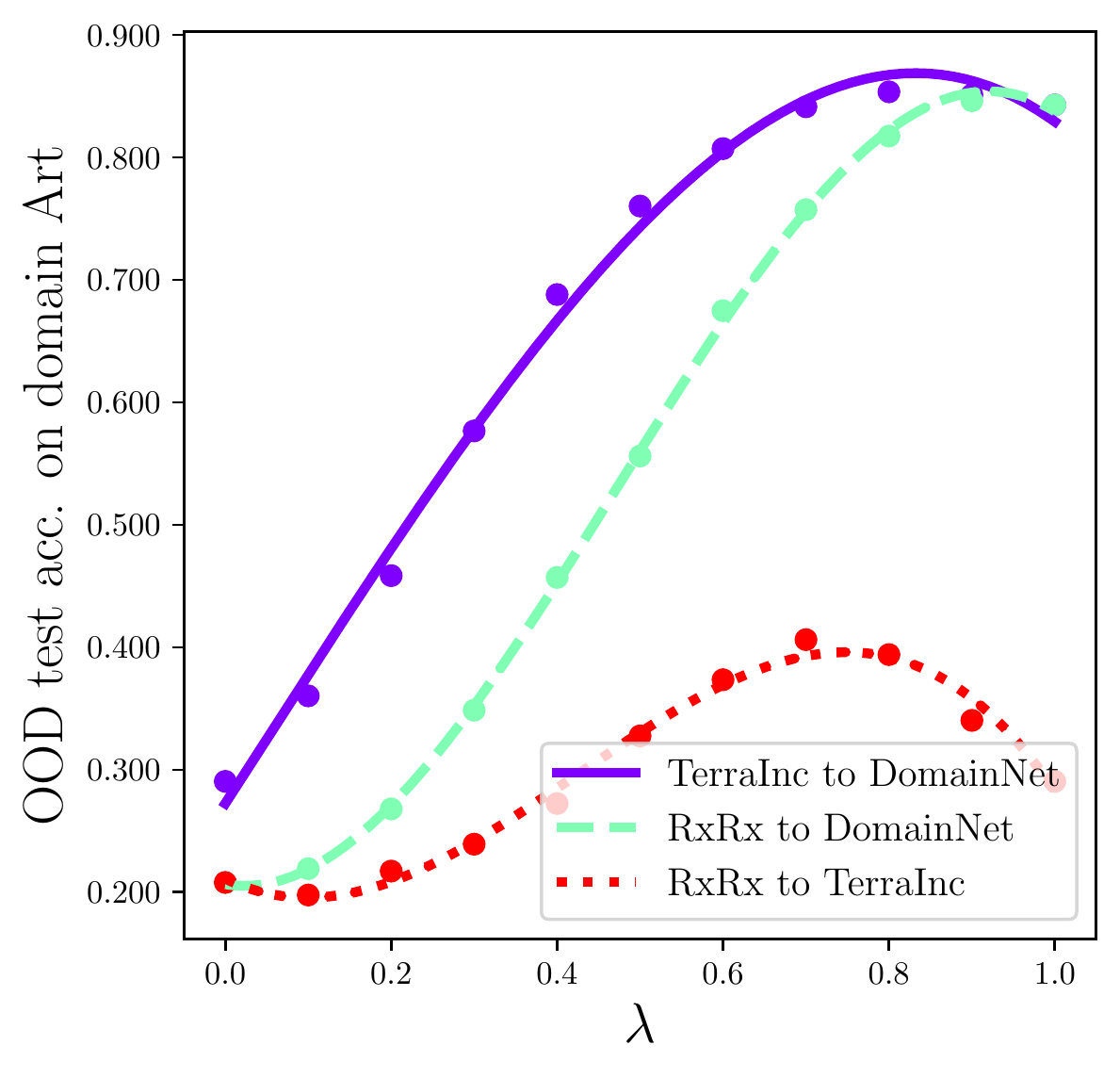}
            \caption{PACS.}
            \label{fig:pacs0_lmc_hyp1_ood}
        \end{subfigure}
        \hfill
        \begin{subfigure}{.19\textwidth}
            \centering
            \includegraphics[width=.95\linewidth]{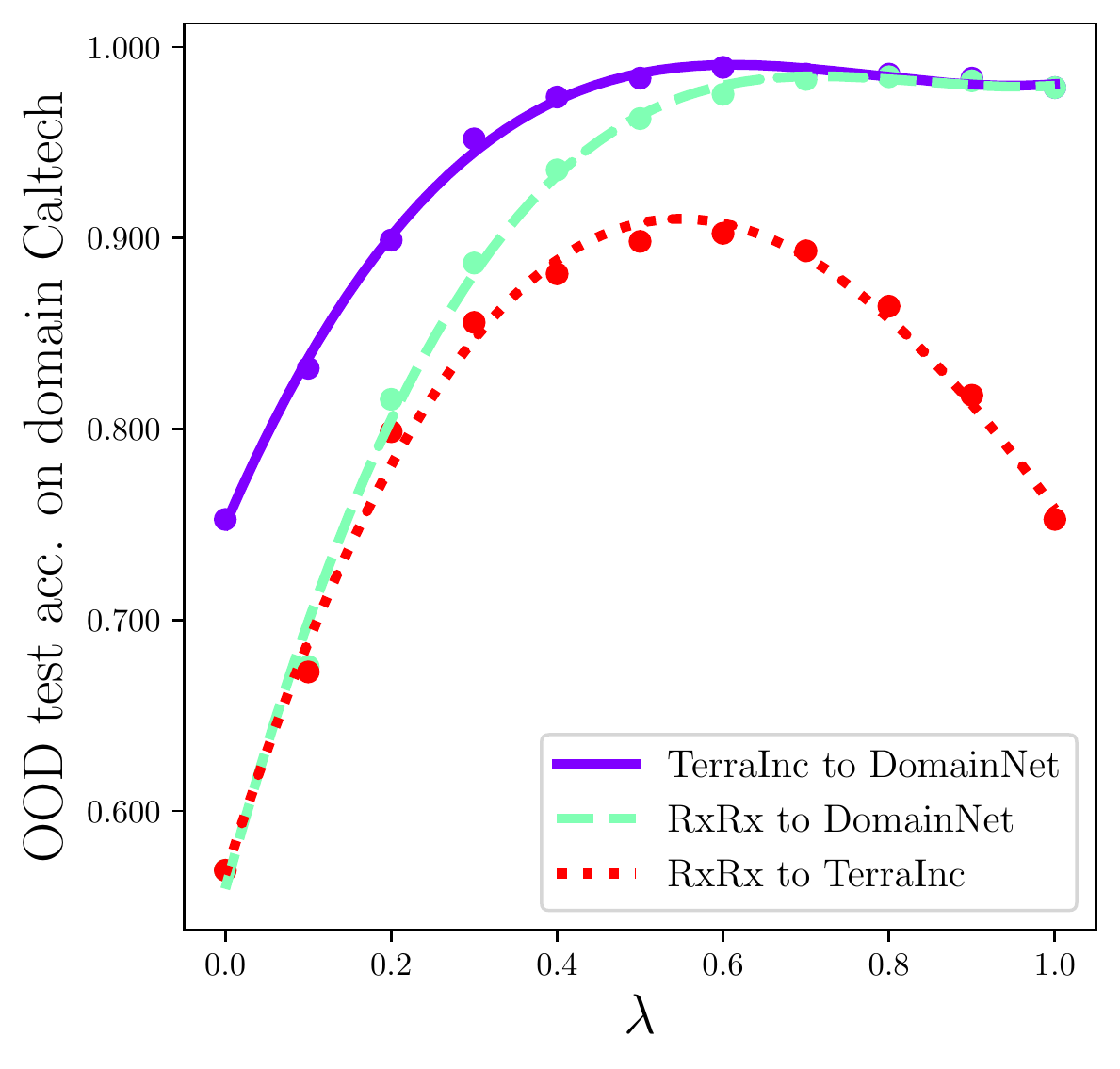}
            \caption{VLCS.}
            \label{fig:vlcs0_lmc_hyp1_ood}
        \end{subfigure}
        \hfill
        \begin{subfigure}{.19\textwidth}
            \centering
            \includegraphics[width=.95\linewidth]{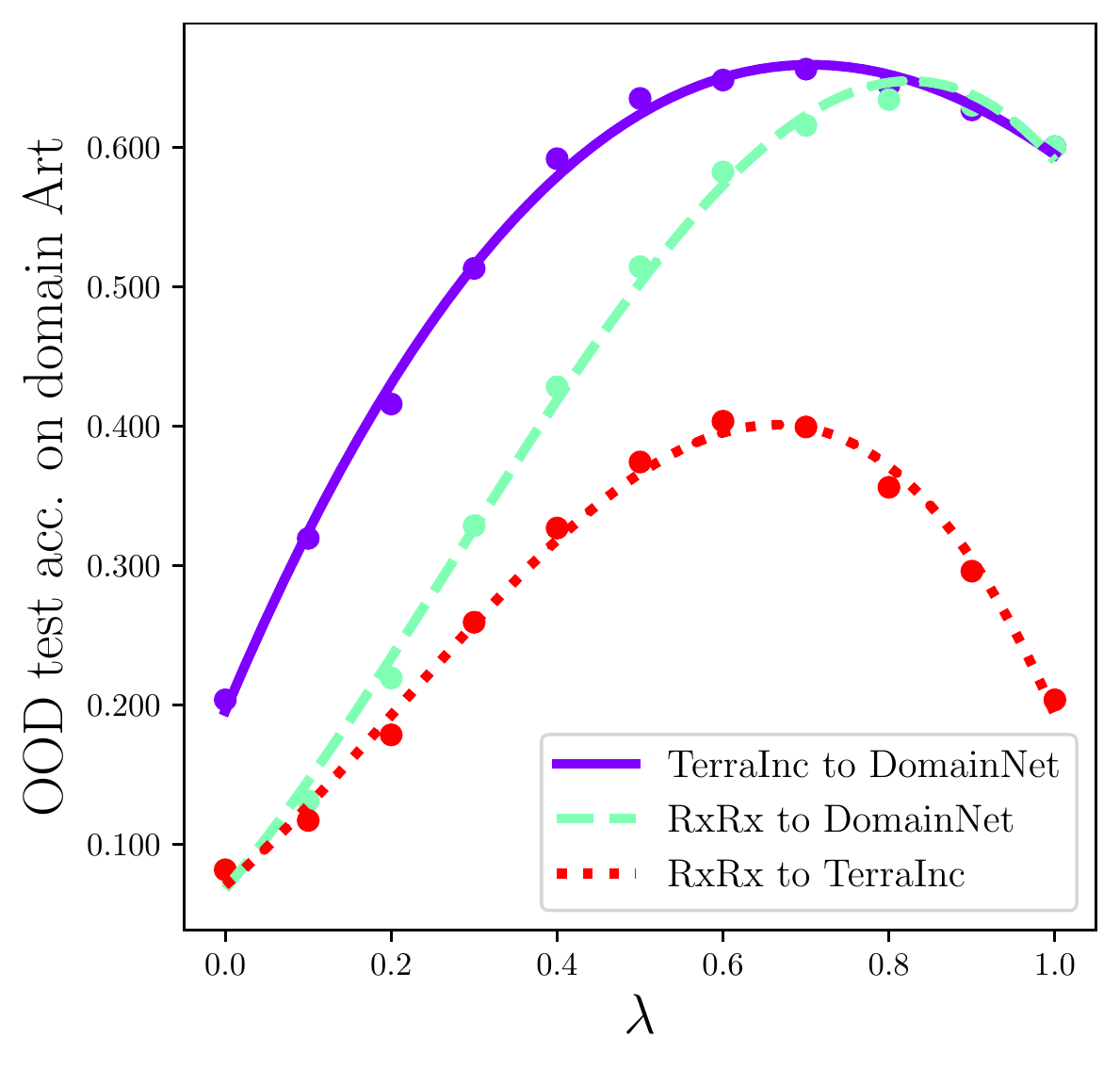}
            \caption{OfficeHome.}
            \label{fig:home0_lmc_hyp1_ood}
        \end{subfigure}
        \hfill
        \begin{subfigure}{.19\textwidth}
            \centering
            \includegraphics[width=.95\linewidth]{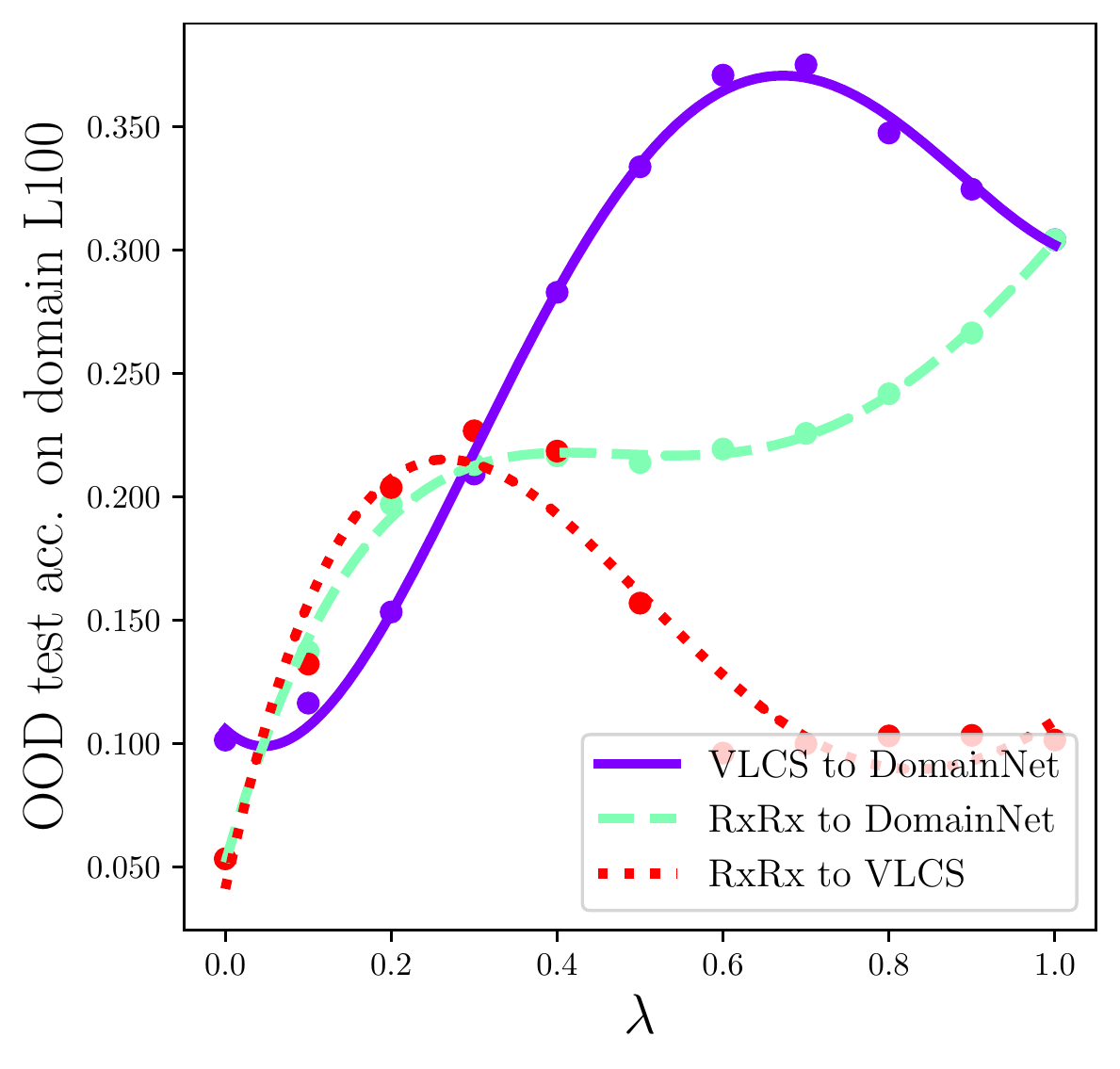}
            \caption{TerraInc.}
            \label{fig:terra0_lmc_hyp1_ood}
        \end{subfigure}
        \hfill
        \begin{subfigure}{.19\textwidth}
            \centering
            \includegraphics[width=.95\linewidth]{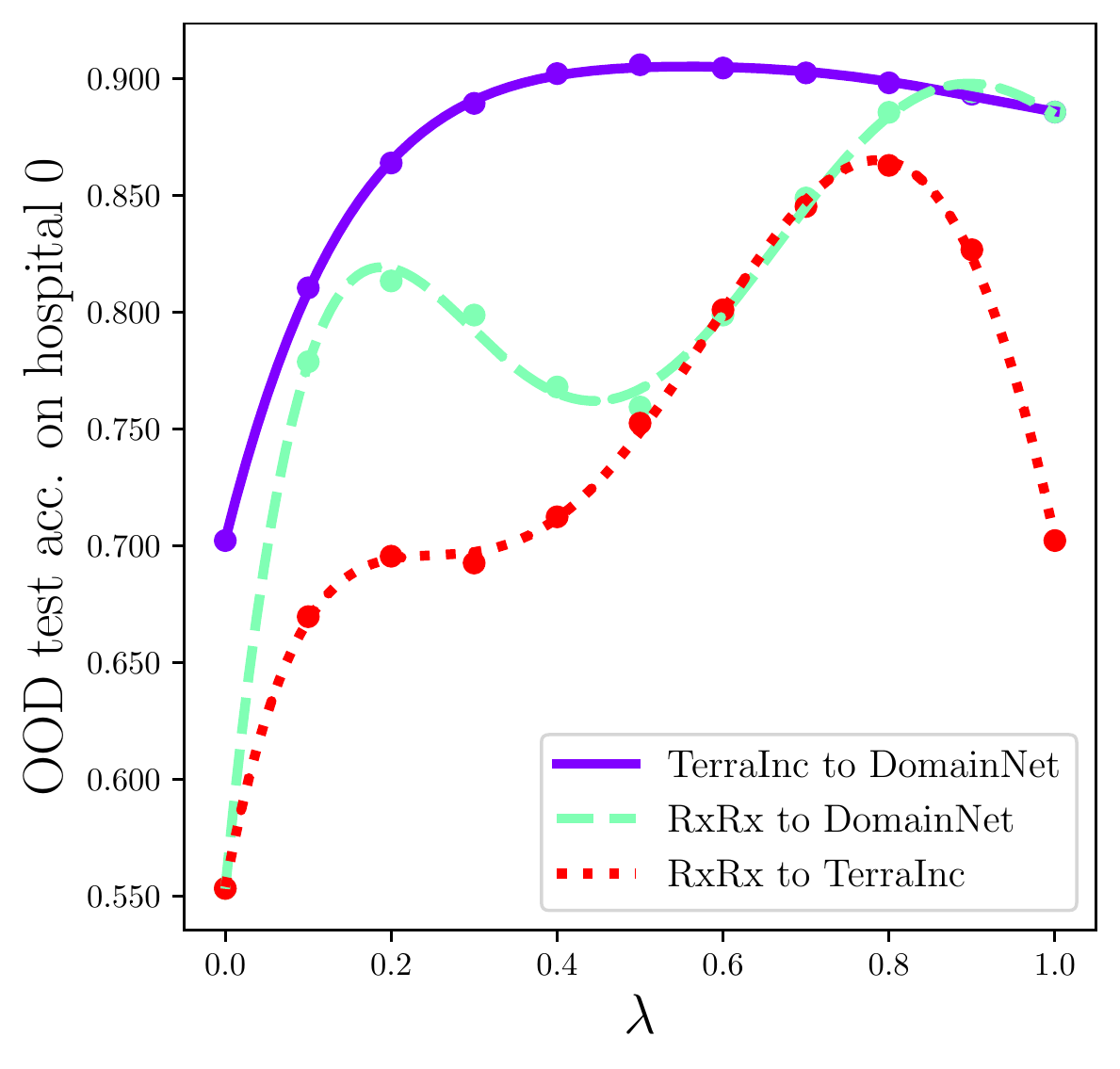}
            \caption{Camelyon.}
            \label{fig:came0_lmc_hyp1_ood}
        \end{subfigure}
    \end{center}
    \begin{center}
        \begin{subfigure}{.19\textwidth}
            \centering
            \includegraphics[width=.95\linewidth]{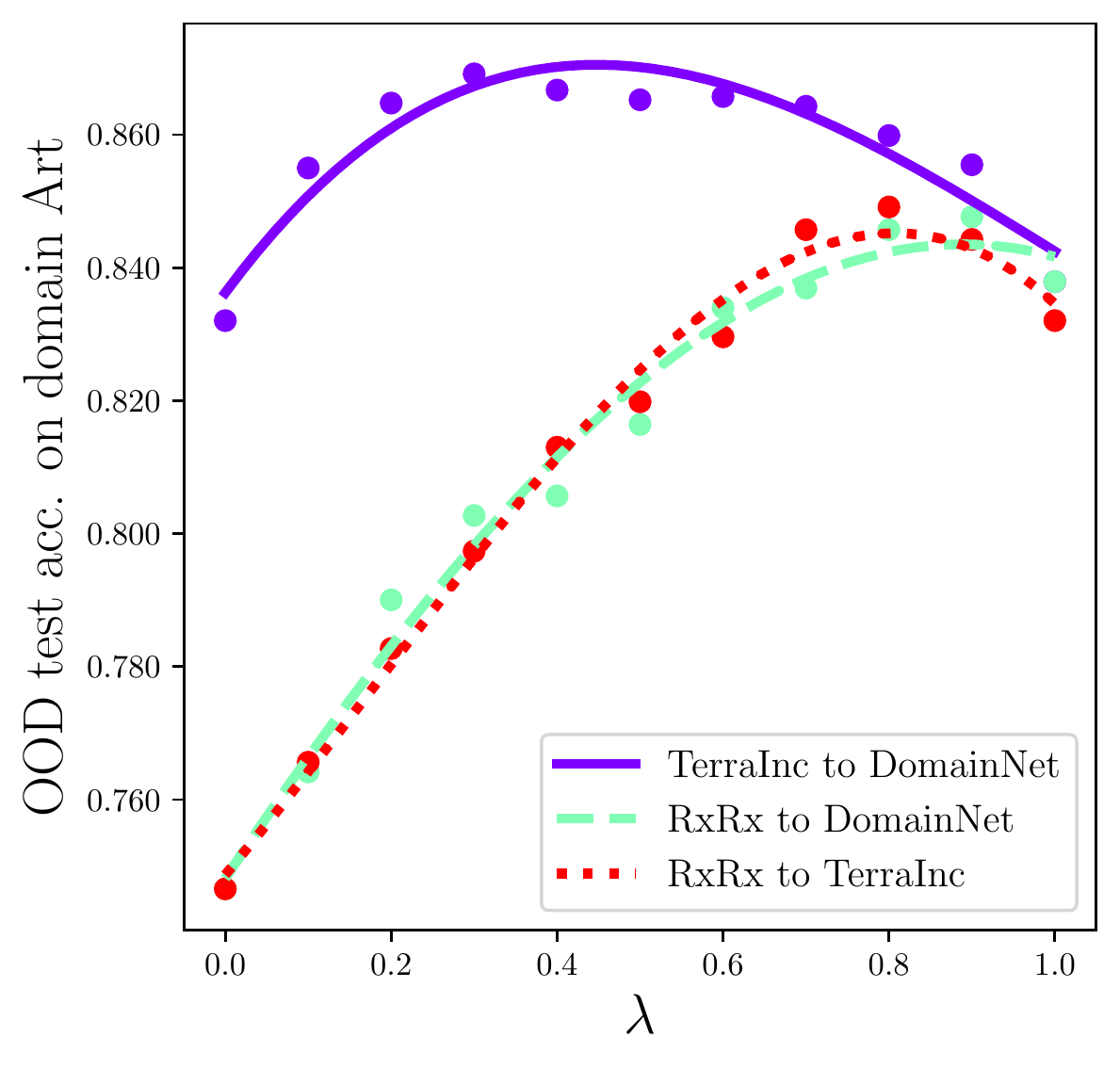}
            \caption{PACS.}
            \label{fig:pacs0_lmc_hyp2_ood}
        \end{subfigure}
        \hfill
        \begin{subfigure}{.19\textwidth}
            \centering
            \includegraphics[width=.95\linewidth]{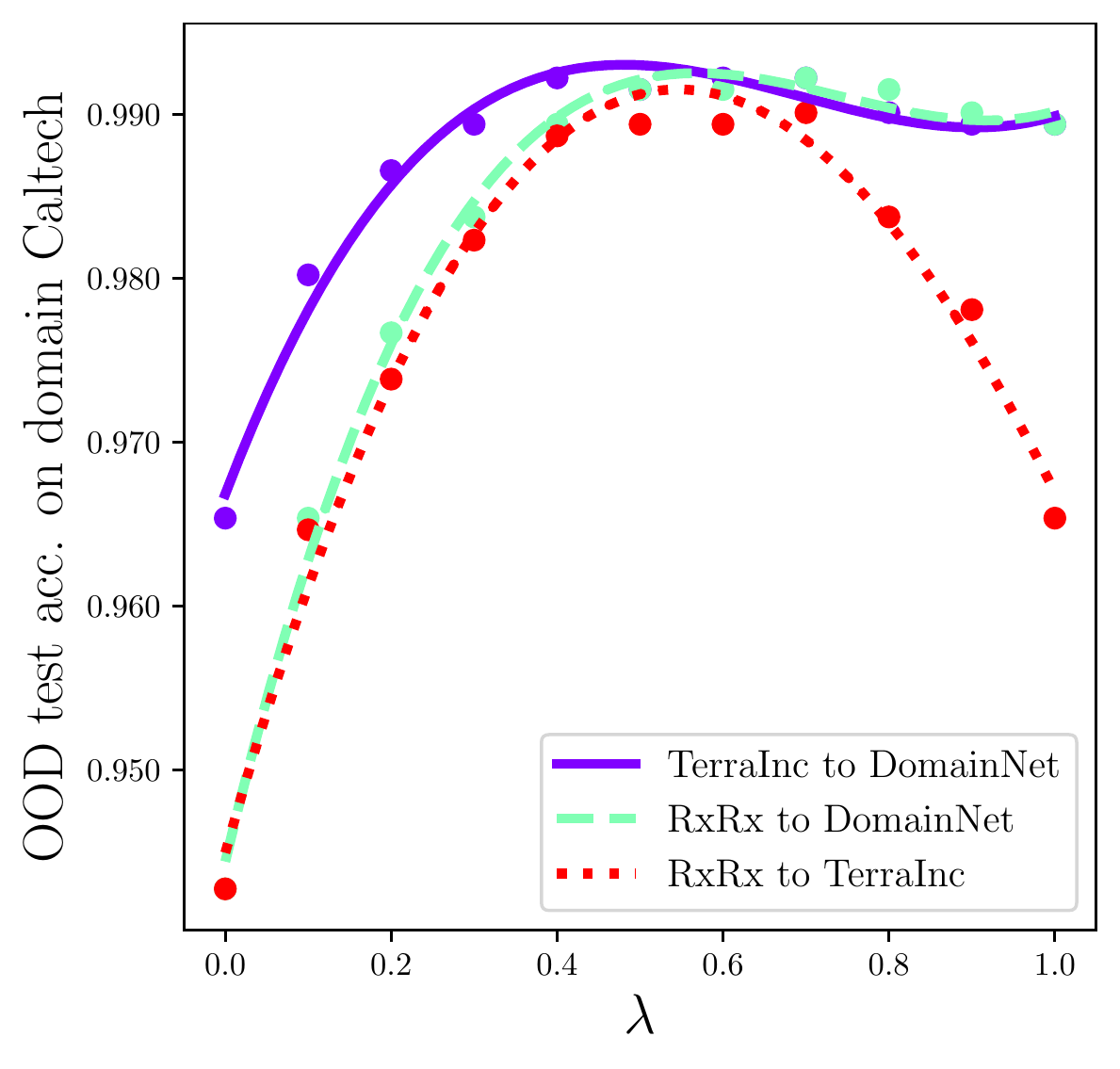}
            \caption{VLCS.}
            \label{fig:vlcs0_lmc_hyp2_ood}
        \end{subfigure}
        \hfill
        \begin{subfigure}{.19\textwidth}
            \centering
            \includegraphics[width=.95\linewidth]{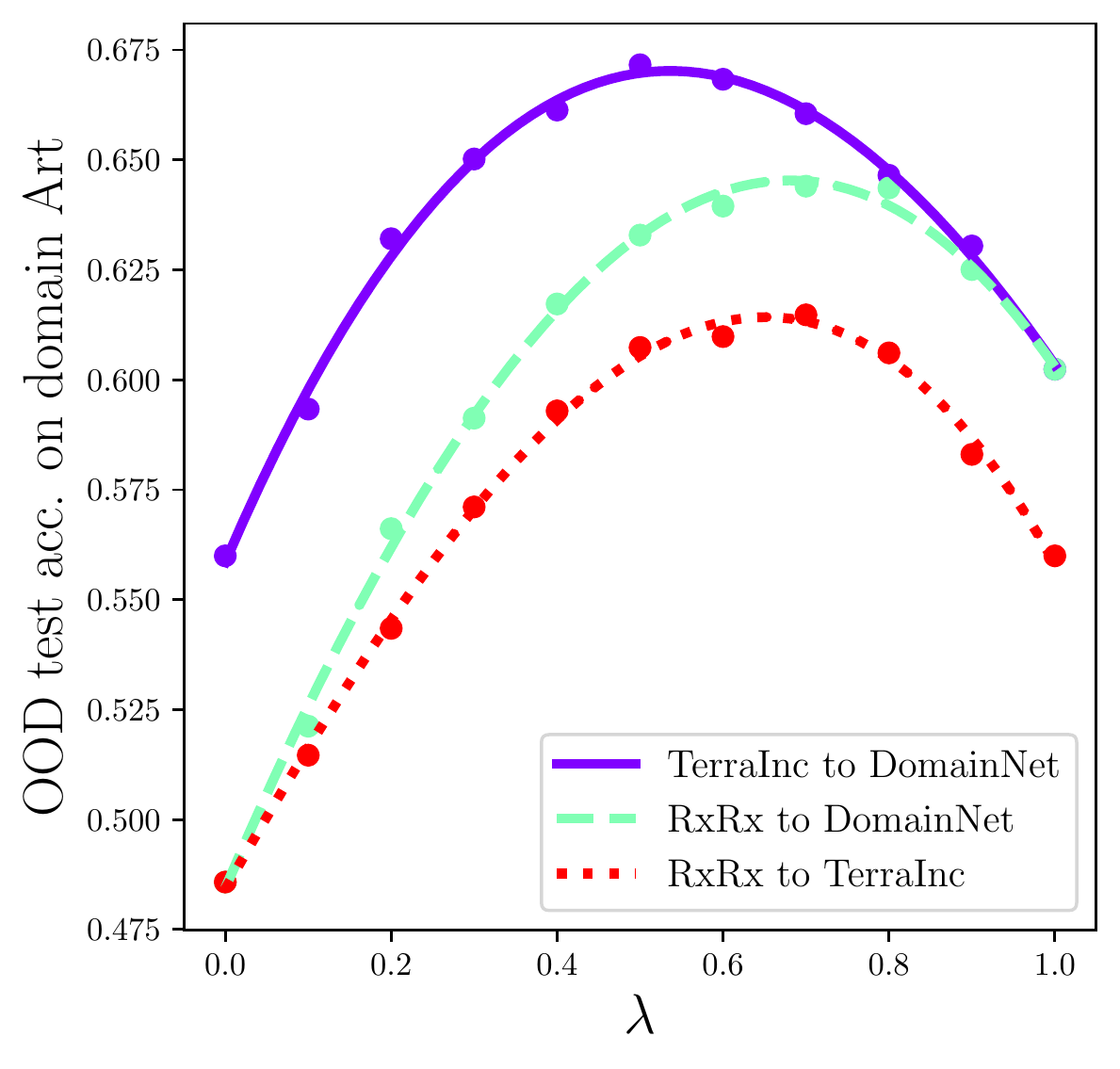}
            \caption{OfficeHome.}
            \label{fig:home0_lmc_hyp2_ood}
        \end{subfigure}
        \hfill
        \begin{subfigure}{.19\textwidth}
            \centering
            \includegraphics[width=.95\linewidth]{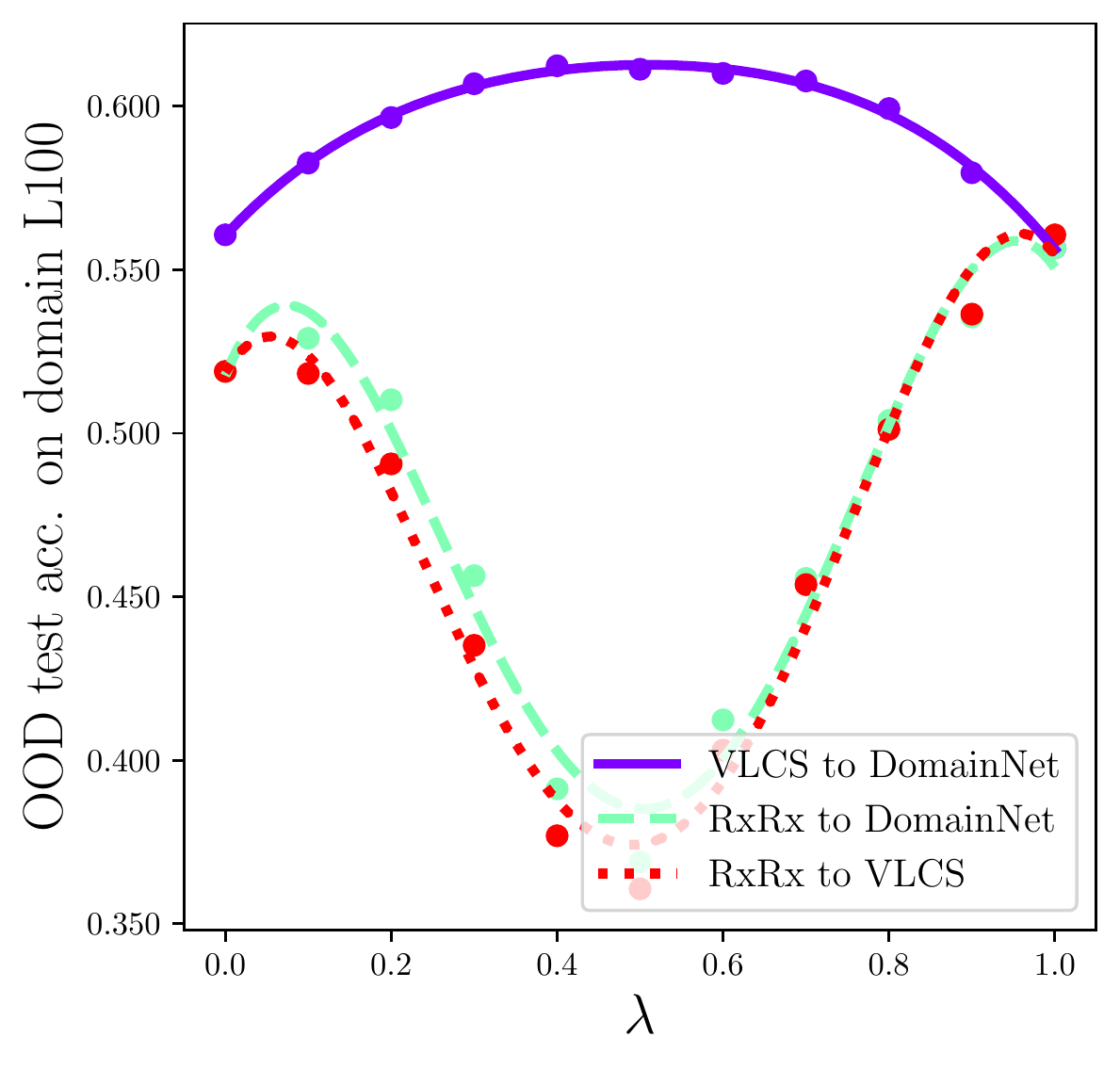}
            \caption{TerraInc.}
            \label{fig:terra0_lmc_hyp2_ood}
        \end{subfigure}
        \hfill
        \begin{subfigure}{.19\textwidth}
            \centering
            \includegraphics[width=.95\linewidth]{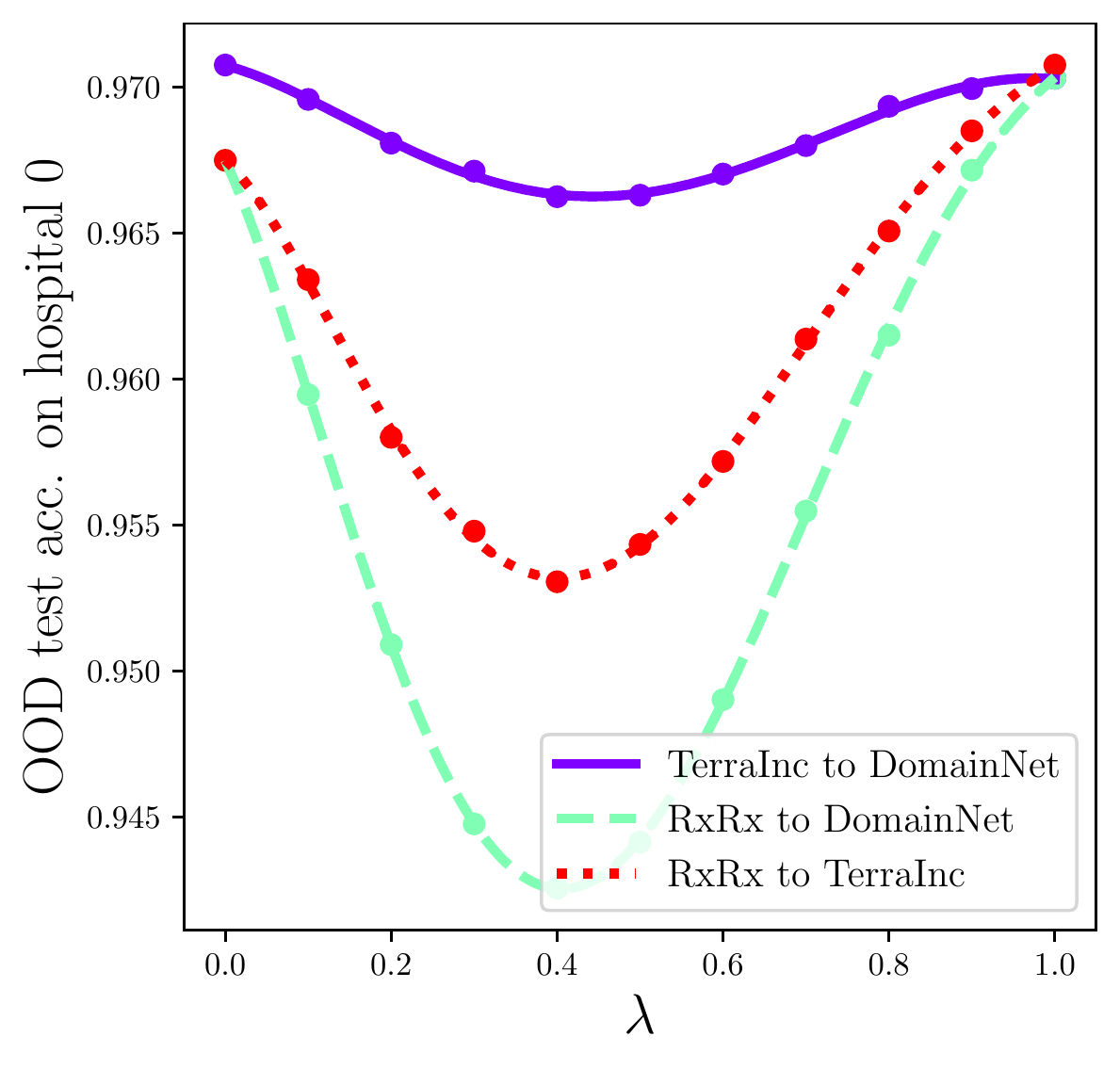}
            \caption{Camelyon.}
            \label{fig:came0_lmc_hyp2_ood}
        \end{subfigure}
    \end{center}
    \caption{\Cref{fig:pacs0_lmc_hyp1_ood,fig:vlcs0_lmc_hyp1_ood,fig:home0_lmc_hyp1_ood,fig:terra0_lmc_hyp1_ood,fig:came0_lmc_hyp1_ood} validate \Cref{hyp:1} by plotting $\lambda \to \mathrm{acc}_\mathrm{te}\left(\left(w^\mathrm{lp}, (1 - \lambda) \cdot \phi_a^{\mathrm{aux}} + \lambda \cdot \phi_b^{\mathrm{aux}}\right)\right)$, where $w^\mathrm{lp}$ is the linear probe of $\phi_{\mathrm{IM}}^{\mathrm{pt}}$, and $\phi_a^{\mathrm{aux}}$ and $\phi_b^{\mathrm{aux}}$ are fine-tuned on the two auxiliary datasets in the legend \enquote{Dataset$_a$ to Dataset$_b$}.
        \Cref{fig:pacs0_lmc_hyp2_ood,fig:vlcs0_lmc_hyp2_ood,fig:home0_lmc_hyp2_ood,fig:terra0_lmc_hyp2_ood,fig:came0_lmc_hyp2_ood} support \Cref{hyp:2} by plotting $\lambda \to \mathrm{acc}_\mathrm{te}\left((1 - \lambda) \cdot \theta_a + \lambda \cdot \theta_b\right)$ where $\theta_a$ and $\theta_b$ are fine-tuned on the target task starting respectively from $(w^\mathrm{lp}, \phi_a^{\mathrm{aux}})$ and $(w^\mathrm{lp}, \phi_b^{\mathrm{aux}})$.
        We encounter two exceptions to \Cref{hyp:2} (\Cref{fig:terra0_lmc_hyp2_ood,fig:came0_lmc_hyp2_ood}), due to the fact that \emph{neither} the auxiliary \emph{nor} the target task bear enough similarity with the pre-training task.}%
    \label{fig:lmcvalid}%
    \begin{center}
        \begin{subfigure}[b]{.23\textwidth}
            \includegraphics[width=1.0\textwidth]{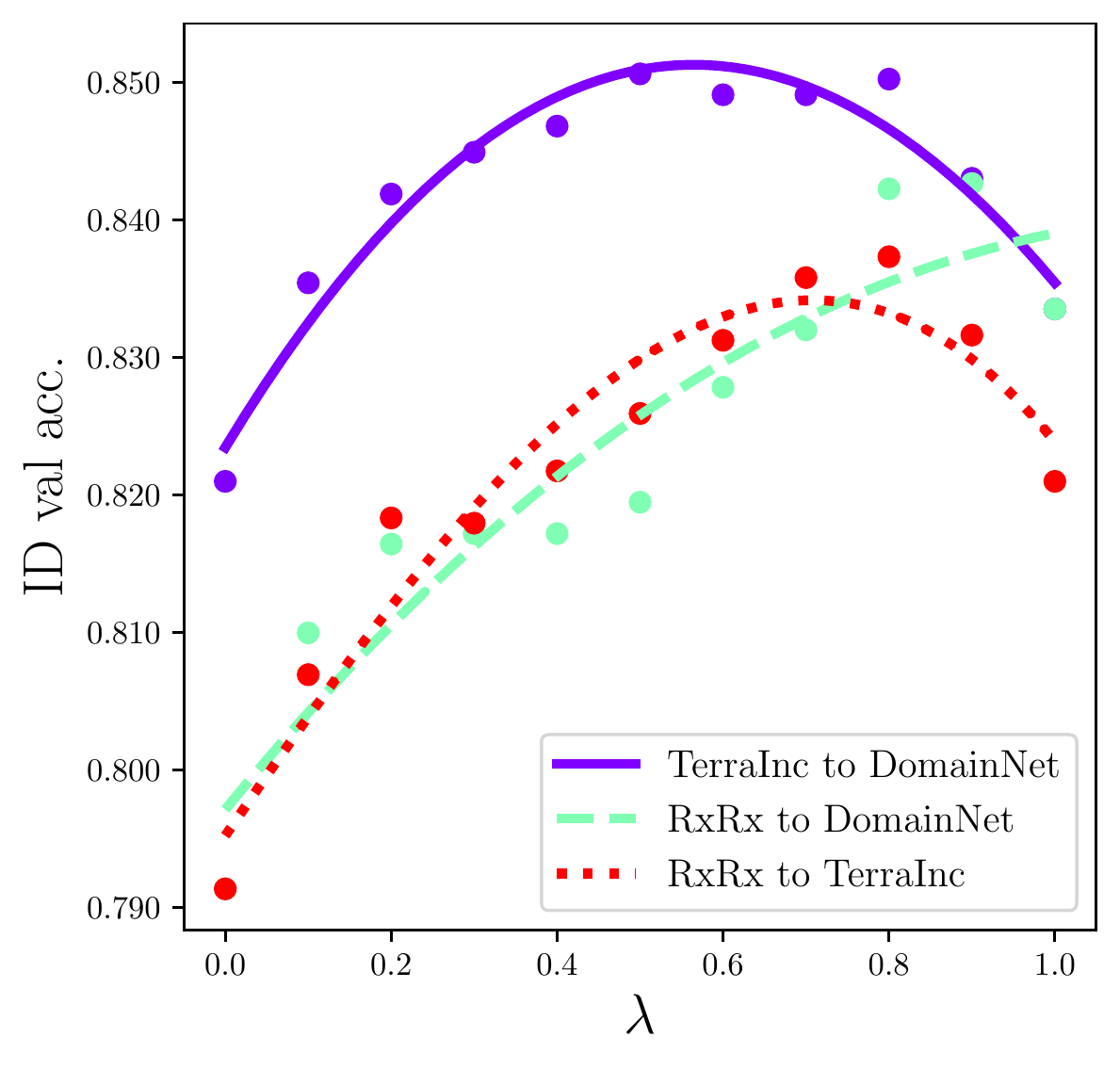}
            \caption{LMC in \iid.}
            \label{fig:home0_lmc_hyp2_iid}
        \end{subfigure}%
        \hfill
        \begin{subfigure}[b]{0.23\textwidth}
            \includegraphics[width=1.0\textwidth]{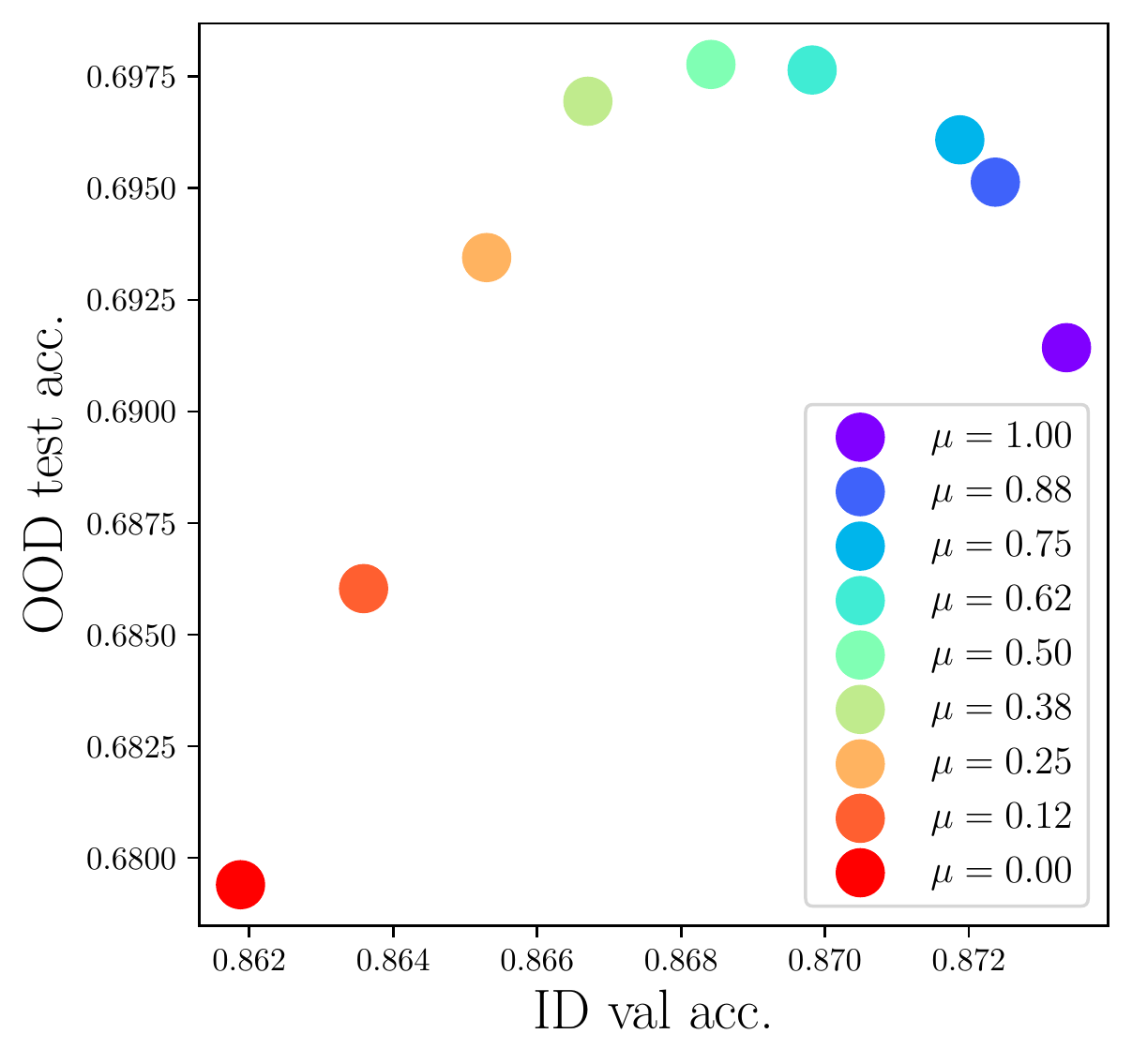}
            \caption{\ood vs. \iid acc.}
            \label{fig:diwa_dnim_iid_vs_ood_home0}
        \end{subfigure}
        \hfill
        \begin{subfigure}[b]{0.23\textwidth}
            \includegraphics[width=1.0\textwidth]{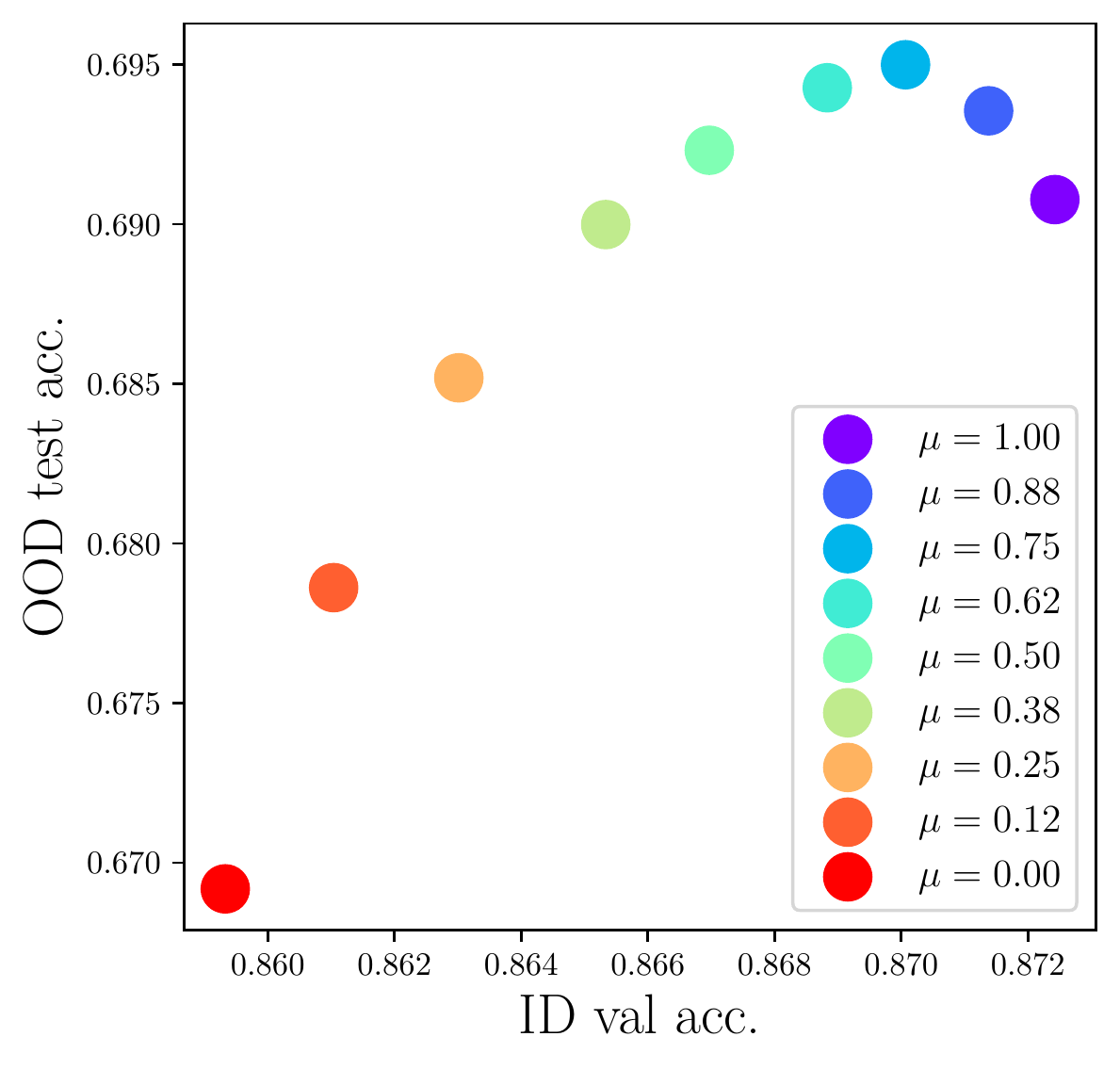}
            \caption{\ood vs. \iid acc.}
            \label{fig:diwa_dnpacs_iid_vs_ood_home0}
        \end{subfigure}
        \hfill
        \begin{subfigure}[b]{0.23\textwidth}
            \includegraphics[width=1.0\textwidth]{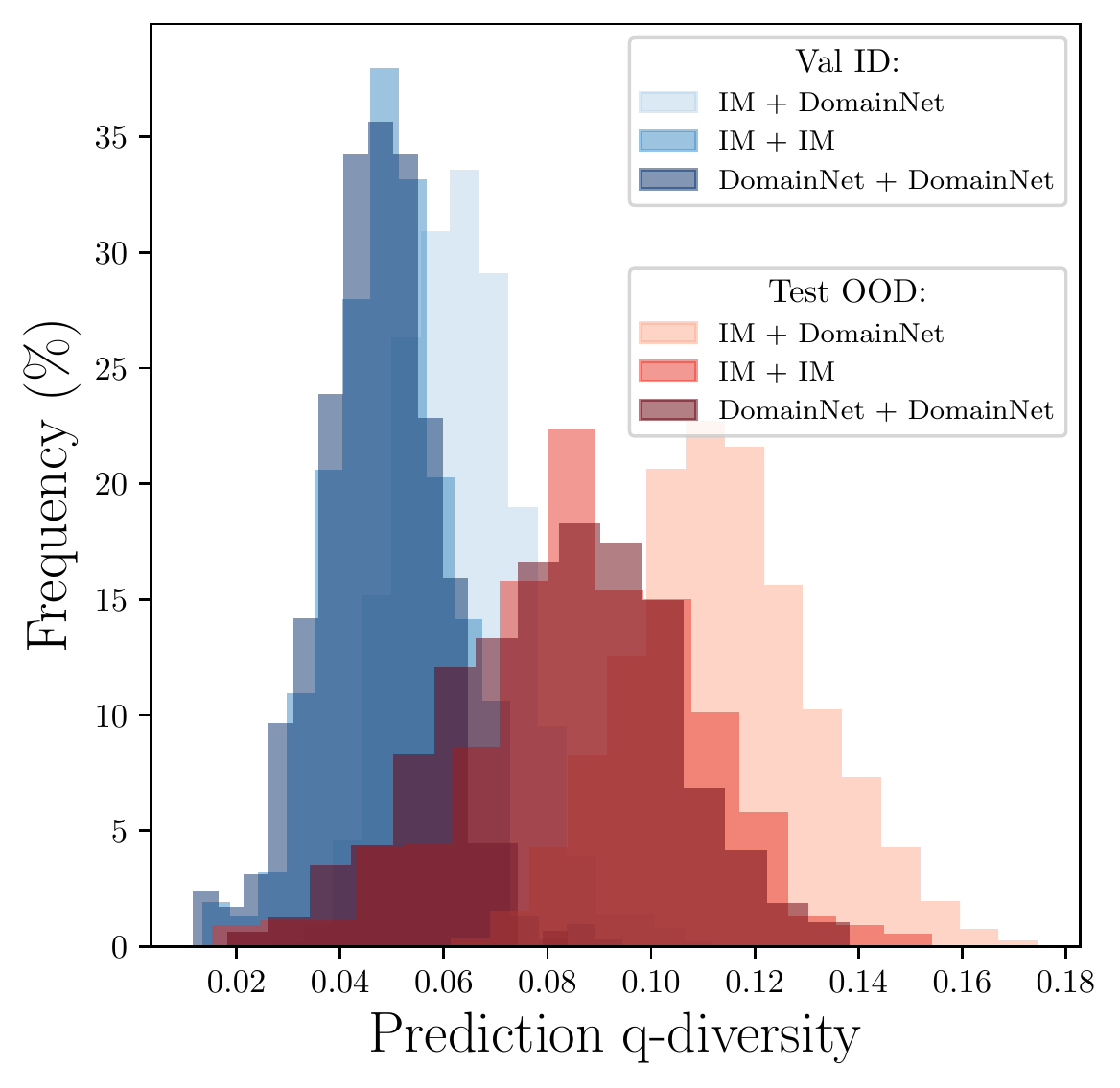}
            \caption{Diversity frequency.}
            \label{fig:dnim_hist_dq_iidood_home0}
        \end{subfigure}
    \end{center}
    \caption{The models were trained on \iid domains \enquote{Clipart}, \enquote{Product}, and \enquote{Photo} from OfficeHome, thus \enquote{Art} is the \ood domain.
        First, in subplot (a), we validate \Cref{hyp:2} on the \iid validation split.
        Then, we analyze the relations between diversity, \iid and \ood accuracies.
        In subplot (b), we report the mean results when averaging $M=8$ weights: $(1 - \mu)$ are fine-tuned on OfficeHome directly from ImageNet, the others $\mu$ are inter-trained on DomainNet. We observe a lack of correlation between \iid and \ood accuracies.
        We observe a similar trend in subplot (c), which mirrors the experiment from subplot (b) with the only difference that the proportion $(1 - \mu)$ are inter-trained on PACS (rather than just transferred from ImageNet).
        In subplot (d), we compute the diversity \cite{kuncheva2003measures} between models either directly fine-tuned from ImageNet, either inter-trained on DomainNet. Though having different initializations increases diversity both in \iid and in \ood, the diversity in \iid remains smaller.
    }%
    \label{fig:iidood}
\end{figure*}
\subsection{Increased Diversity by Recycling}
\label{sec:exps:div}%
In \Cref{fig:diversity}, we investigate how the diversity across models fine-tuned on the target task influences the \ood performance of their weight average.
Here, we measure diversity with the prediction q-diversity~\cite{kuncheva2003measures}, which increases when models fail on different examples; this diversity measure is precisely defined in \Cref{app:diversitymeasures}, where we also arrive at similar conclusions using another diversity measure \cite{aksela2003comparison}.
Following DiWA \cite{rame2022diwa}, let the target task be OfficeHome, with \enquote{Art} as the test \ood domain; we thus train on the \enquote{ClipArt}, \enquote{Product} and \enquote{Photo} domains.
We consider models either only pre-trained on ImageNet or also inter-trained on DomainNet.
These diversity experiments are applied on other DomainBed's datasets in \Cref{app:diversitymore}.

First, we verify that inter-training influences the diversity across fine-tuned models.
Specifically, \Cref{fig:diversity:a} confirms that networks with different initializations are more diverse than networks initialized similarly.
Then, \Cref{fig:diversity:b} verifies that this diversity gain comes from their initialization and remains along fine-tuning on the target task.
Moreover, \Cref{fig:diversity:c} shows that diversity is positively linearly correlated with \ood generalization:
specifically, we observe that having different initializations improves diversity and thus the accuracy of their weight average.
Finally, in \Cref{fig:diversity:d}, we consider averaging $M$ weights: a proportion $(1 - \mu)$ start directly from ImageNet, the others $\mu$ were inter-trained on DomainNet.
In the simplest case $M=2$, using one model from each initialization leads to maximum accuracy; best performances are obtained around $\mu\approx 0.5$, where the final weight average has access to diverse initializations.
In conclusion, each auxiliary task fosters the learning of diverse features \cite{li2021Universal,gontijolopes2022no}.
Model ratatouille increases diversity and improves performance by removing a key limitation of model soups approaches \cite{Wortsman2022ModelSA,rame2022diwa}; the need for all fine-tunings to start from a shared initialization.%
\subsection{Why Ratatouille Works}
\label{sec:exps:lmc}
In \Cref{fig:lmcvalid}, we conclude our experiments by validating \Cref{hyp:1,hyp:2} when considering the five datasets from DomainBed. For the sake of completeness, we also analyze some successes and failure cases in \enquote{extreme} conditions when considering two distant unrelated medical datasets; RxRx~\cite{taylor2019rxrx1} and Camelyon~\cite{pmlr-v139-koh21a} from the WILDS~\cite{pmlr-v139-koh21a} benchmark. For each target task, we consider the first domain as the test \ood; the other domains are used for training.%

We validate \Cref{hyp:1} in \Cref{fig:pacs0_lmc_hyp1_ood,fig:vlcs0_lmc_hyp1_ood,fig:home0_lmc_hyp1_ood,fig:terra0_lmc_hyp1_ood,fig:came0_lmc_hyp1_ood}.
For each dataset, we plot the test \ood accuracy for the weights
$\left(w^\mathrm{lp}, (1 - \lambda) \cdot \phi_a^{\mathrm{aux}} + \lambda \cdot \phi_b^{\mathrm{aux}}\right)$,
where the classifier $w^\mathrm{lp}$ is a linear probe of the ImageNet pre-trained featurizer $\phi_{\mathrm{IM}}^{\mathrm{pt}}$, and $\lambda \in [0, 1]$ interpolates between $\phi_a^{\mathrm{aux}}$ and $\phi_b^{\mathrm{aux}}$, obtained by fine-tuning on two auxiliary tasks initialized from $\phi_{\mathrm{IM}}^{\mathrm{pt}}$.
First, we observe that task similarity influences \ood generalization since the test accuracies in \Cref{fig:home0_lmc_hyp1_ood} agree with the fact that
OfficeHome is most similar to DomainNet, not as similar to TerraIncognita, and most dissimilar to the medical dataset RxRx.
Second, \emph{the accuracy of the interpolated weights is above the interpolated accuracy}: this validates \Cref{hyp:1}. The accuracy is even usually concave in $\lambda$.

Similarly, we empirically support \Cref{hyp:2} in \Cref{fig:pacs0_lmc_hyp2_ood,fig:vlcs0_lmc_hyp2_ood,fig:home0_lmc_hyp2_ood,fig:terra0_lmc_hyp2_ood,fig:came0_lmc_hyp2_ood}.
For each dataset, we plot the test \ood accuracy obtained with weights
$(1 - \lambda) \cdot \theta_a + \lambda \cdot \theta_b$,
where the coefficient $\lambda \in [0, 1]$ interpolates between $\theta_a$ and $\theta_b$, fine-tuned on the target task respectively starting from $(w^\mathrm{lp}, \phi_a^{\mathrm{aux}})$ and $(w^\mathrm{lp}, \phi_b^{\mathrm{aux}})$.
We observe that \Cref{hyp:2} usually holds: for example, even recycling RxRx can help for OfficeHome on \Cref{fig:home0_lmc_hyp2_ood}.
Yet, \Cref{hyp:2} breaks on TerraIncognita and Camelyon in \Cref{fig:terra0_lmc_hyp2_ood,fig:came0_lmc_hyp2_ood} when RxRx is one of the two auxiliary tasks.
In light of these results, we argue \emph{that \Cref{hyp:2} holds as long as either the auxiliary or the target task is sufficiently similar to the pre-training task}. We speculate this prevents feature distortion \cite{kumar2022finetuning} and escaping a shared loss valley.
Better understanding when LMC breaks is a promising research direction \cite{juneja2022linear,lubana2022mechanistic}; among other factors, we speculate that larger pre-training corpus (as in \citet{qin2022exploring}) or larger architectures (as in \citet{li2022branch}) may favor weight averaging strategies.
In \Cref{app:lmc}, we further analyze \Cref{hyp:1,hyp:2}, notably in a more complex setup where the intermediate tasks are successive fine-tunings on several auxiliary datasets.
\subsection{Ratatouille for \iid Tasks}
\label{sec:exps:iid}

Like previous weight averaging strategies \cite{izmailov2018,Wortsman2022ModelSA}, model ratatouille also works for \iid tasks; in particular, we verify in \Cref{fig:home0_lmc_hyp2_iid} and in \Cref{app:iid} that the LMC holds in distribution.
Yet, the gains are smaller in \iid than in \ood, as confirmed by the lack of correlation between \iid and \ood accuracies \cite{teney2022id} in \Cref{fig:diwa_dnim_iid_vs_ood_home0,fig:diwa_dnpacs_iid_vs_ood_home0}.
This is explained by the fact that variance reduction (caused by weight averaging) is less beneficial in \iid than in \ood.
Theoretically, this is because, variance is smaller without distribution shift, as explained in \citet{rame2022diwa}.
Empirically, this is consistent with models' diversity being smaller in \iid, as shown in \Cref{fig:dnim_hist_dq_iidood_home0}.
Overall, diversity procedures are less useful in \iid than in \ood.
Ratatouille performs well OOD thanks to the diversity brought by diverse inter-trainings; for ID, we may sacrifice diversity and select one single optimal initialization.
This finding contrasts with \citet{pmlr-v139-miller21b} and goes against the prescription in \citet{assayingoodwenzel2022} that, \enquote{to make the model more robust on \ood data, the main focus should be to improve the \iid classification error}.

In conclusion, when aiming at \ood with ensembling strategies, our experiments suggest that there exists a trade-off between diversity and \iid accuracy.
This is critical for end-users as \ood is arguably more relevant than \iid to ensure applicability in real-world applications, where train and test hardly ever follow the same distributions.
This also explains occasional failures of the greedy selection (notably for TerraIncognita in \Cref{table:domainbed}): based on the \iid validation accuracy, only a few runs are selected and averaged, causing smaller \ood accuracy than with the uniform selection.

\section{Discussion: Towards Updatable Machine Learning}%
\label{sec:discussion}
In the grand scheme of things, we see model recycling within the emerging \emph{updatable machine learning}~\cite{updatablemachinelearning} paradigm.
The goal is to develop machine learning systems that can be incrementally improved and recombined, allowing for the collaborative creation of increasingly sophisticated AI systems.
The core idea is to consider networks as pieces of software~\cite{software20} and mirror the open-source development of software engineering via version control.
Could it be possible that, someday, we could build decentralized open-source repositories, where we can clone, commit and merge neural networks towards an ever-improving AI system?

Recent works \cite{mergefisher21,li2022branch,choshen2022cold,choshen2022start,rame2022pretrain} and the proposed ratatouille give some primitives to learn neural networks in collaboration.
Here, (i) cloning is simply weights downloading, (ii) commits are fine-tunings performed by individual contributors on their specific tasks, and (iii) branch merging is replaced by weight averaging.
Advanced merging operations \cite{mergefisher21,li2022trainable,jin2022dataless} could help to better select the interpolating coefficients $\lambda_i$; neuron permutations strategies~\cite{entezari2022the,ainsworth2022gitrebasin,jordan2022repair} could remove the need for a shared pre-training, though (so far) these permutations have not improved models' accuracy.

In terms of privacy, such a federated learning setup \cite{li2019federated} where datasets can be kept private does indeed seem desirable.
In terms of computation and sustainability, minimal communication across servers enable embarrassingly simple parallelization \cite{li2022branch,wortsman2022fi} and could reduce costs and CO2 emissions when training on multiple servers.
This paradigm could also leverage the utilization of volunteer computing with single-GPU desktop machines, and complement approaches like Learning@home \cite{ryabinin2020towards} or Petals \cite{borzunov2022petals}.
Finally, the contributors may potentially be incentivized financially through a system similar to blockchain technology \cite{sotamoon2021}.

If collaboration is the way forward, how can we ensure the \emph{recyclability} of the shared models?
In software engineering, practices such as unit tests greatly reduce the failure modes of programs; how can we borrow these ideas to \emph{specify and test} neural networks?
To measure models' shortcomings, we may leverage datasets as \emph{test certificates}~\cite{lopez2022measuring}.
The community would monitor statistics on these datasets, \eg accuracy, forgetting, and robustness against spurious correlations.
Then, the reported scores could guide the choice of what models to clone, fine-tune, and merge.
However, bad actors could directly include these datasets in their training data; then, should these external datasets be watermarked \cite{li2021survey}, or otherwise kept secret by some certifying authority?

These questions are all the more important as traditional foundation models~\cite{bommasani2021opportunities} come with centralization and monetization, raise data privacy concerns, and lack transparency and reproducibility~\cite{reflectionsfoundationmodels}, which may hinder the democratization of AI. The ability to collaboratively improve weights represents a shift from \emph{proprietary network training} to \emph{open-source collaborative network building}, and could lead to the development of more responsible and reliable AI systems. We see this as an exciting possibility for the future of AI.

\paragraph{Acknowledgments.}
AR and MC acknowledge the financial support by the ANR agency in the chair VISA-DEEP (ANR-20-CHIA-0022-01).
\clearpage
\newpage

\bibliography{icml_main}

\begin{thebibliography}{104}
\providecommand{\natexlab}[1]{#1}
\providecommand{\url}[1]{\texttt{#1}}
\expandafter\ifx\csname urlstyle\endcsname\relax
  \providecommand{\doi}[1]{doi: #1}\else
  \providecommand{\doi}{doi: \begingroup \urlstyle{rm}\Url}\fi

\bibitem[Abnar et~al.(2022)Abnar, Dehghani, Neyshabur, and
  Sedghi]{exploringlimits51104}
Abnar, S., Dehghani, M., Neyshabur, B., and Sedghi, H.
\newblock Exploring the limits of large scale pre-training.
\newblock In \emph{ICLR}, 2022.

\bibitem[Ainsworth et~al.(2023)Ainsworth, Hayase, and
  Srinivasa]{ainsworth2022gitrebasin}
Ainsworth, S.~K., Hayase, J., and Srinivasa, S.
\newblock {Git} re-basin: Merging models modulo permutation symmetries.
\newblock In \emph{ICLR}, 2023.

\bibitem[Aksela(2003)]{aksela2003comparison}
Aksela, M.
\newblock Comparison of classifier selection methods for improving committee
  performance.
\newblock In \emph{MCS}, 2003.

\bibitem[Arjovsky et~al.(2019)Arjovsky, Bottou, Gulrajani, and
  Lopez-Paz]{arjovsky2019invariant}
Arjovsky, M., Bottou, L., Gulrajani, I., and Lopez-Paz, D.
\newblock Invariant risk minimization.
\newblock \emph{arXiv preprint}, 2019.

\bibitem[Arpit et~al.(2021)Arpit, Wang, Zhou, and Xiong]{arpit2021ensemble}
Arpit, D., Wang, H., Zhou, Y., and Xiong, C.
\newblock Ensemble of averages: Improving model selection and boosting
  performance in domain generalization.
\newblock In \emph{NeurIPS}, 2021.

\bibitem[Beery et~al.(2018)Beery, Van~Horn, and Perona]{beery2018recognition}
Beery, S., Van~Horn, G., and Perona, P.
\newblock Recognition in {Terra Incognita}.
\newblock In \emph{ECCV}, 2018.

\bibitem[Beery et~al.(2021)Beery, Agarwal, Cole, and
  Birodkar]{beery2021iwildcam}
Beery, S., Agarwal, A., Cole, E., and Birodkar, V.
\newblock The iwildcam 2021 competition dataset.
\newblock \emph{arXiv preprint}, 2021.

\bibitem[Bommasani \& Liang(2021)Bommasani and
  Liang]{reflectionsfoundationmodels}
Bommasani, R. and Liang, P.
\newblock Reflections on foundation models.
\newblock \url{https://hai.stanford.edu/news/reflections-foundation-models},
  2021.

\bibitem[Bommasani et~al.(2021)Bommasani, Hudson, Adeli, Altman, Arora, von
  Arx, Bernstein, Bohg, Bosselut, Brunskill,
  et~al.]{bommasani2021opportunities}
Bommasani, R., Hudson, D.~A., Adeli, E., Altman, R., Arora, S., von Arx, S.,
  Bernstein, M.~S., Bohg, J., Bosselut, A., Brunskill, E., et~al.
\newblock On the opportunities and risks of foundation models.
\newblock \emph{arXiv preprint}, 2021.

\bibitem[Borzunov et~al.(2022)Borzunov, Baranchuk, Dettmers, Ryabinin, Belkada,
  Chumachenko, Samygin, and Raffel]{borzunov2022petals}
Borzunov, A., Baranchuk, D., Dettmers, T., Ryabinin, M., Belkada, Y.,
  Chumachenko, A., Samygin, P., and Raffel, C.
\newblock Petals: Collaborative inference and fine-tuning of large models.
\newblock \emph{arXiv preprint}, 2022.

\bibitem[Cha et~al.(2021)Cha, Chun, Lee, Cho, Park, Lee, and Park]{cha2021wad}
Cha, J., Chun, S., Lee, K., Cho, H.-C., Park, S., Lee, Y., and Park, S.
\newblock {SWAD}: Domain generalization by seeking flat minima.
\newblock In \emph{NeurIPS}, 2021.

\bibitem[Chang \& Lu(2021)Chang and Lu]{chang2021rethinking}
Chang, T.-Y. and Lu, C.-J.
\newblock Rethinking why intermediate-task fine-tuning works.
\newblock \emph{arXiv preprint}, 2021.

\bibitem[Choshen et~al.(2022{\natexlab{a}})Choshen, Venezian, Don-Yehia,
  Slonim, and Katz]{choshen2022start}
Choshen, L., Venezian, E., Don-Yehia, S., Slonim, N., and Katz, Y.
\newblock Where to start? analyzing the potential value of intermediate models.
\newblock \emph{arXiv preprint}, 2022{\natexlab{a}}.

\bibitem[Choshen et~al.(2022{\natexlab{b}})Choshen, Venezian, Slonim, and
  Katz]{choshen2022fusing}
Choshen, L., Venezian, E., Slonim, N., and Katz, Y.
\newblock Fusing finetuned models for better pretraining.
\newblock \emph{arXiv preprint}, 2022{\natexlab{b}}.

\bibitem[DeGrave et~al.(2021)DeGrave, Janizek, and Lee]{degrave2021ai}
DeGrave, A.~J., Janizek, J.~D., and Lee, S.-I.
\newblock {AI} for radiographic {COVID}-19 detection selects shortcuts over
  signal.
\newblock \emph{Nature Machine Intelligence}, 2021.

\bibitem[Don-Yehiya et~al.(2022)Don-Yehiya, Venezian, Raffel, Slonim, Katz, and
  Choshen]{choshen2022cold}
Don-Yehiya, S., Venezian, E., Raffel, C., Slonim, N., Katz, Y., and Choshen, L.
\newblock {ColD} fusion: Collaborative descent for distributed multitask
  finetuning.
\newblock \emph{arXiv preprint}, 2022.

\bibitem[Dosovitskiy et~al.(2021)Dosovitskiy, Beyer, Kolesnikov, Weissenborn,
  Zhai, Unterthiner, Dehghani, Minderer, Heigold, Gelly, Uszkoreit, and
  Houlsby]{dosovitskiy2021an}
Dosovitskiy, A., Beyer, L., Kolesnikov, A., Weissenborn, D., Zhai, X.,
  Unterthiner, T., Dehghani, M., Minderer, M., Heigold, G., Gelly, S.,
  Uszkoreit, J., and Houlsby, N.
\newblock An image is worth 16x16 words: Transformers for image recognition at
  scale.
\newblock In \emph{ICLR}, 2021.

\bibitem[Draxler et~al.(2018)Draxler, Veschgini, Salmhofer, and
  Hamprecht]{Draxler2018}
Draxler, F., Veschgini, K., Salmhofer, M., and Hamprecht, F.
\newblock Essentially no barriers in neural network energy landscape.
\newblock In \emph{ICML}, 2018.

\bibitem[Eeckt et~al.(2022)]{eeckt2022weight}
Eeckt, S.~V. et~al.
\newblock Weight averaging: A simple yet effective method to overcome
  catastrophic forgetting in automatic speech recognition.
\newblock \emph{arXiv preprint}, 2022.

\bibitem[Entezari et~al.(2022)Entezari, Sedghi, Saukh, and
  Neyshabur]{entezari2022the}
Entezari, R., Sedghi, H., Saukh, O., and Neyshabur, B.
\newblock The role of permutation invariance in linear mode connectivity of
  neural networks.
\newblock In \emph{ICLR}, 2022.

\bibitem[Fang et~al.(2022)Fang, Ilharco, Wortsman, Wan, Shankar, Dave, and
  Schmidt]{pmlr-v162-fang22a}
Fang, A., Ilharco, G., Wortsman, M., Wan, Y., Shankar, V., Dave, A., and
  Schmidt, L.
\newblock Data determines distributional robustness in contrastive language
  image pre-training ({CLIP}).
\newblock In \emph{ICML}, 2022.

\bibitem[Fang et~al.(2023)Fang, Kornblith, and Schmidt]{fang2023does}
Fang, A., Kornblith, S., and Schmidt, L.
\newblock Does progress on {ImageNet} transfer to real-world datasets?
\newblock \emph{arXiv preprint}, 2023.

\bibitem[Fang et~al.(2013)Fang, Xu, and Rockmore]{fang2013unbiased}
Fang, C., Xu, Y., and Rockmore, D.~N.
\newblock Unbiased metric learning: On the utilization of multiple datasets and
  web images for softening bias.
\newblock In \emph{ICCV}, 2013.

\bibitem[Frankle et~al.(2020)Frankle, Dziugaite, Roy, and Carbin]{Frankle2020}
Frankle, J., Dziugaite, G.~K., Roy, D.~M., and Carbin, M.
\newblock Linear mode connectivity and the lottery ticket hypothesis.
\newblock In \emph{ICML}, 2020.

\bibitem[Gontijo-Lopes et~al.(2022)Gontijo-Lopes, Dauphin, and
  Cubuk]{gontijolopes2022no}
Gontijo-Lopes, R., Dauphin, Y., and Cubuk, E.~D.
\newblock No one representation to rule them all: Overlapping features of
  training methods.
\newblock In \emph{ICLR}, 2022.

\bibitem[Gulrajani \& Lopez-Paz(2021)Gulrajani and Lopez-Paz]{gulrajani2021in}
Gulrajani, I. and Lopez-Paz, D.
\newblock In search of lost domain generalization.
\newblock In \emph{ICLR}, 2021.

\bibitem[He et~al.(2016)He, Zhang, Ren, and Sun]{he51deep}
He, K., Zhang, X., Ren, S., and Sun, J.
\newblock Deep residual learning for image recognition.
\newblock In \emph{CVPR}, 2016.

\bibitem[Hendrycks \& Dietterich(2019)Hendrycks and
  Dietterich]{hendrycks2018benchmarking}
Hendrycks, D. and Dietterich, T.
\newblock Benchmarking neural network robustness to common corruptions and
  perturbations.
\newblock In \emph{ICLR}, 2019.

\bibitem[Hendrycks et~al.(2021)Hendrycks, Basart, Mu, Kadavath, Wang, Dorundo,
  Desai, Zhu, Parajuli, Guo, et~al.]{hendrycks2021many}
Hendrycks, D., Basart, S., Mu, N., Kadavath, S., Wang, F., Dorundo, E., Desai,
  R., Zhu, T., Parajuli, S., Guo, M., et~al.
\newblock The many faces of robustness: A critical analysis of
  out-of-distribution generalization.
\newblock In \emph{ICCV}, 2021.

\bibitem[Ilharco et~al.(2022)Ilharco, Wortsman, Gadre, Song, Hajishirzi,
  Kornblith, Farhadi, and Schmidt]{ilharco2022patching}
Ilharco, G., Wortsman, M., Gadre, S.~Y., Song, S., Hajishirzi, H., Kornblith,
  S., Farhadi, A., and Schmidt, L.
\newblock Patching open-vocabulary models by interpolating weights.
\newblock In \emph{NeurIPS}, 2022.

\bibitem[Ilharco et~al.(2023)Ilharco, {Tulio Ribeiro}, {Wortsman},
  {Gururangan}, {Schmidt}, {Hajishirzi}, and {Farhadi}]{2022arXiv221204089I}
Ilharco, G., {Tulio Ribeiro}, M., {Wortsman}, M., {Gururangan}, S., {Schmidt},
  L., {Hajishirzi}, H., and {Farhadi}, A.
\newblock Editing models with task arithmetic.
\newblock In \emph{ICLR}, 2023.

\bibitem[Iwasawa \& Matsuo(2021)Iwasawa and Matsuo]{iwasawa2021testtime}
Iwasawa, Y. and Matsuo, Y.
\newblock Test-time classifier adjustment module for model-agnostic domain
  generalization.
\newblock In \emph{NeurIPS}, 2021.

\bibitem[Izmailov et~al.(2018)Izmailov, Podoprikhin, Garipov, Vetrov, and
  Wilson]{izmailov2018}
Izmailov, P., Podoprikhin, D., Garipov, T., Vetrov, D., and Wilson, A.
\newblock Averaging weights leads to wider optima and better generalization.
\newblock In \emph{UAI}, 2018.

\bibitem[Jain et~al.(2022)Jain, Tsipras, and Madry]{pmlr-v162-jain22b}
Jain, S., Tsipras, D., and Madry, A.
\newblock Combining diverse feature priors.
\newblock In \emph{ICML}, 2022.

\bibitem[Jin et~al.(2023)Jin, Ren, Preotiuc-Pietro, and Cheng]{jin2022dataless}
Jin, X., Ren, X., Preotiuc-Pietro, D., and Cheng, P.
\newblock Dataless knowledge fusion by merging weights of language models.
\newblock 2023.

\bibitem[Jordan et~al.(2023)Jordan, Sedghi, Saukh, Entezari, and
  Neyshabur]{jordan2022repair}
Jordan, K., Sedghi, H., Saukh, O., Entezari, R., and Neyshabur, B.
\newblock Repair: Renormalizing permuted activations for interpolation repair.
\newblock In \emph{ICLR}, 2023.

\bibitem[Juneja et~al.(2023)Juneja, Bansal, Cho, Sedoc, and
  Saphra]{juneja2022linear}
Juneja, J., Bansal, R., Cho, K., Sedoc, J., and Saphra, N.
\newblock Linear connectivity reveals generalization strategies.
\newblock In \emph{ICLR}, 2023.

\bibitem[Kaddour(2022)]{kaddour2022stop}
Kaddour, J.
\newblock Stop wasting my time! saving days of imagenet and {BERT} training
  with latest weight averaging.
\newblock In \emph{NeurIPS Workshop}, 2022.

\bibitem[Karpathy(2017)]{software20}
Karpathy, A.
\newblock {Software} 2.0.
\newblock \url{https://karpathy.medium.com/software-2-0-a64152b37c35}, 2017.

\bibitem[Kingma \& Ba(2015)Kingma and Ba]{kingma2014adam}
Kingma, D.~P. and Ba, J.
\newblock Adam: A method for stochastic optimization.
\newblock In \emph{ICLR}, 2015.

\bibitem[Kirsch et~al.(2022)Kirsch, Lakshminarayanan, Hu, Sculley, Phan, Tran,
  Snoek, Liu, Ren, van Amersfoort, Han, Buchanan, Murphy, Collier, Dusenberry,
  Band, Thain, Jenatton, Rudner, Gal, Nado, Mariet, Wang, and
  Ghahramani]{plex2022}
Kirsch, A.~C., Lakshminarayanan, B., Hu, C.~H., Sculley, D., Phan, D., Tran,
  D., Snoek, J.~R., Liu, J., Ren, J.~J., van Amersfoort, J., Han, K., Buchanan,
  K., Murphy, K.~P., Collier, M.~P., Dusenberry, M.~W., Band, N., Thain, N.,
  Jenatton, R., Rudner, T. G.~J., Gal, Y., Nado, Z., Mariet, Z., Wang, Z., and
  Ghahramani, Z.
\newblock Plex: Towards reliability using pretrained large model extensions.
\newblock In \emph{ICML Workshop}, 2022.

\bibitem[Koh et~al.(2021)Koh, Sagawa, Marklund, Xie, Zhang, Balsubramani, Hu,
  Yasunaga, Phillips, Gao, Lee, David, Stavness, Guo, Earnshaw, Haque, Beery,
  Leskovec, Kundaje, Pierson, Levine, Finn, and Liang]{pmlr-v139-koh21a}
Koh, P.~W., Sagawa, S., Marklund, H., Xie, S.~M., Zhang, M., Balsubramani, A.,
  Hu, W., Yasunaga, M., Phillips, R.~L., Gao, I., Lee, T., David, E., Stavness,
  I., Guo, W., Earnshaw, B., Haque, I., Beery, S.~M., Leskovec, J., Kundaje,
  A., Pierson, E., Levine, S., Finn, C., and Liang, P.
\newblock {WILDS}: A benchmark of in-the-wild distribution shifts.
\newblock In \emph{ICML}, 2021.

\bibitem[Kumar et~al.(2022)Kumar, Raghunathan, Jones, Ma, and
  Liang]{kumar2022finetuning}
Kumar, A., Raghunathan, A., Jones, R.~M., Ma, T., and Liang, P.
\newblock Fine-tuning can distort pretrained features and underperform
  out-of-distribution.
\newblock In \emph{ICLR}, 2022.

\bibitem[Kuncheva \& Whitaker(2003)Kuncheva and Whitaker]{kuncheva2003measures}
Kuncheva, L.~I. and Whitaker, C.~J.
\newblock Measures of diversity in classifier ensembles and their relationship
  with the ensemble accuracy.
\newblock \emph{Machine learning}, 2003.

\bibitem[Kuutti et~al.(2020)Kuutti, Bowden, Jin, Barber, and
  Fallah]{kuutti2020survey}
Kuutti, S., Bowden, R., Jin, Y., Barber, P., and Fallah, S.
\newblock A survey of deep learning applications to autonomous vehicle control.
\newblock \emph{T-ITS}, 2020.

\bibitem[Laakom et~al.(2021)Laakom, Raitoharju, Iosifidis, and
  Gabbouj]{laakom2021within}
Laakom, F., Raitoharju, J., Iosifidis, A., and Gabbouj, M.
\newblock Within-layer diversity reduces generalization gap.
\newblock In \emph{ICML Workshop}, 2021.

\bibitem[Lakshminarayanan et~al.(2017)Lakshminarayanan, Pritzel, and
  Blundell]{Lakshminarayanan2017}
Lakshminarayanan, B., Pritzel, A., and Blundell, C.
\newblock Simple and scalable predictive uncertainty estimation using deep
  ensembles.
\newblock In \emph{NeurIPS}, 2017.

\bibitem[Langnickel et~al.(2022)Langnickel, Schulz, Hammer, and
  Fluck]{langnickel2022bert}
Langnickel, L., Schulz, A., Hammer, B., and Fluck, J.
\newblock {BERT WEAVER}: Using {WEight AVERaging} to enable lifelong learning
  for transformer-based models.
\newblock \emph{arXiv preprint}, 2022.

\bibitem[Li et~al.(2017)Li, Yang, Song, and Hospedales]{li2017deeper}
Li, D., Yang, Y., Song, Y.-Z., and Hospedales, T.~M.
\newblock Deeper, broader and artier domain generalization.
\newblock In \emph{ICCV}, 2017.

\bibitem[Li et~al.(2022{\natexlab{a}})Li, Gururangan, Dettmers, Lewis, Althoff,
  Smith, and Zettlemoyer]{li2022branch}
Li, M., Gururangan, S., Dettmers, T., Lewis, M., Althoff, T., Smith, N.~A., and
  Zettlemoyer, L.
\newblock {Branch-Train-Merge}: Embarrassingly parallel training of expert
  language models.
\newblock \emph{arXiv preprint}, 2022{\natexlab{a}}.

\bibitem[Li et~al.(2019)Li, Wen, and He]{li2019federated}
Li, Q., Wen, Z., and He, B.
\newblock Federated learning systems: Vision, hype and reality for data privacy
  and protection.
\newblock \emph{arXiv preprint}, 2019.

\bibitem[Li et~al.(2022{\natexlab{b}})Li, Huang, Tao, Wu, and
  Huang]{li2022trainable}
Li, T., Huang, Z., Tao, Q., Wu, Y., and Huang, X.
\newblock Trainable weight averaging for fast convergence and better
  generalization.
\newblock \emph{arXiv preprint}, 2022{\natexlab{b}}.

\bibitem[Li et~al.(2021{\natexlab{a}})Li, Liu, and Bilen]{li2021Universal}
Li, W.-H., Liu, X., and Bilen, H.
\newblock Universal representation learning from multiple domains for few-shot
  classification.
\newblock In \emph{ICCV}, 2021{\natexlab{a}}.

\bibitem[Li et~al.(2021{\natexlab{b}})Li, Wang, and Barni]{li2021survey}
Li, Y., Wang, H., and Barni, M.
\newblock A survey of deep neural network watermarking techniques.
\newblock \emph{Neurocomputing}, 2021{\natexlab{b}}.

\bibitem[Lopez-Paz et~al.(2022)Lopez-Paz, Bouchacourt, Sagun, and
  Usunier]{lopez2022measuring}
Lopez-Paz, D., Bouchacourt, D., Sagun, L., and Usunier, N.
\newblock Measuring and signing fairness as performance under multiple
  stakeholder distributions.
\newblock \emph{arXiv preprint}, 2022.

\bibitem[Lubana et~al.(2022{\natexlab{a}})Lubana, Bigelow, Dick, Krueger, and
  Tanaka]{lubana2022mechanistic}
Lubana, E.~S., Bigelow, E.~J., Dick, R., Krueger, D., and Tanaka, H.
\newblock Mechanistic lens on mode connectivity.
\newblock In \emph{NeurIPS Workshop}, 2022{\natexlab{a}}.

\bibitem[Lubana et~al.(2022{\natexlab{b}})Lubana, Trivedi, Koutra, and
  Dick]{lubana2021quadratic}
Lubana, E.~S., Trivedi, P., Koutra, D., and Dick, R.~P.
\newblock How do quadratic regularizers prevent catastrophic forgetting: The
  role of interpolation.
\newblock In \emph{CoLLAs}, 2022{\natexlab{b}}.

\bibitem[Marcel \& Rodriguez(2010)Marcel and Rodriguez]{torchvision2010}
Marcel, S. and Rodriguez, Y.
\newblock Torchvision the machine-vision package of {Torch}.
\newblock In \emph{ACM}, 2010.

\bibitem[Maron et~al.(2022)Maron, Hekler, Haggenm{\"u}ller, von Kalle, Utikal,
  M{\"u}ller, Gaiser, Meier, Hobelsberger, Gellrich, et~al.]{maron2022model}
Maron, R.~C., Hekler, A., Haggenm{\"u}ller, S., von Kalle, C., Utikal, J.~S.,
  M{\"u}ller, V., Gaiser, M., Meier, F., Hobelsberger, S., Gellrich, F.~F.,
  et~al.
\newblock Model soups improve performance of dermoscopic skin cancer
  classifiers.
\newblock \emph{European Journal of Cancer}, 2022.

\bibitem[Matena \& Raffel(2022)Matena and Raffel]{mergefisher21}
Matena, M. and Raffel, C.
\newblock Merging models with {Fisher}-weighted averaging.
\newblock In \emph{NeurIPS}, 2022.

\bibitem[Miller et~al.(2020)Miller, Krauth, Recht, and
  Schmidt]{pmlr-v119-miller20a}
Miller, J., Krauth, K., Recht, B., and Schmidt, L.
\newblock The effect of natural distribution shift on question answering
  models.
\newblock In \emph{ICML}, 2020.

\bibitem[Miller et~al.(2021)Miller, Taori, Raghunathan, Sagawa, Koh, Shankar,
  Liang, Carmon, and Schmidt]{pmlr-v139-miller21b}
Miller, J.~P., Taori, R., Raghunathan, A., Sagawa, S., Koh, P.~W., Shankar, V.,
  Liang, P., Carmon, Y., and Schmidt, L.
\newblock Accuracy on the line: on the strong correlation between
  out-of-distribution and in-distribution generalization.
\newblock In \emph{ICML}, 2021.

\bibitem[Mirzadeh et~al.(2021)Mirzadeh, Farajtabar, Gorur, Pascanu, and
  Ghasemzadeh]{mirzadeh2021linear}
Mirzadeh, S.~I., Farajtabar, M., Gorur, D., Pascanu, R., and Ghasemzadeh, H.
\newblock Linear mode connectivity in multitask and continual learning.
\newblock In \emph{ICLR}, 2021.

\bibitem[Monaco(2020)]{ratatouille2020}
Monaco, E.
\newblock The right way to make ratatouille.
\newblock \url{https://www.bbc.com/travel/article/20200812-the-right-way-
  to-make-ratatouille}, 2020.

\bibitem[Nagarajan \& Kolter(2019)Nagarajan and Kolter]{nagarajan2019uniform}
Nagarajan, V. and Kolter, J.~Z.
\newblock Uniform convergence may be unable to explain generalization in deep
  learning.
\newblock \emph{NeurIPS}, 2019.

\bibitem[Nayman et~al.(2022)Nayman, Golbert, Noy, Ping, and
  Zelnik-Manor]{nayman2022diverse}
Nayman, N., Golbert, A., Noy, A., Ping, T., and Zelnik-Manor, L.
\newblock Diverse {ImageNet} models transfer better.
\newblock \emph{arXiv preprint}, 2022.

\bibitem[Neyshabur et~al.(2020)Neyshabur, Sedghi, and Zhang]{Neyshabur2020}
Neyshabur, B., Sedghi, H., and Zhang, C.
\newblock What is being transferred in transfer learning?
\newblock In \emph{NeurIPS}, 2020.

\bibitem[Nguyen et~al.(2022)Nguyen, Ilharco, Wortsman, Oh, and
  Schmidt]{nguyen2022quality}
Nguyen, T., Ilharco, G., Wortsman, M., Oh, S., and Schmidt, L.
\newblock Quality not quantity: On the interaction between dataset design and
  robustness of {CLIP}.
\newblock In \emph{NeurIPS}, 2022.

\bibitem[Oquab et~al.(2014)Oquab, Bottou, Laptev, and Sivic]{oquab2014learning}
Oquab, M., Bottou, L., Laptev, I., and Sivic, J.
\newblock Learning and transferring mid-level image representations using
  convolutional neural networks.
\newblock In \emph{CVPR}, 2014.

\bibitem[Peng et~al.(2019)Peng, Bai, Xia, Huang, Saenko, and
  Wang]{peng2019moment}
Peng, X., Bai, Q., Xia, X., Huang, Z., Saenko, K., and Wang, B.
\newblock Moment matching for multi-source domain adaptation.
\newblock In \emph{ICCV}, 2019.

\bibitem[Phang et~al.(2018)Phang, F{\'e}vry, and Bowman]{phang2018sentence}
Phang, J., F{\'e}vry, T., and Bowman, S.~R.
\newblock Sentence encoders on stilts: Supplementary training on intermediate
  labeled-data tasks.
\newblock \emph{arXiv preprint}, 2018.

\bibitem[Pruksachatkun et~al.(2020)Pruksachatkun, Phang, Liu, Htut, Zhang,
  Pang, Vania, Kann, and Bowman]{pruksachatkun2020intermediate}
Pruksachatkun, Y., Phang, J., Liu, H., Htut, P.~M., Zhang, X., Pang, R.~Y.,
  Vania, C., Kann, K., and Bowman, S.
\newblock Intermediate-task transfer learning with pretrained language models:
  When and why does it work?
\newblock In \emph{ACL}, 2020.

\bibitem[Qin et~al.(2022)Qin, Qian, Yi, Chen, Lin, Han, Liu, Sun, and
  Zhou]{qin2022exploring}
Qin, Y., Qian, C., Yi, J., Chen, W., Lin, Y., Han, X., Liu, Z., Sun, M., and
  Zhou, J.
\newblock Exploring mode connectivity for pre-trained language models.
\newblock In \emph{EMNLP}, 2022.

\bibitem[Raffel(2023)]{updatablemachinelearning}
Raffel, C.
\newblock Building machine learning models like open source software.
\newblock \emph{ACM}, 2023.

\bibitem[Ramé \& Cord(2021)Ramé and Cord]{rame2021dice}
Ramé, A. and Cord, M.
\newblock {DICE}: Diversity in deep ensembles via conditional redundancy
  adversarial estimation.
\newblock In \emph{ICLR}, 2021.

\bibitem[Ramé et~al.(2022{\natexlab{a}})Ramé, Kirchmeyer, Rahier,
  Rakotomamonjy, Gallinari, and Cord]{rame2022diwa}
Ramé, A., Kirchmeyer, M., Rahier, T., Rakotomamonjy, A., Gallinari, P., and
  Cord, M.
\newblock Diverse weight averaging for out-of-distribution generalization.
\newblock In \emph{NeurIPS}, 2022{\natexlab{a}}.

\bibitem[Ramé et~al.(2022{\natexlab{b}})Ramé, Zhang, Bottou, and
  Lopez-Paz]{rame2022pretrain}
Ramé, A., Zhang, J., Bottou, L., and Lopez-Paz, D.
\newblock Pre-train, fine-tune, interpolate: a three-stage strategy for domain
  generalization.
\newblock In \emph{NeurIPS Interpolate Workshop}, 2022{\natexlab{b}}.

\bibitem[Ramé et~al.(2023)Ramé, Couairon, Shukor, Dancette, Gaya, Soulier,
  and Cord]{rame2023rewarded}
Ramé, A., Couairon, G., Shukor, M., Dancette, C., Gaya, J.-B., Soulier, L.,
  and Cord, M.
\newblock Rewarded soups: towards pareto-optimal alignment by interpolating
  weights fine-tuned on diverse rewards.
\newblock \emph{arXiv preprint arXiv:2306.04488}, 2023.

\bibitem[Rebuffi et~al.(2017)Rebuffi, Kolesnikov, Sperl, and
  Lampert]{rebuffi2017icarl}
Rebuffi, S.-A., Kolesnikov, A., Sperl, G., and Lampert, C.~H.
\newblock {iCaRL}: Incremental classifier and representation learning.
\newblock In \emph{CVPR}, 2017.

\bibitem[Russakovsky et~al.(2015)Russakovsky, Deng, Su, Krause, Satheesh, Ma,
  Huang, Karpathy, Khosla, Bernstein, et~al.]{russakovsky2015imagenet}
Russakovsky, O., Deng, J., Su, H., Krause, J., Satheesh, S., Ma, S., Huang, Z.,
  Karpathy, A., Khosla, A., Bernstein, M., et~al.
\newblock {ImageNet} large scale visual recognition challenge.
\newblock In \emph{IJCV}, 2015.

\bibitem[Ryabinin \& Gusev(2020)Ryabinin and Gusev]{ryabinin2020towards}
Ryabinin, M. and Gusev, A.
\newblock Towards crowdsourced training of large neural networks using
  decentralized mixture-of-experts.
\newblock \emph{NeurIPS}, 2020.

\bibitem[Sackfield(2021)]{sotamoon2021}
Sackfield, W.
\newblock {SOTAMoon}.
\newblock \url{https://github.com/8W9aG/SOTAMoon}, 2021.

\bibitem[Shah et~al.(2020)Shah, Tamuly, Raghunathan, Jain, and
  Netrapalli]{NEURIPS2020_6cfe0e61}
Shah, H., Tamuly, K., Raghunathan, A., Jain, P., and Netrapalli, P.
\newblock The pitfalls of simplicity bias in neural networks.
\newblock In \emph{NeurIPS}, 2020.

\bibitem[Shukor et~al.(2023)Shukor, Dancette, Ramé, and
  Cord]{shukor2023unified}
Shukor, M., Dancette, C., Ramé, A., and Cord, M.
\newblock Unified model for image, video, audio and language tasks.
\newblock \emph{arXiv preprint arXiv:2307.16184}, 2023.

\bibitem[Stojanovski et~al.(2022)Stojanovski, Roth, and
  Akata]{stojanovski2022momentum}
Stojanovski, Z., Roth, K., and Akata, Z.
\newblock Momentum-based weight interpolation of strong zero-shot models for
  continual learning.
\newblock In \emph{NeurIPS Interpolate Workshop}, 2022.

\bibitem[Sun et~al.(2016)Sun, Feng, and Saenko]{coral216aaai}
Sun, B., Feng, J., and Saenko, K.
\newblock Return of frustratingly easy domain adaptation.
\newblock In \emph{AAAI}, 2016.

\bibitem[Szegedy et~al.(2016)Szegedy, Vanhoucke, Ioffe, Shlens, and
  Wojna]{szegedy2016rethinking}
Szegedy, C., Vanhoucke, V., Ioffe, S., Shlens, J., and Wojna, Z.
\newblock Rethinking the inception architecture for computer vision.
\newblock In \emph{CVPR}, 2016.

\bibitem[Taori et~al.(2020)Taori, Dave, Shankar, Carlini, Recht, and
  Schmidt]{NEURIPS2020_d8330f85}
Taori, R., Dave, A., Shankar, V., Carlini, N., Recht, B., and Schmidt, L.
\newblock Measuring robustness to natural distribution shifts in image
  classification.
\newblock In \emph{NeurIPS}, 2020.

\bibitem[Taylor et~al.(2016)Taylor, Yudkowsky, LaVictoire, and
  Critch]{taylor2016alignment}
Taylor, J., Yudkowsky, E., LaVictoire, P., and Critch, A.
\newblock Alignment for advanced machine learning systems.
\newblock \emph{Ethics of Artificial Intelligence}, 2016.

\bibitem[Taylor et~al.(2019)Taylor, Earnshaw, Mabey, Victors, and
  Yosinski]{taylor2019rxrx1}
Taylor, J., Earnshaw, B., Mabey, B., Victors, M., and Yosinski, J.
\newblock {RxRx1}: An image set for cellular morphological variation across
  many experimental batches.
\newblock In \emph{ICLR Workshop}, 2019.

\bibitem[Teney et~al.(2022)Teney, Lin, Oh, and Abbasnejad]{teney2022id}
Teney, D., Lin, Y., Oh, S.~J., and Abbasnejad, E.
\newblock {ID} and {OOD} performance are sometimes inversely correlated on
  real-world datasets.
\newblock \emph{arXiv preprint}, 2022.

\bibitem[Vapnik(1992)]{NIPS1991_ff4d5fbb}
Vapnik, V.
\newblock Principles of risk minimization for learning theory.
\newblock In \emph{NeurIPS}, 1992.

\bibitem[Venkateswara et~al.(2017)Venkateswara, Eusebio, Chakraborty, and
  Panchanathan]{venkateswara2017deep}
Venkateswara, H., Eusebio, J., Chakraborty, S., and Panchanathan, S.
\newblock Deep hashing network for unsupervised domain adaptation.
\newblock In \emph{CVPR}, 2017.

\bibitem[Wenzel et~al.(2022)Wenzel, Dittadi, Gehler, Simon-Gabriel, Horn,
  Zietlow, Kernert, Russell, Brox, Schiele, Schölkopf, and
  Locatello]{assayingoodwenzel2022}
Wenzel, F., Dittadi, A., Gehler, P.~V., Simon-Gabriel, C.-J., Horn, M.,
  Zietlow, D., Kernert, D., Russell, C., Brox, T., Schiele, B., Schölkopf, B.,
  and Locatello, F.
\newblock Assaying out-of-distribution generalization in transfer learning.
\newblock In \emph{NeurIPS}, 2022.

\bibitem[Wightman(2019)]{rw2019timm}
Wightman, R.
\newblock {PyTorch Image Models}.
\newblock \url{https://github.com/rwightman/pytorch-image-models}, 2019.

\bibitem[Wolf et~al.(2020)Wolf, Debut, Sanh, Chaumond, Delangue, Moi, Cistac,
  Rault, Louf, Funtowicz, Davison, Shleifer, von Platen, Ma, Jernite, Plu, Xu,
  Le~Scao, Gugger, Drame, Lhoest, and Rush]{wolf-etal-2020-transformers}
Wolf, T., Debut, L., Sanh, V., Chaumond, J., Delangue, C., Moi, A., Cistac, P.,
  Rault, T., Louf, R., Funtowicz, M., Davison, J., Shleifer, S., von Platen,
  P., Ma, C., Jernite, Y., Plu, J., Xu, C., Le~Scao, T., Gugger, S., Drame, M.,
  Lhoest, Q., and Rush, A.
\newblock Transformers: State-of-the-art natural language processing.
\newblock In \emph{EMNLP}, 2020.

\bibitem[Wortsman et~al.(2022{\natexlab{a}})Wortsman, Ilharco, Gadre, Roelofs,
  Gontijo-Lopes, Morcos, Namkoong, Farhadi, Carmon, Kornblith, and
  Schmidt]{Wortsman2022ModelSA}
Wortsman, M., Ilharco, G., Gadre, S.~Y., Roelofs, R., Gontijo-Lopes, R.,
  Morcos, A.~S., Namkoong, H., Farhadi, A., Carmon, Y., Kornblith, S., and
  Schmidt, L.
\newblock Model soups: averaging weights of multiple fine-tuned models improves
  accuracy without increasing inference time.
\newblock In \emph{ICML}, 2022{\natexlab{a}}.

\bibitem[Wortsman et~al.(2022{\natexlab{b}})Wortsman, Ilharco, Kim, Li,
  Hajishirzi, Farhadi, Namkoong, and Schmidt]{Wortsman2022robust}
Wortsman, M., Ilharco, G., Kim, J.~W., Li, M., Hajishirzi, H., Farhadi, A.,
  Namkoong, H., and Schmidt, L.
\newblock Robust fine-tuning of zero-shot models.
\newblock In \emph{CVPR}, 2022{\natexlab{b}}.

\bibitem[Wortsman et~al.(2023)Wortsman, Gururangan, Li, Farhadi, Schmidt,
  Rabbat, and Morcos]{wortsman2022fi}
Wortsman, M., Gururangan, S., Li, S., Farhadi, A., Schmidt, L., Rabbat, M., and
  Morcos, A.~S.
\newblock lo-fi: distributed fine-tuning without communication.
\newblock \emph{TMLR}, 2023.

\bibitem[Ye et~al.(2022)Ye, Li, Hong, Bai, Chen, Zhou, and Li]{ye2021odbench}
Ye, N., Li, K., Hong, L., Bai, H., Chen, Y., Zhou, F., and Li, Z.
\newblock Ood-bench: Benchmarking and understanding out-of-distribution
  generalization datasets and algorithms.
\newblock \emph{CVPR}, 2022.

\bibitem[Yule(1900)]{yule1900vii}
Yule, G.~U.
\newblock On the association of attributes in statistics.
\newblock \emph{Philosophical Transactions of the Royal Society of London.},
  1900.

\bibitem[Zech et~al.(2018)Zech, Badgeley, Liu, Costa, Titano, and
  Oermann]{Zech2018}
Zech, J.~R., Badgeley, M.~A., Liu, M., Costa, A.~B., Titano, J.~J., and
  Oermann, E.~K.
\newblock Variable generalization performance of a deep learning model to
  detect pneumonia in chest radiographs: A cross-sectional study.
\newblock \emph{PLOS Medicine}, 2018.

\bibitem[Zhang \& Bottou(2022)Zhang and Bottou]{anonymous2023learning}
Zhang, J. and Bottou, L.
\newblock Learning useful representations for shifting tasks and distributions.
\newblock \emph{arXiv preprint}, 2022.

\bibitem[Zhang et~al.(2022)Zhang, Lopez-Paz, and Bottou]{zhang2022rich}
Zhang, J., Lopez-Paz, D., and Bottou, L.
\newblock Rich feature construction for the optimization-generalization
  dilemma.
\newblock In \emph{ICML}, 2022.

\end{thebibliography}
\bibliographystyle{icml2023}

\newpage
\appendix
\onecolumn
\hrule

\begin{center}
\LARGE Model Ratatouille:\\ Recycling Diverse Models for Out-of-Distribution Generalization
\end{center}

\begin{center}
\large Supplementary material
\end{center}

\hrule
\vskip 1cm
This supplementary material is organized as follows:
\begin{itemize}
    \item \Cref{app:equations} describes the different fine-tuning strategies as equations.
    \item \Cref{app:trainingcost} analyzes ratatouille's components: \Cref{app:numbertasks} ablates the number of auxiliary tasks, \Cref{app:numbersteps} ablates the number of target fine-tuning steps and \Cref{app:numberruns} ablates the number of target fine-tuning runs.
    \item \Cref{app:diversity} enriches our diversity experiments.
    \item \Cref{app:lmc} further empirically analyzes the validity of \Cref{hyp:1,hyp:2} on additional setups.
    \item \Cref{app:robustintertraining} introduces a new robust ratatouille strategy to (slightly) further improve performance.
    \item \Cref{app:domainbed} describes and enriches our experiments on DomainBed \cite{gulrajani2021in}.
\end{itemize}
\section{Fine-Tuning Strategies as Equations}
\label{app:equations}
In \Cref{fig:intro}, we illustrated the different fine-tuning strategies. In \Cref{eq:strategies}, we now provide an analytical formulation of these strategies with equations, where $\theta$ represents the weights, $T_i$ the auxiliary tasks and $T$ the target task.
\begin{equation}
    \begin{aligned}
        \theta & = \mathrm{Train}\left(\theta^{\mathrm{pt}}, T\right),                                                            & \text{[Vanilla fine-tuning~\cite{oquab2014learning}]}                    \\
        \theta & = \mathrm{Train}\left(\theta^{\mathrm{pt}}, T, \mathrm{collect\_ckpts=True}\right),                              & \text{[Moving average~\cite{izmailov2018}]}                              \\
        \theta & = (1-\lambda) \cdot \mathrm{Train}\left(\theta^{\mathrm{pt}}, T\right) + \lambda \cdot \theta^{\mathrm{pt}},     & \text{[WiSE fine-tuning~\cite{Wortsman2022robust}]}                      \\
        \theta & = \frac{1}{M} \sum_{i=0}^{M-1} \mathrm{Train}\left(\theta^{\mathrm{pt}}, T\right),                                 & \text{[Model soups~\cite{Wortsman2022ModelSA}/DiWA~\cite{rame2022diwa}]} \\
        \theta & = \mathrm{Train}\left(\mathrm{Train}\left(\theta^{\mathrm{pt}}, T_i\right), T\right),                            & \text{[Inter-training \cite{phang2018sentence}]}                         \\
        \theta & = \mathrm{Train}\left(\sum_i \lambda_i \cdot \mathrm{Train}\left(\theta^{\mathrm{pt}}, T_i\right), T\right),      & \text{[Fusing \cite{choshen2022fusing}]}                                 \\
        \theta & = \frac{1}{M} \sum_{i=0}^{M-1} \mathrm{Train}\left(\mathrm{Train}\left(\theta^{\mathrm{pt}}, T_i\right), T\right). & \text{[Model ratatouille (ours)]}                                           \\
    \end{aligned} \label{eq:strategies}%
\end{equation}%

\section{Ratatouille's Components Analysis}
\label{app:trainingcost}
In this section we try to refine our understanding of the importance of various components in ratatouille.
\begin{remark}
    If auxiliary weights are shared by the community, recycling strategies cost no more than other fine-tuning strategies: recycling simply benefits from weights that would otherwise ignore each other and be discarded.
\end{remark}
\subsection{Analysis of the Number of Auxiliary Tasks}
\label{app:numbertasks}
\begin{figure}[h!]
    \begin{center}
        \begin{subfigure}[b]{0.24\textwidth}
            \includegraphics[width=\textwidth]{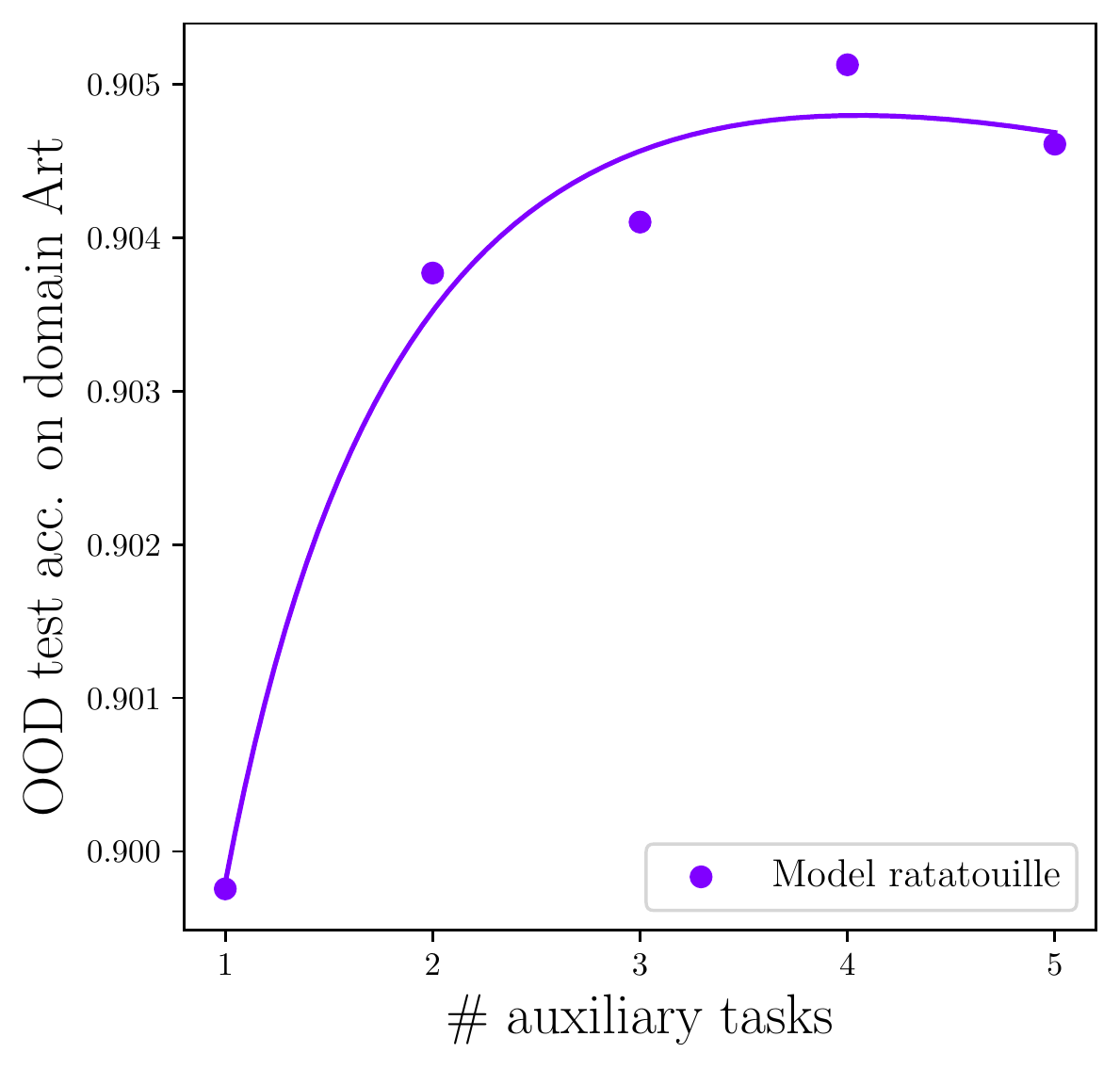}
            \caption{PACS.}
            \label{fig:countaux:a}
        \end{subfigure}
        \hfill
        \begin{subfigure}[b]{0.24\textwidth}
            \includegraphics[width=\textwidth]{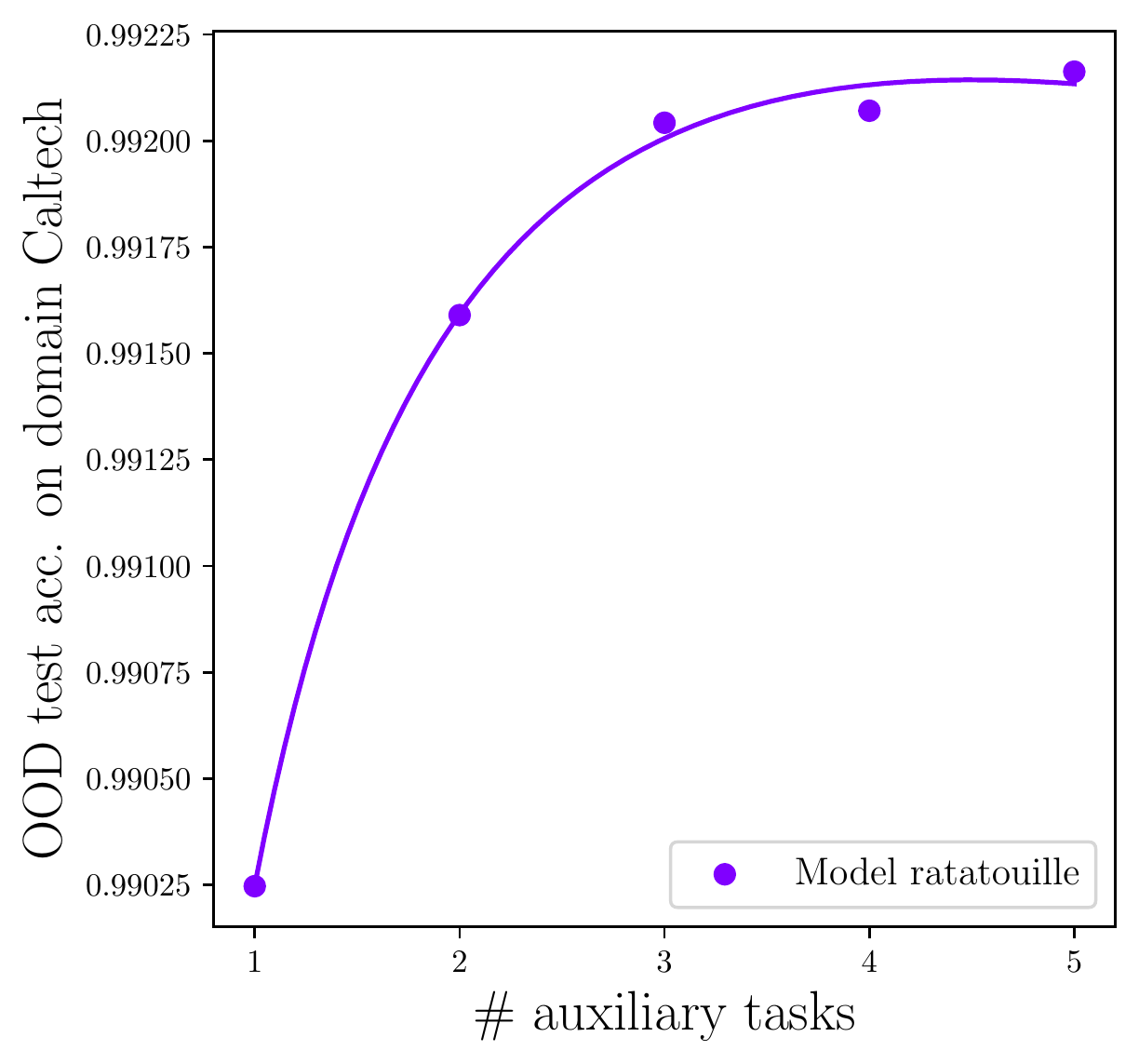}
            \caption{VLCS.}
            \label{fig:countaux:b}
        \end{subfigure}
        \hfill
        \begin{subfigure}[b]{0.24\textwidth}
            \includegraphics[width=\textwidth]{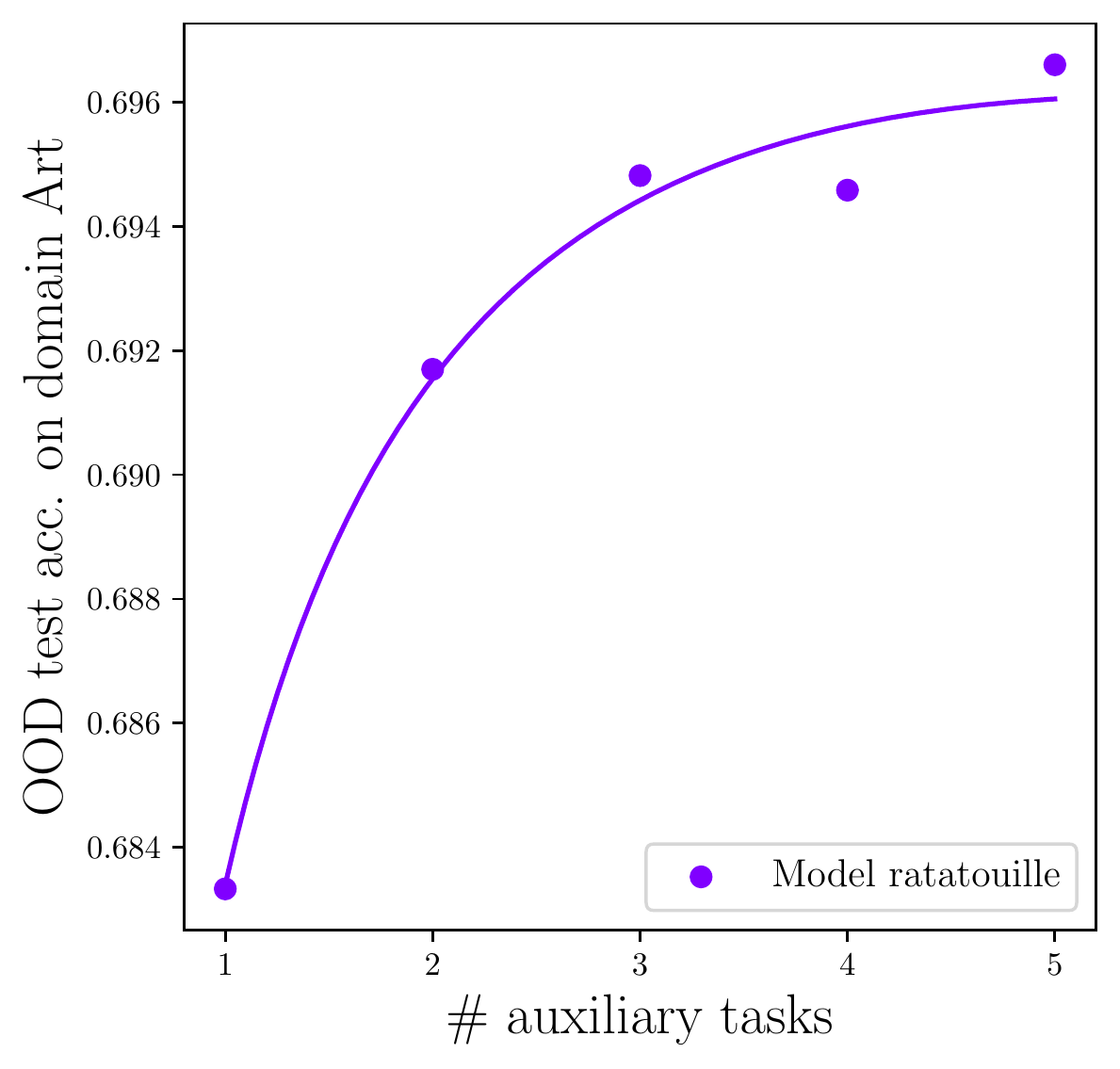}
            \caption{OfficeHome.}
            \label{fig:countaux:c}
        \end{subfigure}
        \hfill
        \begin{subfigure}[b]{0.24\textwidth}
            \includegraphics[width=\textwidth]{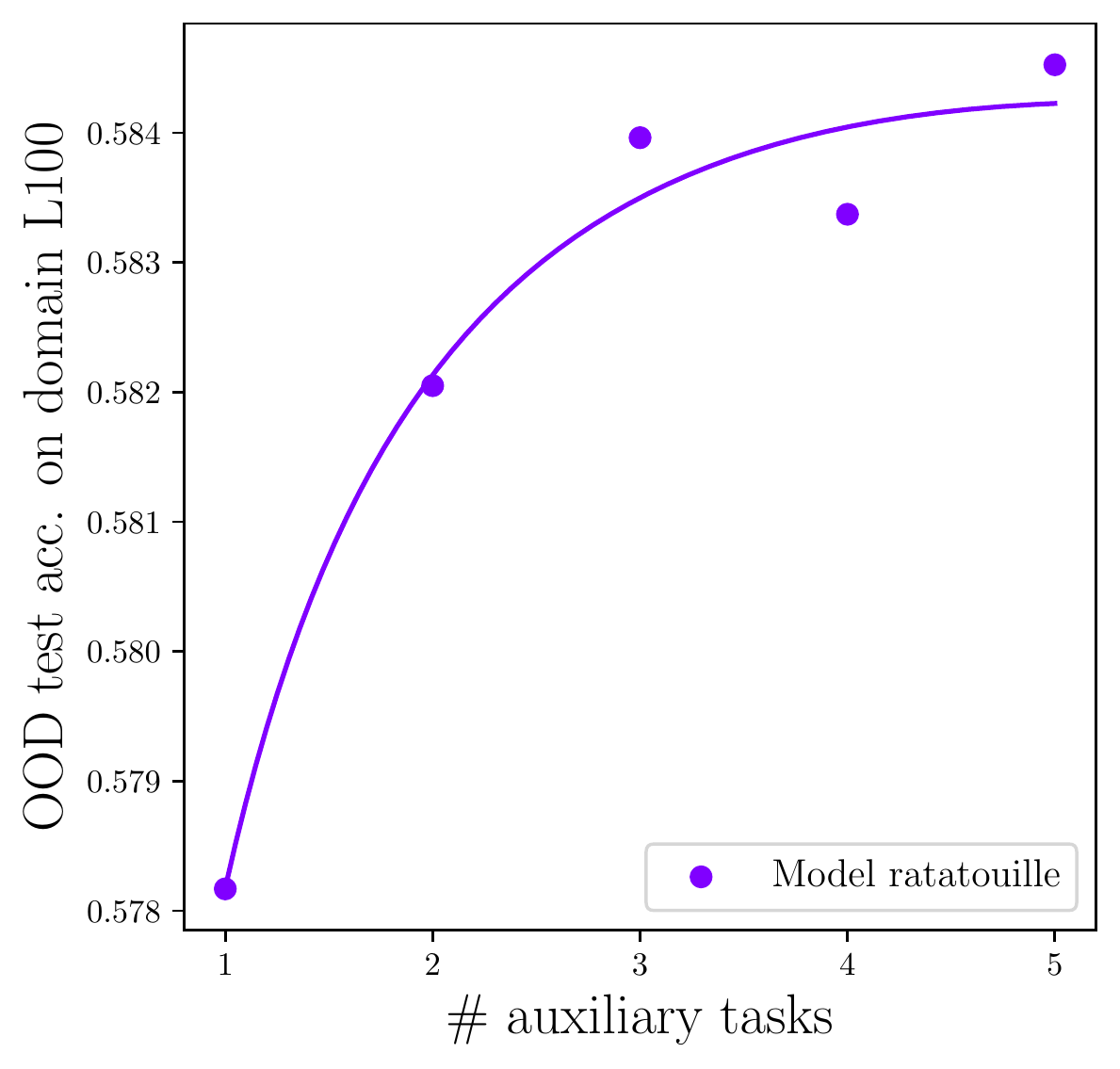}
            \caption{TerraIncognita.}
            \label{fig:countaux:d}
        \end{subfigure}
    \end{center}
    \vskip -0.4cm
    \caption{\ood accuracy ($\uparrow$) for model ratatouille when increasing the number of auxiliary tasks and uniformly averaging all fine-tuned weights. For each target task, we consider the first domain as the test \ood; the other domains are used for training.}
    \label{fig:countaux}
\end{figure}

In our \Cref{table:domainbed}, ratatouille leverages $5$ auxiliary tasks for simplicity: ImageNet (which we consider as the auxiliary task \enquote{number zero}), and the $4$ other datasets from DomainBed (out of the $5$, as we leave out the target task to prevent any information leakage). In following \Cref{fig:countaux}, we report the scores obtained using $1$ to $5$ auxiliary tasks: we always average $M=20$ weights, the only difference is how they were initialized.
When we have $1$ auxiliary task, they were all inter-trained on this auxiliary task: when we have $2$ auxiliary tasks, $10$ are inter-trained on the first auxiliary task, $10$ on the second: and etc.
This validates that a greater number of auxiliary tasks leads to an increase in expected \ood accuracy.
In the paper, we argue that this improvement is a result of the diversity gained through different specializations on different auxiliary tasks.
We expect that further increasing the number of auxiliary datasets---beyond those from DomainBed---would further improve results.
\begin{wrapfigure}[10]{hR!}{0.24\textwidth}
    \centering%
    \includegraphics[width=0.24\textwidth]{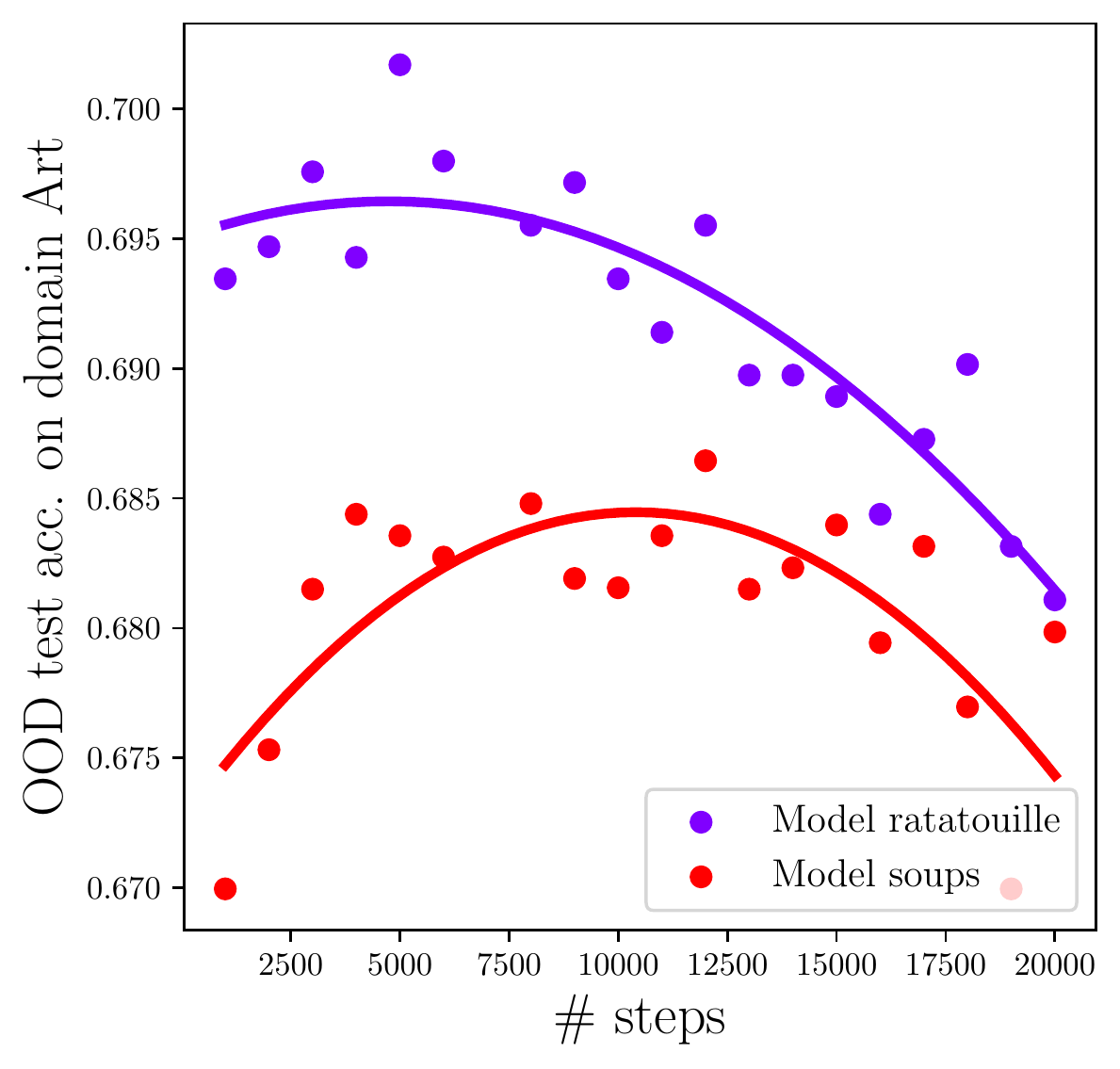}%
    \caption{OfficeHome \ood accuracy with uniform averaging at different training steps.}
    \label{fig:trainingcoststeps}
\end{wrapfigure}%
\subsection{Analysis of the Number of Target Fine-Tuning Steps}
\label{app:numbersteps}
One could argue that recycling auxiliary weights only benefit from longer training, part of which is delegated to the community.
To invalidate this hypothesis, we ablate the number of training steps for model soups and ratatouille in \Cref{fig:trainingcoststeps}, on OfficeHome with \enquote{Art} as the \ood domain.
We observe that even with unlimited number of training steps, model soups can not beat ratatouille. Therefore ratatouille's gains are made possible by fine-tuning on auxiliary datasets. We also observe that after a large number of epochs, the initialization becomes less important (as previously suggested in \Cref{fig:diversity:b,fig:rdiversity:b}) and thus model ratatouille's gain over model soups decreases.
In short, using the standard number of training steps ($5000$) provided by Domainbed is close to optimal.
\subsection{Analysis of the Number of Target Fine-Tuning Runs}
\label{app:numberruns}
\begin{figure}[h!]
    \begin{center}
        \begin{subfigure}[b]{0.24\textwidth}
            \includegraphics[width=\textwidth]{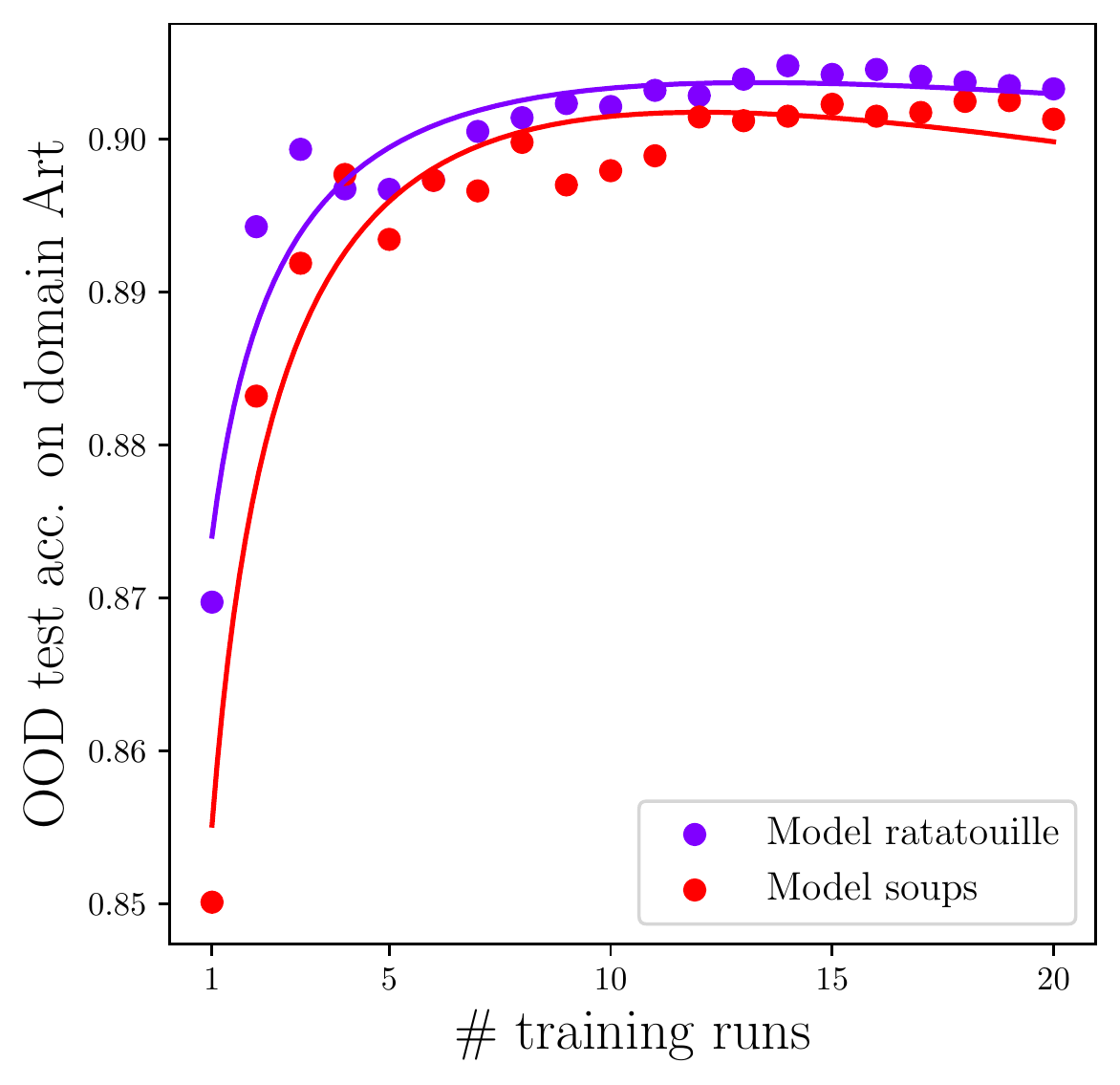}
            \caption{PACS.}
            \label{fig:trainingcost:a}
        \end{subfigure}
        \hfill
        \begin{subfigure}[b]{0.24\textwidth}
            \includegraphics[width=\textwidth]{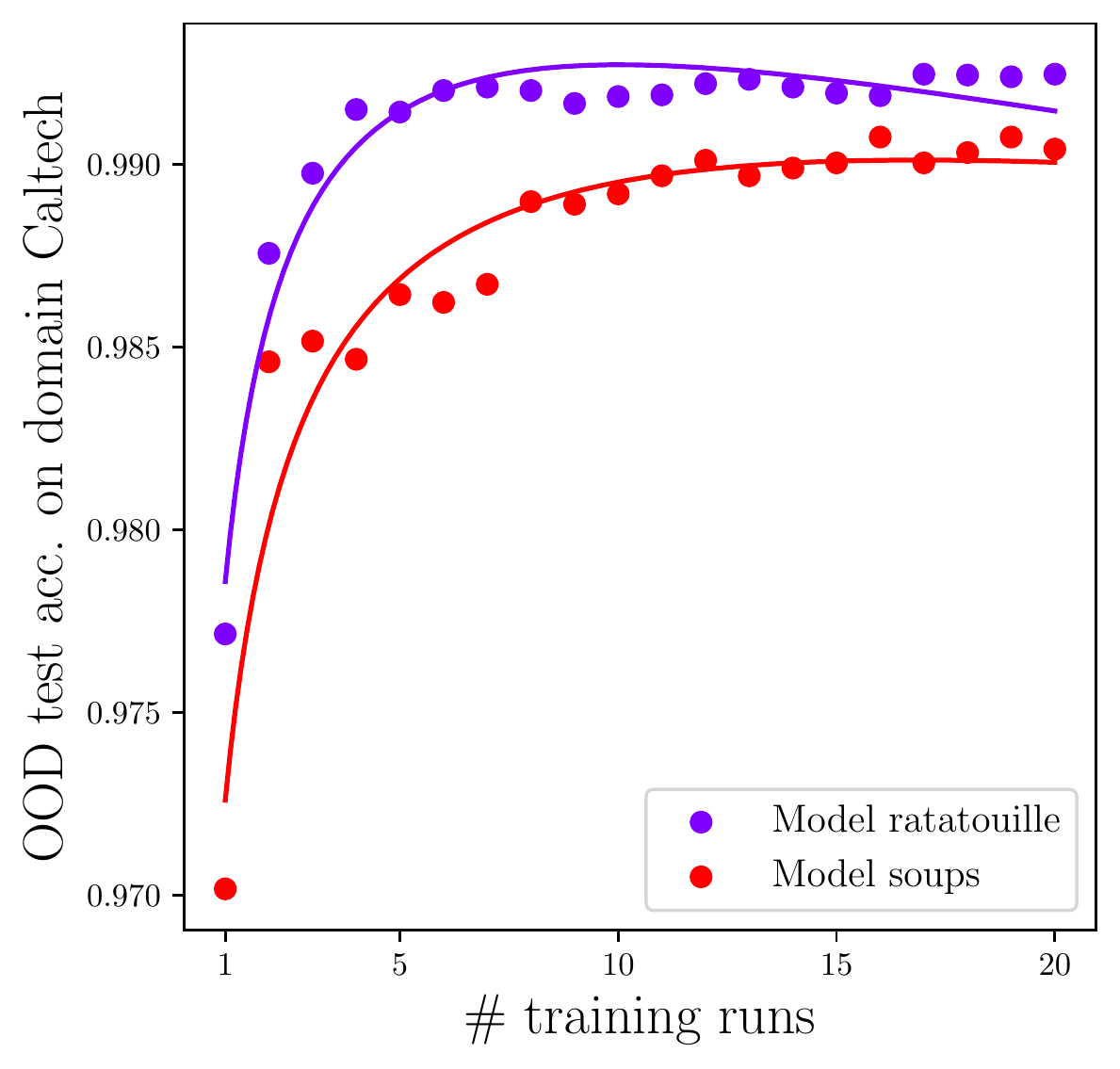}
            \caption{VLCS.}
            \label{fig:trainingcost:b}
        \end{subfigure}
        \hfill
        \begin{subfigure}[b]{0.24\textwidth}
            \includegraphics[width=\textwidth]{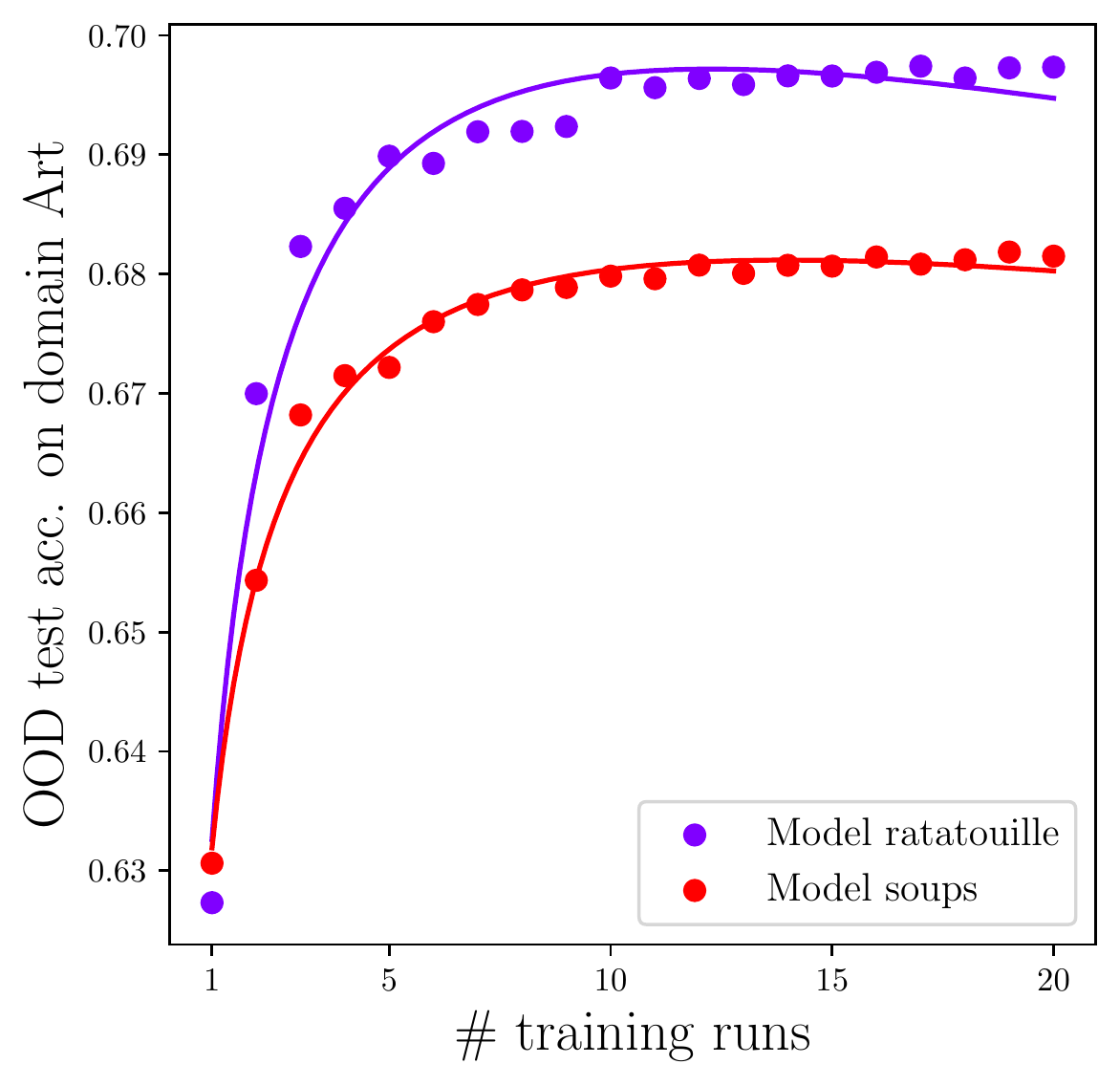}
            \caption{OfficeHome.}
            \label{fig:trainingcost:c}
        \end{subfigure}
        \hfill
        \begin{subfigure}[b]{0.24\textwidth}
            \includegraphics[width=\textwidth]{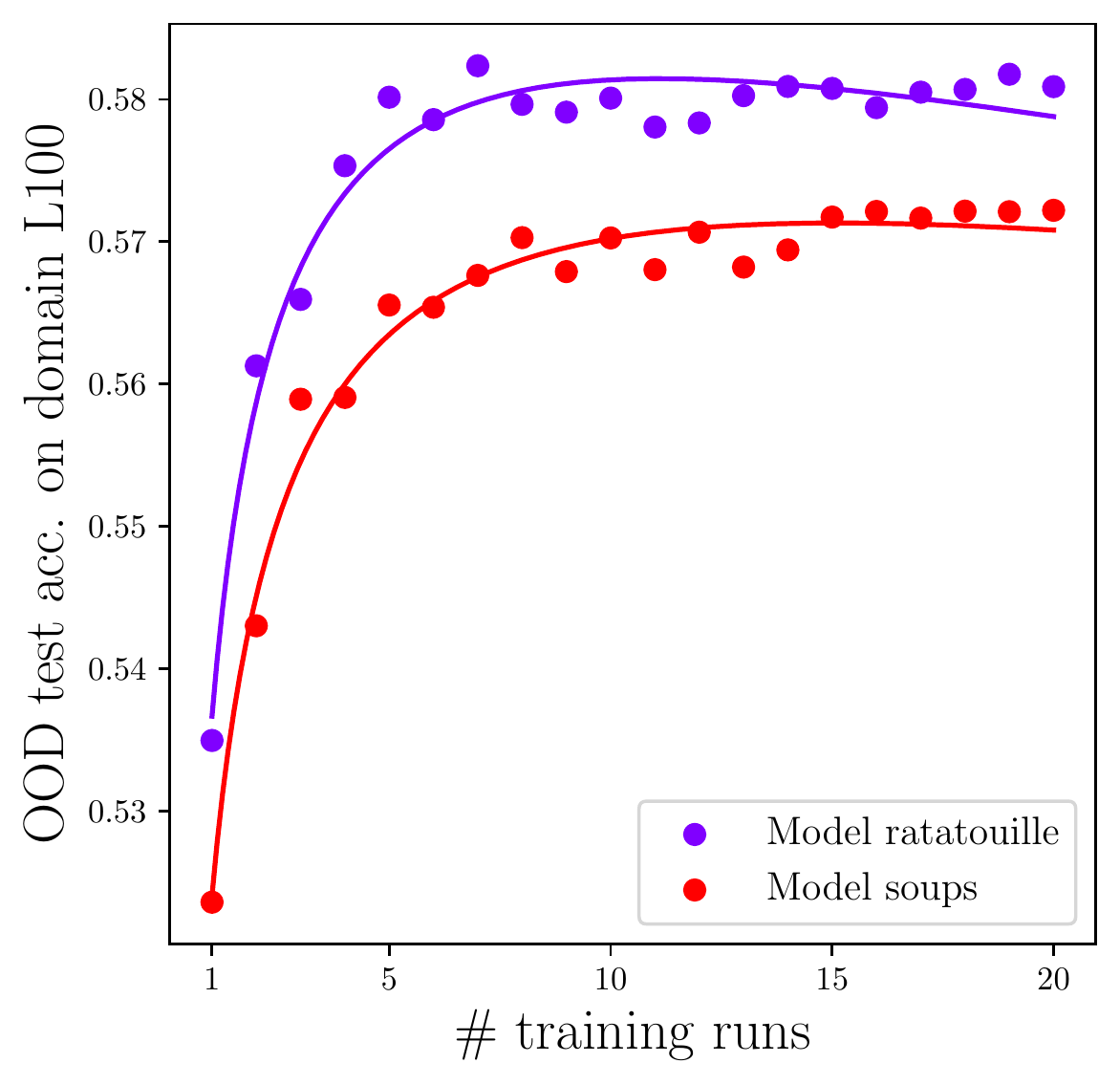}
            \caption{TerraIncognita.}
            \label{fig:trainingcost:d}
        \end{subfigure}
    \end{center}
    \caption{\ood accuracy ($\uparrow$) for model ratatouille and model soups, when increasing the number of training runs and uniformly averaging all fine-tuned weights. For each target task, we consider the first domain as the test \ood; the other domains are used for training.}
    \label{fig:trainingcost}
\end{figure}

In our main experiment from \Cref{table:domainbed}, we train and average $M=20$ independent weights, as $20$ is the standard number of hyperparameter trials in DomainBed \cite{gulrajani2021in}.
In \Cref{fig:trainingcost} we ablate this value. We observe that a larger number of runs improves performance. If reducing the training budget is critical, one could already benefit from significant gains over model soups (and vanilla fine-tuning) with only $5$ runs on the target task.

\FloatBarrier

\section{Diversity Experiments}
\label{app:diversity}

\subsection{Diversity Measures}
\label{app:diversitymeasures}

As stated in \enquote{Measures of Diversity in Classifier Ensembles and Their Relationship with the Ensemble Accuracy} \cite{kuncheva2003measures}, \enquote{measuring diversity is not straightforward because there is no generally accepted formal definition}.
In \Cref{fig:diversity}, we leverage the q-statistics $Q$, introduced in \citet{yule1900vii}, brought up to date in \citet{kuncheva2003measures} and also used in \citet{rame2021dice}.
Specifically, it is defined by $Q=\frac{N^{11} N^{00}-N^{01} N^{10}}{N^{11} N^{00}+N^{01} N^{10}}$, where $N^{ij}$ is the number of times that the first classifier is (correct if $i=1$ or wrong if $i=0$) and the second classifier is (correct if $j=1$ or wrong if $j=0$). For example, $N^{10}$ is the number of times that the first classifier is correct but not the second.
Overall, classifiers which commit errors on different objects render $Q$ small.
To transform this similarity into a diversity measure that increases for more diverse classifiers, we report $1$ minus the q-statistics, \ie the r-diversity is $1-Q$.

\begin{figure}[h!]
    \begin{center}
        \begin{subfigure}[b]{0.32\textwidth}
            \includegraphics[width=\textwidth]{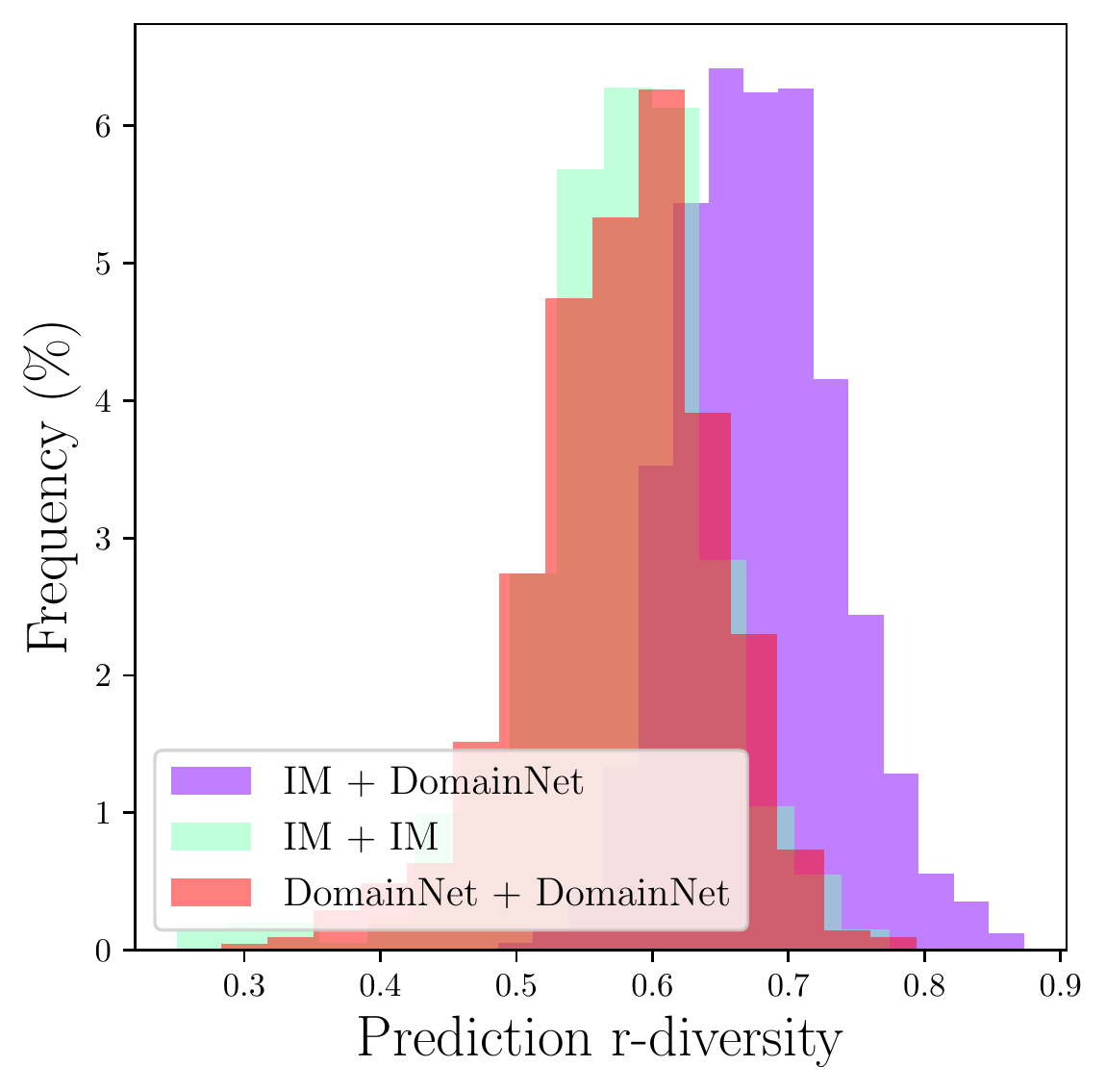}
            \caption{R-diversity frequency.}
            \label{fig:rdiversity:a}
        \end{subfigure}
        \hfill
        \begin{subfigure}[b]{0.32\textwidth}
            \includegraphics[width=\textwidth]{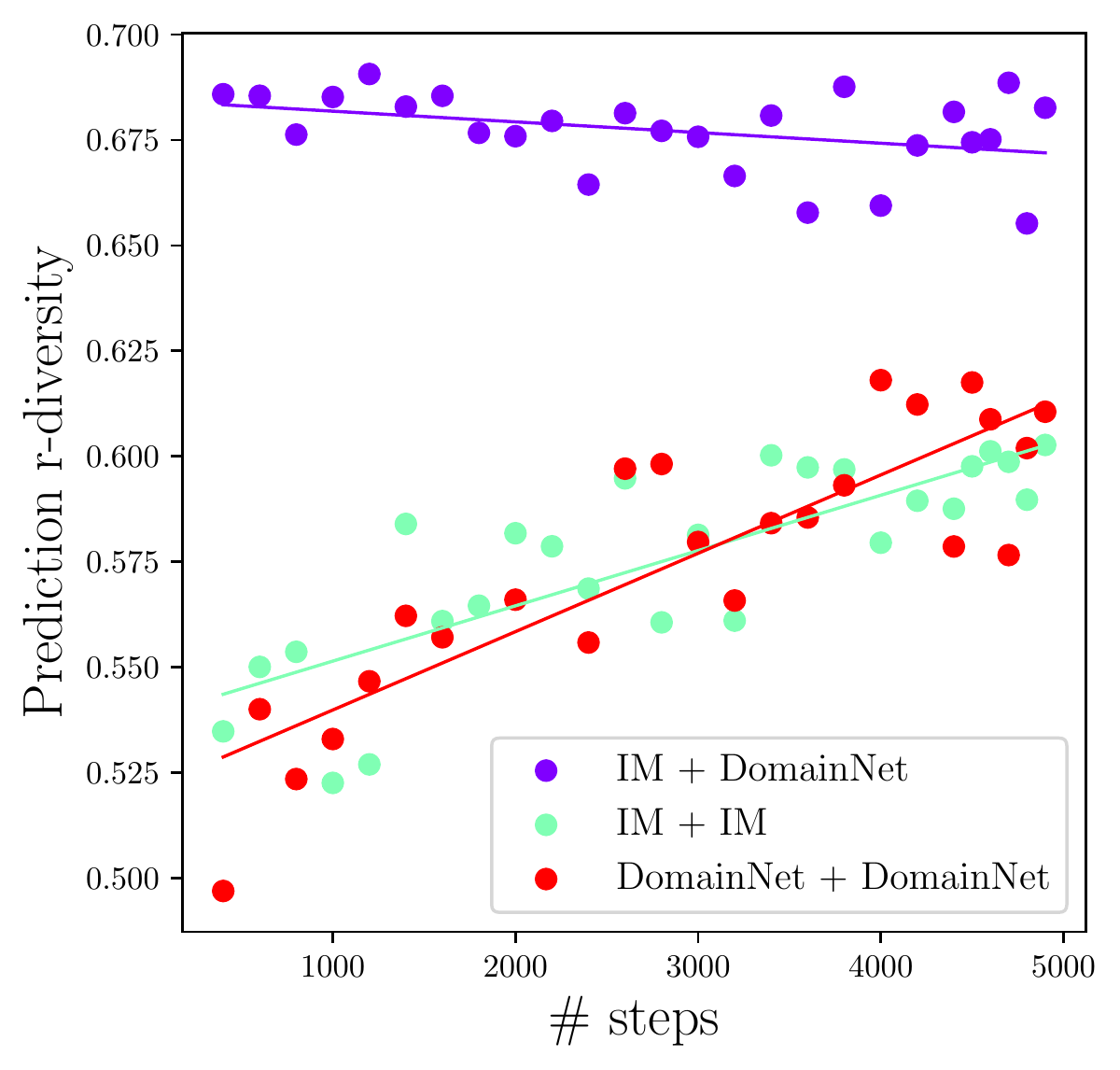}
            \caption{R-diversity vs. fine-tuning steps.}
            \label{fig:rdiversity:b}
        \end{subfigure}
        \hfill
        \begin{subfigure}[b]{0.32\textwidth}
            \includegraphics[width=\textwidth]{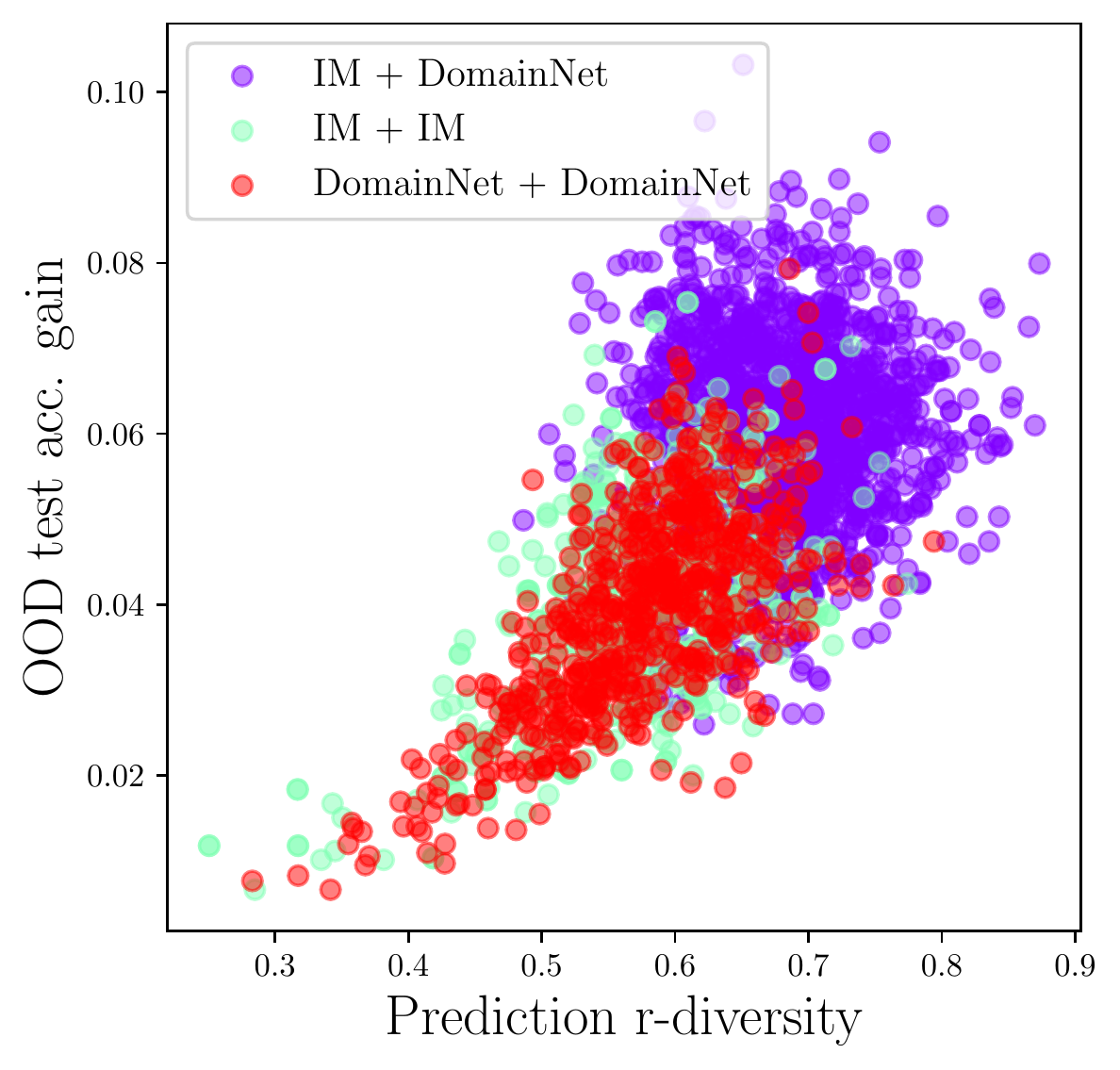}
            \caption{Acc. gain vs. r-diversity.}
            \label{fig:rdiversity:c}
        \end{subfigure}
    \end{center}
    \caption{We reproduce \Cref{fig:diversity} leveraging the ratio-error \cite{aksela2003comparison} r-diversity measure.}
    \label{fig:rdiversity}
\end{figure}

In \Cref{fig:rdiversity}, we leverage another diversity measure, the ratio-error ($\uparrow$), introduced in \citet{aksela2003comparison}, brought up to date in \citet{kuncheva2003measures} and also used in \citet{rame2021dice}.
This ratio-error 
This r-diversity leads to similar conclusions as with the q-diversity.

\subsection{Additional Diversity Results}
\label{app:diversitymore}
\begin{figure}[h!]
    \begin{center}
        \begin{subfigure}[b]{0.24\textwidth}
            \includegraphics[width=\textwidth]{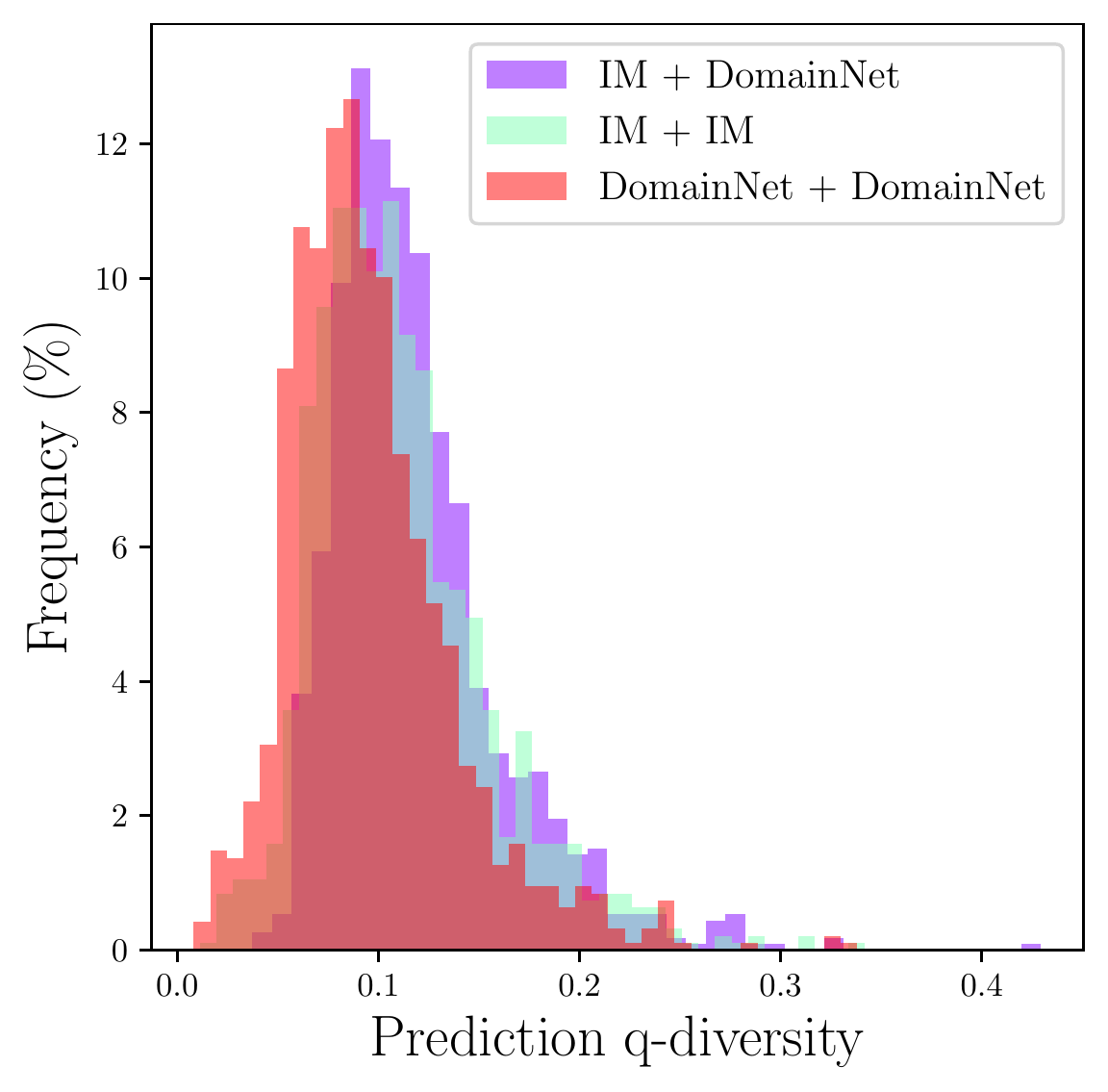}
            \caption{PACS.}
            \label{fig:divq:a}
        \end{subfigure}
        \hfill
        \begin{subfigure}[b]{0.24\textwidth}
            \includegraphics[width=\textwidth]{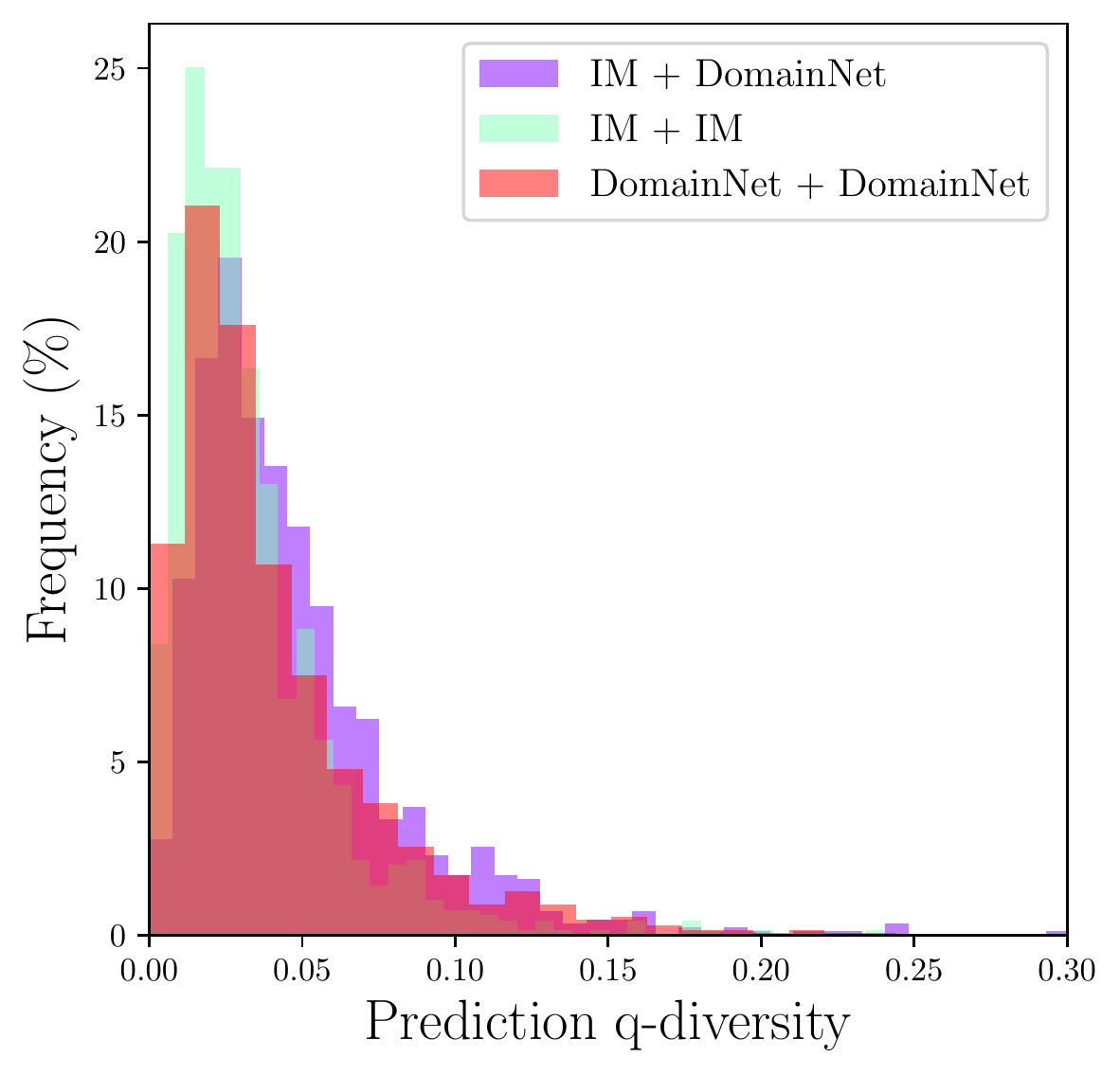}
            \caption{VLCS.}
            \label{fig:divq:b}
        \end{subfigure}
        \hfill
        \begin{subfigure}[b]{0.24\textwidth}
            \includegraphics[width=\textwidth]{images/filesdevfair/diwa/fig_dnim_hist_dq_home0.pdf}
            \caption{OfficeHome.}
            \label{fig:divq:c}
        \end{subfigure}
        \hfill
        \begin{subfigure}[b]{0.24\textwidth}
            \includegraphics[width=\textwidth]{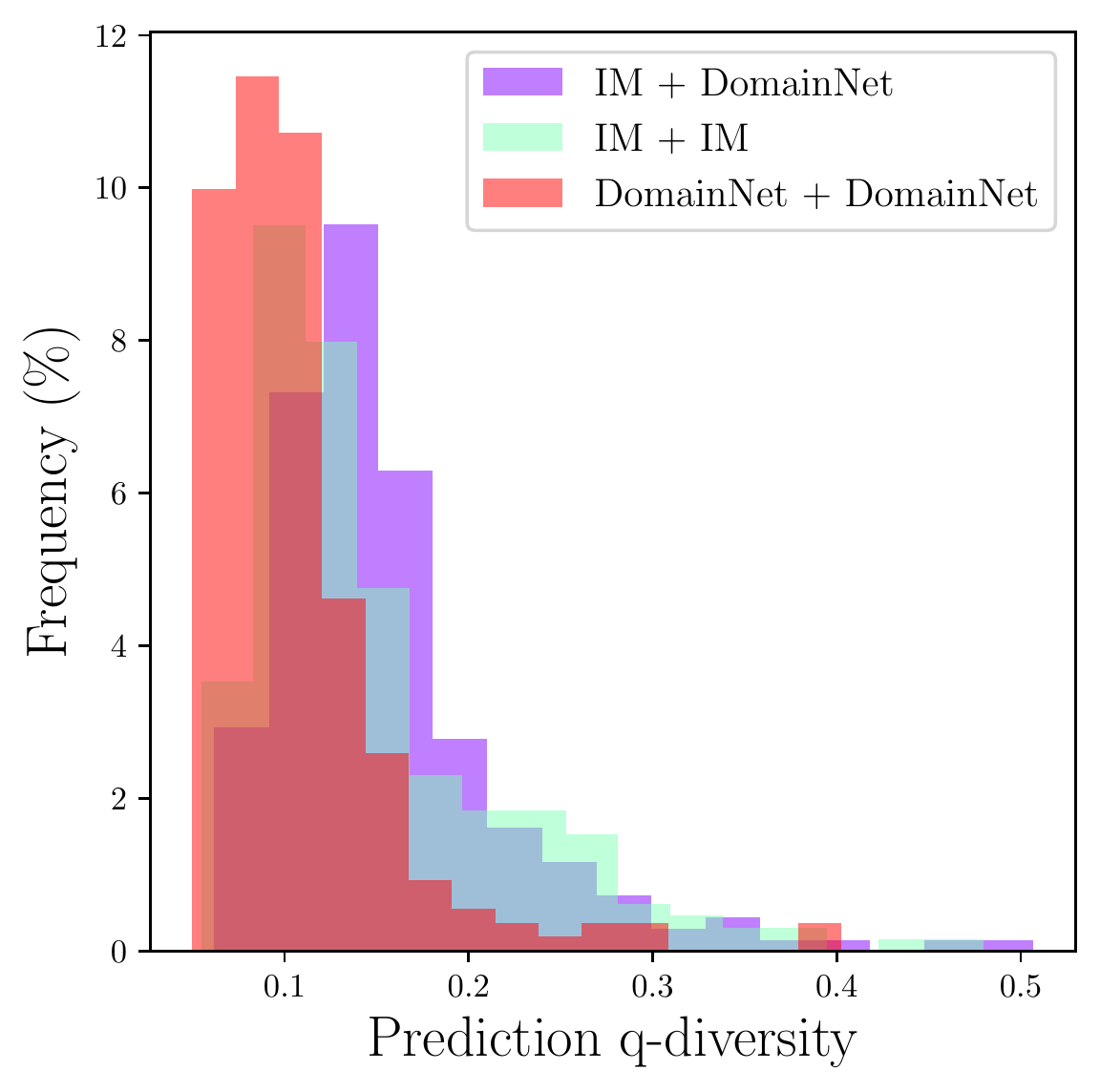}
            \caption{TerraIncognita.}
            \label{fig:divq:d}
        \end{subfigure}
    \end{center}
    \vskip -0.5cm
    \caption{Q-diversity in \ood.}
    \label{fig:divq}
\end{figure}

\begin{figure}[h!]
    \begin{center}
        \begin{subfigure}[b]{0.24\textwidth}
            \includegraphics[width=\textwidth]{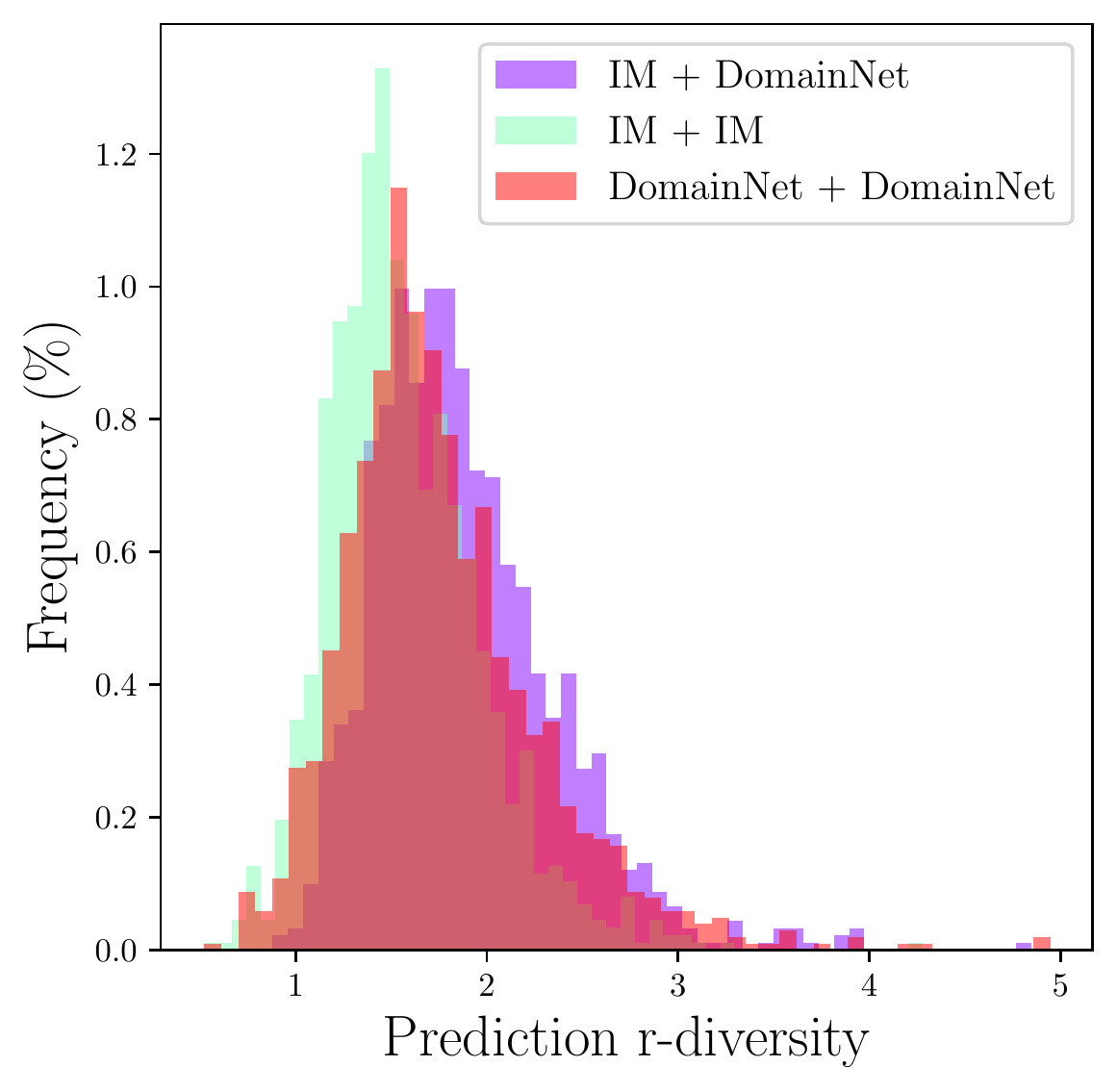}
            \caption{PACS.}
            \label{fig:divr:a}
        \end{subfigure}
        \hfill
        \begin{subfigure}[b]{0.24\textwidth}
            \includegraphics[width=\textwidth]{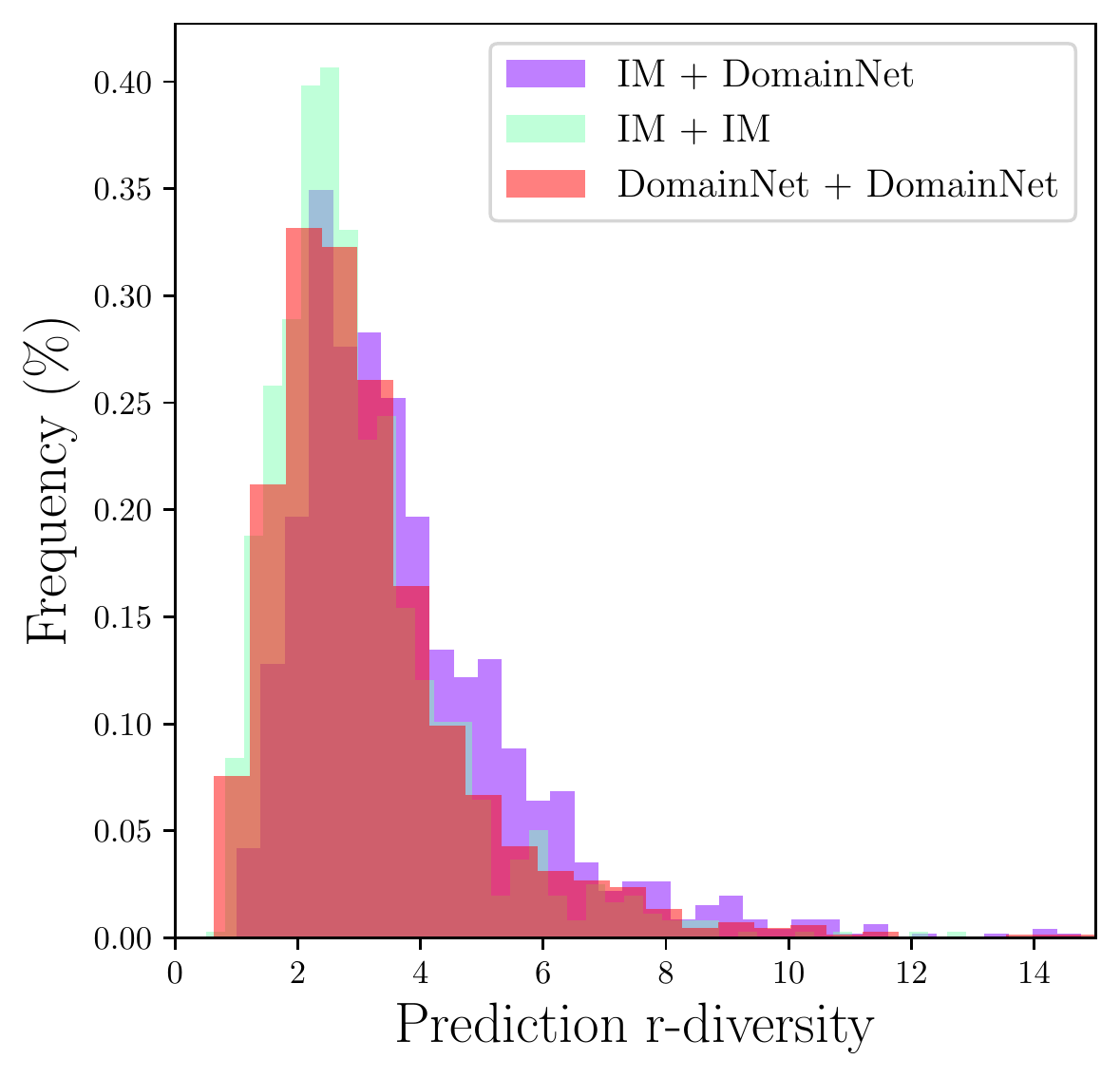}
            \caption{VLCS.}
            \label{fig:divr:b}
        \end{subfigure}
        \hfill
        \begin{subfigure}[b]{0.24\textwidth}
            \includegraphics[width=\textwidth]{images/filesdevfair/diwa/fig_dnim_hist_dr_home0.pdf}
            \caption{OfficeHome.}
            \label{fig:divr:c}
        \end{subfigure}
        \hfill
        \begin{subfigure}[b]{0.24\textwidth}
            \includegraphics[width=\textwidth]{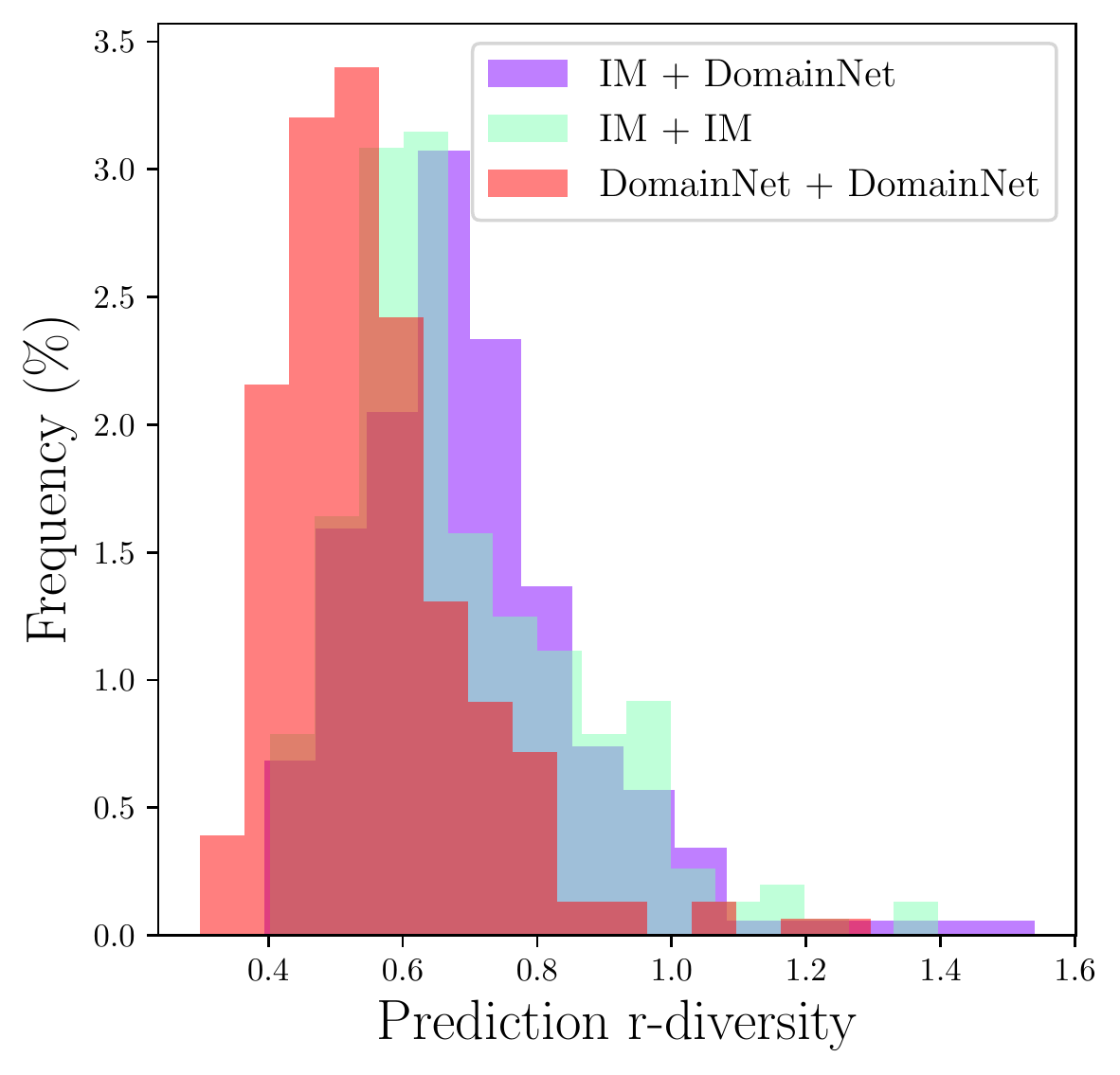}
            \caption{TerraIncognita.}
            \label{fig:divr:d}
        \end{subfigure}
    \end{center}
    \vskip -0.5cm
    \caption{R-diversity in \ood.}
    \label{fig:divr}
\end{figure}

We now apply our diversity analysis on other DomainBed's datasets, where we consider the first domain as the test \ood; the other domains are used for training.
Then, we compare the diversity---either measured with the q-statistics (in \Cref{fig:divq}) or in ratio-error (in \Cref{fig:divr})---between two networks, either both directly transferred from ImageNet, either both inter-trained on DomainNet, either one directly transferred from ImageNet and the other inter-trained on DomainNet.
Across all plots, we consistently observe that having different initializatons increases diversity, with the most pronounced shift seen on the OfficeHome and TerraIncognita datasets.

\FloatBarrier
\section{Linear Mode Connectivity Experiments}
\label{app:lmc}
\subsection{Linear Mode Connectivity per Target Dataset and Domain}
\begin{figure}[h!]
    \begin{center}
        \begin{subfigure}{.24\textwidth}
            \centering
            \includegraphics[width=\textwidth]{images/filesdevfair/lmc/vlcs0_lmc_hyp2_ood.pdf}
            \caption{\enquote{Caltech101} as test.}
        \end{subfigure}%
        \begin{subfigure}{.24\textwidth}
            \centering
            \includegraphics[width=\textwidth]{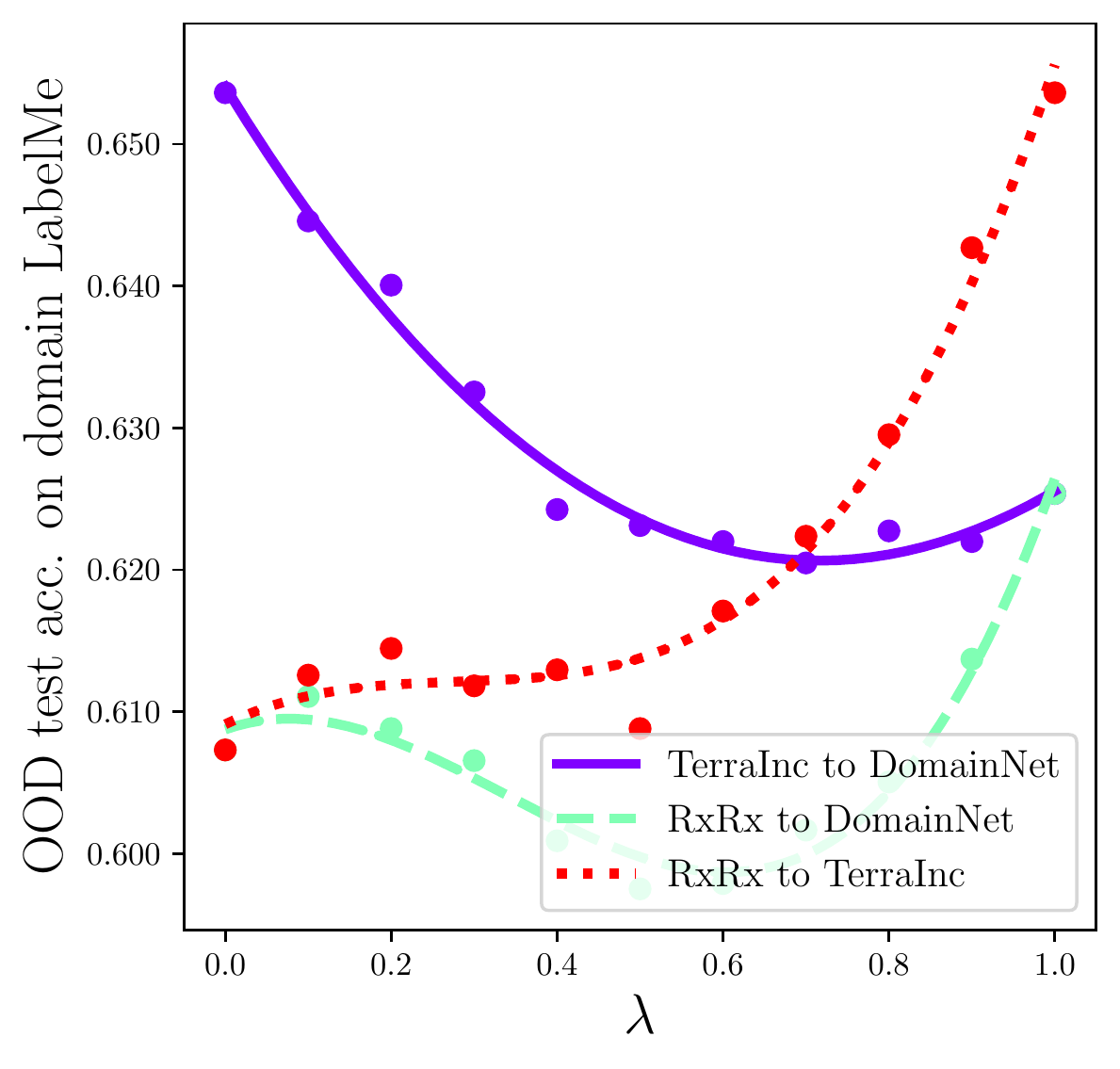}
            \caption{\enquote{LabelMe} as test.}
            \label{fig:vlcs1_lmc_hyp2_ood}
        \end{subfigure}
        \begin{subfigure}{.24\textwidth}
            \centering
            \includegraphics[width=\textwidth]{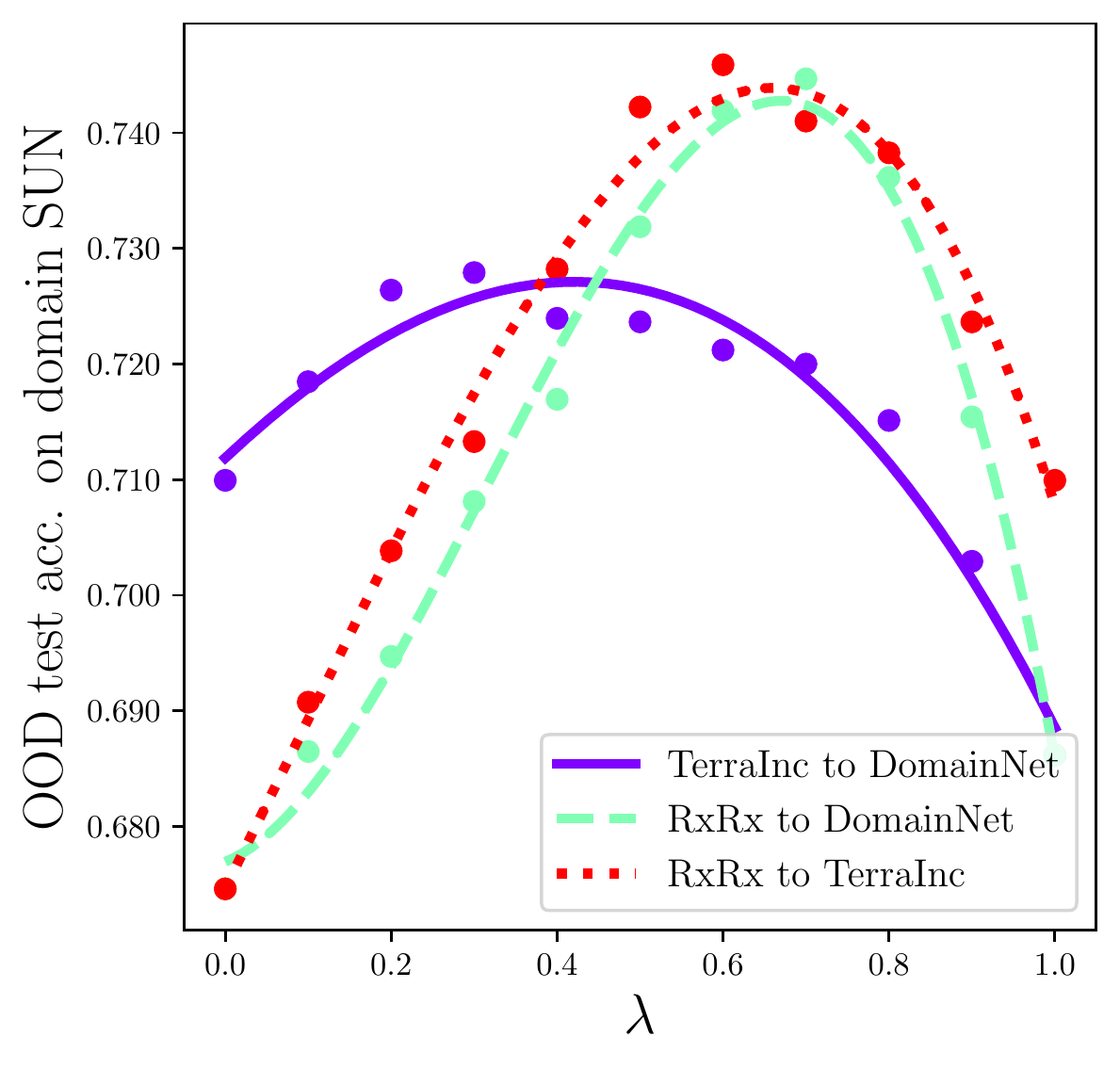}
            \caption{\enquote{SUN09} as test.}
        \end{subfigure}
        \begin{subfigure}{.24\textwidth}
            \centering
            \includegraphics[width=\textwidth]{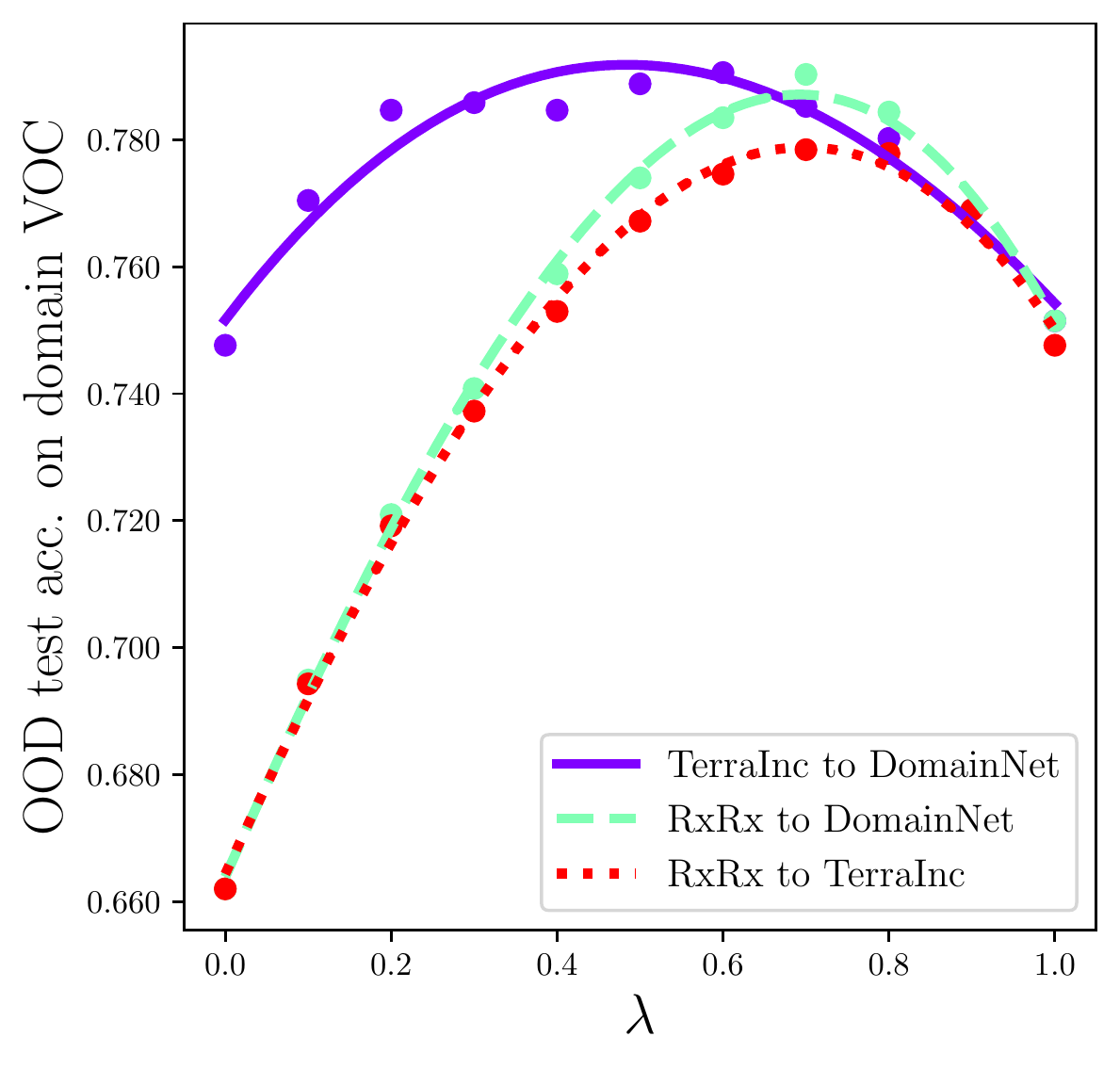}
            \caption{\enquote{VOC2007} as test.}
        \end{subfigure}
    \end{center}
    \vskip -0.5cm
    \caption{Empirical analysis of \Cref{hyp:2} on VLCS.}%
    \label{fig:vlcs_lmc_hyp2_ood}
\end{figure}

We further empirically analyze our \Cref{hyp:2}.
In particular, we observe that the LMC usually holds except in two cases: (i) when the \ood test domain is the \enquote{LabelMe} domain from VLCS (in \Cref{fig:vlcs1_lmc_hyp2_ood}), and (ii) when both the target and the auxiliary tasks are distant from the pre-trained task, for example when tackling TerraIncognita or Camelyon with RxRx as an auxiliary task.
We want to emphasize that we selected the \enquote{extreme} RxRx dataset precisely to test the empirical limits of the \Cref{hyp:2}, but that, in practice, milder auxiliary tasks selection already helps for \ood generalization (and notably reaches SoTA performance in \Cref{sec:exps:ood}).
\FloatBarrier
\begin{figure}[h!]
    \begin{center}
        \begin{subfigure}{.24\textwidth}
            \centering
            \includegraphics[width=\textwidth]{images/filesdevfair/lmc/pacs0_lmc_hyp2_ood.pdf}
            \caption{\enquote{Art} as test.}
        \end{subfigure}%
        \begin{subfigure}{.24\textwidth}
            \centering
            \includegraphics[width=\textwidth]{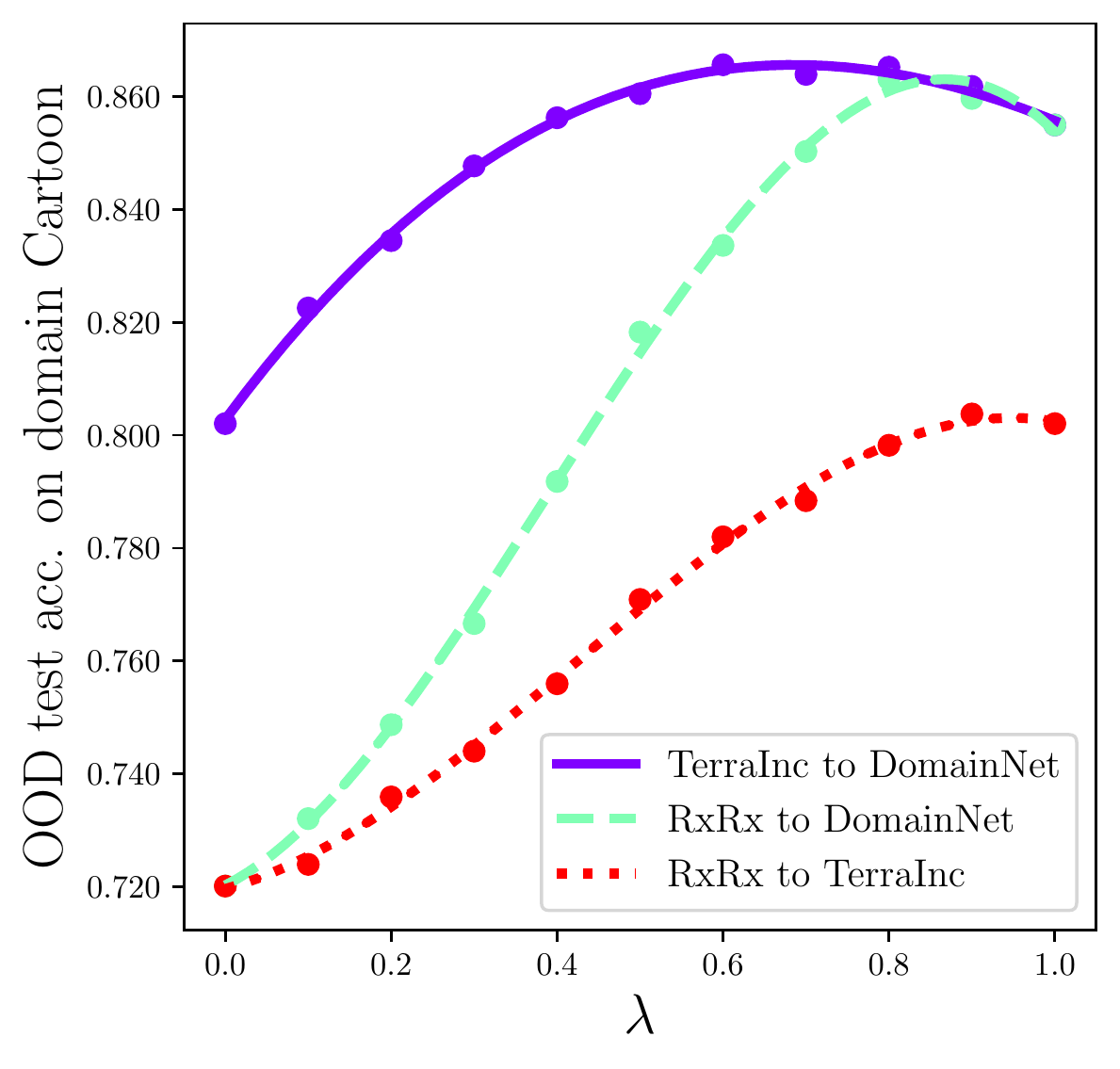}
            \caption{\enquote{Cartoon} as test.}
        \end{subfigure}
        \begin{subfigure}{.24\textwidth}
            \centering
            \includegraphics[width=\textwidth]{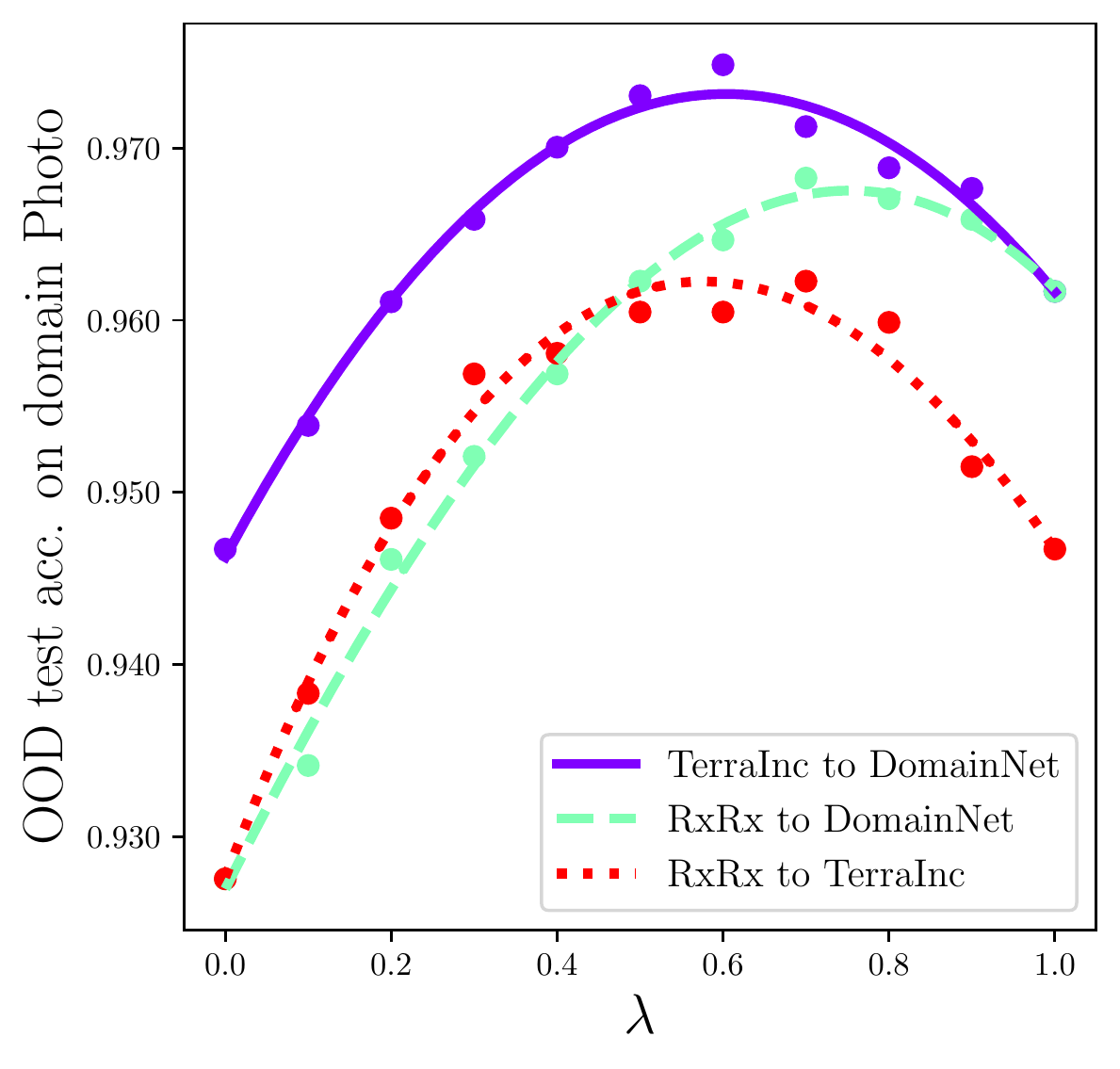}
            \caption{\enquote{Photo} as test.}
        \end{subfigure}
        \begin{subfigure}{.24\textwidth}
            \centering
            \includegraphics[width=\textwidth]{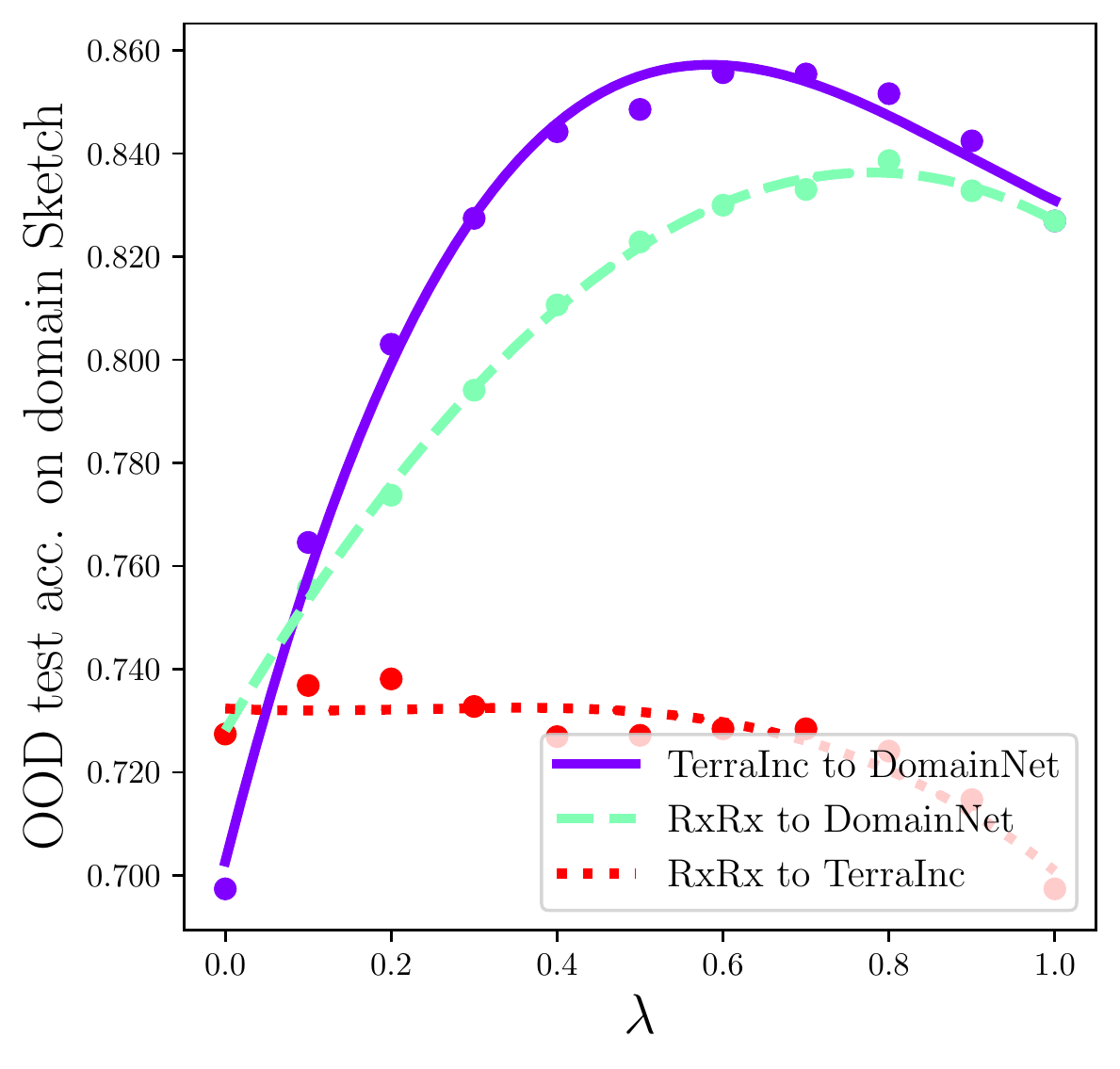}
            \caption{\enquote{Sketch} as test.}
        \end{subfigure}
    \end{center}
    \vskip -0.5cm
    \caption{Empirical analysis of \Cref{hyp:2} on PACS.}%
    \label{fig:pacs_lmc_hyp2_ood}
\end{figure}

\begin{figure}[h!]
    \begin{center}
        \begin{subfigure}{.24\textwidth}
            \centering
            \includegraphics[width=\textwidth]{images/filesdevfair/lmc/home0_lmc_hyp2_ood.pdf}
            \caption{\enquote{Art} as test.}
        \end{subfigure}%
        \begin{subfigure}{.24\textwidth}
            \centering
            \includegraphics[width=\textwidth]{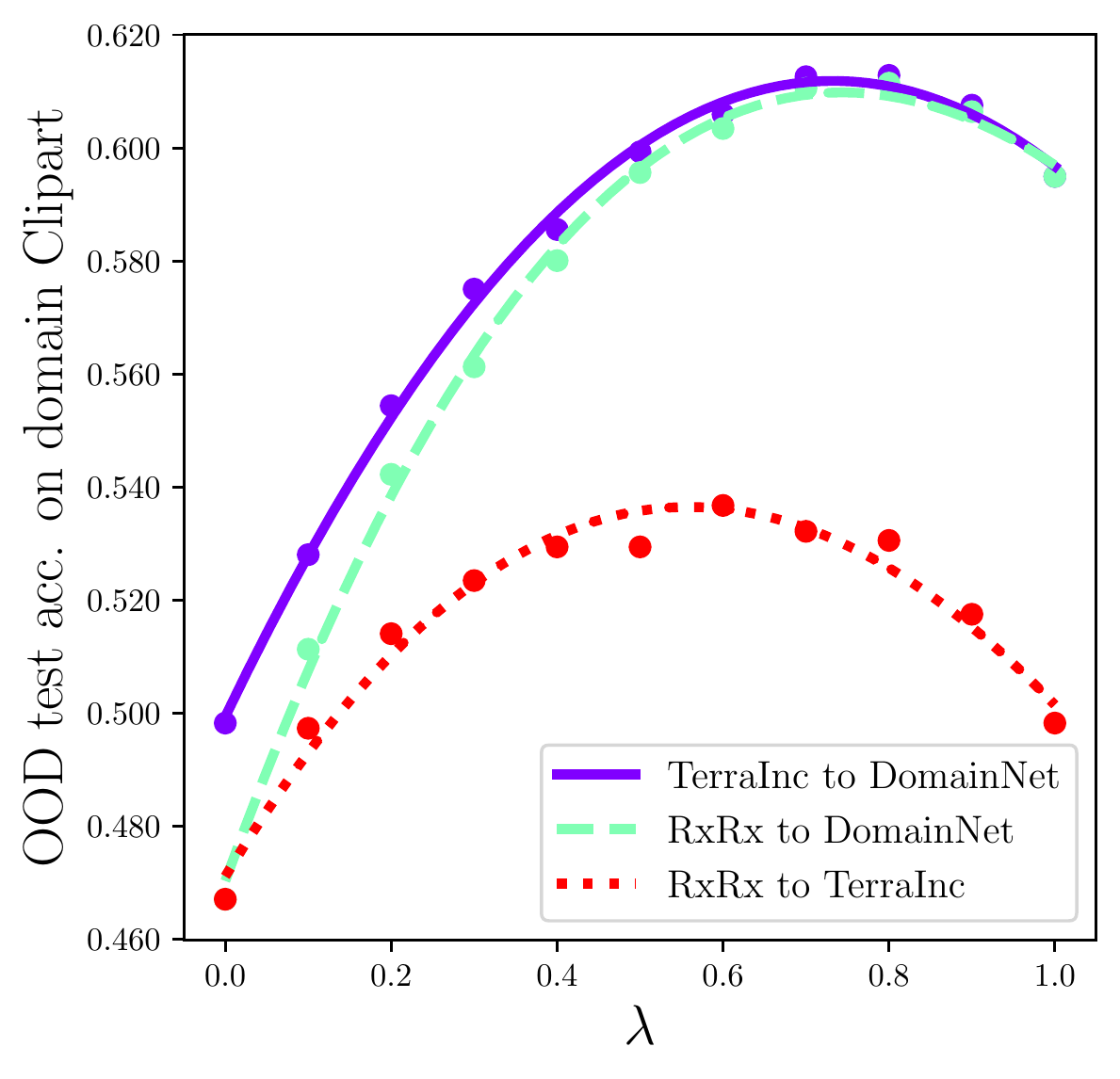}
            \caption{\enquote{Clipart} as test.}
        \end{subfigure}
        \begin{subfigure}{.24\textwidth}
            \centering
            \includegraphics[width=\textwidth]{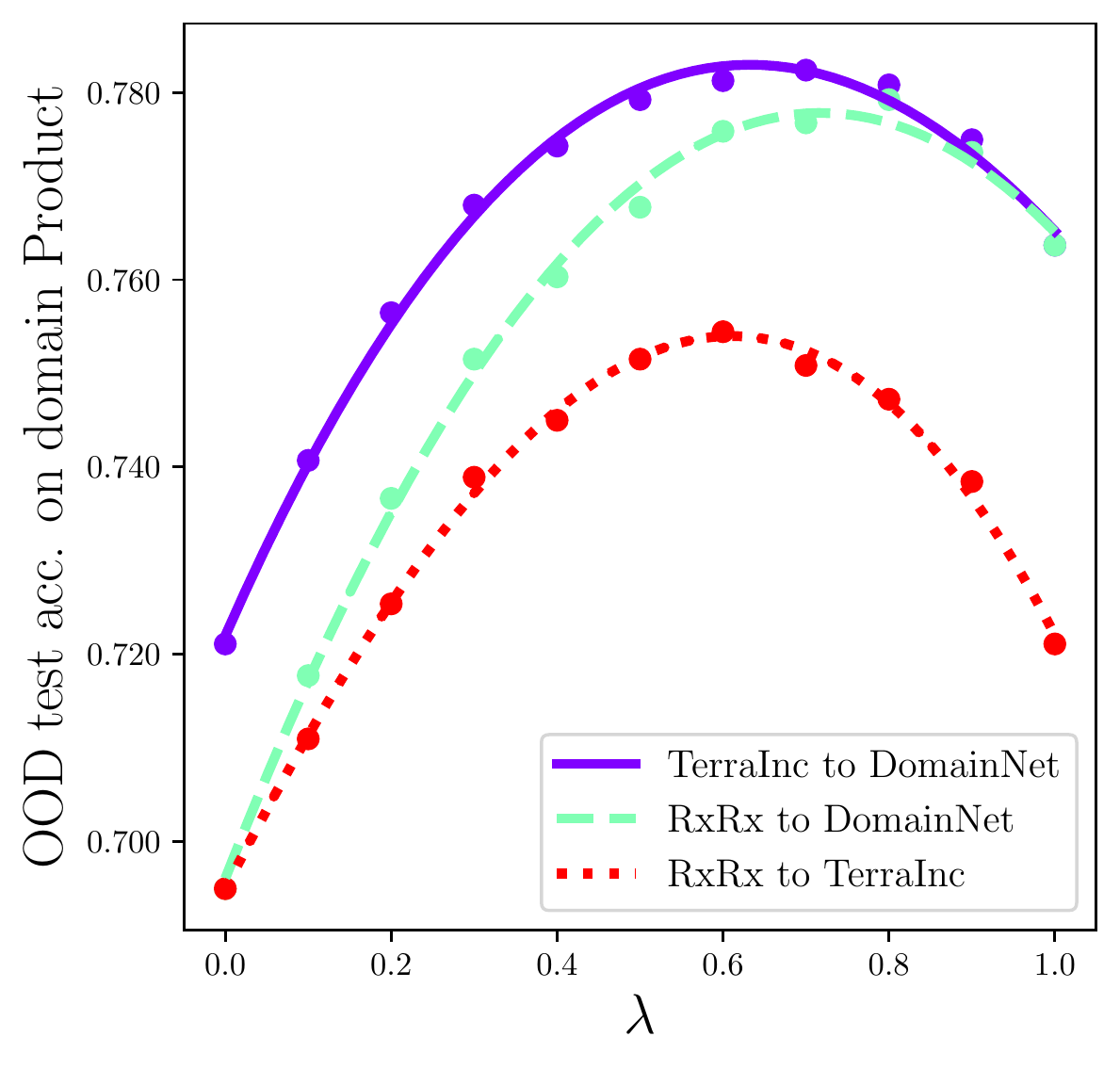}
            \caption{\enquote{Product} as test.}
        \end{subfigure}
        \begin{subfigure}{.24\textwidth}
            \centering
            \includegraphics[width=\textwidth]{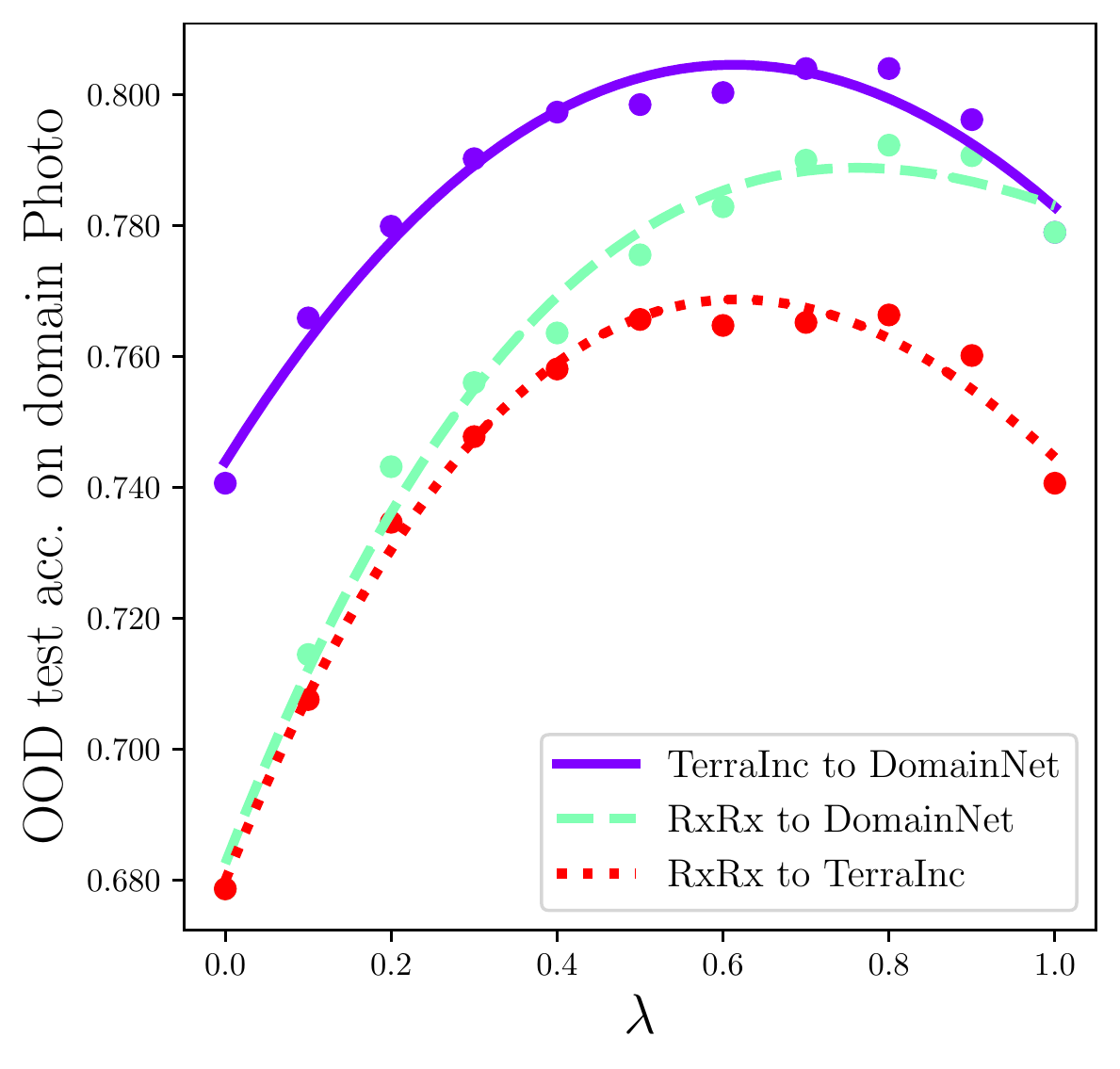}
            \caption{\enquote{Photo} as test.}
        \end{subfigure}
    \end{center}
    \vskip -0.5cm
    \caption{Empirical analysis of \Cref{hyp:2} on OfficeHome.}%
    \label{fig:home_lmc_hyp2_ood}
\end{figure}

\begin{figure}[h!]
    \begin{center}
        \begin{subfigure}{.24\textwidth}
            \centering
            \includegraphics[width=\textwidth]{images/filesdevfair/lmc/terra0_lmc_hyp2_ood.pdf}
            \caption{\enquote{L100} as test.}
        \end{subfigure}%
        \begin{subfigure}{.24\textwidth}
            \centering
            \includegraphics[width=\textwidth]{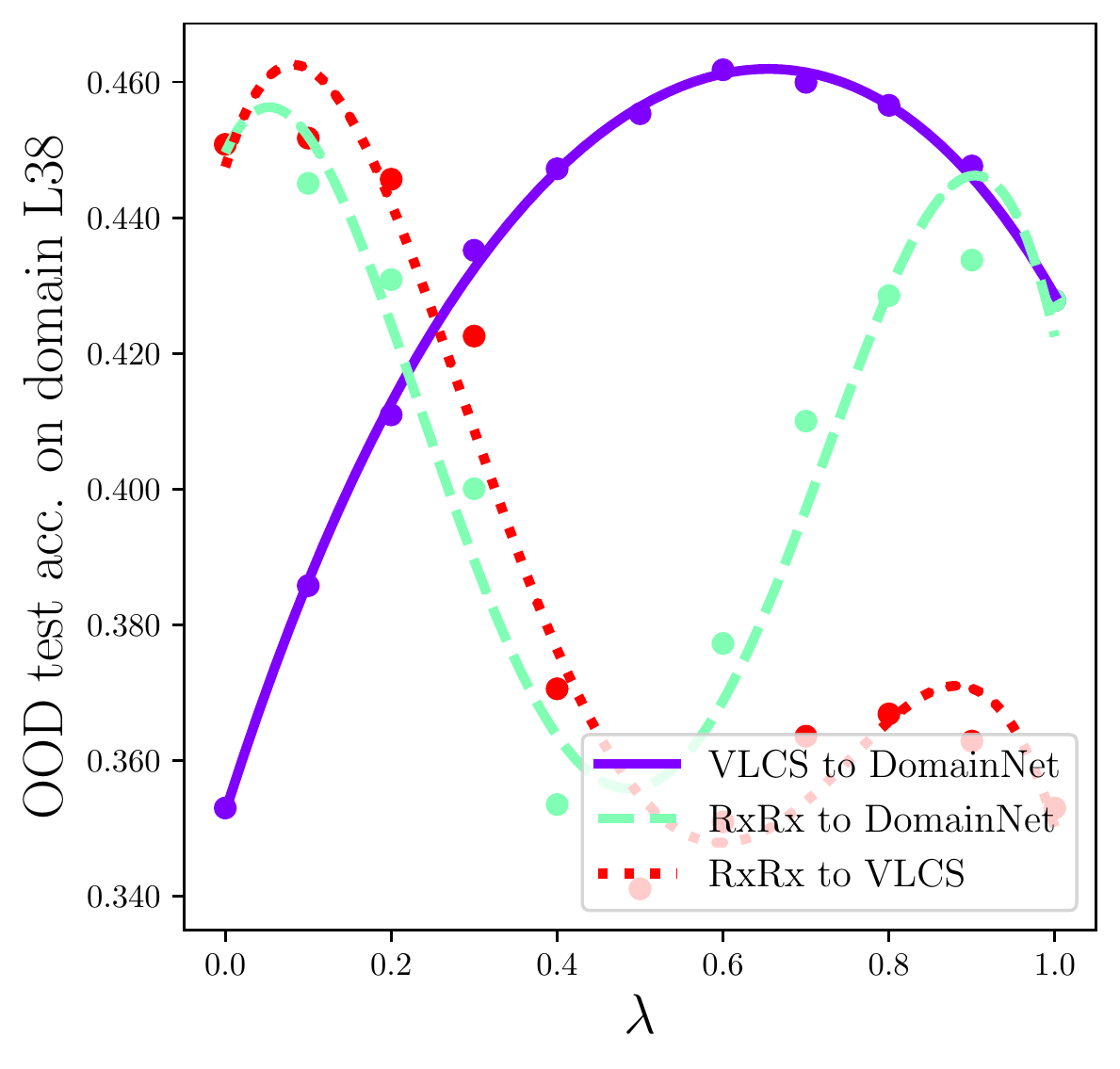}
            \caption{\enquote{L38} as test.}
        \end{subfigure}
        \begin{subfigure}{.24\textwidth}
            \centering
            \includegraphics[width=\textwidth]{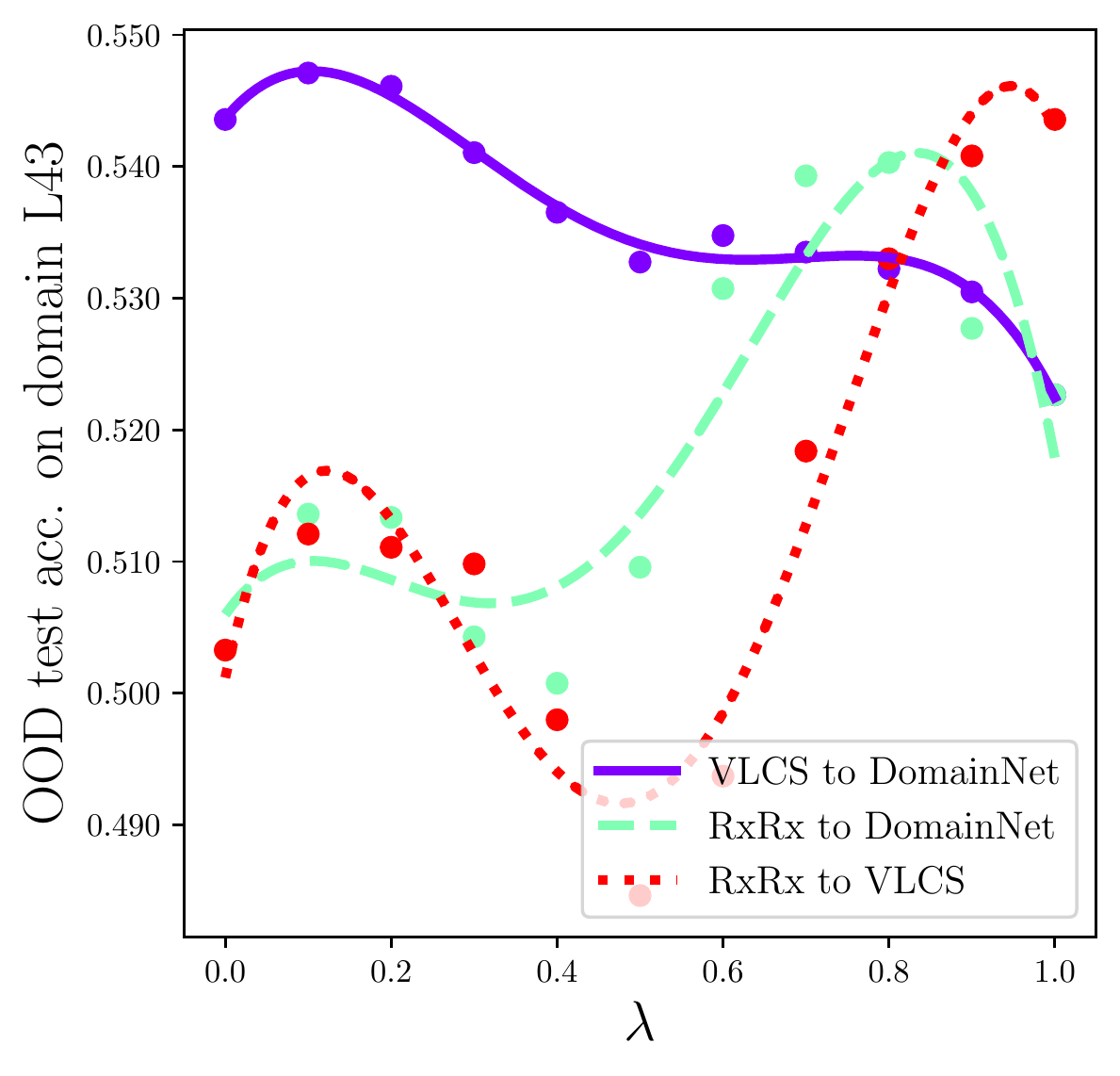}
            \caption{\enquote{L43} as test.}
        \end{subfigure}
        \begin{subfigure}{.24\textwidth}
            \centering
            \includegraphics[width=\textwidth]{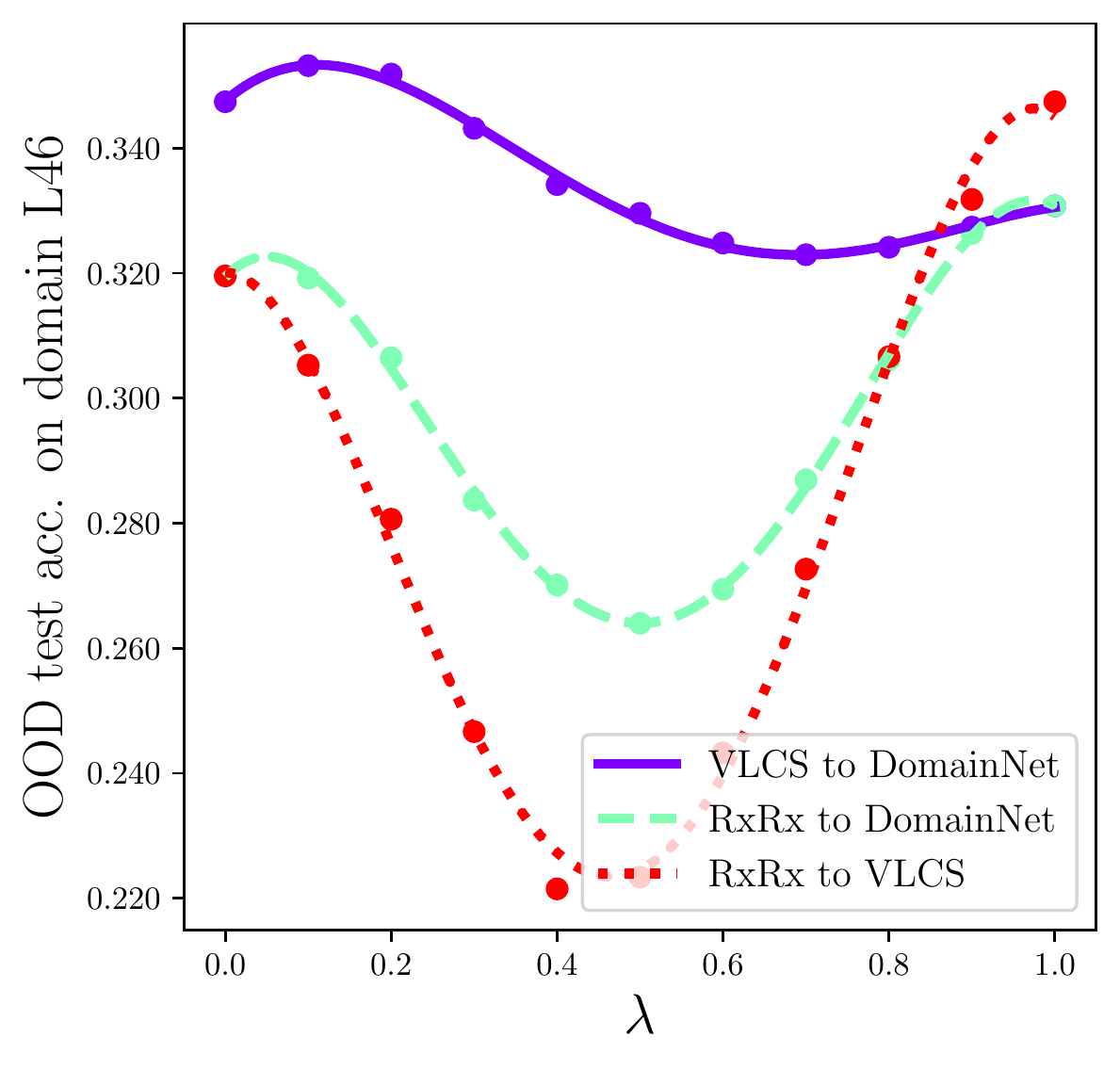}
            \caption{\enquote{L46} as test.}
        \end{subfigure}
    \end{center}
    \vskip -0.5cm
    \caption{Empirical analysis of \Cref{hyp:2} on TerraIncognita.}%
    \label{fig:terra_lmc_hyp2_ood}
\end{figure}

\FloatBarrier

\begin{figure}[h!]
    \centering
    \begin{subfigure}{.24\textwidth}
        \centering
        \includegraphics[width=.95\linewidth,left]{images/filesdevfair/lmc/came0_lmc_hyp2_ood.pdf}
        \caption{\enquote{Hospital 0} as test.}
    \end{subfigure}%
    \begin{subfigure}{.24\textwidth}
        \centering
        \includegraphics[width=.95\linewidth]{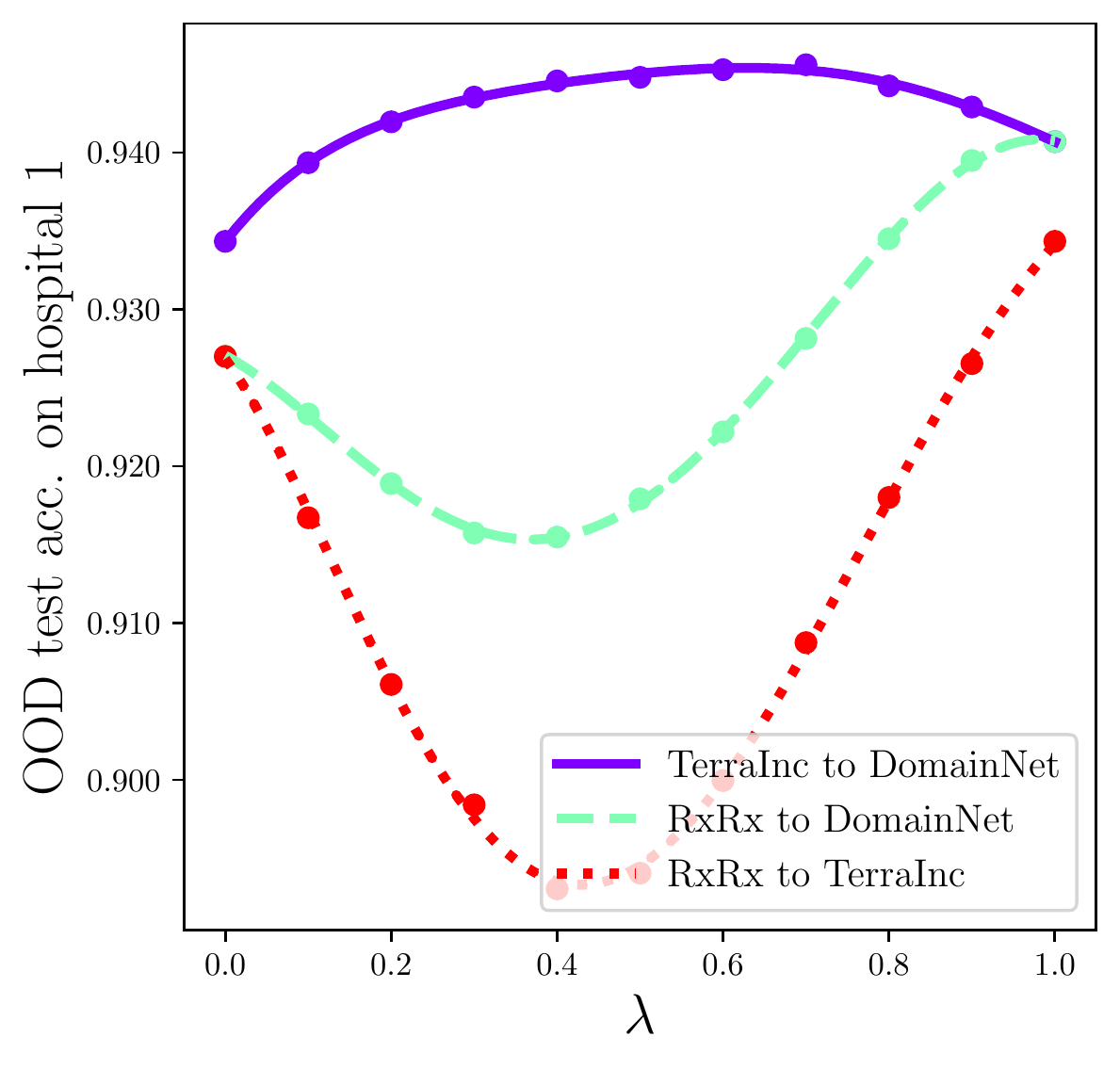}
        \caption{\enquote{Hospital 1} as test.}
    \end{subfigure}
    \begin{subfigure}{.24\textwidth}
        \centering
        \includegraphics[width=.95\linewidth]{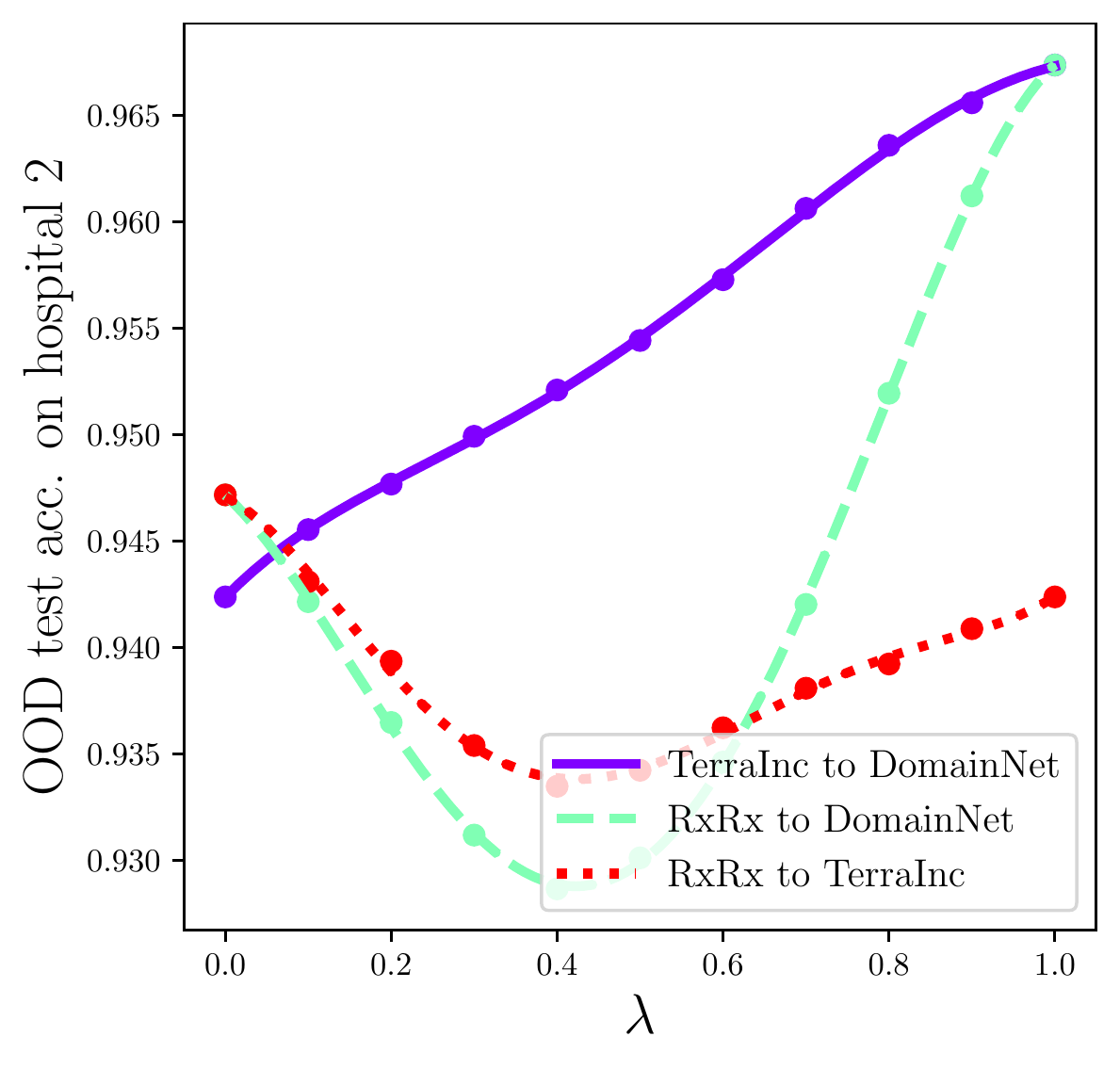}
        \caption{\enquote{Hospital 2} as test.}
    \end{subfigure}
    \begin{subfigure}{.24\textwidth}
        \centering
        \includegraphics[width=.95\linewidth]{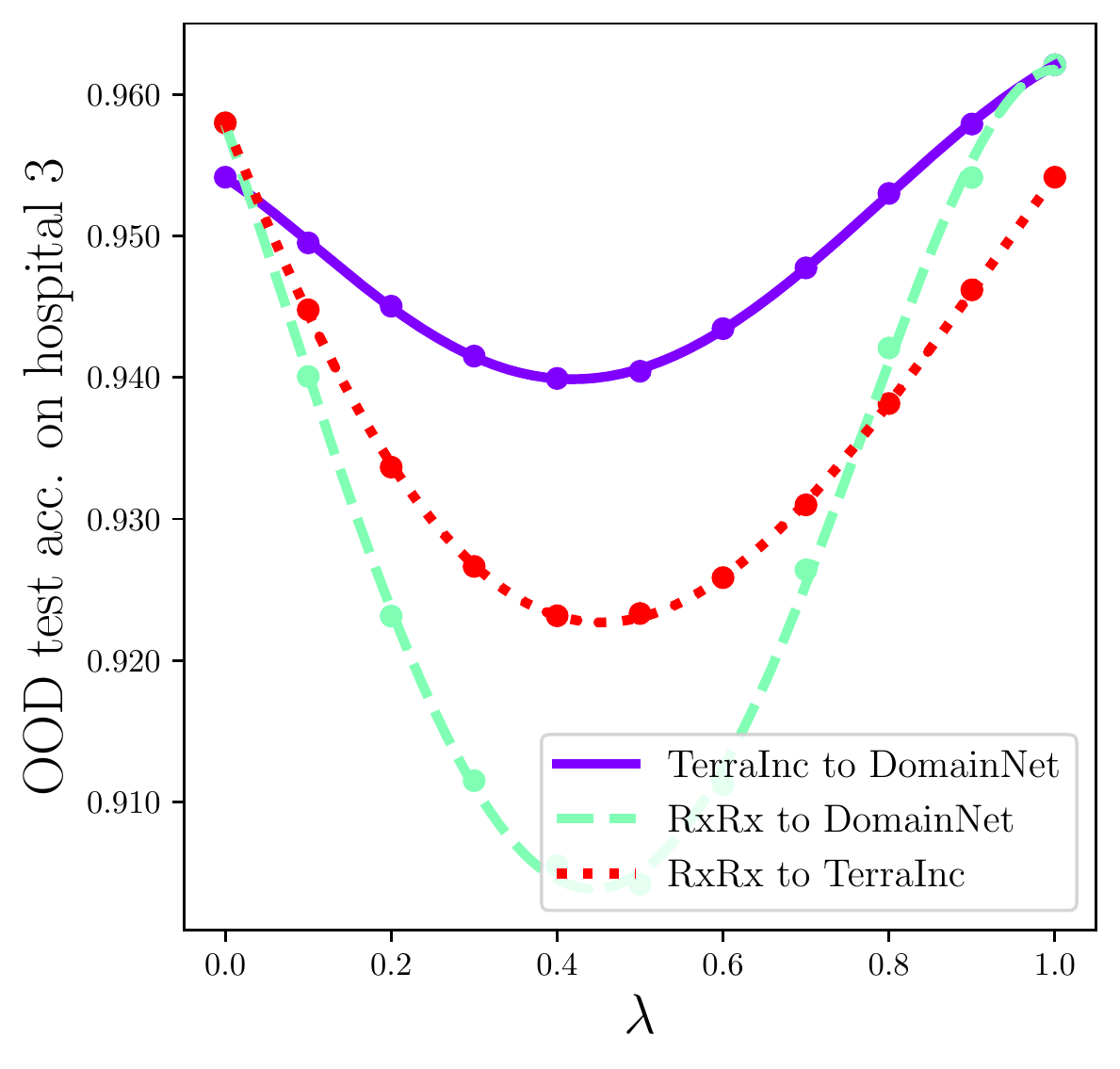}
        \caption{\enquote{Hospital 3} as test.}
    \end{subfigure}
    \caption{Empirical analysis of \Cref{hyp:2} on Camelyon.}%
    \label{fig:came_lmc_hyp2_ood}
\end{figure}

\FloatBarrier
\subsection{Linear Mode Connectivity across Three Weights}
\begin{figure}[h!]
    \begin{center}
        \begin{subfigure}{.19\textwidth}
            \centering
            \includegraphics[width=.95\linewidth]{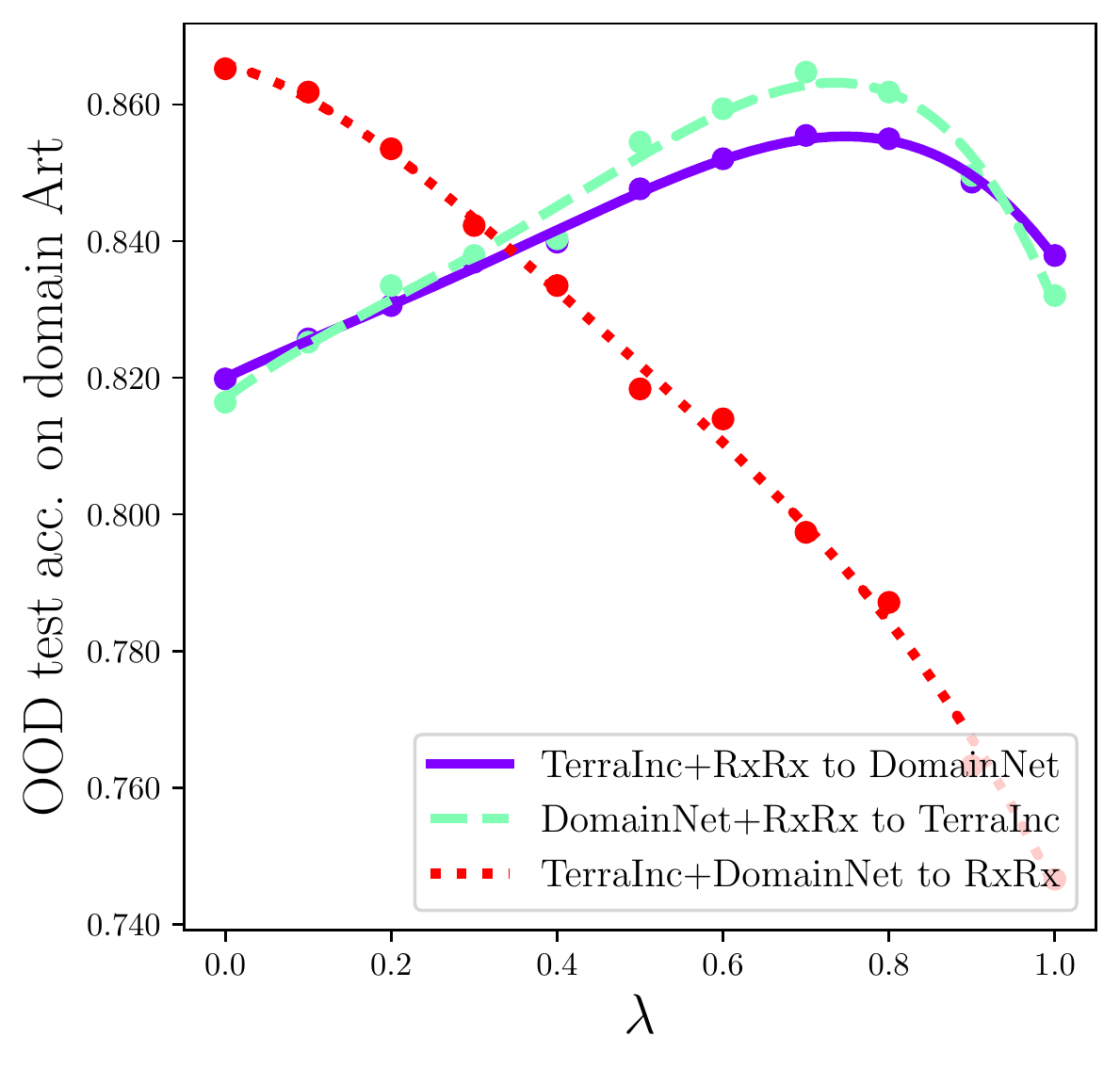}
            \caption{PACS.}
            \label{fig:pacs0_lmc_hyp3_ood}
        \end{subfigure}
        \begin{subfigure}{.19\textwidth}
            \centering
            \includegraphics[width=.95\linewidth]{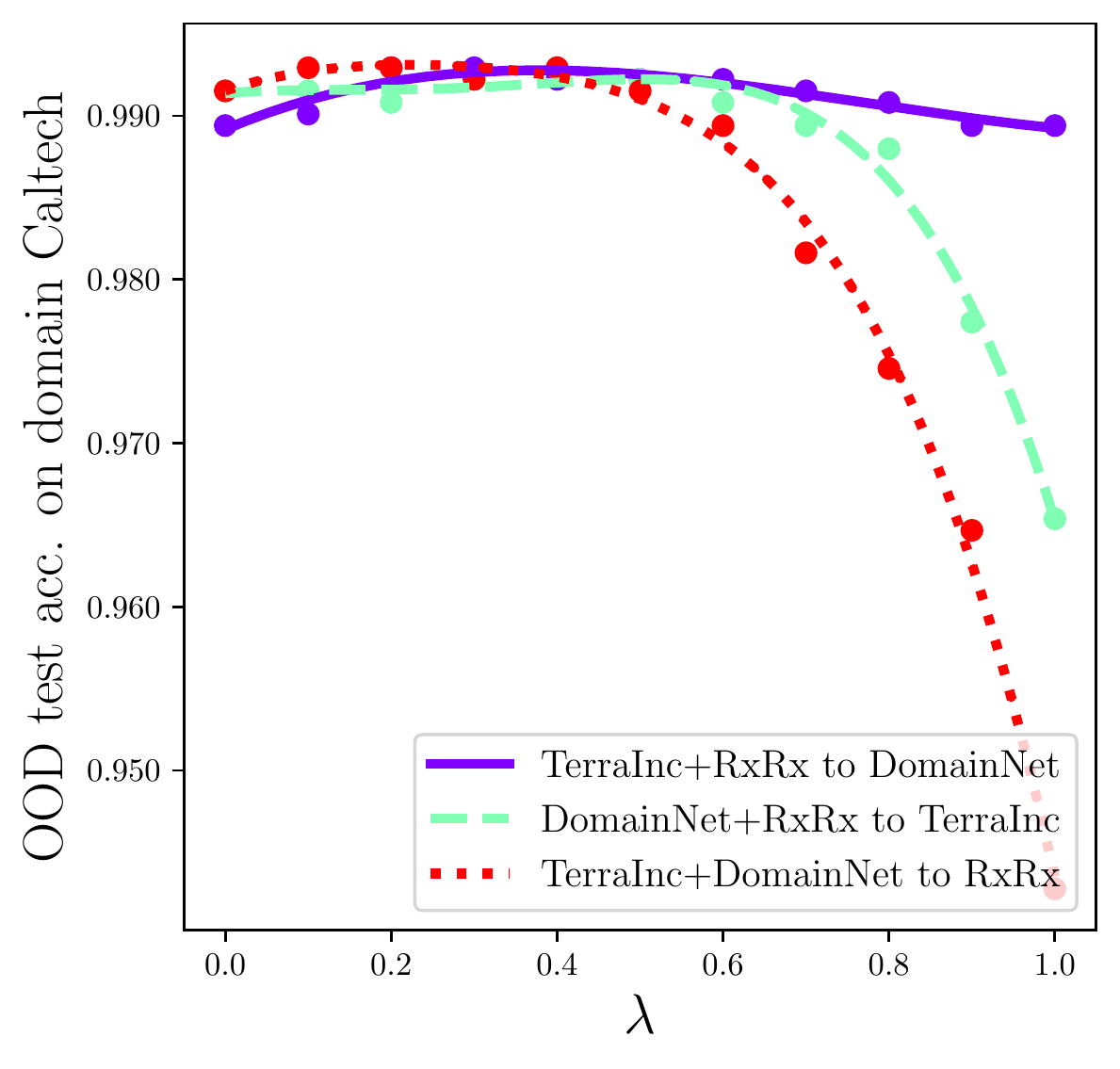}
            \caption{VLCS.}
            \label{fig:vlcs0_lmc_hyp3_ood}
        \end{subfigure}
        \begin{subfigure}{.19\textwidth}
            \centering
            \includegraphics[width=.95\linewidth]{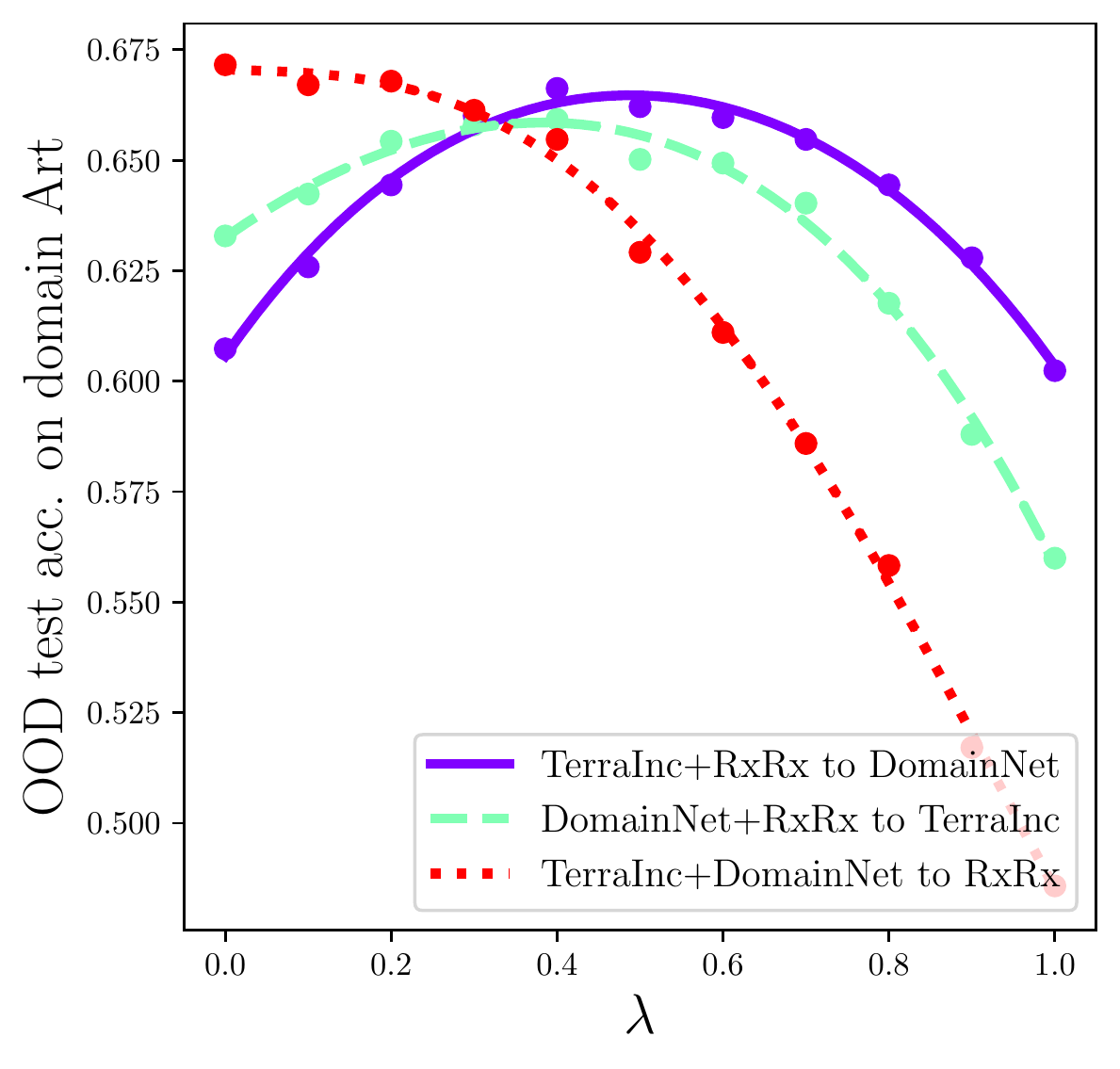}
            \caption{OfficeHome.}
            \label{fig:home0_lmc_hyp3_ood}
        \end{subfigure}%
        \begin{subfigure}{.19\textwidth}
            \centering
            \includegraphics[width=.95\linewidth]{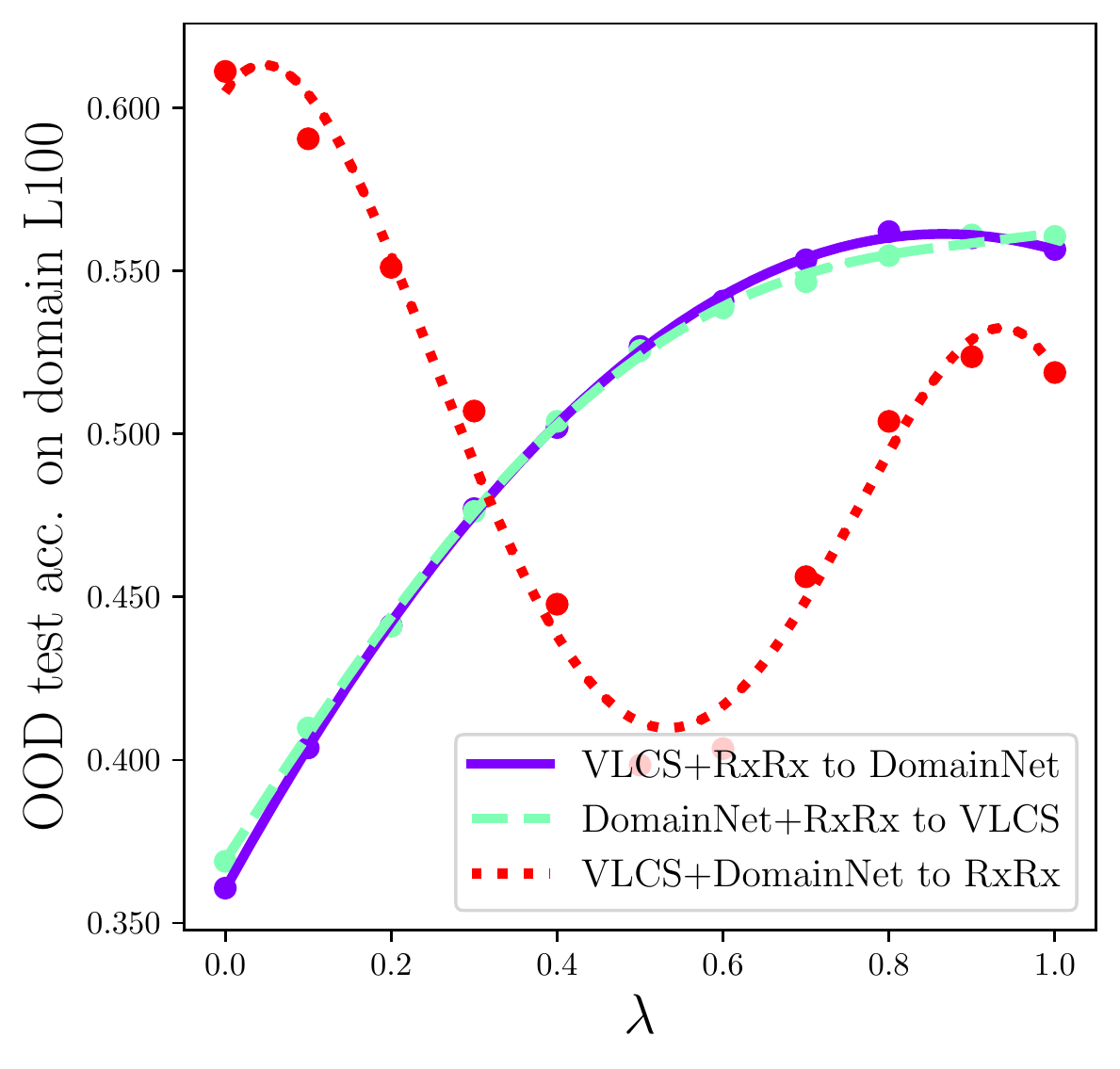}
            \caption{TerraIncognita.}
            \label{fig:terra0_lmc_hyp3_ood}
        \end{subfigure}
        \begin{subfigure}{.19\textwidth}
            \centering
            \includegraphics[width=.95\linewidth]{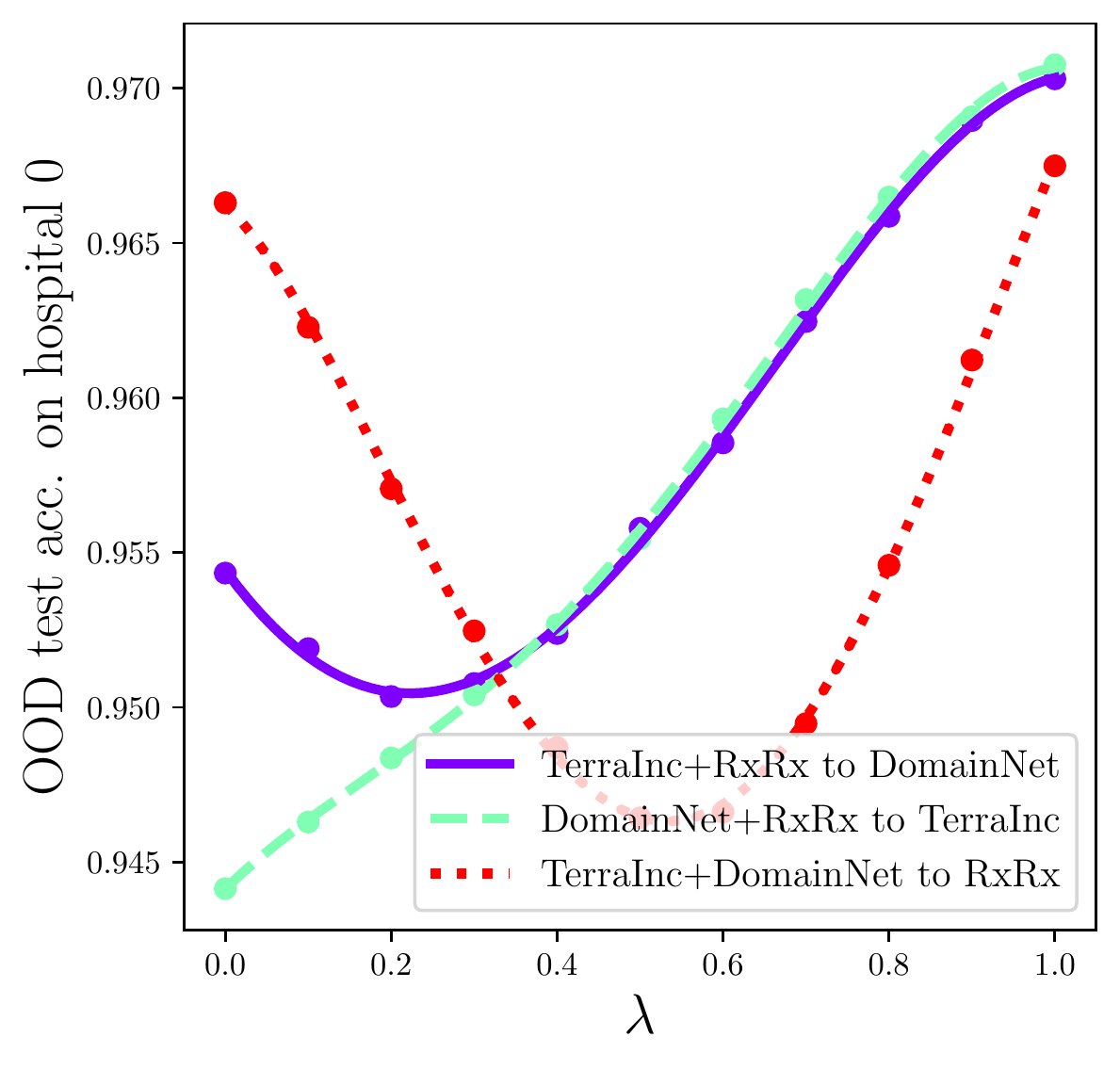}
            \caption{Camelyon.}
            \label{fig:came0_lmc_hyp3_ood}
        \end{subfigure}
    \end{center}
    \caption{Empirical analysis of LMC when combining three weights. \enquote{Dataset$_a$+Dataset$_b$ to Dataset$_c$} means that the model for $\lambda=0$ is the uniform weight average of $\theta_a$ and $\theta_b$ (fine-tuned on Dataset$_a$ and respectively on Dataset$_b$ before the target task) while the model $\theta_c$ for $\lambda=1$ was fine-tuned on Dataset$_c$ before the target task; $0<\lambda<1$ interpolates between those three fine-tuned weights as $(1-\lambda)/2 \cdot \theta_a + (1-\lambda)/2 \cdot \theta_b + \lambda \cdot  \theta_c $. On each target task, we consider the first domain as the test \ood domain.}
    \label{fig:all_lmc_hyp3_ood}
\end{figure}

In the practical settings from \Cref{sec:exps:ood}, model ratatouille averages more than two weight  inter-trained on different auxiliary tasks. For consistency, in \Cref{fig:all_lmc_hyp3_ood} we thus analyze LMC when interpolating across three fine-tuned weights. We observe the same successes, but also the same occasional failures when the target and task datasets are both simultaneously distant from the pre-training task, \ie with RxRx as the auxiliary target and either TerraIncognita or Camelyon as the target task.

\subsection{Recycling of Weights Fine-tuned Sequentially on Multiple Datasets}

\begin{figure}[h!]
    \begin{center}
        \begin{subfigure}{.19\textwidth}
            \centering
            \includegraphics[width=.95\linewidth]{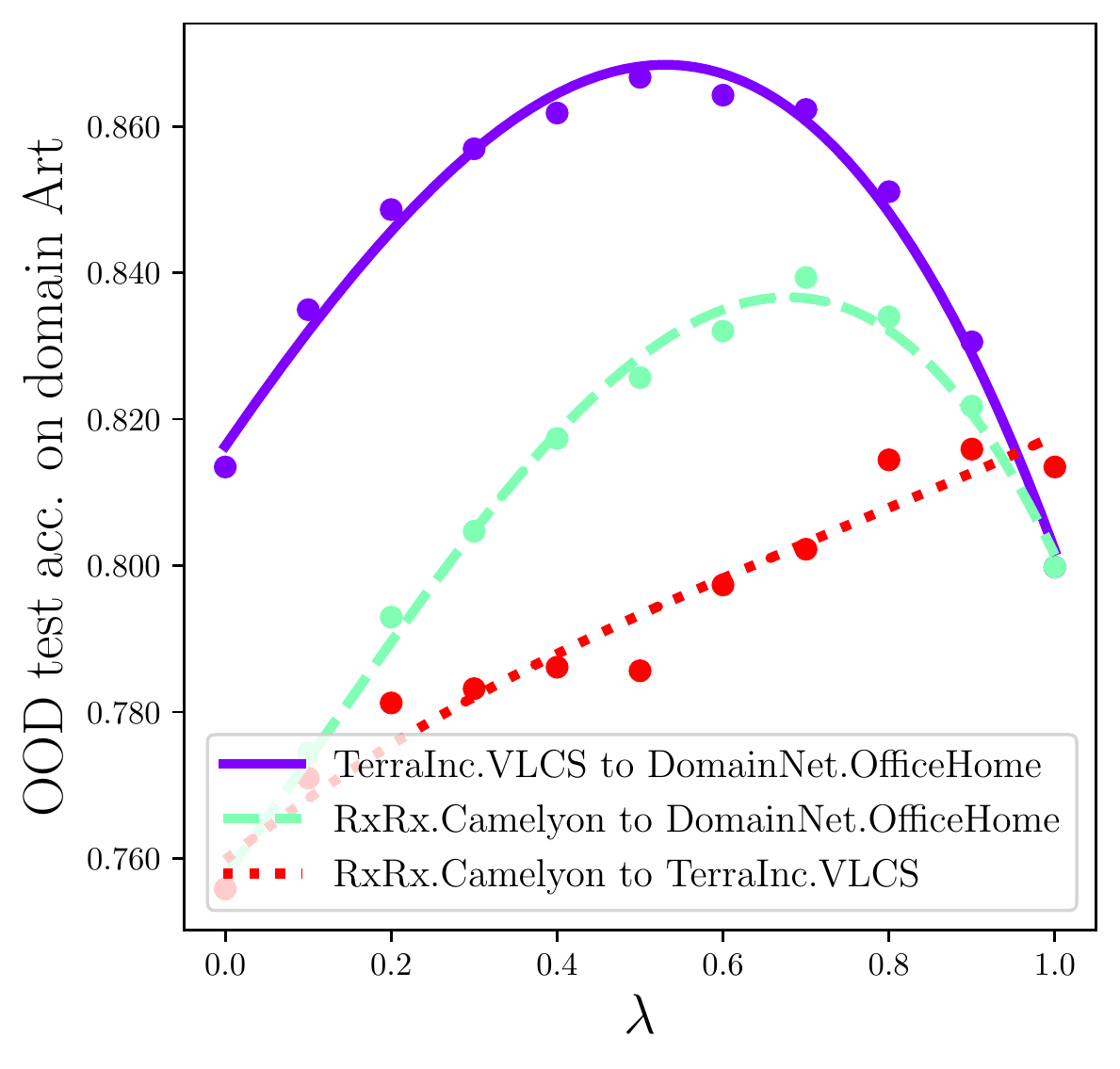}
            \caption{PACS.}
            \label{fig:pacs0_lmc_hyp2h_ood}
        \end{subfigure}
        \begin{subfigure}{.19\textwidth}
            \centering
            \includegraphics[width=.95\linewidth]{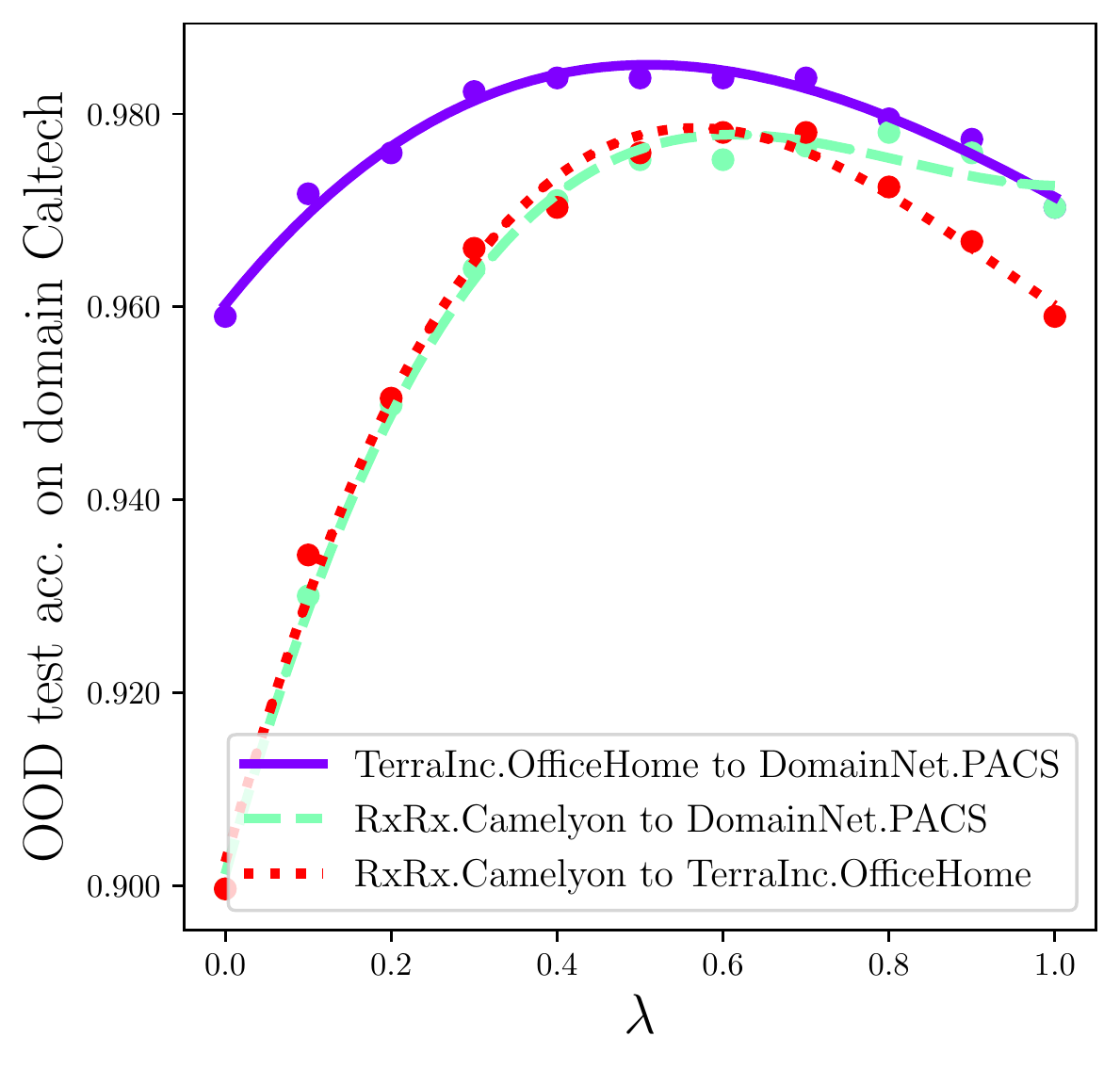}
            \caption{VLCS.}
            \label{fig:vlcs0_lmc_hyp2h_ood}
        \end{subfigure}
        \begin{subfigure}{.19\textwidth}
            \centering
            \includegraphics[width=.95\linewidth]{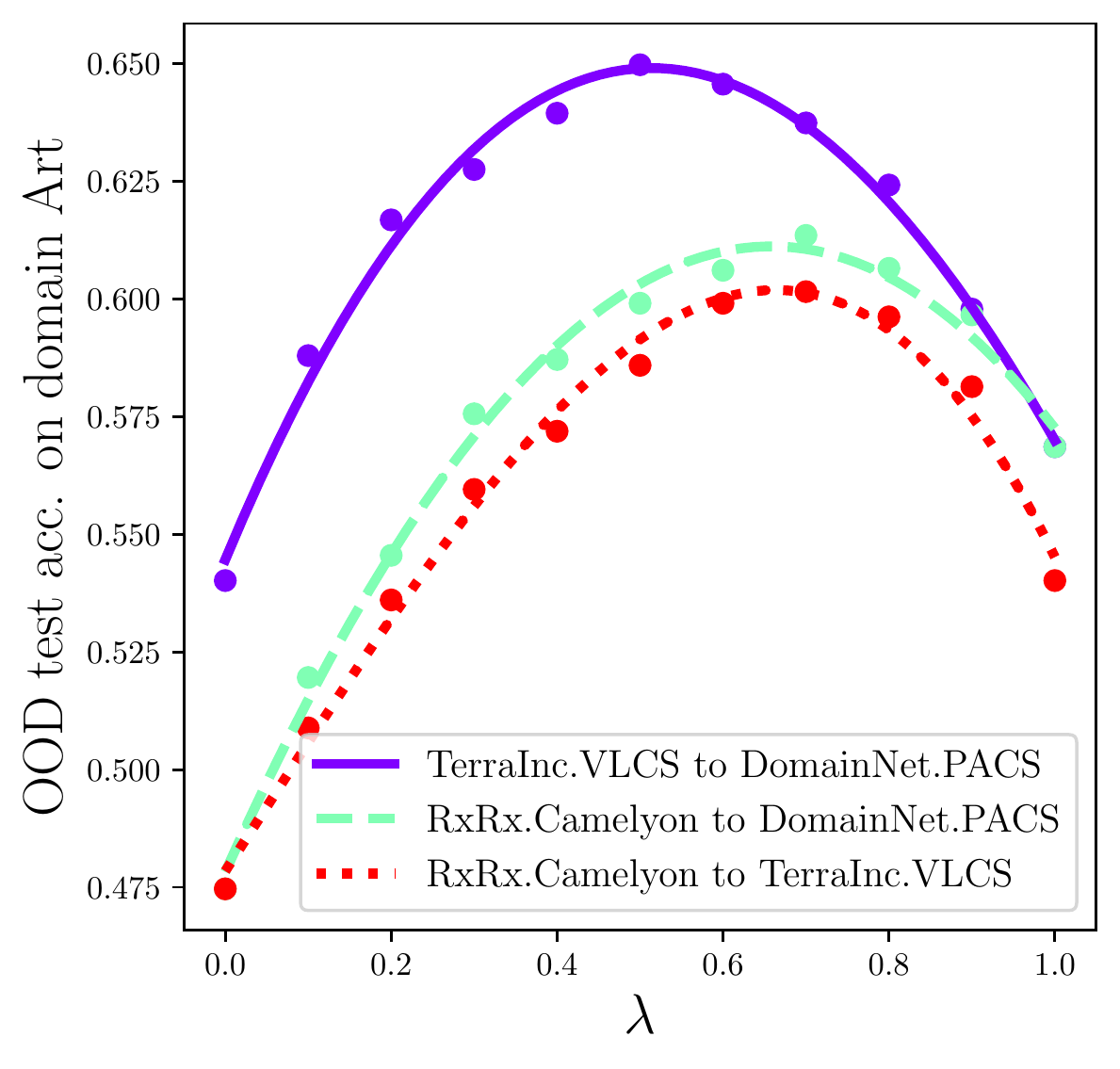}
            \caption{OfficeHome.}
            \label{fig:home0_lmc_hyp2h_ood}
        \end{subfigure}%
        \begin{subfigure}{.19\textwidth}
            \centering
            \includegraphics[width=.95\linewidth]{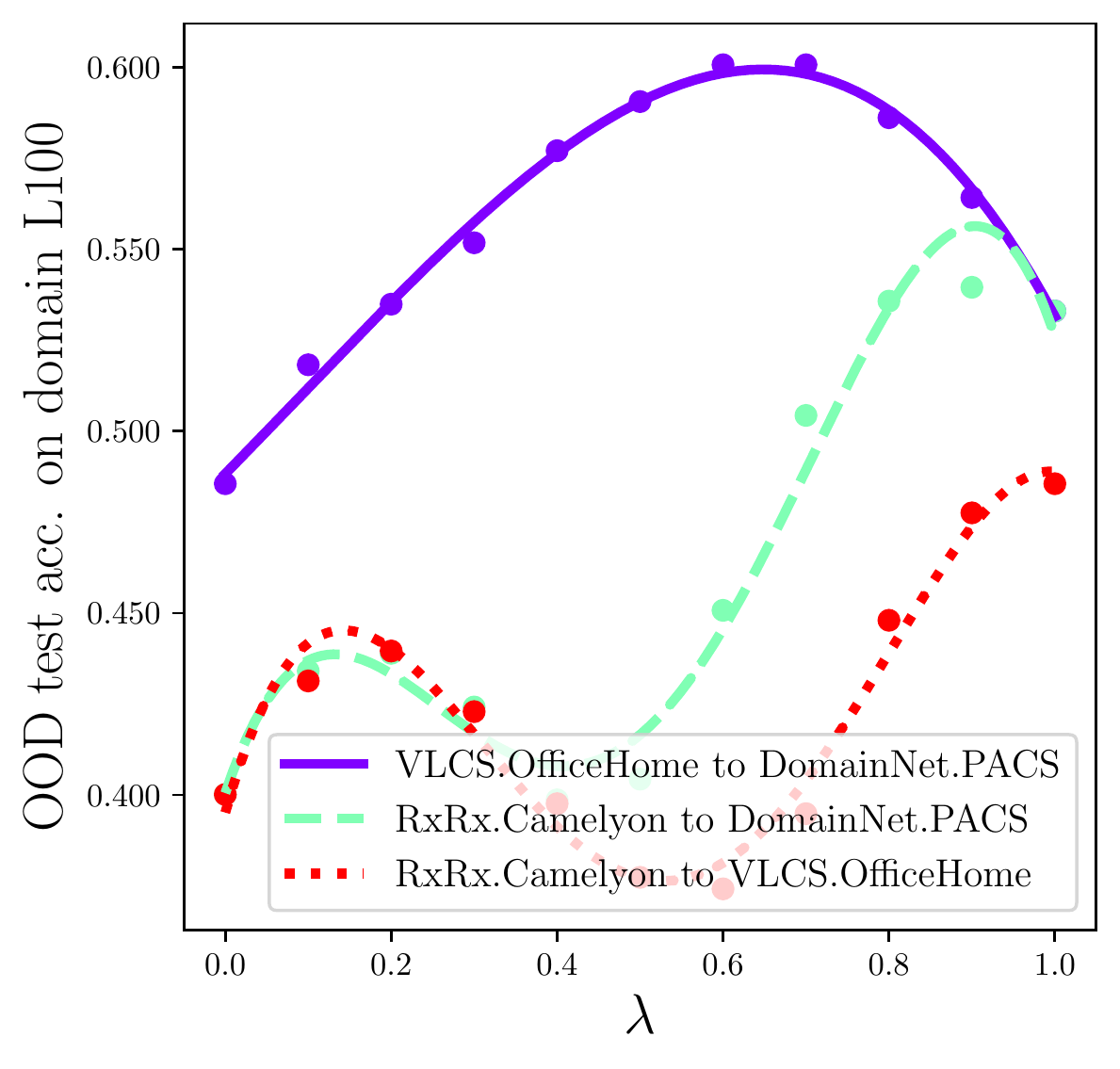}
            \caption{TerraIncognita.}
            \label{fig:terra0_lmc_hyp2h_ood}
        \end{subfigure}
        \begin{subfigure}{.19\textwidth}
            \centering
            \includegraphics[width=.95\linewidth]{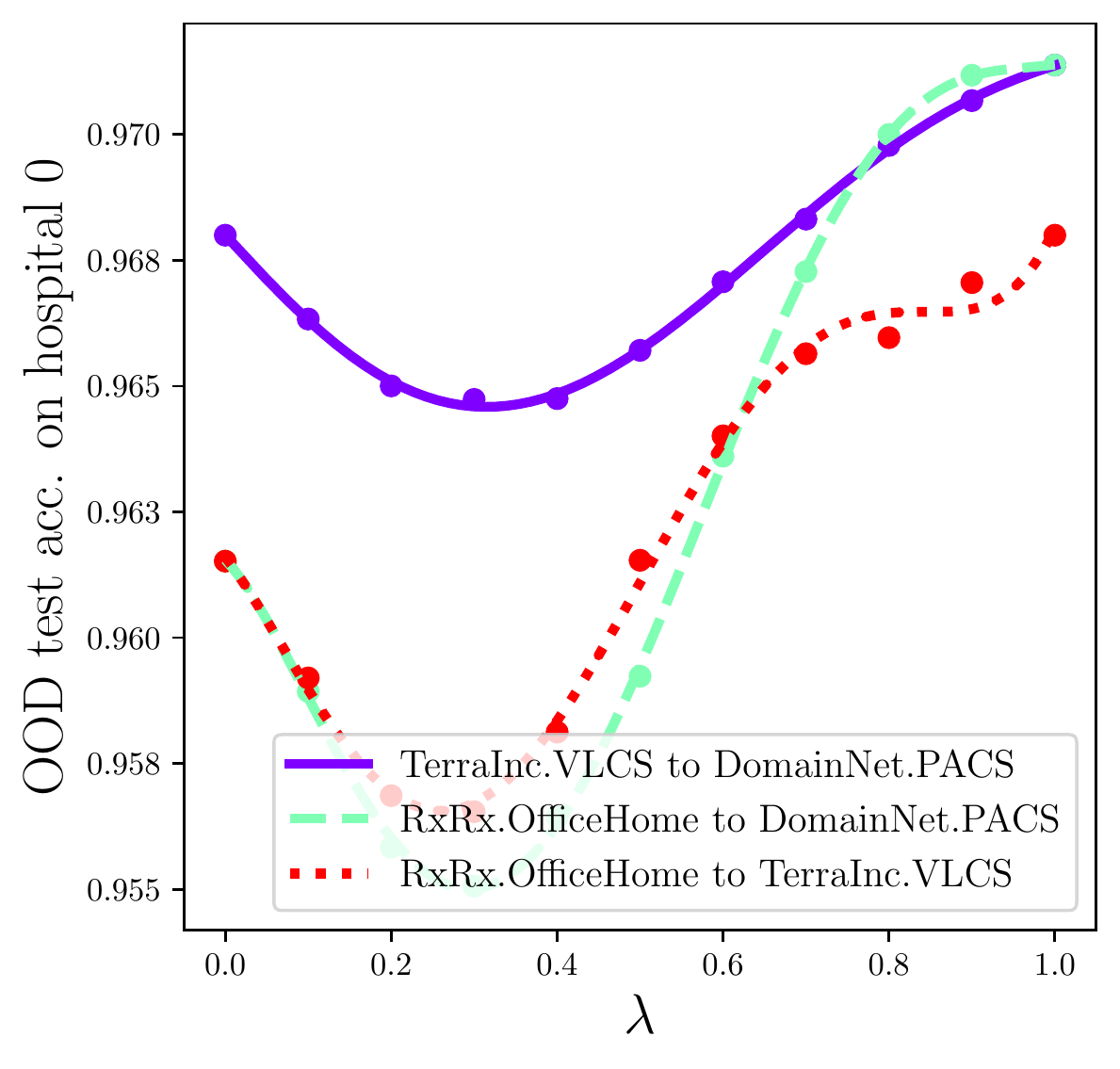}
            \caption{Camelyon.}
            \label{fig:came0_lmc_hyp2h_ood}
        \end{subfigure}
    \end{center}
    \vskip -0.4cm
    \caption{Empirical analysis of \Cref{hyp:2} when the intermediate tasks are themselves several successive fine-tunings on different auxiliary datasets. \enquote{Dataset$_a$.Dataset$_b$ to Dataset$_c$.Dataset$_d$} means that the model for $\lambda=0$ was sequentially fine-tuned on Dataset$_a$ then Dataset$_b$ before fine-tuning on the target task, while the model for $\lambda=1$ was sequentially fine-tuned on Dataset$_c$ then Dataset$_d$ before fine-tuning on the target task; $0<\lambda<1$ interpolates between those two fine-tuned weights.}%
    \label{fig:all_lmc_hyp2h_ood}
\end{figure}

In \Cref{fig:all_lmc_hyp2h_ood}, we empirically analyze \Cref{hyp:2} when the intermediate tasks are themselves several successive trainings on different auxiliary datasets.
Thus the initialization for \enquote{TerraInc.VLCS} in \Cref{fig:home0_lmc_hyp2h_ood} was sequentially fine-tuned on two auxiliary tasks (TerraIncognita and then on VLCS) before tackling the target task (OfficeHome). The concavity of the curves validates the LMC in most setups.
It hints towards a more general inheritance property of LMC: if two initializations satisfy the LMC, then the two fine-tuned weights too.
Yet, analysis of this inheritance property is best left for future work.

\FloatBarrier

\subsection{Ratatouille with VITs architecture}
\begin{figure}[h!]
    \begin{center}
        \begin{subfigure}{.24\textwidth}
            \includegraphics[width=1\linewidth]{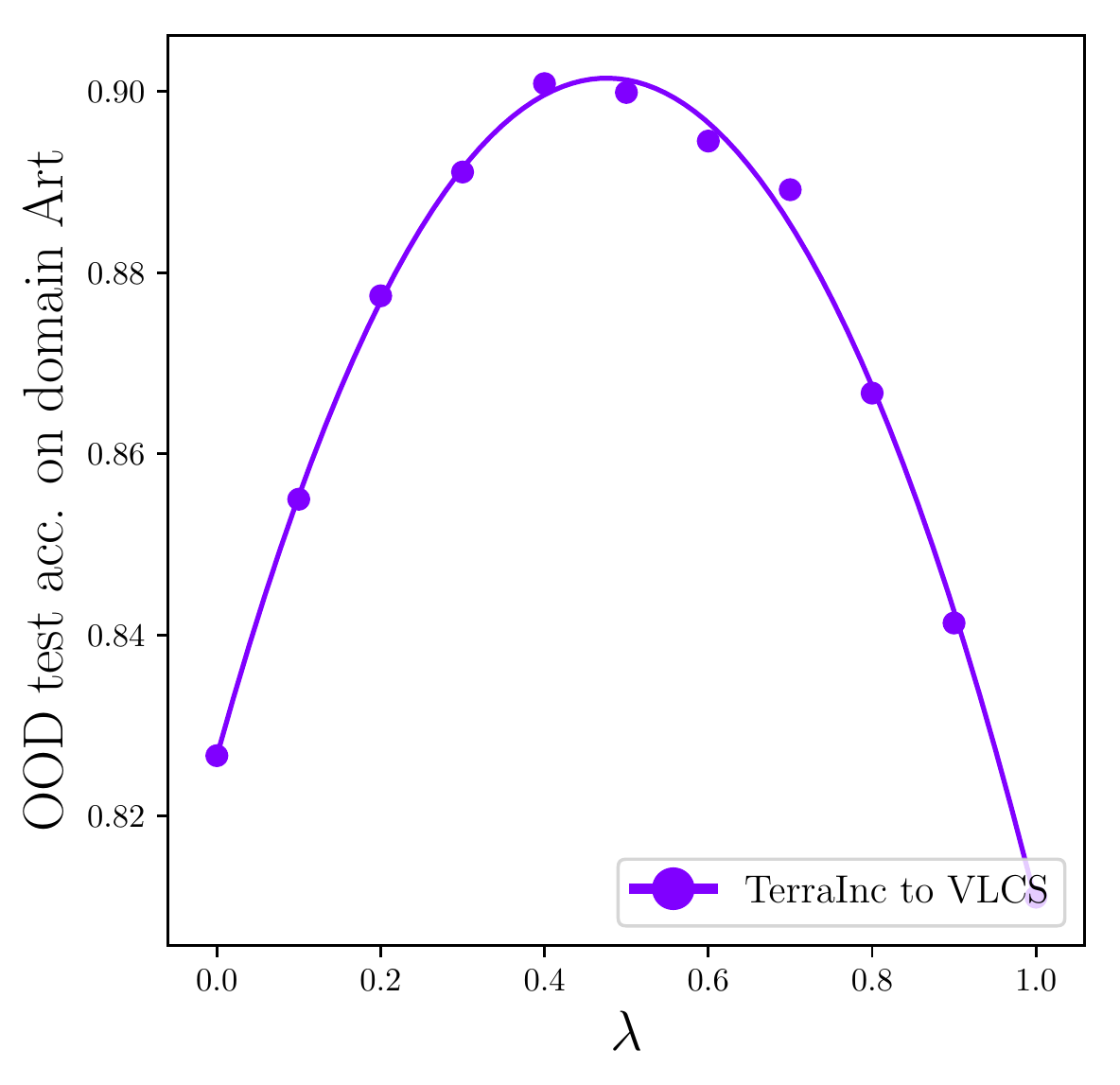}
            \caption{PACS.}
            \label{fig:pacs0_lmc_hyp2_ood_vit}
        \end{subfigure}
        \begin{subfigure}{.24\textwidth}
            \includegraphics[width=1\linewidth]{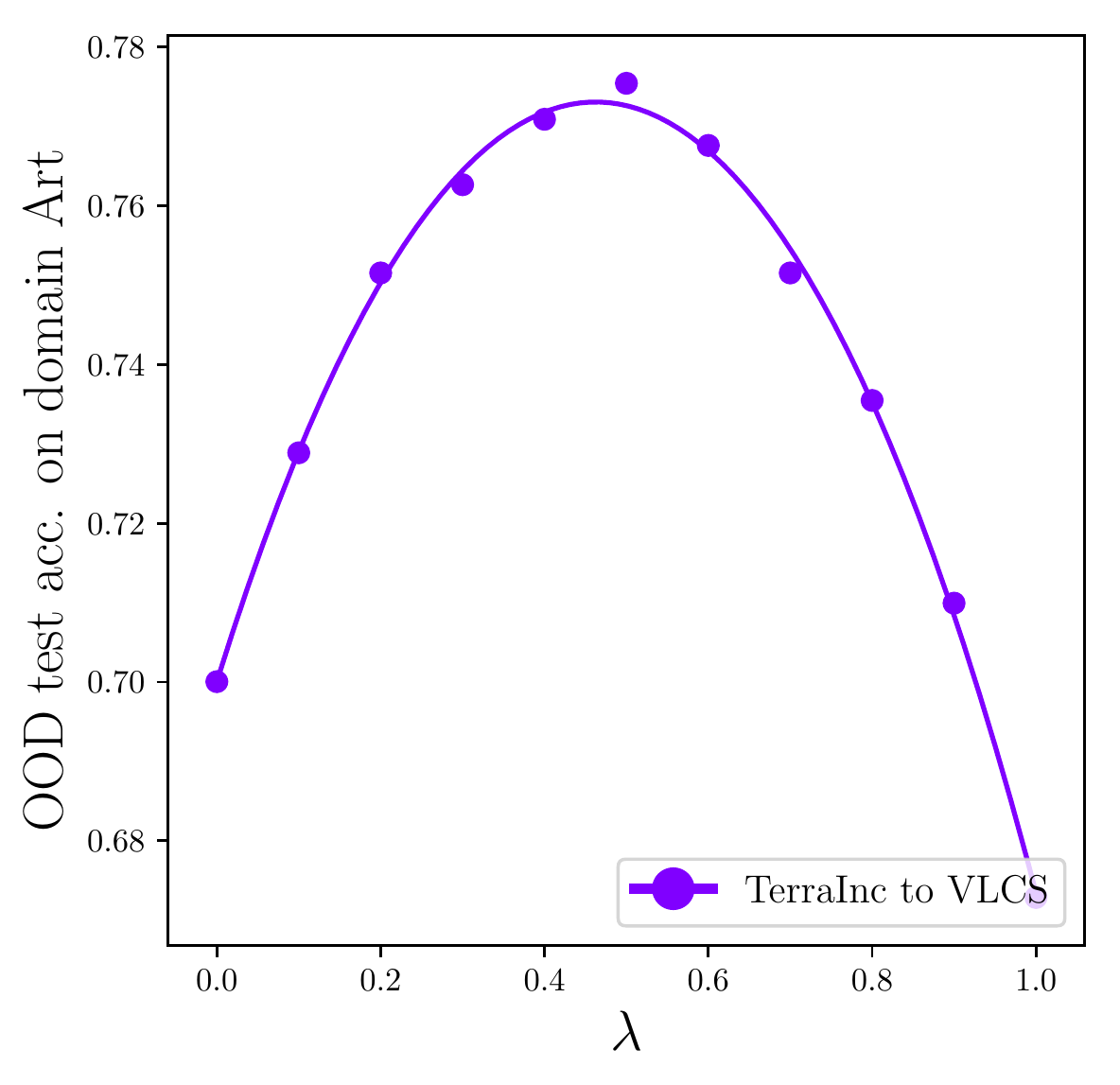}
            \caption{OfficeHome.}
            \label{fig:home0_lmc_hyp2_ood_vit}
        \end{subfigure}
    \end{center}
    \caption{Empirical validation of \Cref{hyp:2} with VITs.}%
    \label{fig:all_lmc_hyp2_vit}
\end{figure}

We previously have experimented with the ResNet-50 architecture, the standard for DomainBed on which the OOD generalization community relies, enabling reproducibility and fair comparisons with concurrent papers.
This exact same ResNet-50 architecture was the one used in the seminal works on LMC \cite{Neyshabur2020,Frankle2020}. Yet, the LMC is architecture agnostic.
For the sake of completeness, we show in \Cref{fig:all_lmc_hyp2_vit} that the LMC holds with vision transformers \cite{dosovitskiy2021an}, namely the ViT-B16 \enquote{vit\_base\_patch16\_224\_in21k} from \texttt{timm} \cite{rw2019timm}, following the setup from \citet{iwasawa2021testtime}.

\subsection{LMC in \iid}%
\label{app:iid}
In this section, we validate the LMC on \iid samples, without distribution shift between train and test.
The LMC holds in \iid, except sometimes when RxRx is the auxiliary task, and with curves less concave than in \ood.
These smaller gains when interpolating are because variance reduction via weight averaging is less beneficial in \iid than in \ood.

\begin{figure}[h!]
    \begin{center}
        \begin{subfigure}{.19\textwidth}
            \includegraphics[width=.95\linewidth]{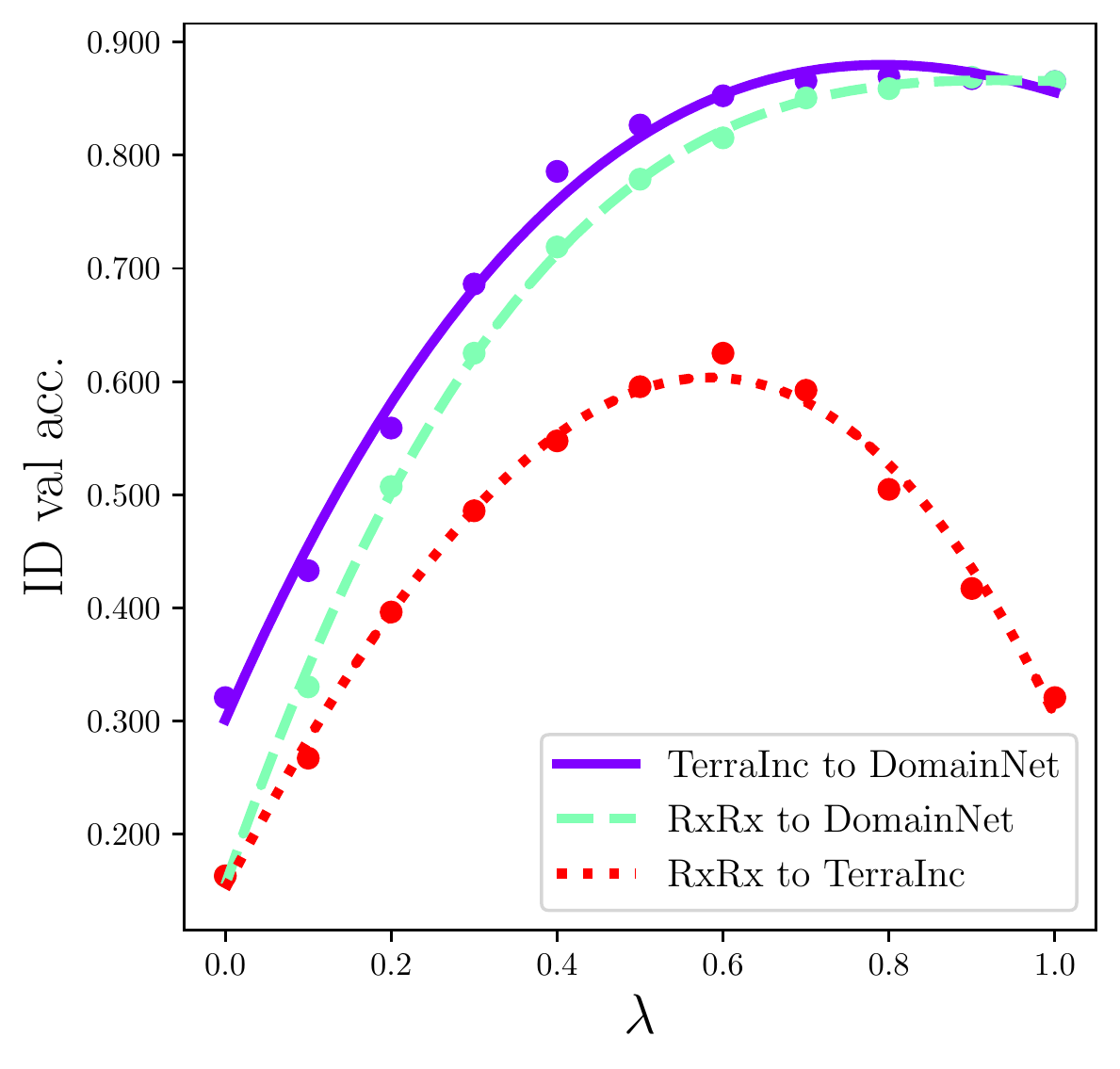}
            \caption{PACS.}
            \label{fig:pacs0_lmc_hyp1_iid}
        \end{subfigure}
        \begin{subfigure}{.19\textwidth}
            \includegraphics[width=.95\linewidth]{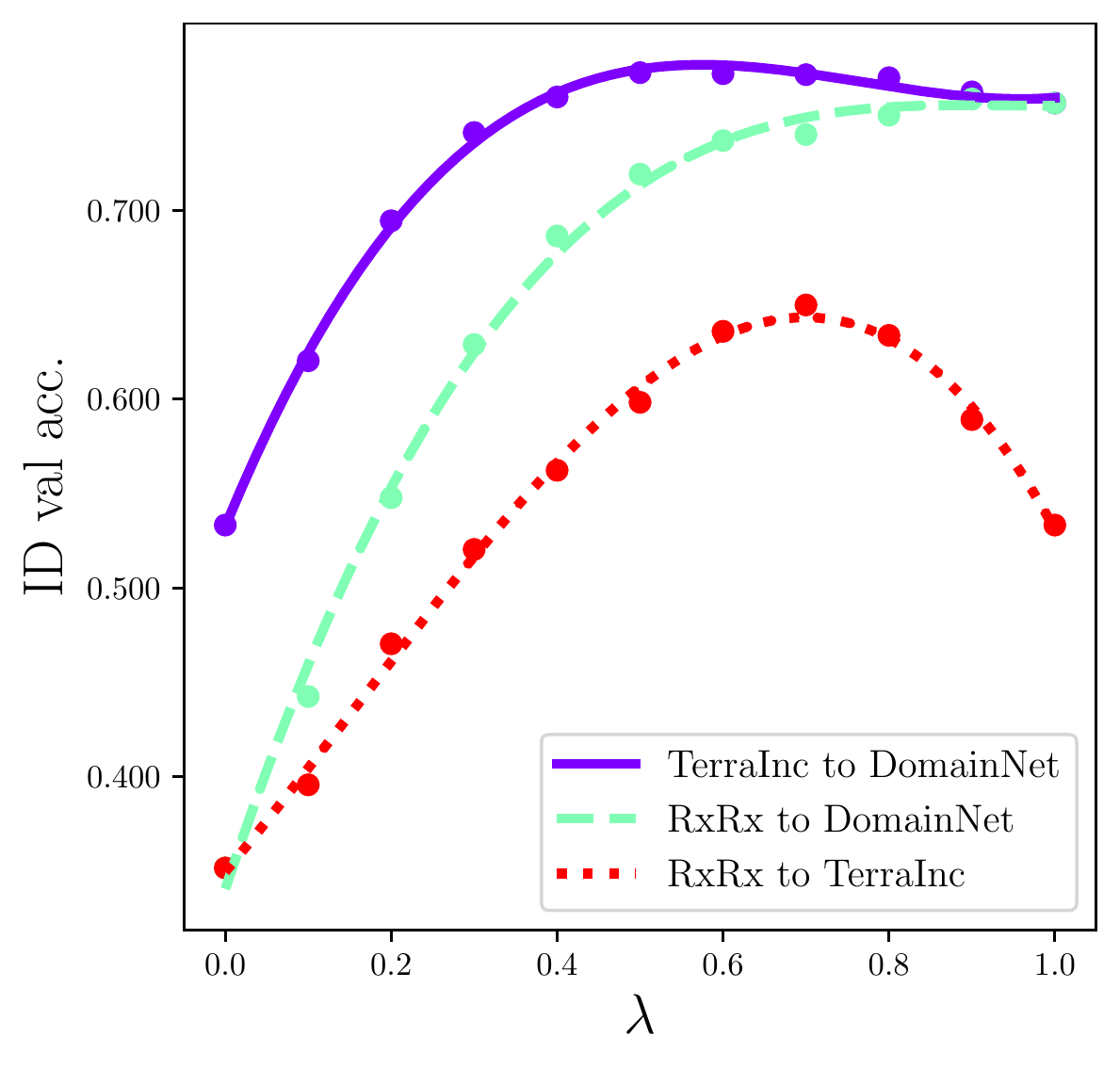}
            \caption{VLCS.}
            \label{fig:vlcs0_lmc_hyp1_iid}
        \end{subfigure}
        \begin{subfigure}{.19\textwidth}
            \includegraphics[width=.95\linewidth]{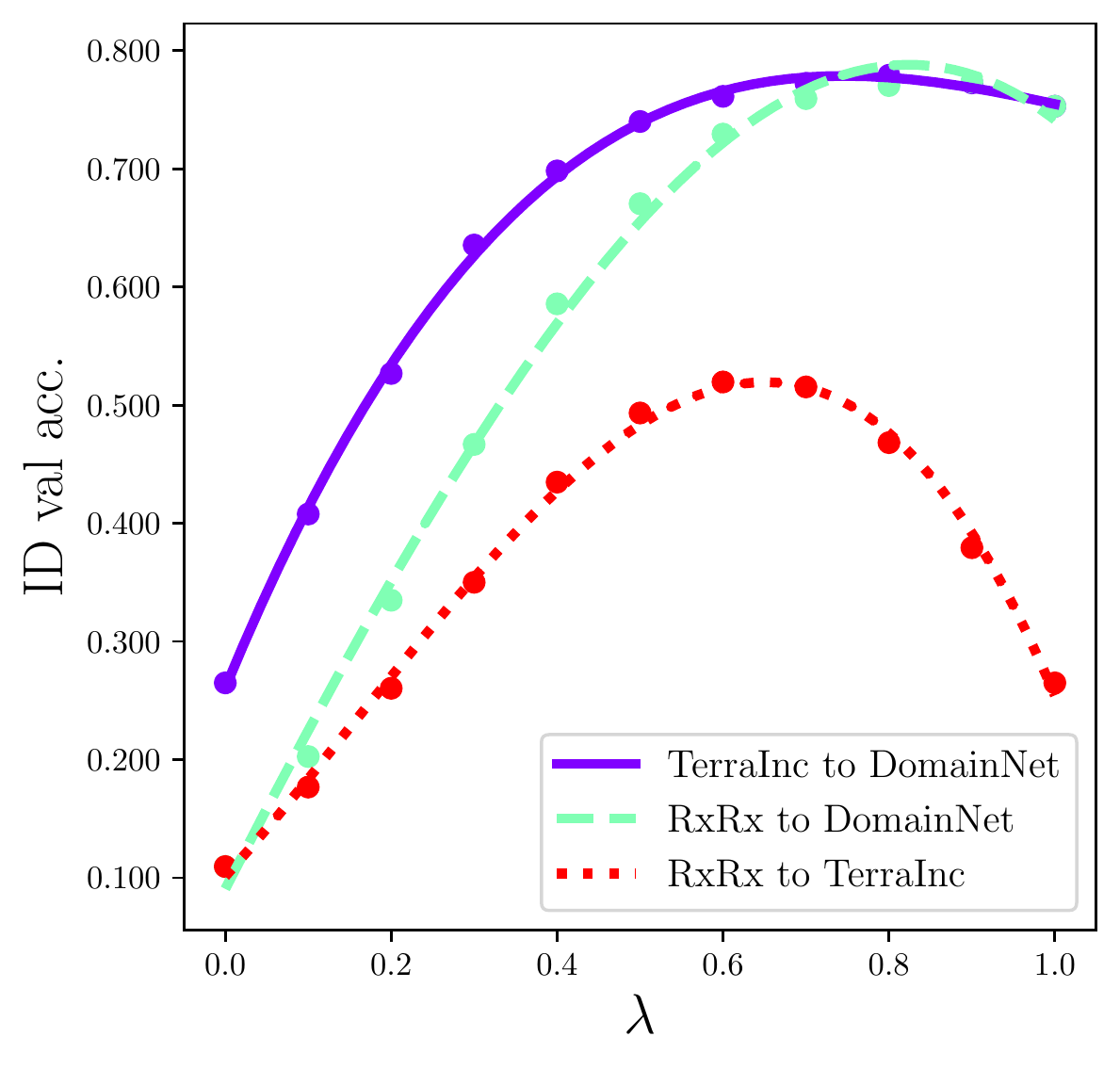}
            \caption{OfficeHome.}
            \label{fig:home0_lmc_hyp1_iid}
        \end{subfigure}%
        \begin{subfigure}{.19\textwidth}
            \includegraphics[width=.95\linewidth]{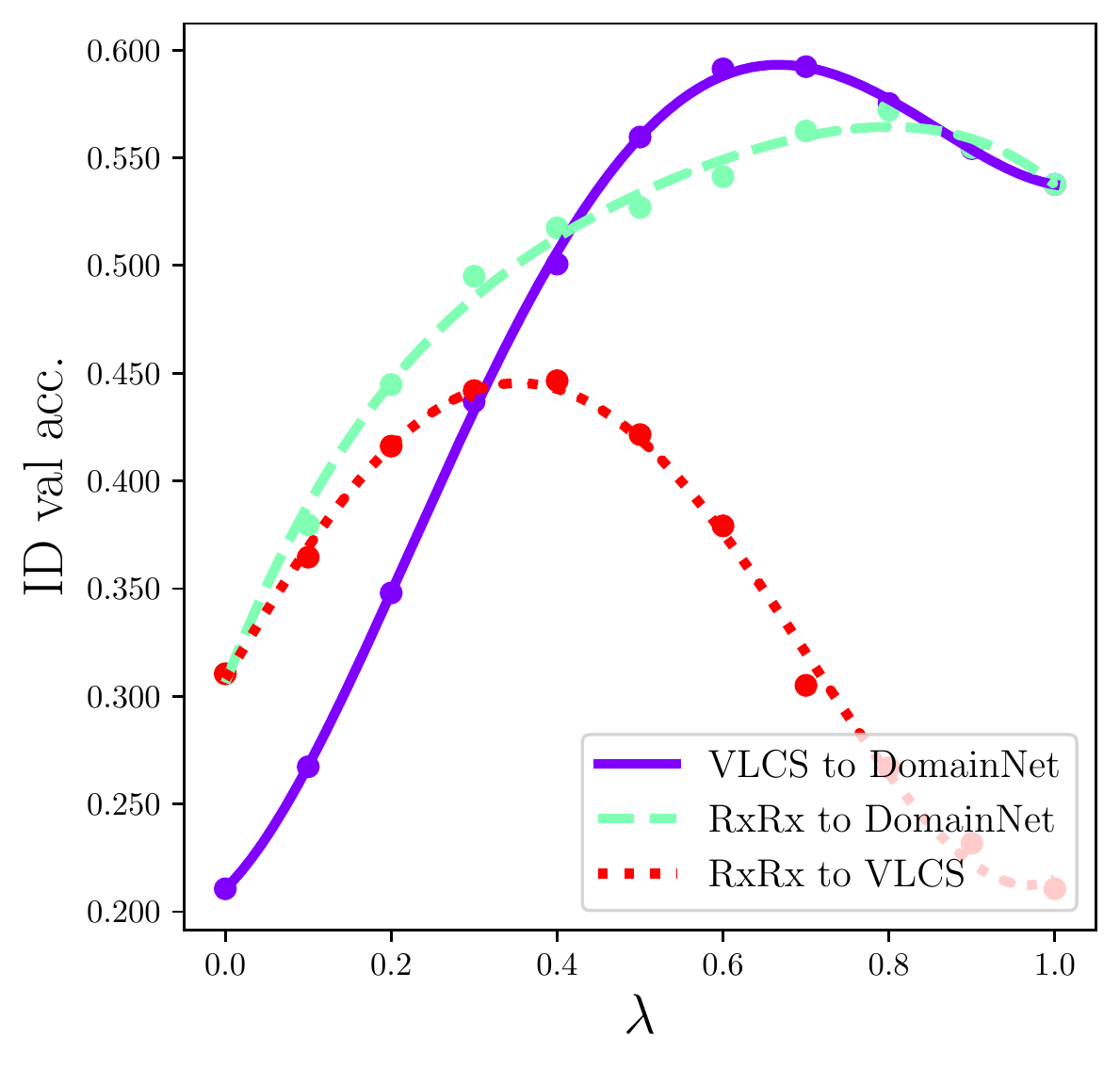}
            \caption{TerraIncognita.}
            \label{fig:terra0_lmc_hyp1_iid}
        \end{subfigure}
        \begin{subfigure}{.19\textwidth}
            \includegraphics[width=.95\linewidth]{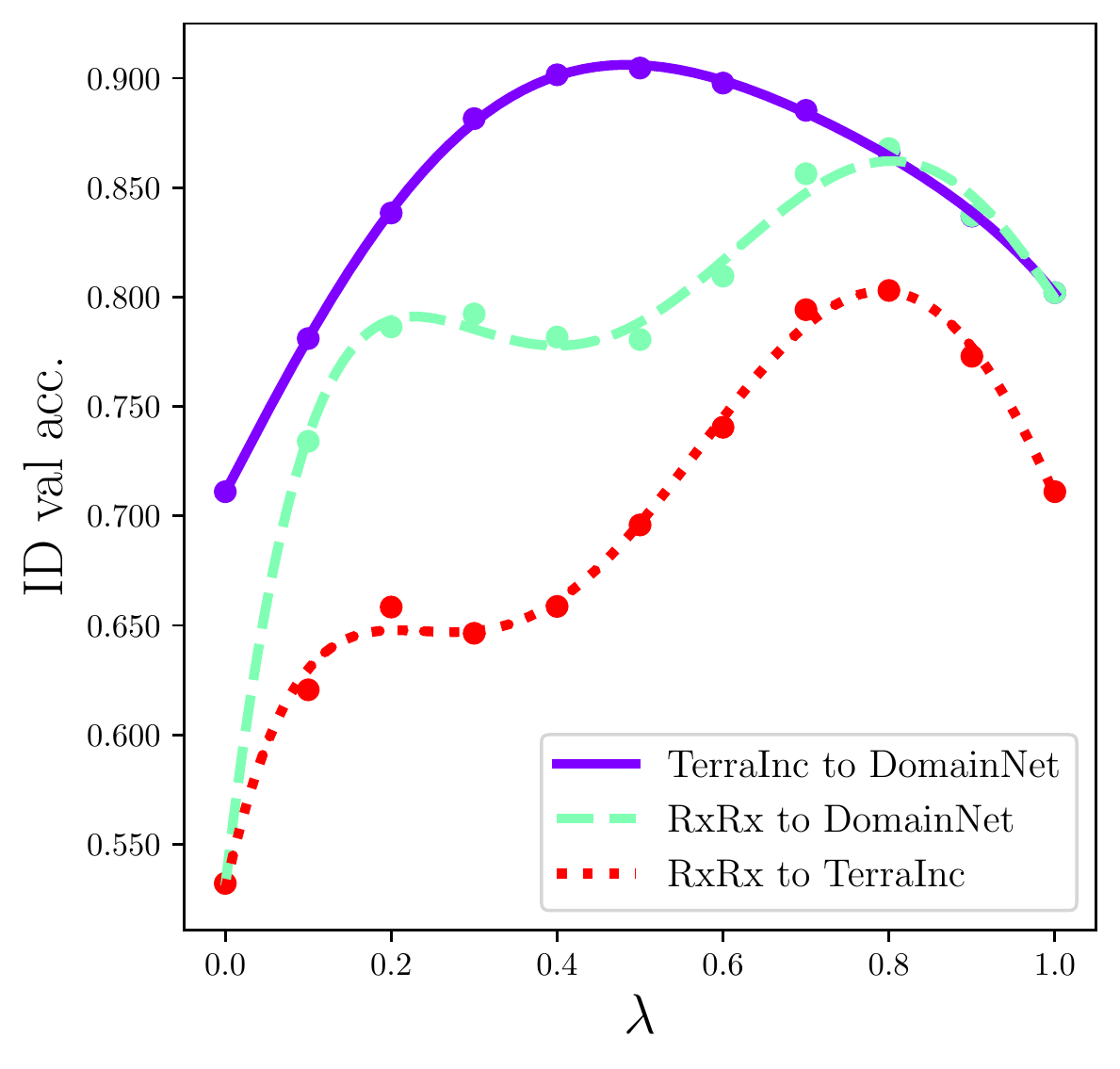}
            \caption{Camelyon.}
            \label{fig:came0_lmc_hyp1_iid}
        \end{subfigure}
    \end{center}
    \caption{Empirical analysis of \Cref{hyp:1} on the \iid validation split. This mirrors the setup from \Cref{fig:pacs0_lmc_hyp1_ood,fig:vlcs0_lmc_hyp1_ood,fig:home0_lmc_hyp1_ood,fig:terra0_lmc_hyp1_ood,fig:came0_lmc_hyp1_ood}.}%
    \label{fig:all_lmc_hyp1_iid}
\end{figure}

\begin{figure}[h!]
    \begin{center}
        \begin{subfigure}{.19\textwidth}
            \includegraphics[width=.95\linewidth]{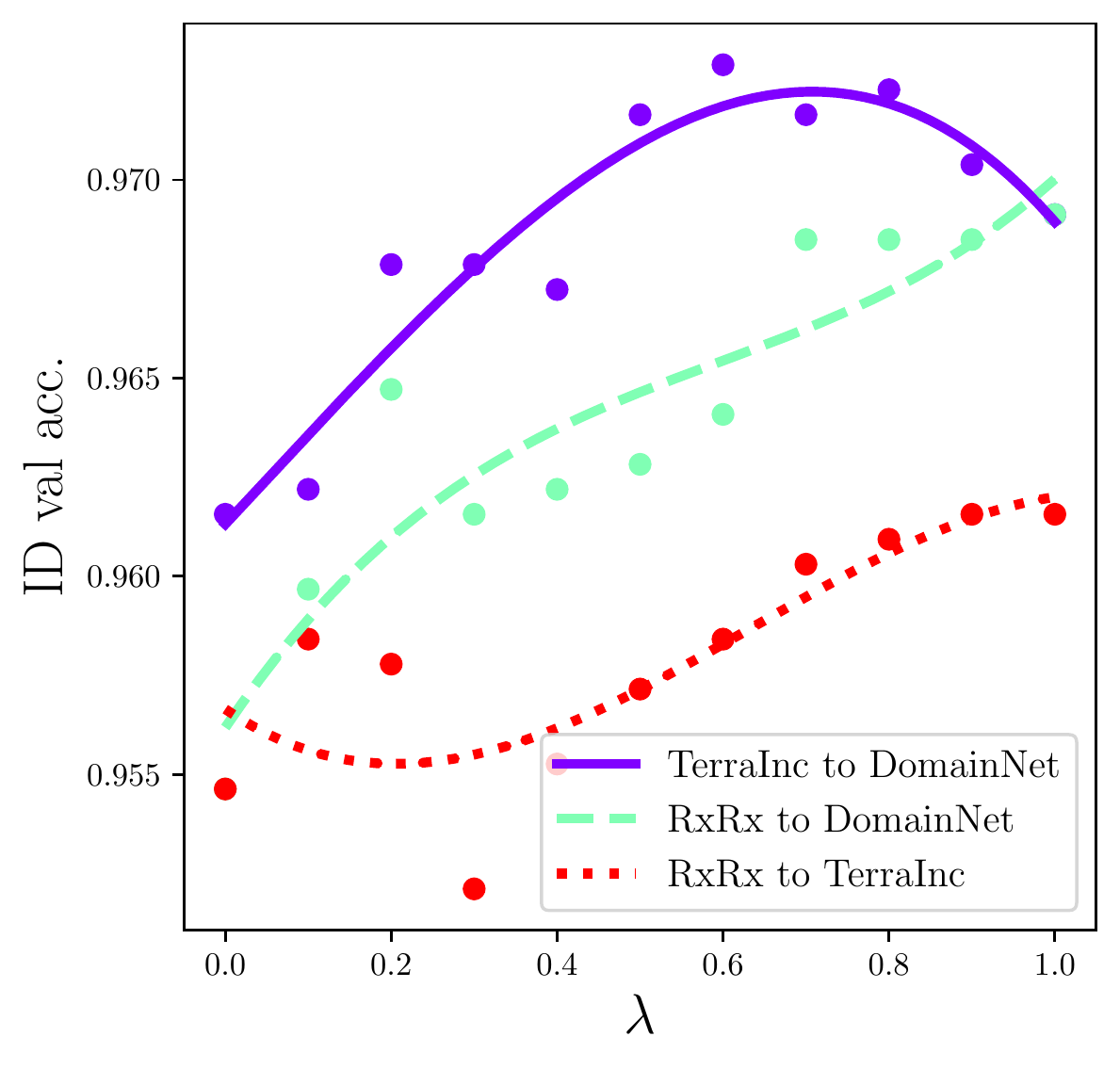}
            \caption{PACS.}
            \label{fig:pacs0_lmc_hyp2_iid}
        \end{subfigure}
        \begin{subfigure}{.19\textwidth}
            \includegraphics[width=.95\linewidth]{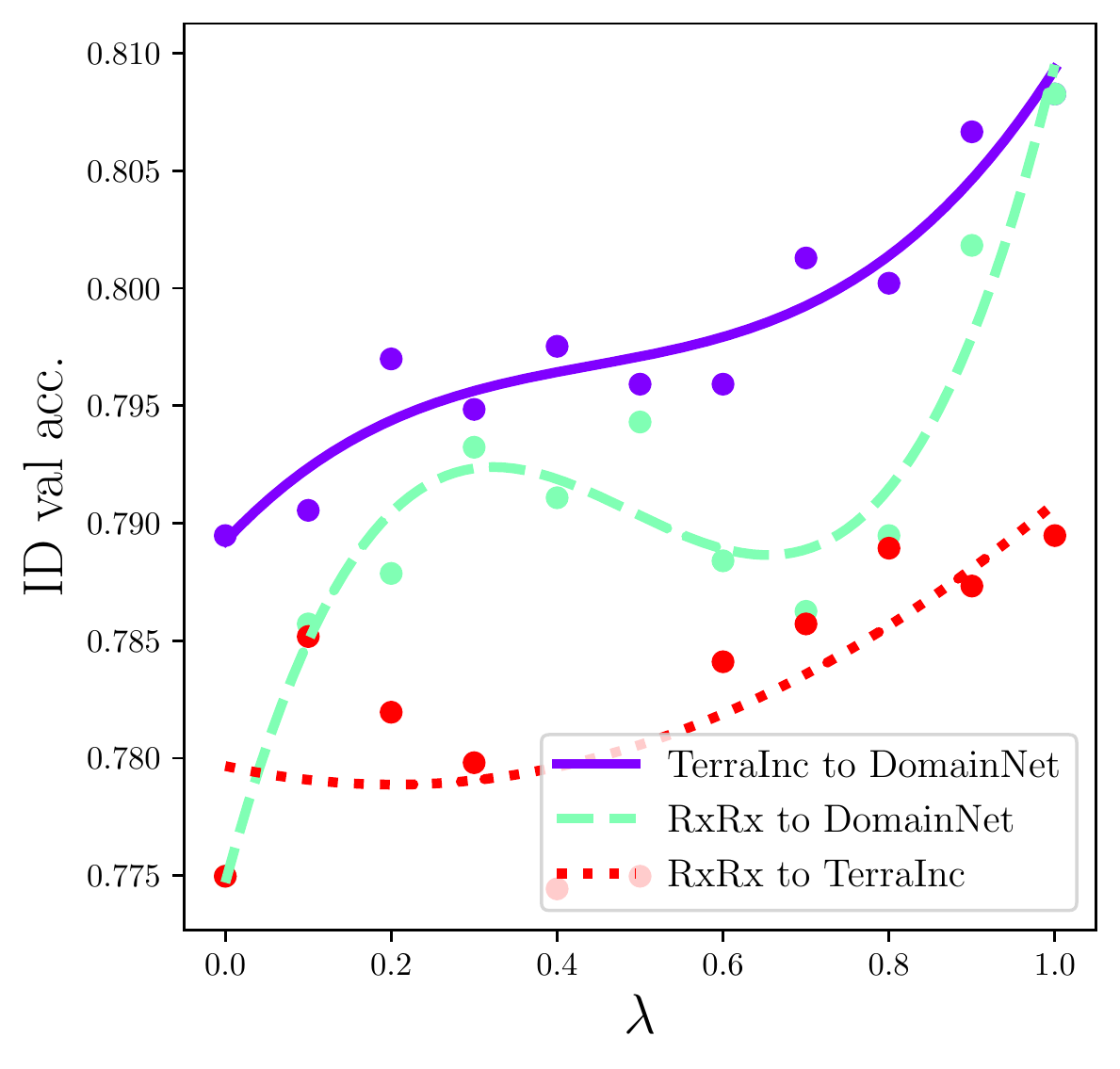}
            \caption{VLCS.}
            \label{fig:vlcs0_lmc_hyp2_iid}
        \end{subfigure}
        \begin{subfigure}{.19\textwidth}
            \includegraphics[width=.95\linewidth]{images/filesdevfair/lmc/home0_lmc_hyp2_iid.pdf}
            \caption{OfficeHome.}
            \label{fig:home0_lmc_hyp2_iidv2}
        \end{subfigure}%
        \begin{subfigure}{.19\textwidth}
            \includegraphics[width=.95\linewidth]{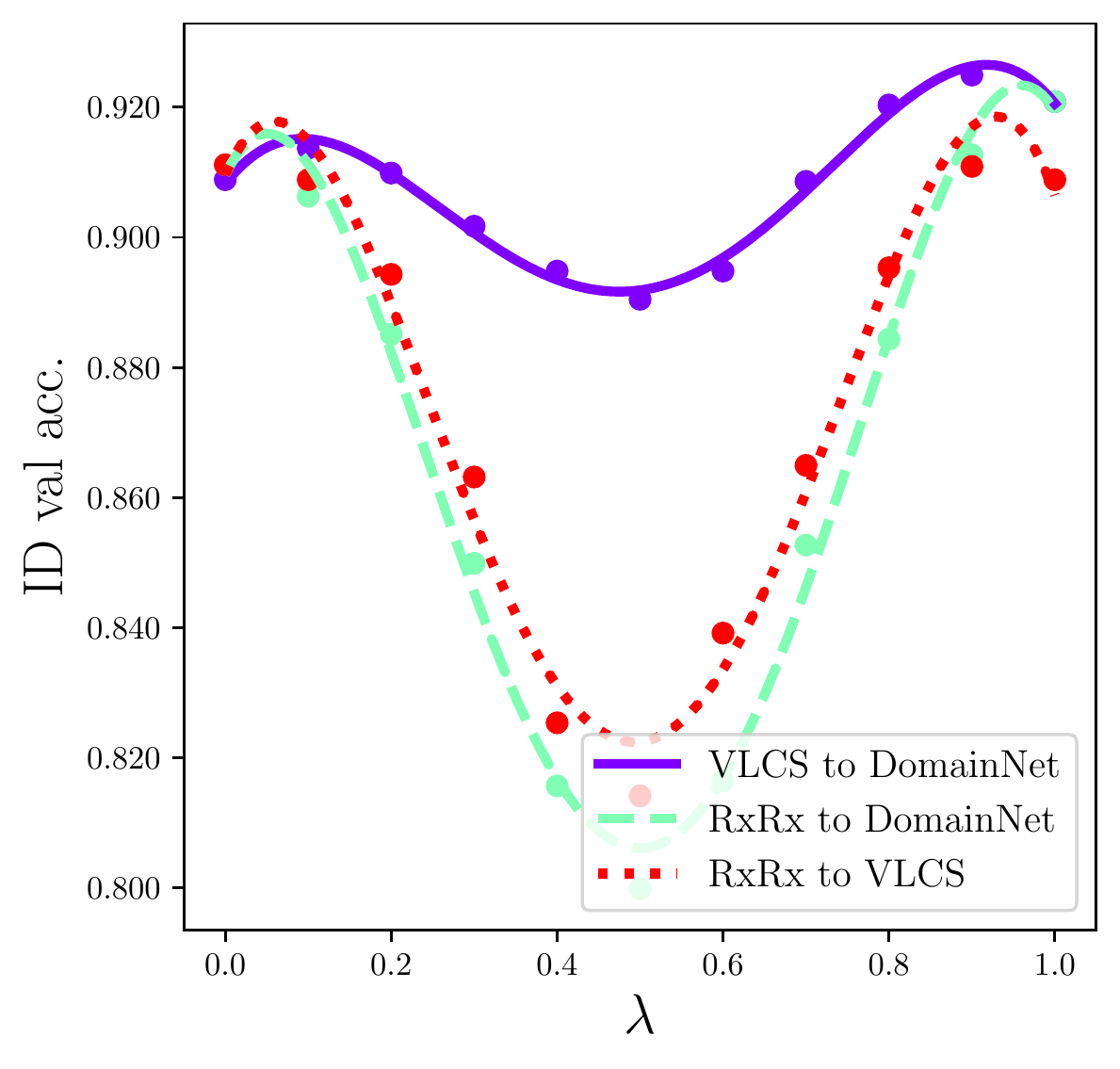}
            \caption{TerraIncognita.}
            \label{fig:terra0_lmc_hyp2_iid}
        \end{subfigure}
        \begin{subfigure}{.19\textwidth}
            \includegraphics[width=.95\linewidth]{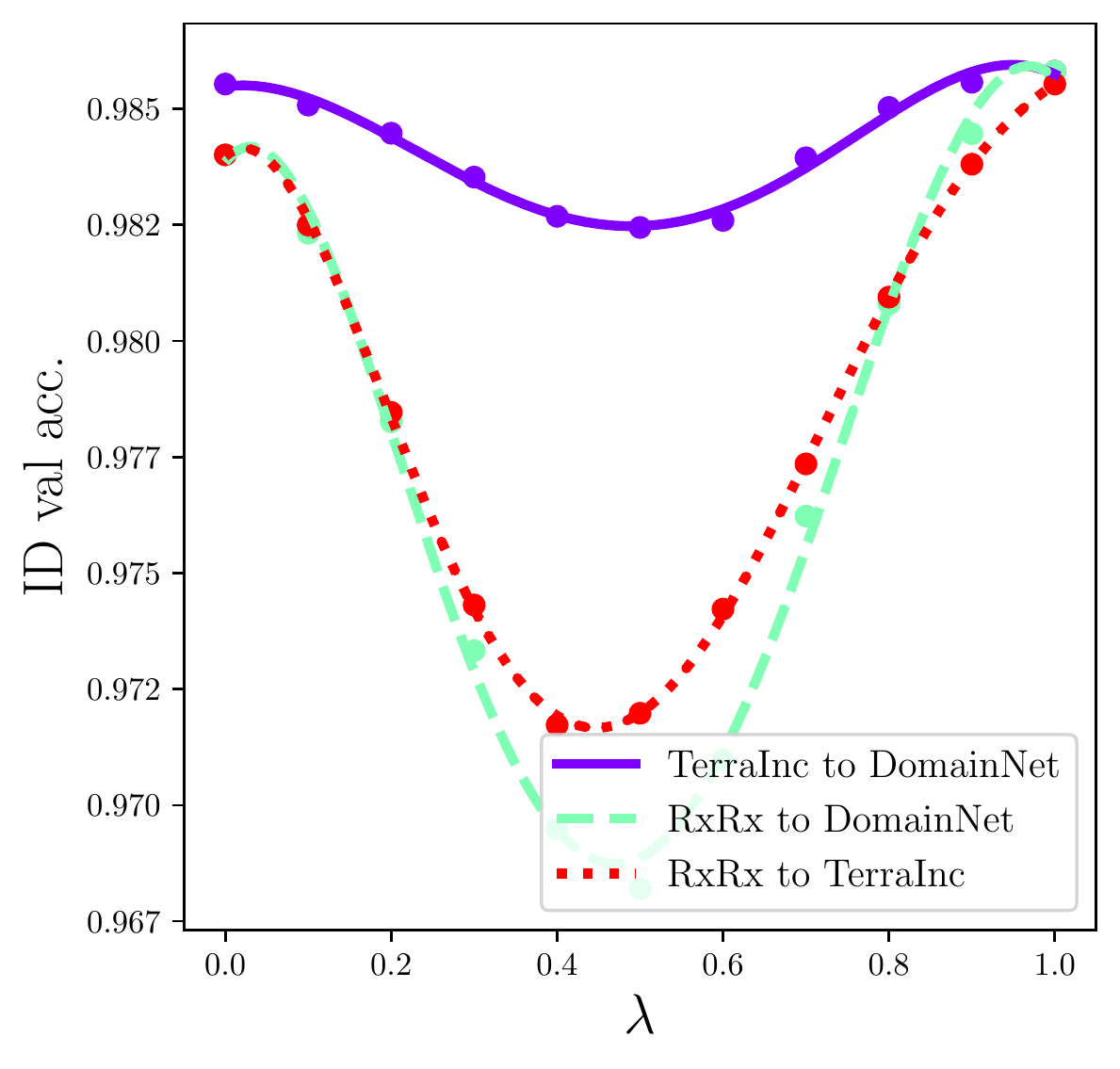}
            \caption{Camelyon.}
            \label{fig:came0_lmc_hyp2_iid}
        \end{subfigure}
    \end{center}
    \caption{Empirical analysis of \Cref{hyp:2} on the \iid validation split. This mirrors the setup from \Cref{fig:pacs0_lmc_hyp2_ood,fig:vlcs0_lmc_hyp2_ood,fig:home0_lmc_hyp2_ood,fig:terra0_lmc_hyp2_ood,fig:came0_lmc_hyp2_ood}.}%
    \label{fig:all_lmc_hyp2_iid}
\end{figure}

\FloatBarrier
\section{Robust Inter-Training}
\label{app:robustintertraining}
Recycled soups leverage weights fine-tuned on various auxiliary tasks; these starting points may sometimes may too specialized, and less general than the initial pre-trained weights.
To preserve the pre-trained general knowledge, in this section we consider a \emph{robust inter-training} strategy where the initializations are robustified via moving average \cite{szegedy2016rethinking,izmailov2018,Wortsman2022robust} along the auxiliary fine-tuning.
This follows recent evidence that moving average can reduce catastrophic forgetting \cite{lubana2021quadratic,eeckt2022weight,stojanovski2022momentum,langnickel2022bert}.
As in \Cref{eq:strategies}, these new robust strategies can be written as:
\begin{equation}
    \begin{aligned}
        \theta & = \mathrm{Train}\big(\mathrm{Train}\left(\theta^{\mathrm{pt}}, T_i, \mathrm{collect\_ckpts=True}\right), T\big),                            & \text{[Robust inter-training]}    \\
        \theta & = \frac{1}{M} \sum_{i=0}^{M-1} \mathrm{Train}\big(\mathrm{Train}\left(\theta^{\mathrm{pt}}, T_i, \mathrm{collect\_ckpts=True}\right), T\big). & \text{[Robust model ratatouille]}
    \end{aligned} \label{eq:strategiesrobust}%
\end{equation}%
As we show in \Cref{table:domainbedrobust}, moving average improves the initializations and thus the transfer abilities of inter-training, and as a consequence also improves model ratatouille.
Better understanding how to further improve auxiliary initializations is an interesting research direction, already discussed in \citet{choshen2022start}.
\begin{table}[h!]%
    \begin{center}
        \caption{Accuracies ($\%,\uparrow$) on the DomainBed \cite{gulrajani2021in} benchmark.}
        \resizebox{0.7\textwidth}{!}{%
            \begin{tabular}{llccccccc}
                \toprule
                \textbf{Algorithm}                      & \textbf{Selection} & \textbf{PACS}              & \textbf{VLCS}              & \textbf{OfficeHome}        & \textbf{TerraInc}  & \textbf{DomainNet}         & \textbf{Avg}    \\
                \midrule
                Vanilla fine-tuning                                     & \iid val           & $85.9 \pm 0.6$             & $78.1 \pm 0.5$             & $69.4 \pm 0.2$             & $50.4 \pm 1.8$     & $44.3 \pm 0.2$             & $65.6$          \\
                Model soups                             & Uniform            & $88.7 \pm 0.2$             & $78.4 \pm 0.2$             & $72.1 \pm 0.2$             & $51.4 \pm 0.6$     & $47.4 \pm 0.2$             & $67.6$          \\
                Model soups                             & Greedy             & $88.0 \pm 0.3$             & $78.5 \pm 0.1$             & $71.5 \pm 0.2$             & $51.6 \pm 0.9$     & $\underline{47.7} \pm 0.1$ & $67.5$          \\
                Model soups$^\dagger$                   & Uniform$^\dagger$  & $89.0$                     & $78.6$                     & $72.8$                     & $51.9$             & $\underline{47.7}$         & $68.0$          \\
                \midrule
                Inter-training \cite{phang2018sentence} & \iid val           & $89.0 \pm 0.0$             & $77.7 \pm 0.0$             & $69.9 \pm 0.6$             & $46.7 \pm 0.1$     & $44.5 \pm 0.1$             & $65.6$          \\
                Model ratatouille                       & Uniform            & $89.5 \pm 0.1$             & $78.5 \pm 0.1$             & $73.1 \pm 0.1$             & $51.8 \pm 0.4$     & $47.5 \pm 0.1$             & $68.1$          \\
                Model ratatouille                       & Greedy             & $\underline{90.5} \pm 0.2$ & $78.7 \pm 0.2$             & $\underline{73.4} \pm 0.3$ & $49.2 \pm 0.9$     & $\underline{47.7} \pm 0.0$ & $67.9$          \\
                Model ratatouille$^{\dagger}$           & Uniform$^\dagger$  & $89.8$                     & $78.3$                     & $\textbf{73.5}$            & $\underline{52.0}$ & $\underline{47.7}$         & $\textbf{68.3}$ \\
                \midrule
                Robust inter-training                   & \iid val           & $88.9 \pm 0.5$             & $77.8 \pm 0.1$             & $71.8 \pm 0.6$             & $47.3 \pm 0.5$     & $44.5 \pm 0.2$             & $66.1$          \\
                Robust model ratatouille                & Uniform            & $89.7 \pm 0.1$             & $\underline{78.6} \pm 0.2$ & $73.0 \pm 0.1$             & $51.9 \pm 0.2$     & $47.4 \pm 0.1$             & $68.2$          \\
                Robust model ratatouille                & Restricted         & $\textbf{90.7} \pm 0.1$    & $\textbf{78.8} \pm 0.2$    & $\underline{73.4} \pm 0.2$ & $50.6 \pm 0.3$     & $47.6 \pm 0.1$             & $68.2$          \\
                Robust model ratatouille$^{\dagger}$    & Uniform$^\dagger$  & $89.8$                     & $\underline{78.6}$         & $\underline{73.4}$         & $\textbf{52.1}$    & $\textbf{47.8}$            & $\textbf{68.3}$ \\
                \bottomrule
                \label{table:domainbedrobust}
            \end{tabular}}%
    \end{center}%
\end{table}%

\FloatBarrier
\section{DomainBed}
\label{app:domainbed}
\subsection{Experimental Details}%
\label{app:domainbeddetails}

\textbf{Datasets.}
We consider PACS \cite{li2017deeper}, VLCS \cite{fang2013unbiased}, OfficeHome \cite{venkateswara2017deep}, TerraIncognita \cite{beery2018recognition} and DomainNet \cite{peng2019moment}.
Domains are split into $80\%$ (used as training and evaluation) and $20\%$ (used as validation).
When considered as the target task, each domain is successively considered as the test domain while others are for training.
When considered as an auxiliary task, we train on all domains.
Critically, the procedure to obtain the pool of initializations is agnostic to the target task or the test domain, and thus is done only once.

\textbf{Training protocol.}
In all cases, we follow the training protocol from DomainBed.
For each dataset, we perform a random search of $20$ trials on the mild hyperparameter distributions described in Table \ref{tab:hyperparam}. We use a ResNet-50 \cite{he51deep} pre-trained on ImageNet, with a dropout layer before the newly added dense layer and fine-tuned with frozen batch normalization layers. The optimizer is Adam \cite{kingma2014adam}.
The linear probe classifier are obtained with default hyperparameters from \Cref{tab:hyperparam} and features extracted from the ImageNet pre-trained featurizer.
All runs are trained for $5$k steps, except on DomainNet for $15$k steps as done in concurrent works \cite{arpit2021ensemble,cha2021wad,rame2022diwa}.
When the featurizer was inter-trained on auxiliary datasets, it remains frozen during the first $200$ steps to prevent feature distortion \cite{kumar2022finetuning}.
As in \citet{rame2022diwa,cha2021wad}, validation accuracy is calculated every $50$ steps for VLCS, $500$ steps for DomainNet and $100$ steps for others. Our code is released at \urlcode.
\begin{table}[h!]%
    \centering%
    \caption{Hyperparameters, their default values and distributions for random search.}%
    \centering
    \adjustbox{max width=0.90\textwidth}{%
    \begin{tabular}{cccc}%
        \toprule
        \multirow{2}{*}{\textbf{Hyperparameter}} & \multirow{2}{*}{\textbf{Default value}} & \multicolumn{2}{c}{\textbf{Random distribution}}                          \\
                                                 &                                         & (DomainBed)                                      & (Ours, DiWA and SWAD)  \\
        \midrule
        Learning rate                            & $5\cdot 10^{-5}$                        & $10^{\U(-5,-3.5)}$                               & $[1,3,5]\cdot 10^{-5}$ \\
        Batch size                               & $32$                                    & $2^{\U(3,5.5)}$                                  & $32$                   \\
        ResNet dropout                           & $0$                                     & $[0,0.1,0.5]$                                    & $[0, 0.1, 0.5]$        \\
        Weight decay                             & $0$                                     & $10^{\U(-6,-2)}$                                 & $[10^{-6}, 10^{-4}]$   \\
        \bottomrule
    \end{tabular}}
    \label{tab:hyperparam}
\end{table}%
\FloatBarrier
\textbf{Baselines.}
Vanilla fine-tuning was named Empirical Risk Minimization in previous papers; CORAL \cite{coral216aaai} is the best invariance-based approach; their scores are taken from DomainBed \cite{gulrajani2021in}.
MA \cite{arpit2021ensemble} and SWAD \cite{cha2021wad} average weights along the trajectory of a vanilla fine-tuning; their scores are taken from their respective papers.
Deep ensembles$^{*}$ averages the predictions of $M=6$ models, each trained with different classifier initializations on different data splits; the scores are taken from \citet{arpit2021ensemble}.
Model soups \cite{Wortsman2022ModelSA} averages the weights obtained from different vanilla fine-tunings; for fair comparison, we report the scores achieved in DiWA \cite{rame2022diwa} with linear probing.
Fusing averages at initialization $5$ auxiliary weights $\phi_{i}^{\mathrm{aux}}$; for each of the $20$ runs and $0\leq i <5$, we sample $\kappa_i \sim \text{Unif}(0,4)$ and choose $\lambda_i=\frac{e^{\kappa_i}}{\sum_{j=0}^{4} e^{\kappa_j}}$, \ie the featurizer is initialized from $\sum_{i=0}^{4} \frac{e^{\kappa_i}}{\sum_{j=0}^{4} e^{\kappa_j}} \phi_{i}^{\mathrm{aux}}$.

\textbf{Model and weight selection.}
We consider the training-domain validation set protocol.
From each run, we thus take the weights at the epoch with maximum accuracy on the \iid validation dataset.
The greedy weight selection is also based on this \iid validation set.
This greedy strategy is not possible for $\dagger$ approaches, that average uniformly the $M=20\times3=60$ weights from the $3$ data splits: indeed, there is no shared \iid validation dataset.

\subsection{Results per Target Dataset and Domain}
\label{app:domainbedperdataset}
Tables below detail results per domain for the $5$ datasets from DomainBed. The average scores were reported in \Cref{table:domainbed}.
\FloatBarrier
\begin{table}[h]%
    \caption{Accuracy ($\%,\uparrow$) on PACS (best in \textbf{bold} and second \underline{underlined}).}%
    \centering
    \adjustbox{max width=0.90\textwidth}{%
        \begin{tabular}{lll|ccccc}
            \toprule
                                                                          & \textbf{Algorithm}                            & \textbf{Selection}    & \textbf{Art}               & \textbf{Cartoon}        & \textbf{Photo}             & \textbf{Sketch}         & \textbf{Avg}            \\
            \midrule
                                                                          & Vanilla fine-tuning                                           & \iid val              & $84.7 \pm 0.4$             & $80.8 \pm 0.6$          & $97.2 \pm 0.3$             & $79.3 \pm 1.0$          & $85.5 \pm 0.2$          \\
                                                                          & CORAL \cite{coral216aaai}                     & \iid val              & $88.3 \pm 0.2$             & $80.0 \pm 0.5$          & $97.5 \pm 0.3$             & $78.8 \pm 1.3$          & $86.2 \pm 0.3$          \\
                                                                          & SWAD \cite{cha2021wad}                        & Loss-aware trajectory & $89.3 \pm 0.5$             & $83.4 \pm 0.6$          & $97.3 \pm 0.3$             & $82.5 \pm 0.8$          & $88.1 \pm 0.1$          \\
                                                                          & MA \cite{arpit2021ensemble}                   & Uniform trajectory    & $89.1 \pm 0.1$             & $82.6 \pm 0.2$          & $97.6 \pm 0.0$             & $80.5 \pm 0.9$          & $87.5 \pm 0.2$          \\
                                                                          & Deep ensembles$^{*}$ \cite{arpit2021ensemble} & Uniform               & $88.3$                     & $83.6$                  & $96.5$                     & $81.9$                  & $87.6$                  \\
            \midrule
            \multirow{5}{*}{\begin{turn}{90}{\small DiWA runs}\end{turn}} & Vanilla fine-tuning                                           & \iid val              & $86.8 \pm 0.8$             & $80.6 \pm 1.0$          & $97.4 \pm 0.4$             & $78.7 \pm 2.0$          & $85.9 \pm 0.6$          \\
                                                                          & Ensemble$^{*}$                                & Uniform               & $89.6 \pm 0.2$             & $81.6 \pm 0.3$          & $97.8 \pm 0.2$             & $83.5 \pm 0.5$          & $88.1 \pm 0.3$          \\
                                                                          & Model soups                                   & Uniform               & $90.1 \pm 0.2$             & $82.8 \pm 0.6$          & $98.3 \pm 0.1$             & $83.3 \pm 0.4$          & $88.7 \pm 0.2$          \\
                                                                          & Model soups                                   & Greedy                & $89.3 \pm 0.2$             & $82.8 \pm 0.2$          & $98.0 \pm 0.1$             & $82.0 \pm 0.9$          & $88.0 \pm 0.3$          \\
                                                                          & Model soups$^\dagger$                         & Uniform$^\dagger$     & $90.6$                     & $83.4$                  & $98.2$                     & $83.8$                  & $89.0$                  \\
            \midrule
            \multirow{6}{*}{\begin{turn}{90}{\small Our runs}\end{turn}}  & Inter-training \cite{phang2018sentence}       & \iid val              & $89.2 \pm 1.0$             & $85.3 \pm 0.7$          & $97.5 \pm 0.0$             & $84.2 \pm 0.2$          & $89.0 \pm 0.0$          \\
                                                                          & Ensemble$^{*}$ of inter-training              & Uniform               & $90.4 \pm 0.2$             & $83.7 \pm 0.3$          & $97.9 \pm 0.2$             & $84.9 \pm 0.3$          & $89.2 \pm 0.1$          \\
                                                                          & Fusing \cite{choshen2022fusing}               & \iid val              & $\underline{90.8} \pm 0.1$ & $79.1 \pm 1.4$          & $98.0 \pm 0.4$             & $84.1 \pm 2.1$          & $88.0 \pm 1.0$          \\
                                                                          & Model ratatouille                             & Uniform               & $90.3 \pm 0.2$             & $84.4 \pm 0.1$          & $\underline{98.7} \pm 0.1$ & $84.8 \pm 0.1$          & $89.5 \pm 0.1$          \\
                                                                          & Model ratatouille                             & Greedy                & $\textbf{90.9} \pm 0.1$    & $\textbf{86.5} \pm 1.1$ & $98.6 \pm 0.0$             & $\textbf{85.9} \pm 0.4$ & $\textbf{90.5} \pm 0.2$ \\
                                                                          & Model ratatouille$^{\dagger}$                 & Uniform$^\dagger$     & $90.6$                     & $\underline{84.7}$      & $\textbf{98.8}$            & $\underline{85.0}$      & $\underline{89.8}$      \\
            \bottomrule
        \end{tabular}}
\end{table}

\begin{table}[h]%
    \caption{Accuracy ($\%,\uparrow$) on VLCS (best in \textbf{bold} and second \underline{underlined}).}%
    \centering
    \adjustbox{max width=0.90\textwidth}{%
        \begin{tabular}{lll|ccccc}
            \toprule
                                                                          & \textbf{Algorithm}                            & \textbf{Selection}    & \textbf{Caltech}        & \textbf{LabelMe}           & \textbf{SUN}               & \textbf{VOC}            & \textbf{Avg}               \\
            \midrule
                                                                          & Vanilla fine-tuning                                           & \iid val              & $97.7 \pm 0.4$          & $64.3 \pm 0.9$             & $73.4 \pm 0.5$             & $74.6 \pm 1.3$          & $77.5 \pm 0.4$             \\
                                                                          & CORAL \cite{coral216aaai}                     & \iid val              & $98.3 \pm 0.1$          & $\textbf{66.1} \pm 1.2$    & $73.4 \pm 0.3$             & $77.5 \pm 1.2$          & $78.8 \pm 0.6$             \\
                                                                          & SWAD \cite{cha2021wad}                        & Loss-aware trajectory & $98.8 \pm 0.1$          & $63.3 \pm 0.3$             & $\textbf{75.3} \pm 0.5$    & $79.2 \pm 0.6$          & $\textbf{79.1} \pm 0.1$    \\
                                                                          & MA \cite{arpit2021ensemble}                   & Uniform trajectory    & $99.0 \pm 0.2$          & $63.0 \pm 0.2$             & $\underline{74.5} \pm 0.3$ & $76.4 \pm 1.1$          & $78.2 \pm 0.2$             \\
                                                                          & Deep ensembles$^{*}$ \cite{arpit2021ensemble} & Uniform               & $98.7$                  & $64.5$                     & $72.1$                     & $78.9$                  & $78.5$                     \\
            \midrule
            \multirow{5}{*}{\begin{turn}{90}{\small DiWA runs}\end{turn}} & Vanilla fine-tuning                                           & \iid val              & $98.1 \pm 0.3$          & $64.4 \pm 0.3$             & $72.5 \pm 0.5$             & $77.7 \pm 1.3$          & $78.1 \pm 0.5$             \\
                                                                          & Ensemble$^{*}$                                & Uniform               & $98.5 \pm 0.1$          & $\underline{64.9} \pm 0.1$ & $73.4 \pm 0.4$             & $77.2 \pm 0.4$          & $78.5 \pm 0.1$             \\
                                                                          & Model soups                                   & Uniform               & $98.8 \pm 0.1$          & $62.8 \pm 0.2$             & $73.9 \pm 0.3$             & $78.3 \pm 0.1$          & $78.4 \pm 0.2$             \\
                                                                          & Model soups                                   & Greedy                & $98.4 \pm 0.0$          & $64.1 \pm 0.2$             & $73.3 \pm 0.4$             & $78.1 \pm 0.8$          & $78.5 \pm 0.1$             \\
                                                                          & Model soups$^\dagger$                         & Uniform$^\dagger$     & $98.9$                  & $62.4$                     & $73.9$                     & $78.9$                  & $78.6$                     \\
            \midrule
            \multirow{6}{*}{\begin{turn}{90}{\small Our runs}\end{turn}}  & Inter-training \cite{phang2018sentence}       & \iid val              & $98.2 \pm 0.0$          & $63.8 \pm 0.5$             & $72.3 \pm 0.5$             & $76.6 \pm 0.2$          & $77.7 \pm 0.0$             \\
                                                                          & Ensemble$^{*}$ of inter-training              & Uniform               & $98.9 \pm 0.1$          & $64.7 \pm 0.4$             & $73.8 \pm 0.5$             & $78.6 \pm 0.2$          & $\underline{79.0} \pm 0.2$ \\
                                                                          & Fusing \cite{choshen2022fusing}               & \iid val              & $98.4 \pm 0.4$          & $64.8 \pm 1.2$             & $72.2 \pm 0.9$             & $78.5 \pm 0.6$          & $78.5 \pm 0.8$             \\
                                                                          & Model ratatouille                             & Uniform               & $\textbf{99.3} \pm 0.0$ & $60.8 \pm 0.3$             & $74.3 \pm 0.3$             & $\textbf{79.5} \pm 0.3$ & $78.5 \pm 0.1$             \\
                                                                          & Model ratatouille                             & Greedy                & $99.0 \pm 0.0$          & $62.4 \pm 0.5$             & $73.8 \pm 0.3$             & $\textbf{79.5} \pm 0.1$ & $78.7 \pm 0.2$             \\
                                                                          & Model ratatouille$^{\dagger}$                 & Uniform$^\dagger$     & $\textbf{99.3}$         & $60.4$                     & $73.9$                     & $\textbf{79.5}$         & $78.3$                     \\
            \bottomrule
        \end{tabular}}
\end{table}

\begin{table}[h]%
    \caption{Accuracy ($\%,\uparrow$) on OfficeHome (best in \textbf{bold} and second \underline{underlined}).}%
    \centering
    \adjustbox{max width=0.90\textwidth}{%
        \begin{tabular}{lll|ccccc}
            \toprule
                                                                          & \textbf{Algorithm}                            & \textbf{Selection}    & \textbf{Art}               & \textbf{Clipart}        & \textbf{Product}        & \textbf{Photo}     & \textbf{Avg}               \\
            \midrule
                                                                          & Vanilla fine-tuning                                           & \iid val              & $61.3 \pm 0.7$             & $52.4 \pm 0.3$          & $75.8 \pm 0.1$          & $76.6 \pm 0.3$     & $66.5 \pm 0.3$             \\
                                                                          & CORAL \cite{coral216aaai}                     & \iid val              & $65.3 \pm 0.4$             & $54.4 \pm 0.5$          & $76.5 \pm 0.1$          & $78.4 \pm 0.5$     & $68.7 \pm 0.3$             \\
                                                                          & SWAD \cite{cha2021wad}                        & Loss-aware trajectory & $66.1 \pm 0.4$             & $57.7 \pm 0.4$          & $78.4 \pm 0.1$          & $80.2 \pm 0.2$     & $70.6 \pm 0.2$             \\
                                                                          & MA \cite{arpit2021ensemble}                   & Uniform trajectory    & $66.7 \pm 0.5$             & $57.1 \pm 0.1$          & $78.6 \pm 0.1$          & $80.0 \pm 0.0$     & $70.6 \pm 0.1$             \\
                                                                          & Deep ensembles$^{*}$ \cite{arpit2021ensemble} & Uniform               & $65.6$                     & $58.5$                  & $78.7$                  & $80.5$             & $70.8$                     \\
            \midrule
            \multirow{5}{*}{\begin{turn}{90}{\small DiWA runs}\end{turn}} & Vanilla fine-tuning                                           & \iid val              & $63.9 \pm 1.2$             & $54.8 \pm 0.6$          & $78.7 \pm 0.1$          & $80.4 \pm 0.2$     & $69.4 \pm 0.2$             \\
                                                                          & Ensemble$^{*}$                                & Uniform               & $67.0 \pm 0.1$             & $57.9 \pm 0.4$          & $80.0 \pm 0.2$          & $81.7 \pm 0.3$     & $71.7 \pm 0.1$             \\
                                                                          & Model soups                                   & Uniform               & $68.4 \pm 0.2$             & $58.2 \pm 0.5$          & $80.0 \pm 0.1$          & $81.7 \pm 0.3$     & $72.1 \pm 0.2$             \\
                                                                          & Model soups                                   & Greedy                & $67.8 \pm 0.5$             & $57.2 \pm 0.5$          & $79.6 \pm 0.1$          & $81.4 \pm 0.4$     & $71.5 \pm 0.2$             \\
                                                                          & Model soups$^\dagger$                         & Uniform$^\dagger$     & $69.2$                     & $59.0$                  & $\textbf{80.6}$         & $\underline{82.2}$ & $72.8$                     \\
            \midrule
            \multirow{6}{*}{\begin{turn}{90}{\small Our runs}\end{turn}}  & Inter-training \cite{phang2018sentence}       & \iid val              & $65.3 \pm 0.3$             & $55.8 \pm 2.2$          & $78.6 \pm 0.1$          & $80.1 \pm 0.2$     & $69.9 \pm 0.6$             \\
                                                                          & Ensemble$^{*}$ of inter-training              & Uniform               & $67.8 \pm 0.1$             & $60.5 \pm 0.1$          & $80.5 \pm 0.2$          & $82.0 \pm 0.2$     & $72.7 \pm 0.1$             \\
                                                                          & Fusing \cite{choshen2022fusing}               & \iid val              & $66.4 \pm 0.5$             & $59.8 \pm 1.2$          & $78.8 \pm 0.2$          & $81.0 \pm 0.3$     & $71.5 \pm 0.5$             \\
                                                                          & Model ratatouille                             & Uniform               & $69.8 \pm 0.1$             & $60.3 \pm 0.2$          & $80.4 \pm 0.1$          & $81.8 \pm 0.2$     & $73.1 \pm 0.1$             \\
                                                                          & Model ratatouille                             & Greedy                & $\underline{70.0} \pm 0.2$ & $\textbf{60.8} \pm 1.0$ & $\textbf{80.6} \pm 0.1$ & $82.0 \pm 0.2$     & $\underline{73.4} \pm 0.3$ \\
                                                                          & Model ratatouille$^{\dagger}$                 & Uniform$^\dagger$     & $\textbf{70.4}$            & $\underline{60.7}$      & $\textbf{80.6}$         & $\textbf{82.3}$    & $\textbf{73.5}$            \\
            \bottomrule
        \end{tabular}}
    \label{apptab:home}
\end{table}

\begin{table}[h]%
    \caption{Accuracy ($\%,\uparrow$) on TerraIncognita (best in \textbf{bold} and second \underline{underlined}).}%
    \centering
    \adjustbox{max width=0.90\textwidth}{%
        \begin{tabular}{lll|ccccc}
            \toprule
                                                                          & \textbf{Algorithm}                            & \textbf{Selection}    & \textbf{L100}              & \textbf{L38}               & \textbf{L43}            & \textbf{L46}            & \textbf{Avg}       \\
            \midrule
                                                                          & Vanilla fine-tuning                                           & \iid val              & $49.8 \pm 4.4$             & $42.1 \pm 1.4$             & $56.9 \pm 1.8$          & $35.7 \pm 3.9$          & $46.1 \pm 1.8$     \\
                                                                          & CORAL \cite{coral216aaai}                     & \iid val              & $51.6 \pm 2.4$             & $42.2 \pm 1.0$             & $57.0 \pm 1.0$          & $39.8 \pm 2.9$          & $47.6 \pm 1.0$     \\
                                                                          & SWAD \cite{cha2021wad}                        & Loss-aware trajectory & $55.4 \pm 0.0$             & $44.9 \pm 1.1$             & $59.7 \pm 0.4$          & $39.9 \pm 0.2$          & $50.0 \pm 0.3$     \\
                                                                          & MA \cite{arpit2021ensemble}                   & Uniform trajectory    & $54.9 \pm 0.4$             & $45.5 \pm 0.6$             & $60.1 \pm 1.5$          & $40.5 \pm 0.4$          & $50.3 \pm 0.5$     \\
                                                                          & Deep ensembles$^{*}$ \cite{arpit2021ensemble} & Uniform               & $53.0$                     & $42.6$                     & $60.5$                  & $40.8$                  & $49.2$             \\
            \midrule
            \multirow{5}{*}{\begin{turn}{90}{\small DiWA runs}\end{turn}} & Vanilla fine-tuning                                           & \iid val              & $\textbf{59.9} \pm 4.2$    & $46.9 \pm 0.9$             & $54.6 \pm 0.3$          & $40.1 \pm 2.2$          & $50.4 \pm 1.8$     \\
                                                                          & Ensemble$^{*}$                                & Uniform               & $55.6 \pm 1.4$             & $45.4 \pm 0.4$             & $\textbf{61.0} \pm 0.4$ & $\textbf{41.3} \pm 0.3$ & $50.8 \pm 0.5$     \\
                                                                          & Model soups                                   & Uniform               & $56.3 \pm 1.9$             & $49.4 \pm 0.7$             & $59.9 \pm 0.4$          & $39.8 \pm 0.5$          & $51.4 \pm 0.6$     \\
                                                                          & Model soups                                   & Greedy                & $\underline{58.5} \pm 2.2$ & $48.2 \pm 0.3$             & $58.5 \pm 0.3$          & $41.1 \pm 1.2$          & $51.6 \pm 0.9$     \\
                                                                          & Model soups$^\dagger$                         & Uniform$^\dagger$     & $57.2$                     & $\underline{50.1}$         & $60.3$                  & $39.8$                  & $\underline{51.9}$ \\
            \midrule
            \multirow{6}{*}{\begin{turn}{90}{\small Our runs}\end{turn}}  & Inter-training \cite{phang2018sentence}       & \iid val              & $49.9 \pm 1.7$             & $44.3 \pm 1.6$             & $54.7 \pm 0.4$          & $37.9 \pm 1.1$          & $46.7 \pm 0.1$     \\
                                                                          & Ensemble$^{*}$ of inter-training              & Uniform               & $58.1 \pm 0.2$             & $43.8 \pm 0.4$             & $\textbf{61.0} \pm 0.2$ & $\textbf{41.3} \pm 0.4$ & $51.1 \pm 0.3$     \\
                                                                          & Fusing \cite{choshen2022fusing}               & \iid val              & $52.8 \pm 3.2$             & $43.2 \pm 2.3$             & $55.2 \pm 1.3$          & $35.5 \pm 0.3$          & $46.7 \pm 1.8$     \\
                                                                          & Model ratatouille                             & Uniform               & $57.9 \pm 0.2$             & $\underline{50.1} \pm 0.7$ & $59.8 \pm 0.1$          & $38.9 \pm 0.5$          & $51.8 \pm 0.4$     \\
                                                                          & Model ratatouille                             & Greedy                & $54.0 \pm 2.0$             & $47.7 \pm 0.8$             & $57.3 \pm 0.8$          & $37.9 \pm 1.2$          & $49.2 \pm 0.9$     \\
                                                                          & Model ratatouille$^{\dagger}$                 & Uniform$^\dagger$     & $57.9$                     & $\textbf{50.6}$            & $60.2$                  & $39.2$                  & $\textbf{52.0}$    \\
            \bottomrule
        \end{tabular}}
\end{table}
\begin{table}[h]%
    \caption{Accuracy ($\%,\uparrow$) on DomainNet (best in \textbf{bold} and second \underline{underlined}).}%
    \centering
    \adjustbox{max width=0.90\textwidth}{%
        \begin{tabular}{lll|ccccccc}
            \toprule
                                                                          & \textbf{Algorithm}                            & \textbf{Selection}    & \textbf{Clipart}           & \textbf{Info}           & \textbf{Painting}  & \textbf{QuickDraw}      & \textbf{Photo}     & \textbf{Sketch}            & \textbf{Avg}            \\
            \midrule
                                                                          & Vanilla fine-tuning                                           & \iid val              & $58.1 \pm 0.3$             & $18.8 \pm 0.3$          & $46.7 \pm 0.3$     & $12.2 \pm 0.4$          & $59.6 \pm 0.1$     & $49.8 \pm 0.4$             & $40.9 \pm 0.1$          \\
                                                                          & CORAL \cite{coral216aaai}                     & \iid val              & $59.2 \pm 0.1$             & $19.7 \pm 0.2$          & $46.6 \pm 0.3$     & $13.4 \pm 0.4$          & $59.8 \pm 0.2$     & $50.1 \pm 0.6$             & $41.5 \pm 0.1$          \\
                                                                          & SWAD \cite{cha2021wad}                        & Loss-aware trajectory & $66.0 \pm 0.1$             & $22.4 \pm 0.3$          & $53.5 \pm 0.1$     & $16.1 \pm 0.2$          & $65.8 \pm 0.4$     & $55.5 \pm 0.3$             & $46.5 \pm 0.1$          \\
                                                                          & MA \cite{arpit2021ensemble}                   & Uniform trajectory    & $64.4 \pm 0.3$             & $22.4 \pm 0.2$          & $53.4 \pm 0.3$     & $15.4 \pm 0.1$          & $64.7 \pm 0.2$     & $55.5 \pm 0.1$             & $46.0 \pm 0.1$          \\
                                                                          & Deep ensembles$^{*}$ \cite{arpit2021ensemble} & Uniform               & $68.3$                     & $23.1$                  & $54.5$             & $16.3$                  & $66.9$             & $\textbf{57.0}$            & $\textbf{47.7}$         \\
            \midrule
            \multirow{5}{*}{\begin{turn}{90}{\small DiWA runs}\end{turn}} & Vanilla fine-tuning                                           & \iid val              & $63.4 \pm 0.2$             & $21.1 \pm 0.4$          & $50.7 \pm 0.3$     & $13.5 \pm 0.4$          & $64.8 \pm 0.4$     & $52.4 \pm 0.1$             & $44.3 \pm 0.2$          \\
                                                                          & Ensemble$^{*}$                                & Uniform               & $\underline{66.7} \pm 0.4$ & $22.2 \pm 0.1$          & $54.1 \pm 0.2$     & $15.1 \pm 0.2$          & $68.4 \pm 0.1$     & $55.7 \pm 0.2$             & $47.0 \pm 0.2$          \\
                                                                          & Model soups                                   & Uniform               & $65.9 \pm 0.4$             & $23.0 \pm 0.2$          & $55.0 \pm 0.3$     & $16.1 \pm 0.2$          & $68.4 \pm 0.1$     & $55.7 \pm 0.4$             & $47.4 \pm 0.2$          \\
                                                                          & Model soups                                   & Greedy                & $\underline{66.7} \pm 0.2$ & $\textbf{23.3} \pm 0.2$ & $55.3 \pm 0.1$     & $16.3 \pm 0.2$          & $68.2 \pm 0.0$     & $\underline{56.2} \pm 0.1$ & $\textbf{47.7} \pm 0.1$ \\
                                                                          & Model soups$^\dagger$                         & Uniform$^\dagger$     & $66.2$                     & $\textbf{23.3}$         & $\underline{55.4}$ & $16.5$                  & $\textbf{68.7}$    & $56.0$                     & $\textbf{47.7}$         \\
            \midrule
            \multirow{6}{*}{\begin{turn}{90}{\small Our runs}\end{turn}}  & Inter-training \cite{phang2018sentence}       & \iid val              & $63.5 \pm 0.1$             & $21.1 \pm 0.1$          & $51.2 \pm 0.2$     & $14.2 \pm 0.2$          & $64.7 \pm 0.3$     & $52.1 \pm 0.1$             & $44.5 \pm 0.1$          \\
                                                                          & Ensemble$^{*}$ of inter-training              & Uniform               & $\textbf{66.8} \pm 0.2$    & $22.3 \pm 0.0$          & $54.2 \pm 0.2$     & $15.4 \pm 0.2$          & $68.3 \pm 0.0$     & $55.8 \pm 0.2$             & $47.2 \pm 0.1$          \\
                                                                          & Fusing \cite{choshen2022fusing}               & \iid val              & $63.6 \pm 0.1$             & $21.3 \pm 0.1$          & $51.4 \pm 0.2$     & $14.0 \pm 0.2$          & $64.1 \pm 0.4$     & $52.1 \pm 0.3$             & $44.4 \pm 0.2$          \\
                                                                          & Model ratatouille                             & Uniform               & $65.9 \pm 0.2$             & $23.0 \pm 0.1$          & $55.1 \pm 0.0$     & $16.5 \pm 0.1$          & $68.3 \pm 0.0$     & $55.8 \pm 0.0$             & $47.5 \pm 0.1$          \\

                                                                          & Model ratatouille                             & Greedy                & $66.5 \pm 0.1$             & $23.2 \pm 0.1$          & $55.3 \pm 0.0$     & $\textbf{16.7} \pm 0.1$ & $68.0 \pm 0.0$     & $56.0 \pm 0.0$             & $\textbf{47.7} \pm 0.0$ \\
                                                                          & Model ratatouille$^{\dagger}$                 & Uniform$^\dagger$     & $66.1$                     & $23.1$                  & $\textbf{55.5}$    & $\textbf{16.7}$         & $\underline{68.5}$ & $56.0$                     & $\textbf{47.7}$         \\
            \bottomrule
        \end{tabular}}
\end{table}

\FloatBarrier
\subsection{Additional experiments}
\label{app:expe}

\subsubsection{Improved TerraIncognita}
Ratatouille has significant gains over model soups on some datasets. Yet, the gains are indeed moderate on DomainNet (47.4\% to 47.5\% with uniform selection) and TerraIncognita (51.4\% to 51.8\%). In particular for TerraIncognita, this small gain is because other tasks from DomainBed are distant from photos of animals in the wild and even detrimental, explaining the very low performances of inter-trainings (46.7\% versus 50.4\% for ERM); ratatouille manages to fill the gap by increased diversity across fine-tunings. This is an evidence of ratatouille's robustness to the choice of auxiliary tasks. To validate that more similar auxiliary tasks can help on TerraIncognita, we run an additional experiment with iWildCam \cite{beery2021iwildcam} as a (similar) auxiliary task: as detailed in \Cref{table:terraiwilcam}, we reach 52.9\% averaged accuracy.

\begin{table}[h!]%
    \begin{center}
        \caption{Accuracies ($\%,\uparrow$) on TerraIncongita with uniform selection.}
        \resizebox{0.8\textwidth}{!}{%
            \begin{tabular}{llccccc}
                \toprule
                \textbf{Algorithm} & \textbf{Auxiliary datasets} & \textbf{L100} & \textbf{L38}  & \textbf{L43}  & \textbf{L46}  & \textbf{Avg}  \\
                \midrule
                Soups              & \xmark                      & 56.3          & 49.4          & 59.9          & 39.8          & 51.4          \\
                Ratatouille        & DomainBed's                 & 57.9          & 50.1          & 59.8          & 38.9          & 51.8          \\
                Ratatouille        & iWildCam                    & \textbf{59.8} & \textbf{50.3} & \textbf{60.0} & \textbf{41.4} & \textbf{52.9} \\
                \bottomrule
                \label{table:terraiwilcam}
            \end{tabular}}%
    \end{center}%
\end{table}%

\subsubsection{Camelyon}

We conduct some experiments on the Camelyon~\cite{pmlr-v139-koh21a} dataset from the WILDS~\cite{pmlr-v139-koh21a} benchmark, where the task is to classify \enquote{breast cancer metastases in whole-slide images of histological lymph node sections}, with each hospital successively considered as the test while others are for training. The results in \Cref{table:camelyon} show that model ratatouille consistently beats model soups on Camelyon for histopathology. These results may facilitate the adoption of ratatouille in the medical community \cite{maron2022model}.

\begin{table}[h!]%
    \begin{center}
        \caption{Accuracies ($\%,\uparrow$) on Camelyon.}
        \resizebox{\textwidth}{!}{%
            \begin{tabular}{llcccccc}
                \toprule
                \textbf{Selection}       & \textbf{Algorithm} & \textbf{Hospital 1} & \textbf{Hospital 2} & \textbf{Hospital 3} & \textbf{Hospital 4} & \textbf{Hospital 5} & \textbf{Avg} \\
                \midrule
                \multirow{2}{*}{Uniform} & Soups              & 96.4                & 94.3                & 96.1                & 94.2                & 90.4                & 94.3         \\
                                         & Ratatouille        & 97.1                & 94.4                & 96.1                & 94.8                & 90.5                & 94.6         \\
                \midrule
                \multirow{2}{*}{Greedy}  & Soups              & 97.4                & 95.1                & 96.5                & 96.1                & 90.6                & 95.1         \\
                                         & Ratatouille        & 97.5                & 95.3                & 96.7                & 96.6                & 90.8                & 95.4         \\
                \bottomrule
                \label{table:camelyon}
            \end{tabular}}%
    \end{center}%
\end{table}%

\end{document}